\newcommand{\comment}[1]{}
\comment{Submission Version
}

\documentclass[twoside,11pt]{article}
\usepackage{jmlr2e}

\usepackage{subfigure}
\usepackage{graphicx}
\usepackage{amsmath}

\usepackage{algorithm}
\usepackage{algorithmic}

\newcommand{\INPUT}{\STATE {\bfseries input}~}
\newcommand{\OUTPUT}{\STATE {\bfseries output}~}

\usepackage{url}
\usepackage{color}
\usepackage[colorlinks,citecolor=blue]{hyperref}

\input{Definitions}
\graphicspath{{./}}    

\renewcommand{\inner}[2]{{#1}^{\!\top}\!{#2}}  
\newcommand{\samefootnote}{\addtocounter{footnote}{-1}\footnotemark{}}

\newcommand{\DIRINIT}{Line 3}
\newcommand{\DIRSTOP}{Line 6}
\newcommand{\DIRUPDATE}{Line 9}

\newcommand{\DIRRETURN}{Line 18}

\newcommand{\DIRCOE}{Line 7}
\newcommand{\AGGREGATEGRAD}{Line 8}
\newcommand{\SUBBFGSGRAD}{Line 12}
\newcommand{\SUBBFGSS}{Line 14}
\newcommand{\SUBBFGSSTEP}{Line 9}
\newcommand{\BLSCALHINGE}{Lines 4--5}
\newcommand{\MLSUPDATESLOPE}{Lines 20--23}
\newcommand{\MULTILSREVERSESTACK}{Lines 6--8}

\jmlrheading{11}{2010}{1--57}{11/08}{-/10}{
Jin Yu, S.V$\!.\,$N. Vishwanathan, Simon G\"unter, and Nicol N. Schraudolph}

\ShortHeadings{Quasi-Newton Approach to Nonsmooth Convex
  Optimization}{Yu, Vishwanathan, G\"unter, and Schraudolph}

\firstpageno{1}

\begin{document}

\title{A Quasi-Newton Approach to Nonsmooth \\
Convex Optimization Problems in Machine Learning}

\author{\name Jin Yu \email jin.yu@adelaide.edu.au \\
  \addr School of Computer Science \\
   The University of Adelaide \\
   Adelaide SA 5005, Australia \\[1ex]
  \name S.V\!.\,N. Vishwanathan \email  vishy@stat.purdue.edu \\
  \addr Departments of Statistics and Computer Science \\
  Purdue University \\
  West Lafayette, IN 47907-2066 USA \\[1ex]
  \name Simon  G\"unter \email guenter\_simon@hotmail.com \\
  \addr DV Bern AG \\
  Nussbaumstrasse 21, CH-3000 Bern 22, Switzerland \\[1ex]
  \name Nicol N. Schraudolph \email jmlr@schraudolph.org \\
  \addr adaptive tools AG \\
   Canberra ACT 2602, Australia}
  
\editor{Sathiya Keerthi}

\maketitle
\begin{abstract}
  We extend the well-known BFGS quasi-Newton method and its
  memory-limited variant LBFGS to the optimization of nonsmooth convex
  objectives. This is done in a rigorous fashion by generalizing three
  components of BFGS to subdifferentials: the local quadratic model,
  the identification of a descent direction, and the Wolfe line search
  conditions. We prove that under some technical conditions, the
  resulting subBFGS algorithm is globally convergent in objective
  function value.  We apply its memory-limited variant (subLBFGS)
  to $L_2$-regularized risk minimization with the binary hinge
  loss. To extend our algorithm to the multiclass and multilabel
  settings, we develop a new, efficient, exact line search
  algorithm. We prove its worst-case time complexity bounds, and show
  that our line search can also be used to extend a recently developed
  bundle method to the multiclass and multilabel settings.
  We also apply the direction-finding component of our algorithm to
  $L_1$-regularized risk minimization with logistic loss. In all these
  contexts our methods perform comparable to or better than
  specialized state-of-the-art solvers on a number of publicly
  available datasets.  An open source implementation of our
  algorithms is freely available.
\end{abstract}
\vspace{0.1in}

\begin{keywords}
  BFGS, Variable Metric Methods, Wolfe Conditions, Subgradient, Risk
  Minimization, Hinge Loss, Multiclass, Multilabel, Bundle Methods,
  BMRM, OCAS, OWL-QN
\end{keywords}

\section{Introduction}
\label{sec:intro}

\begin{figure}[tb]
  \centering
  \includegraphics[width=0.6\textwidth]{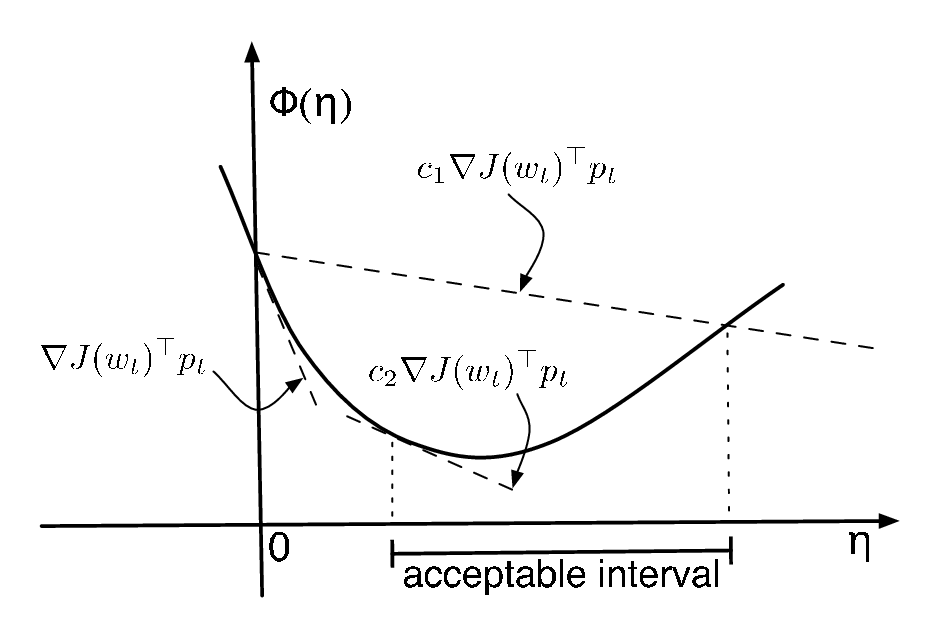}
  \caption{Geometric illustration of the Wolfe conditions
    \eqref{eq:wolfe1} and \eqref{eq:wolfe2}.}
  \label{fig:wolfe}
\end{figure}

The BFGS quasi-Newton method \citep{NocWri99} and its memory-limited
LBFGS variant are widely regarded as the workhorses of smooth nonlinear
optimization due to their combination of computational efficiency and
good asymptotic convergence. Given a smooth objective function $J: \RR^d
\rightarrow \RR$ and a current iterate $\vw_t \in \RR^d$, BFGS forms a
local quadratic model of $J$: 
\begin{align}
  \label{eq:quadratic-model}
  Q_t(\vp) ~:=~ J(\vw_t) + \half \inner{\vp}{\Bmat_t^{-1} \vp} +
  \inner{\nabla J(\vw_t)}{\vp} \,,
\end{align}
where $\Bmat_t \succ 0$ is a positive-definite estimate of the inverse
Hessian of $J$, and $\nabla J$ denotes the gradient. Minimizing
$Q_t(\vp)$ gives the quasi-Newton direction
\begin{align}
\label{eq:quasi-newton-dir}
 \vp_t := -\Bmat_t\nabla J(\vw_t),
\end{align}
which is used for the parameter update:
\begin{align}
  \label{eq:update}
  \vw_{t+1} = \vw_t + \eta_t \vp_t.
\end{align}
The step size $\eta_t > 0$ is normally determined by a line search
obeying the \citet{Wolfe69} conditions:
\begin{align}
  \label{eq:wolfe1}
  J(\vw_{t+1}) ~\le~ J(\vw_t) + c_1\eta_t & \inner{\nabla
  J(\vw_t)}{\vp_t} \mbox{~~~~~~~~(sufficient decrease)} \\ \mbox{and~~~} 
  \inner{\nabla J(\vw_{t+1})}{\vp_t} ~\ge~ c_2 & \inner{\nabla
  J(\vw_t)}{\vp_t} \mbox{~~~~~~~(curvature)}
  \label{eq:wolfe2}
\end{align}
with $0 < c_1 < c_2 < 1$. Figure~\ref{fig:wolfe} illustrates these
conditions geometrically. The matrix $\Bmat_t$ is then modified
via the incremental rank-two update
\begin{align}
  \label{eq:bfgs-update}
  \Bmat_{t+1} = (\one - \rho_t \vs_t \vy_t^{\top}) \Bmat_t (\one -
  \rho_t \vy_t \vs_t^{\top}) + \rho_t \vs_t \vs_t^{\top},
\end{align}
where $\vs_t := \vw_{t+1} - \vw_t$ and $\vy_t := \nabla J(\vw_{t+1}) -
\nabla J(\vw_{t})$ denote the most recent step along the optimization
trajectory in parameter and gradient space, respectively, and $\rho_t
:= (\inner{\vy_t}{\vs_t})^{-1}$. The BFGS update \eqref{eq:bfgs-update}
enforces the secant equation $\Bmat_{t+1}\vy_t = \vs_t$.
Given a descent direction $\vp_t$, the
Wolfe conditions ensure that $(\forall t) ~\vs_t^{\top} \!\vy_t > 0$ and
hence $\Bmat_0 \succ 0 \implies (\forall t) ~\Bmat_t \succ 0$.

Limited-memory BFGS \citep[LBFGS,][]{LiuNoc89} is a variant of BFGS designed for
high-dimensional optimization problems where the $O(d^2)$ cost of storing and
updating $\Bmat_t$ would be prohibitive. LBFGS approximates the quasi-Newton
direction \eqref{eq:quasi-newton-dir} directly from the last $m$ pairs of
$\vs_t$ and $\vy_t$ via a matrix-free approach, reducing the cost to $O(md)$
space and time per iteration, with $m$ freely chosen.

There have been some attempts to apply (L)BFGS directly to nonsmooth
optimization problems, in the hope that they would perform well on
nonsmooth functions that are convex and differentiable almost
everywhere. Indeed, it has been noted that in cases where BFGS
(\emph{resp.}  LBFGS) does not encounter any nonsmooth point, it often
converges to the optimum \citep{Lemarechal82,LewOve08a}. However,
\citet{LukVlc99}, \citet{Haarala04}, and \citet{LewOve08b} also report
catastrophic failures of (L)BFGS on nonsmooth functions. Various fixes
can be used to avoid this problem, but only in an ad-hoc
manner. Therefore, subgradient-based approaches such as subgradient
descent \citep{NedBer00} or bundle methods
\citep{Joachims06,FraSon08,TeoVisSmoLe09} have gained considerable
attention for minimizing nonsmooth objectives.

\begin{figure}[tb]
  \centering
  \includegraphics[width=0.6\textwidth]{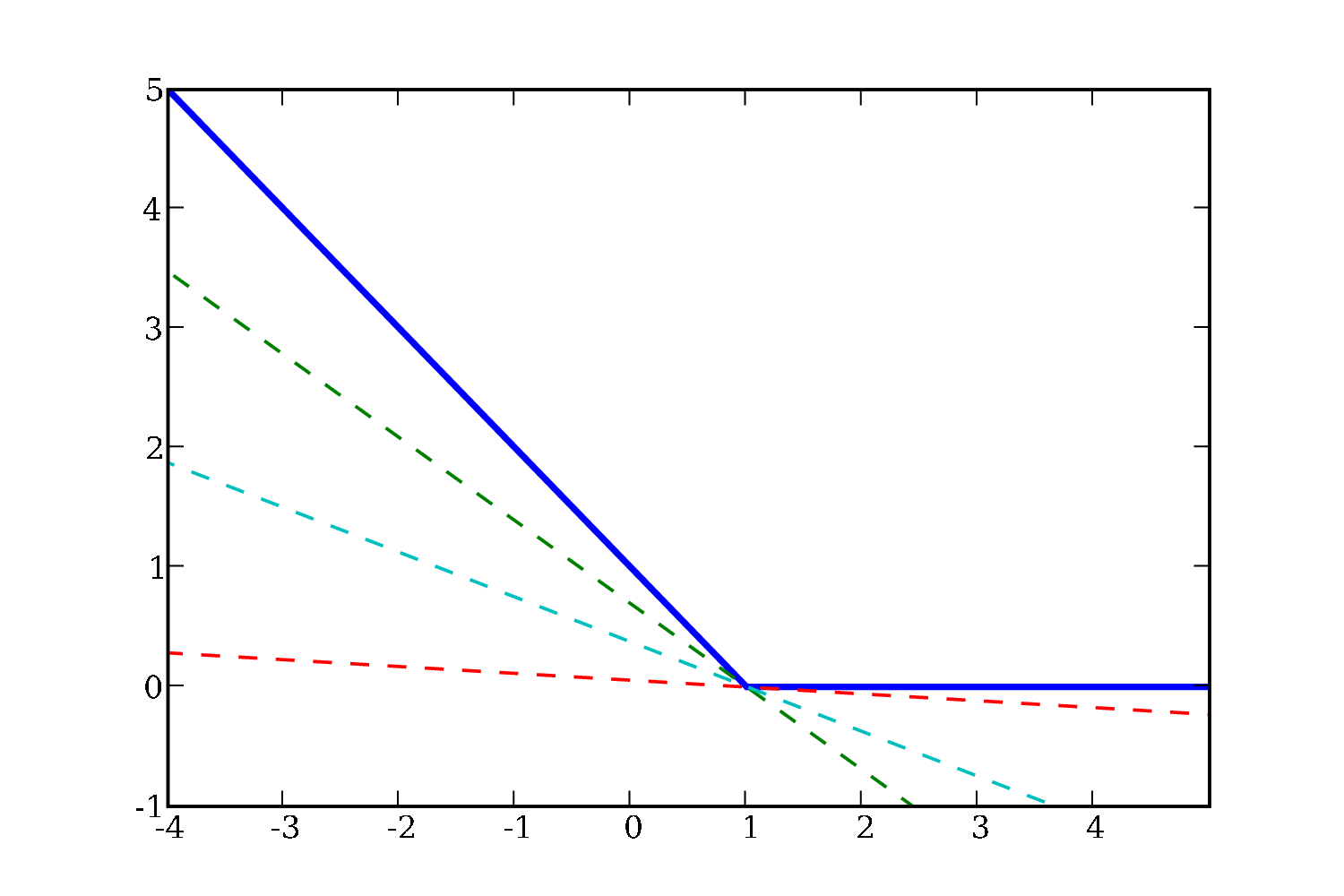}
  \caption{Geometric interpretation of subgradients. The dashed lines are
    tangential to the hinge function (solid blue line); the slopes of
    these lines are subgradients.}
  \label{fig:subgrad}
\end{figure}

Although a convex function might not be differentiable everywhere, a subgradient
always exists \citep{HirLem93}. Let $\vw$ be a point where a convex function $J$
is finite. Then a subgradient is the normal vector of any tangential supporting
hyperplane of $J$ at $\vw$. Formally, $\vg$ is called a subgradient of $J$ at
$\vw$ if and only if \citep[Definition VI.1.2.1]{HirLem93}
\begin{align}
  \label{eq:subgrad-def}
  (\forall \vw') ~~J(\vw') ~\geq~ J(\vw) + (\vw' - \vw)^{\top} \!\vg .
\end{align}
The set of all subgradients at a point is called the subdifferential, and is
denoted $\partial J(\vw)$. If this set is not empty then $J$ is said to be {\em
  subdifferentiable at} $\vw$. If it contains exactly one element, \ie $\partial
J(\vw) = \{\nabla J(\vw)\}$, then $J$ is \emph{differentiable} at
$\vw$. Figure~\ref{fig:subgrad} provides the geometric interpretation of
\eqref{eq:subgrad-def}.

The aim of this paper is to develop principled and robust quasi-Newton
methods that are amenable to subgradients. This results in subBFGS and
its memory-limited variant subLBFGS, two new subgradient quasi-Newton
methods that are applicable to nonsmooth convex optimization
problems. In particular, we apply our algorithms to a variety
of machine learning problems, exploiting knowledge about the
subdifferential of the binary hinge loss and its generalizations to the
multiclass and multilabel settings.

In the next section we motivate our work by illustrating the difficulties of
LBFGS on nonsmooth functions, and the advantage of incorporating BFGS' curvature
estimate into the parameter update. In Section~\ref{sec:subbfgs} we develop our
optimization algorithms generically, before discussing their application to
$L_{2}$-regularized risk minimization with the hinge loss in
Section~\ref{sec:subbfgs-hingeloss}. We describe a new efficient algorithm to
identify the nonsmooth points of a one-dimensional pointwise maximum of linear
functions in Section~\ref{sec:Minimization1DConvex}, then use it to develop an
exact line search that extends our optimization algorithms to the multiclass and
multilabel settings
(Section~\ref{sec:subbfgs-multiloss}). Section~\ref{sec:related-work} compares
and contrasts our work with other recent efforts in this area. We report our
experimental results on a number of public datasets in
Section~\ref{sec:results}, and conclude with a discussion and outlook in
Section~\ref{sec:discuss}.

\section{Motivation}
\label{sec:motivation}

The application of standard (L)BFGS to nonsmooth optimization is
problematic since the quasi-Newton direction generated at a
nonsmooth point is not necessarily a descent direction. Nevertheless,
BFGS' inverse Hessian estimate can provide an effective model of the
overall shape of a nonsmooth objective; incorporating it into the
parameter update can therefore be beneficial. We discuss these two aspects
of (L)BFGS to motivate our work on developing new quasi-Newton methods
that are amenable to subgradients while preserving the fast
convergence properties of standard (L)BFGS.

\subsection{Problems of (L)BFGS on Nonsmooth Objectives}
\label{sec:problems-lbfgs}

Smoothness of the objective function is essential for classical
(L)BFGS because both the local quadratic model
\eqref{eq:quadratic-model} and the Wolfe conditions
(\ref{eq:wolfe1},\,\ref{eq:wolfe2}) require the existence of the
gradient $\nabla J$ at every point. As pointed out by \citet[Remark
VIII.2.1.3]{HirLem93}, even though nonsmooth convex functions are
differentiable everywhere except on a set of Lebesgue measure zero, it
is unwise to just use a smooth optimizer on a nonsmooth convex problem
under the assumption that ``it should work almost surely.'' Below we
illustrate this on both a toy example and real-world machine learning
problems.

\subsubsection{A Toy Example}
\begin{figure}
  \centering
    \begin{tabular}{@{$\!\!$}c@{$\!\!$}c@{$\!$}c}
      \includegraphics[width=0.35\textwidth]{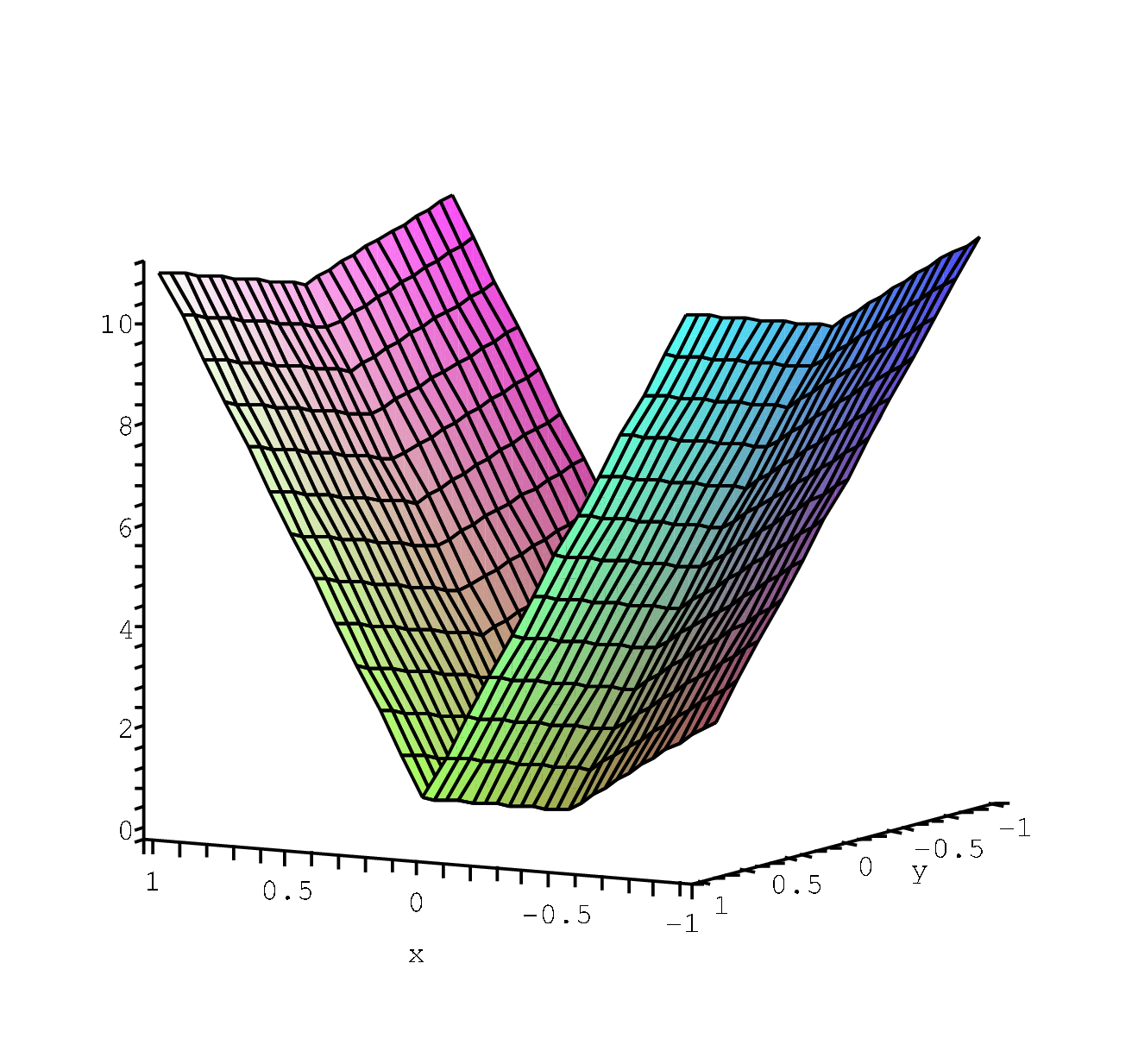} &
      \includegraphics[width=0.34\textwidth]{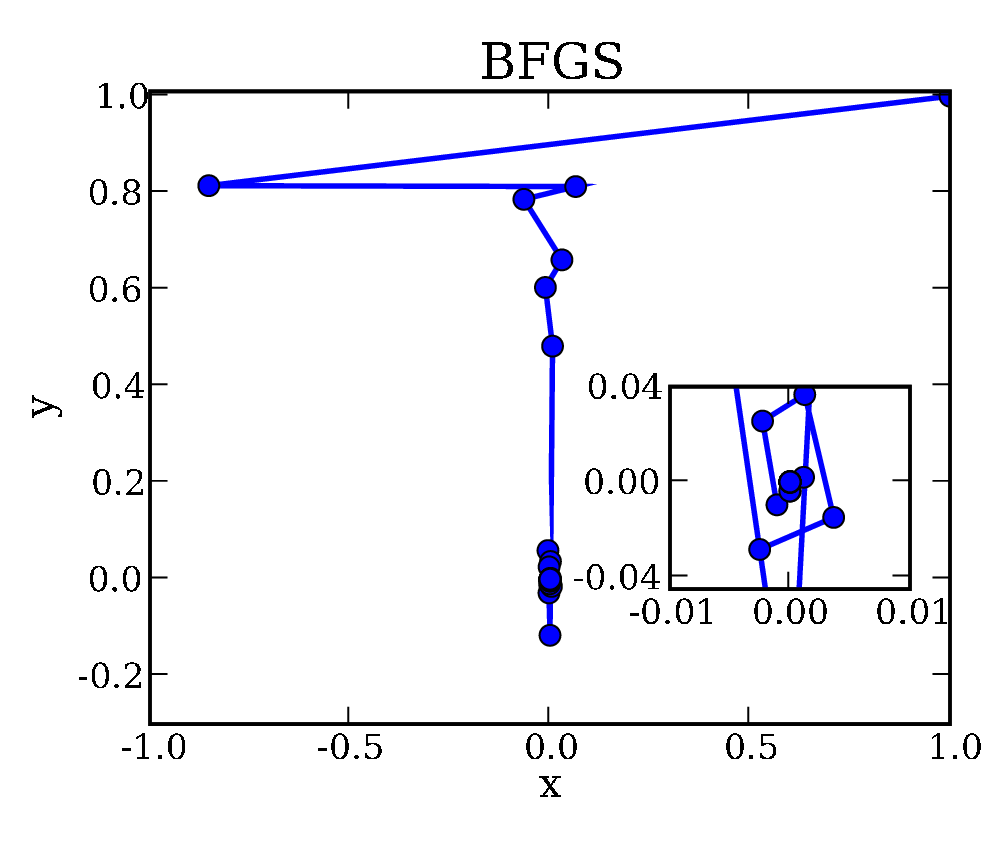} &
      \includegraphics[width=0.34\textwidth]{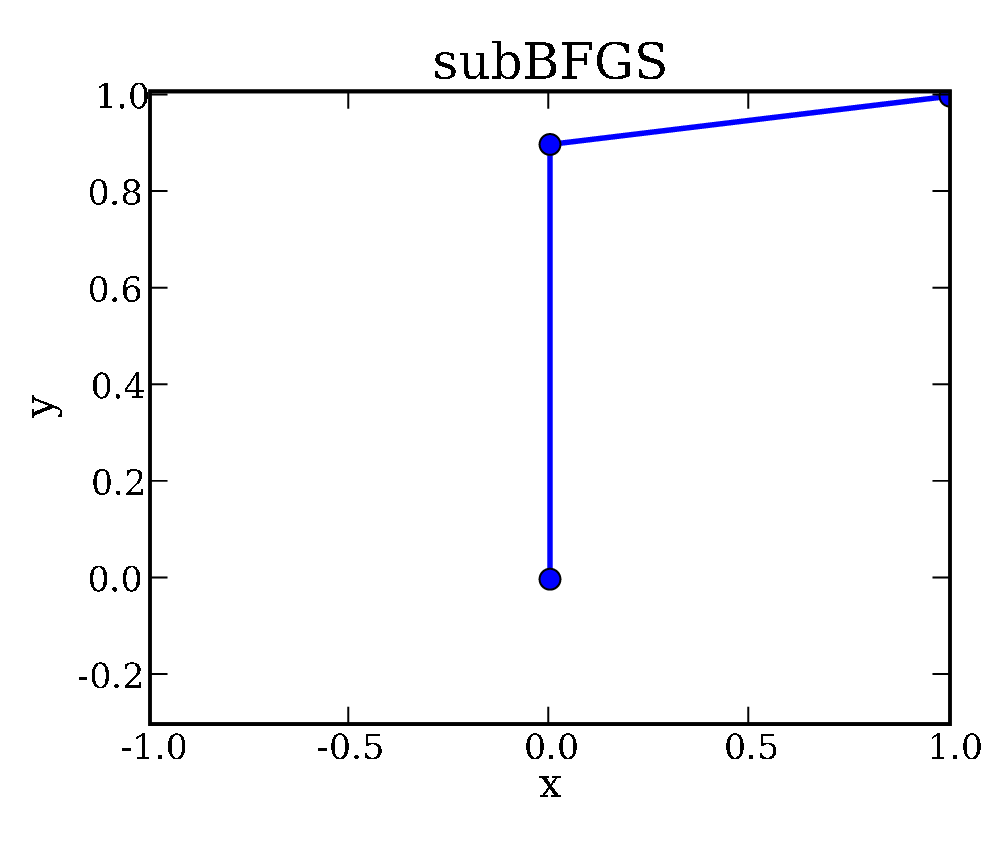}\\
    \end{tabular}
  \caption{Left: the nonsmooth convex function
    \eqref{eq:toy-problem}; optimization trajectory of BFGS with
    inexact line search (center) and subBFGS (right) on this function.}
  \label{fig:subbfgs-toy}
\end{figure}

The following simple example demonstrates the problems faced by BFGS
when working with a nonsmooth objective function, and how our subgradient
BFGS (subBFGS) method (to be introduced in Section~\ref{sec:subbfgs}) with
exact line search overcomes these problems. Consider the task of
minimizing
\begin{align}
  \label{eq:toy-problem}
  f(x,y) = 10 \, |x| + |y|
\end{align}
with respect to $x$ and $y$.  Clearly, $f(x, y)$ is convex but nonsmooth,
with the minimum located at $(0, 0)$ (Figure~\ref{fig:subbfgs-toy}, left).
It is subdifferentiable whenever $x$ or $y$ is zero:
\begin{align}
\partial_{x} f(0, \cdot) = [-10, 10] \mbox{~~and~~} \partial_{y}
f(\cdot, 0) = [-1, 1].
\end{align}
We call such lines of subdifferentiability in parameter space \emph{hinges}.

We can minimize \eqref{eq:toy-problem} with the standard BFGS algorithm,
employing a backtracking line search \citep[Procedure 3.1]{NocWri99}
that starts with a step size that obeys the curvature condition
\eqref{eq:wolfe2}, then exponentially decays it until both Wolfe
conditions (\ref{eq:wolfe1},\,\ref{eq:wolfe2}) are
satisfied.\footnote{We set $c_1 = 10^{-3}$ in \eqref{eq:wolfe1} and
  $c_2=0.8$ in \eqref{eq:wolfe2}, and used a decay factor of 0.9.}
The curvature condition forces BFGS to jump across at least one hinge,
thus ensuring that the gradient displacement vector $\vy_t$ in
\eqref{eq:bfgs-update} is non-zero; this prevents BFGS from diverging.
Moreover, with such an \emph{inexact} line search BFGS will generally
not step on any hinges directly, thus avoiding (in an ad-hoc manner) the
problem of non-differentia\-bility. Although this algorithm quickly
decreases the objective from the starting point $(1, 1)$, it is then
slowed down by heavy oscillations around the optimum
(Figure~\ref{fig:subbfgs-toy}, center), caused by the utter mismatch
between BFGS' quadratic model and the actual function.

A generally sensible strategy is to use an exact line search
that finds the optimum along a given descent direction (\cf
Section~\ref{sec:exact-ls}). However, this line optimum will often lie on a
hinge (as it does in our toy example), where the function is not
differentiable. If an arbitrary subgradient is supplied instead, the
BFGS update \eqref{eq:bfgs-update} can produce a search direction
which is not a descent direction, causing the next line search to
fail. In our toy example, standard BFGS with exact line search
consistently fails after the first step, which takes it to the hinge
at $x = 0$.

Unlike standard BFGS, our subBFGS method can handle hinges and thus
reap the benefits of an exact line search. As
Figure~\ref{fig:subbfgs-toy} (right) shows, once the first iteration
of subBFGS lands it on the hinge at $x = 0$, its direction-finding
routine (Algorithm~\ref{alg:find-descent-dir-cg}) finds a descent
direction for the next step. In fact, on this simple example
Algorithm~\ref{alg:find-descent-dir-cg} yields a vector with zero $x$
component, which takes subBFGS straight to the optimum at the second
step.\footnote{This is achieved for any choice of initial subgradient
  $\vg^{(1)}$ (\DIRINIT~of Algorithm~\ref{alg:find-descent-dir-cg}).} 


\subsubsection{Typical Nonsmooth Optimization Problems in Machine Learning}

\begin{figure}[tb]
  \centering
   \begin{tabular}{@{$\!\!\!$}c@{}c@{}c}
     \includegraphics[width=0.34\linewidth]{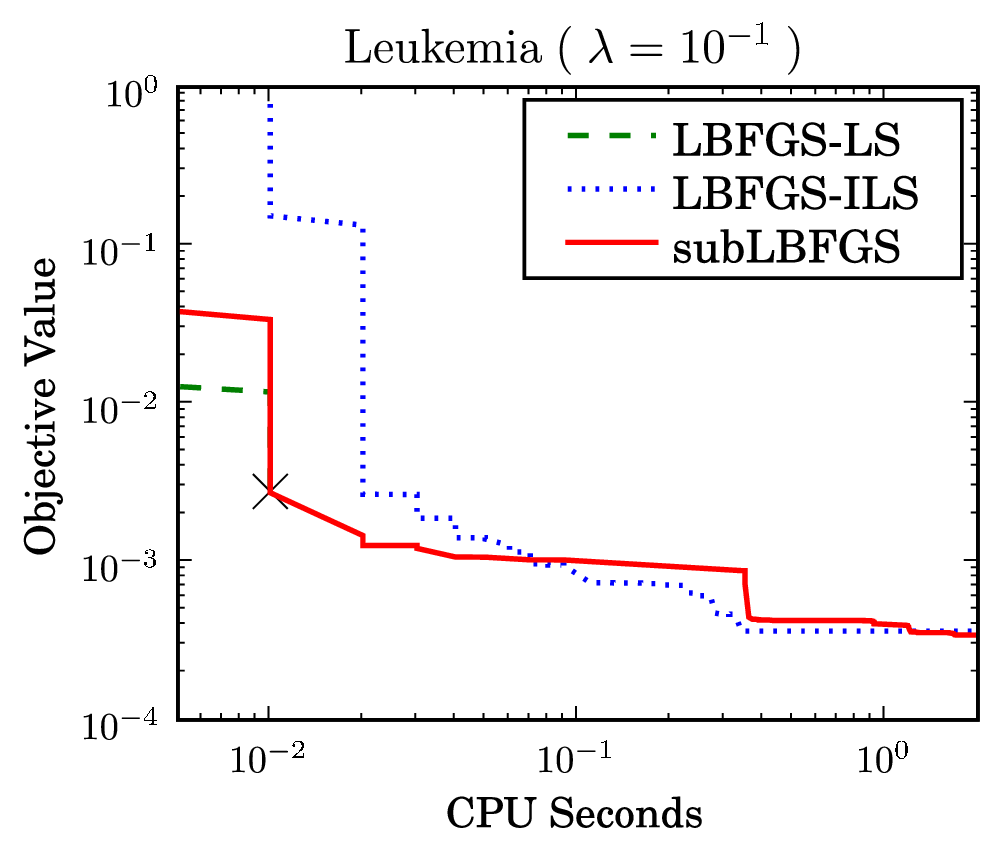}&
     \includegraphics[width=0.34\linewidth]{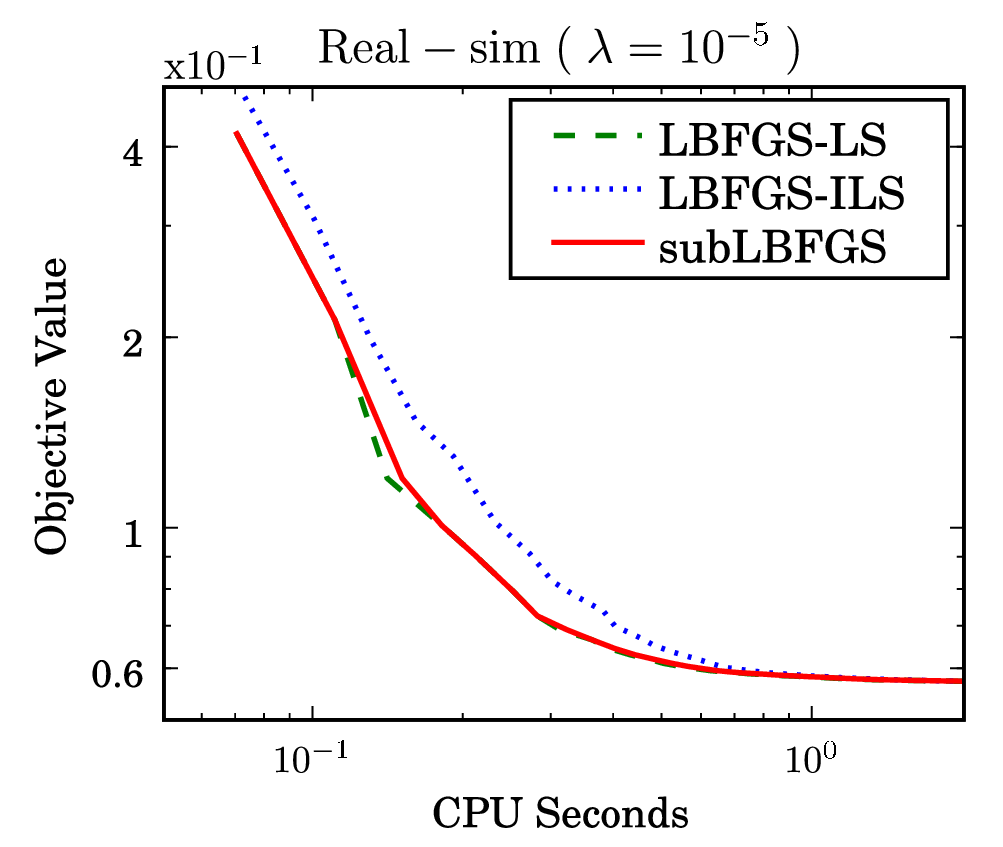}&
     \includegraphics[width=0.34\linewidth]{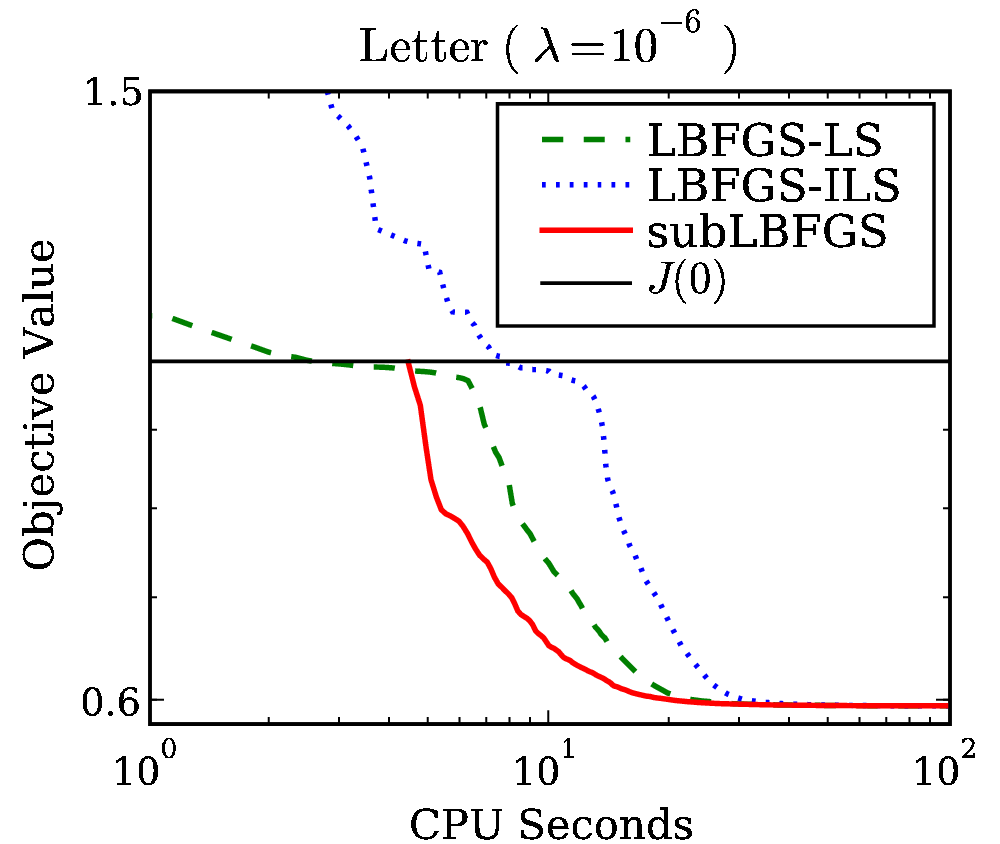}\\
   \end{tabular}
   \caption{Performance of subLBFGS (solid) and standard LBFGS with
     exact (dashed) and inexact (dotted) line search methods on sample
     $L_2$-regularized risk minimization problems with the binary
     (left and center) and multiclass hinge losses (right). LBFGS with
     exact line search (dashed) fails after 3 iterations (marked as
     $\times$) on the Leukemia dataset (left).}
 \label{fig:lbfgs}
\end{figure}

The problems faced by smooth quasi-Newton methods on nonsmooth objectives
are not only encountered in cleverly constructed toy examples, but also in real-world
applications. To show this, we apply
LBFGS to $L_2$-regularized risk minimization problems
\eqref{eq:regrisk} with binary hinge loss \eqref{eq:binaryloss}, a
typical nonsmooth optimization problem encountered in machine
learning. For this particular objective function, an exact line search
is cheap and easy to compute (see Section~\ref{sec:exact-ls} for
details). Figure~\ref{fig:lbfgs} (left \& center) shows the behavior
of LBFGS with this exact line search (LBFGS-LS) on two datasets,
namely Leukemia and Real-sim.\footnote{Descriptions of these datasets
  can be found in Section~\ref{sec:results}.}  It can be seen that
LBFGS-LS converges on Real-sim but diverges on the
Leukemia dataset. This is because using an exact line search on a
nonsmooth objective function increases the chance of landing on
nonsmooth points, a situation that standard BFGS (\emph{resp.}\ LBFGS)
is not designed to deal with. To prevent (L)BFGS' sudden breakdown, a
scheme that actively avoids nonsmooth points must be used. One such
possibility is to use an inexact line search that obeys the Wolfe
conditions. Here we used an efficient inexact line search that uses a
caching scheme specifically designed for $L_2$-regularized hinge loss
(\cf end of Section~\ref{sec:line-search}). This implementation of
LBFGS (LBFGS-ILS) converges on both datasets shown here but may fail
on others. It is also slower, due to the inexactness of its line
search.

For the multiclass hinge loss \eqref{eq:multi-lossdef} we encounter another
problem: if we follow the usual
practice of initializing $\vw = \vzero$, which happens to be a
non-differentiable point, then LBFGS stalls. One way to get around this is to
force LBFGS to take a unit step along its search direction to escape this
nonsmooth point. However, as can be seen on the Letter dataset\samefootnote{}
in Figure~\ref{fig:lbfgs} (right),
such an ad-hoc fix increases the value of the objective above $J(\vzero)$ (solid
horizontal line), and it takes several CPU seconds for the optimizers to recover
from this.  In all cases shown in Figure~\ref{fig:lbfgs}, our subgradient LBFGS
(subLBFGS) method (as will be introduced later) performs comparable to or better
than the best implementation of LBFGS.

\subsection{Advantage of Incorporating BFGS' Curvature Estimate}
\label{sec:quasi_vs_gd}

\begin{figure}
  \centering
  \begin{tabular}{@{$\!\!\!$}c@{}c@{}c}
    \includegraphics[width=0.34\linewidth]{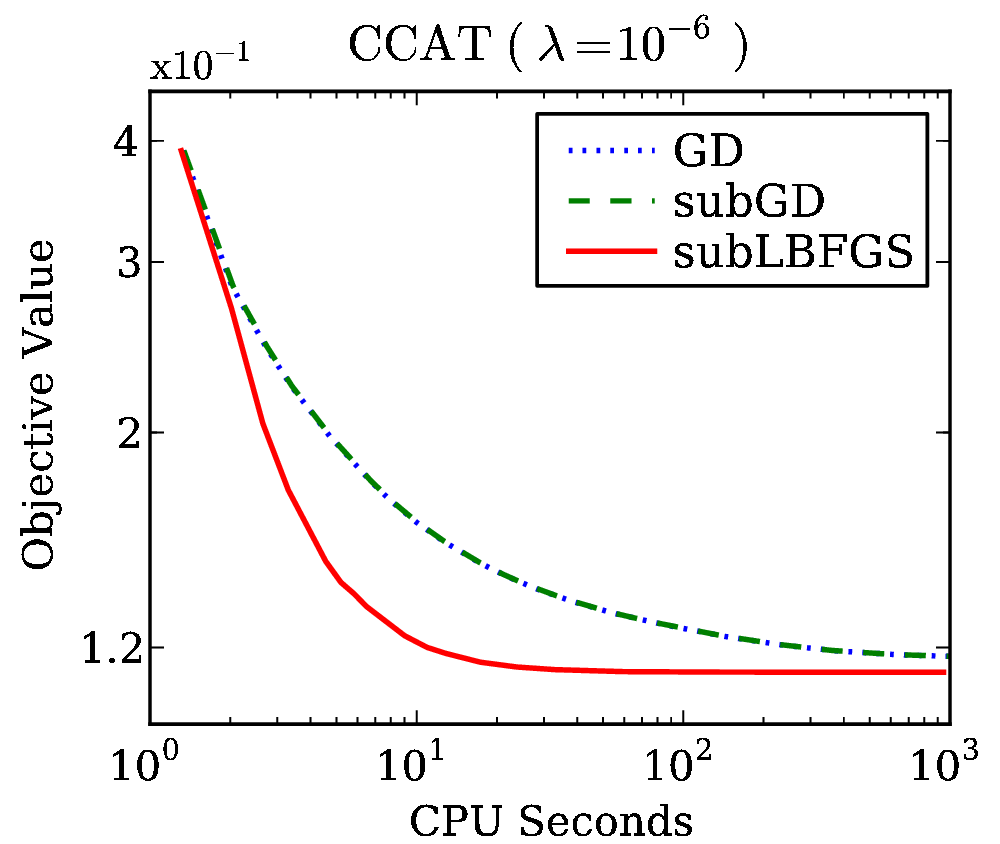} &
    \includegraphics[width=0.34\linewidth]{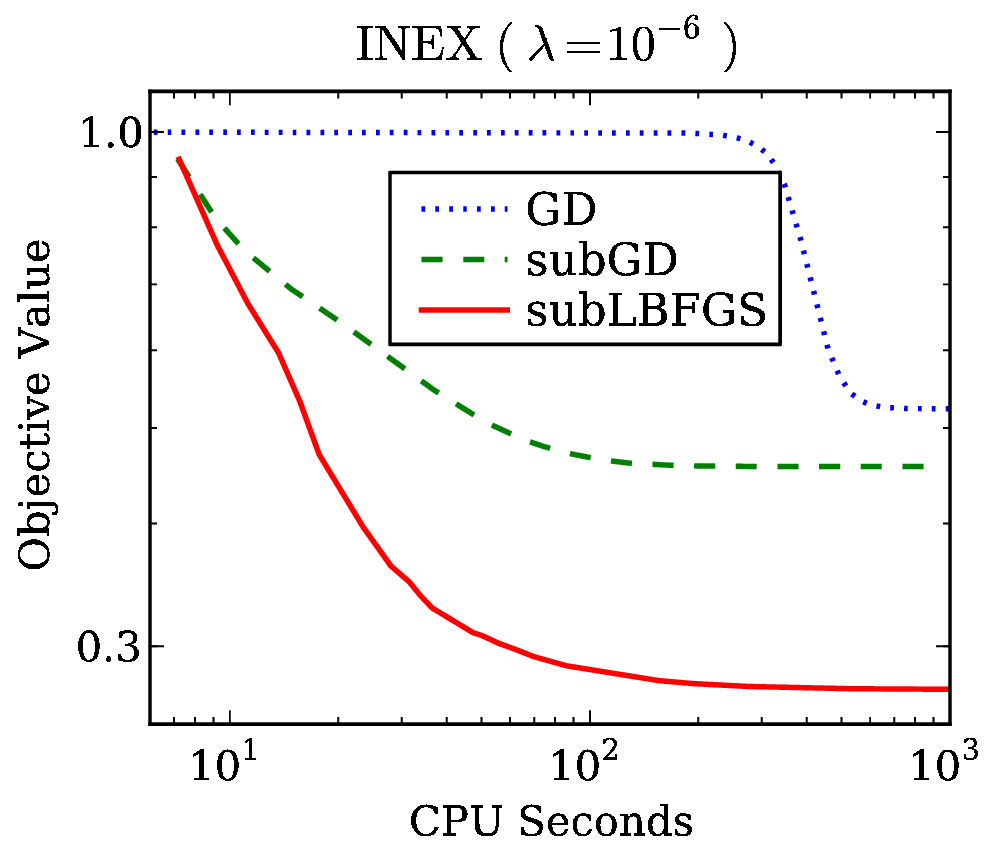} &
    \includegraphics[width=0.34\linewidth]{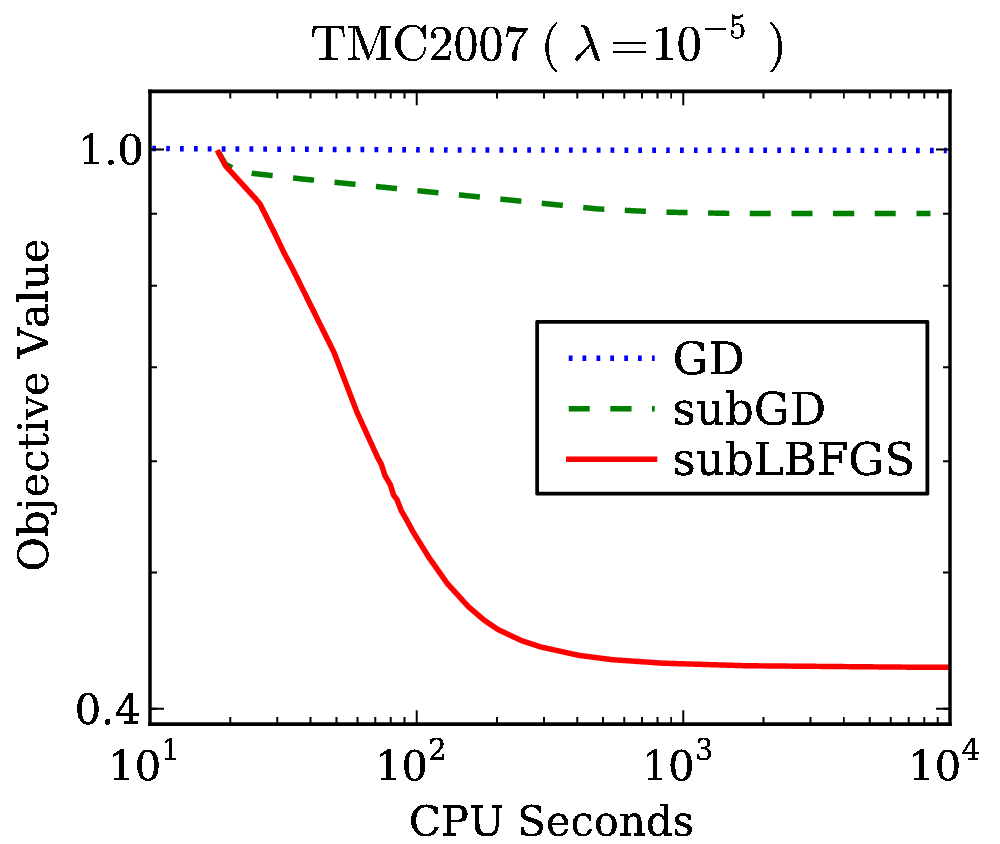} \\
  \end{tabular}
  \caption{Performance of subLBFGS, GD, and subGD on sample
    $L_2$-regularized risk minimization problems with binary
    (left), multiclass (center), and multilabel (right) hinge losses.}
  \label{fig:qn-vs-gd}
\end{figure}

In machine learning one often encounters $L_2$-regularized risk minimization
problems \eqref{eq:regrisk} with various hinge losses (\ref{eq:binaryloss},
\ref{eq:multi-lossdef}, \ref{eq:multilabel-lossdef}). Since the Hessian of
those objective functions at differentiable points equals $\lambda \one$
(where $\lambda$ is the regularization constant), one might be tempted to
argue that for such problems, BFGS' approximation $\Bmat_t$ to the inverse
Hessian should be simply set to $\lambda^{-1} \one$. This would
reduce the quasi-Newton direction $\vp_t = -\Bmat_t\vg_t, ~\vg_t
\in \partial J(\vw_t)$ to simply a scaled subgradient direction.

To check if doing so is beneficial, we compared the
performance of our subLBFGS method with two implementations of
subgradient descent: a vanilla gradient descent method (denoted GD)
that uses a random subgradient for its parameter update, and an
improved subgradient descent method (denoted subGD) whose parameter
is updated in the direction produced by our
direction-finding routine (Algorithm~\ref{alg:find-descent-dir-cg})
with $\Bmat_t = \one$. All algorithms used exact line search, except
that GD took a unit step for the first update in order to avoid the
nonsmooth point $\vw_0 = \vzero$ (\cf the discussion in Section
\ref{sec:problems-lbfgs}). As can be seen in
Figure~\ref{fig:qn-vs-gd}, on all sample $L_2$-regularized hinge loss
minimization problems,
subLBFGS (solid) converges significantly faster than GD (dotted) and subGD
(dashed). This indicates that BFGS' $\Bmat_t$ matrix is able to model the
objective function, including its hinges, better than simply setting $\Bmat_{t}$
to a scaled identity matrix.

\begin{figure}
  \centering
  \begin{tabular}{cc}
    \includegraphics[width=0.45\linewidth]{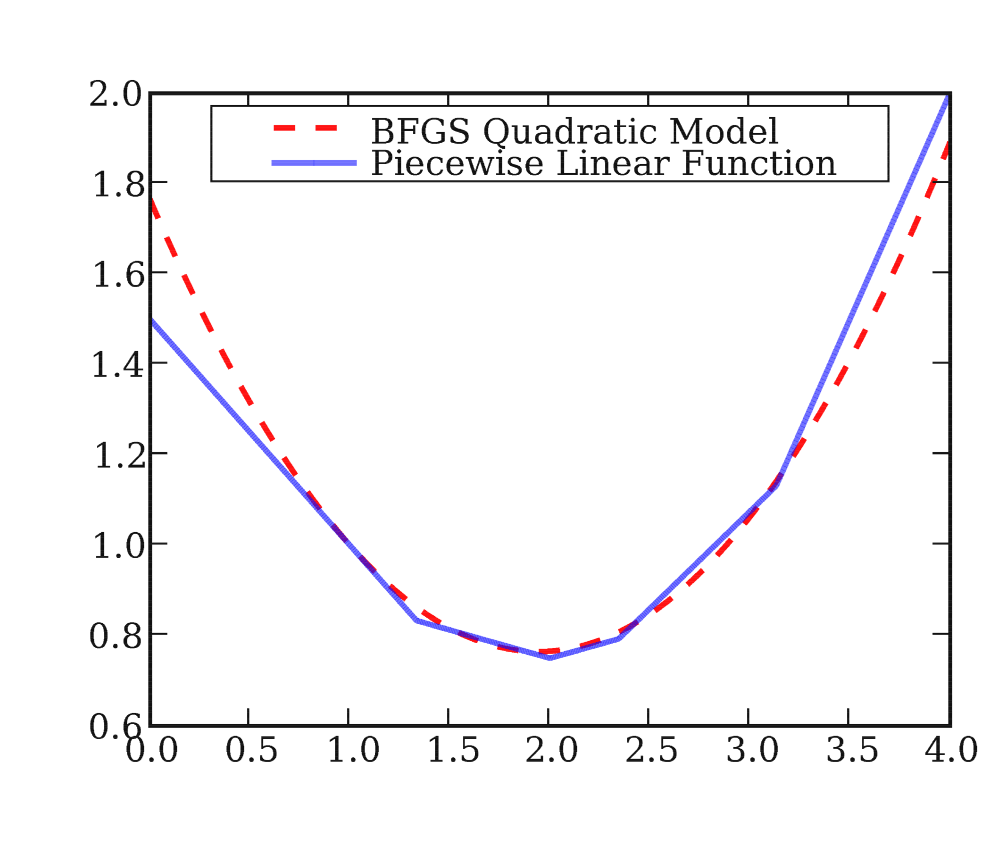} &
    \includegraphics[width=0.45\linewidth]{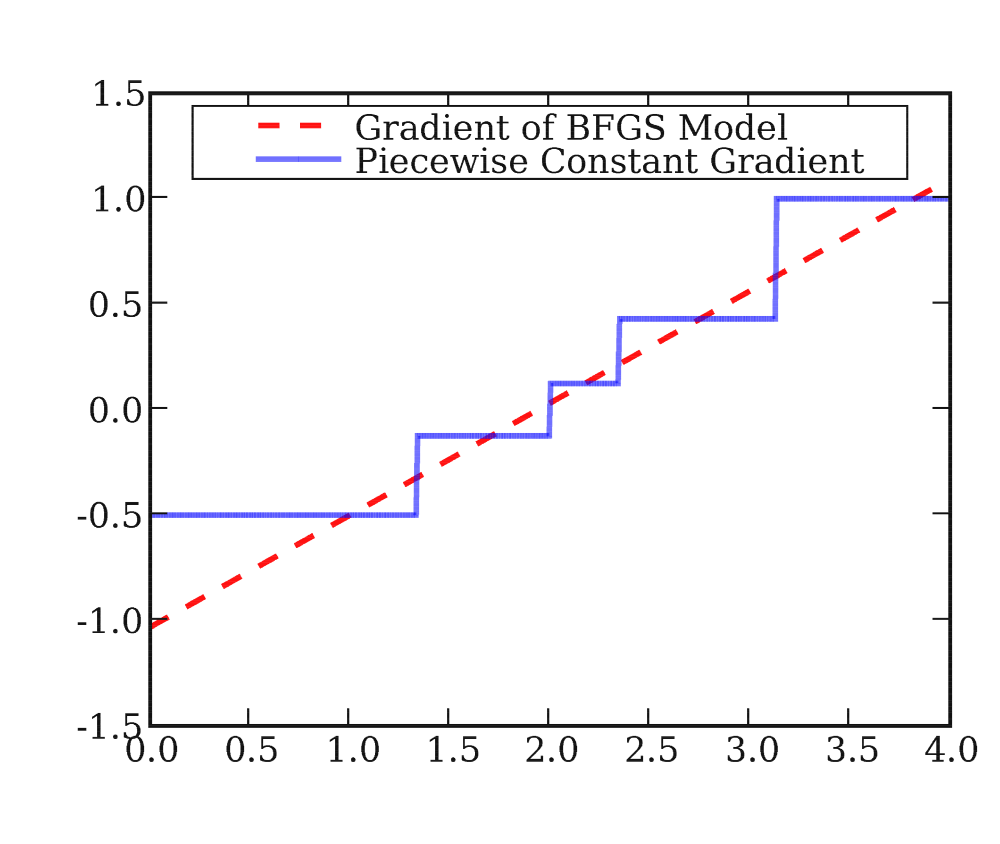} \\
  \end{tabular}
  \caption{BFGS' quadratic approximation to a piecewise linear function (left),
    and its estimate of the gradient of this function (right). }
  \label{fig:bfgs-model}
\end{figure}

We believe that BFGS' curvature update \eqref{eq:bfgs-update} plays an
important role in the performance of subLBFGS seen in
Figure~\ref{fig:qn-vs-gd}.  Recall that \eqref{eq:bfgs-update}
satisfies the secant condition $\Bmat_{t+1}\vy_t = \vs_t$, where
$\vs_t$ and $\vy_t$ are displacement vectors in parameter and gradient
space, respectively. The secant condition in fact implements a
\emph{finite differencing} scheme: for a one-dimensional
objective function $J: \RR \rightarrow \RR$, we have
\begin{align}
  \Bmat_{t+1} = \frac{(w+p) - w}{\nabla J(w+p) - \nabla J(w)}.
  \label{eq:finite-diff}
\end{align}
Although the original motivation behind the secant condition was to
approximate the inverse Hessian, the finite differencing scheme
\eqref{eq:finite-diff} allows BFGS to model the global
  curvature (\ie overall shape) of the objective function from
first-order information. For instance, Figure~\ref{fig:bfgs-model}
(left) shows that the BFGS quadratic model\footnote{For ease of
  exposition, the model was constructed at a differentiable point.}
\eqref{eq:quadratic-model} fits a piecewise linear function quite well
despite the fact that the actual Hessian in this case is zero almost
everywhere, and infinite (in the limit) at nonsmooth points.
Figure~\ref{fig:bfgs-model} (right) reveals that BFGS captures the
global trend of the gradient rather than its infinitesimal variation,
\ie the Hessian. This is beneficial for nonsmooth problems, where
Hessian does not fully represent the overall curvature of the
objective function.

\section{Subgradient BFGS Method}
\label{sec:subbfgs}

We modify the standard BFGS algorithm to derive our new algorithm
(subBFGS, Algorithm~\ref{alg:subbfgs}) for nonsmooth convex
optimization, and its memory-limited variant (subLBFGS).
Our modifications can be grouped into three areas, which
we elaborate on in turn: generalizing the local quadratic model,
finding a descent direction, and finding a step size that obeys a
subgradient reformulation of the Wolfe conditions.
We then show that our algorithm's estimate of the inverse Hessian
has a bounded spectrum, which allows us to prove its convergence.

\begin{algorithm}
  \caption{~Subgradient BFGS (subBFGS)}
  \label{alg:subbfgs}
  \begin{algorithmic}[1]
    \STATE Initialize: $t := 0, \vw_0 = \vzero, \Bmat_0 = \one$
    \STATE Set: direction-finding tolerance $\epsilon \ge 0$,
    iteration limit $k_{\text{max}} > 0$, \\ ~~~~~~
    lower bound $h > 0$ on $\frac{\vs_t^{\top} \!\vy_t}{\vy_t^{\top} \!\vy_t}$  (\cf
    discussion in Section~\ref{sec:bounded-b})
    \STATE Compute subgradient $\vg_0\in \partial J(\vw_0)$
    \WHILE{not converged} 
    \STATE $\vp_t = {\tt descentDirection}(\vg_t, \epsilon,
    k_{\text{max}})$
    \hfill (Algorithm~\ref{alg:find-descent-dir-cg})
    \IF{$\vp_t = \text{failure}$}
    \STATE Return $\vw_{t}$
    \ENDIF
    \STATE Find $\eta_t$ that obeys \eqref{eq:subwolfe-decrease} and \eqref{eq:subwolfe-curvature} 
    \hfill (\eg Algorithm~\ref{alg:exact-ls} or \ref{alg:exact-multi-ls})
    \STATE $\vs_t = \eta_t\vp_t$
    \STATE $\vw_{t+1} = \vw_t + \vs_t$
    \STATE Choose subgradient $\vg_{t+1} \in \partial J(\vw_{t+1}) :
      \vs_t^{\top} \!(\vg_{t+1} - \vg_t ) > 0$
    \STATE   $\vy_t := \vg_{t+1} - \vg_t$
    \STATE $\vs_t := \vs_t + \max \!\left( 0,~ h - \frac{\vs_t^{\top} \!\vy_t}{
    \vy_t^{\top} \!\vy_t} \right) \vy_t$ \hfill{(ensure $\frac{\vs_t^{\top} \!\vy_t}{
      \vy_t^{\top} \!\vy_t} \ge h$)}
    \STATE Update $\Bmat_{t+1}$ via \eqref{eq:bfgs-update}
    \STATE $t := t + 1 $
    \ENDWHILE
  \end{algorithmic}
\end{algorithm}

\subsection{Generalizing the Local Quadratic Model}
\label{sec:gen-model}

\begin{figure}
  \includegraphics[width=0.5\textwidth]{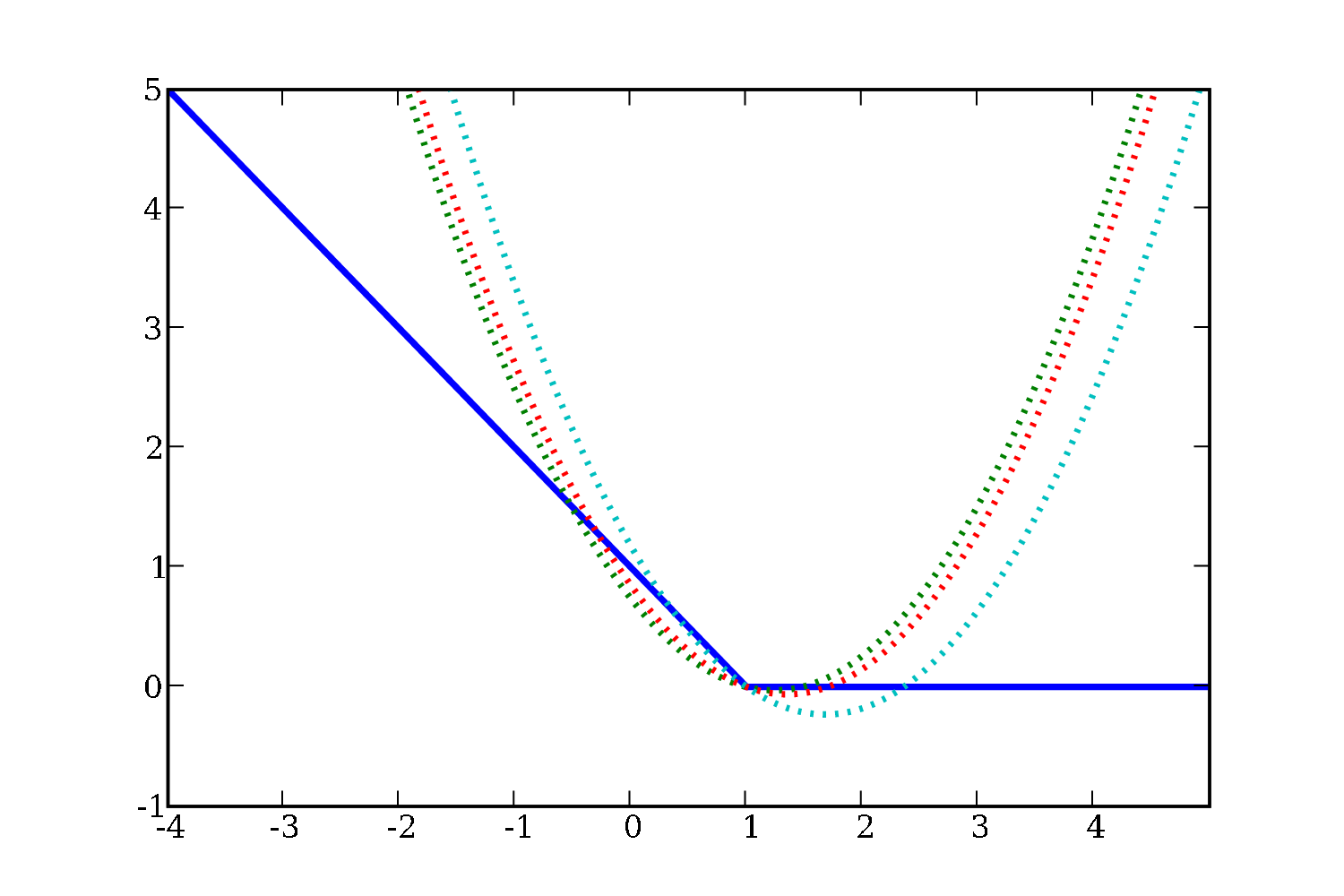}
  \includegraphics[width=0.5\textwidth]{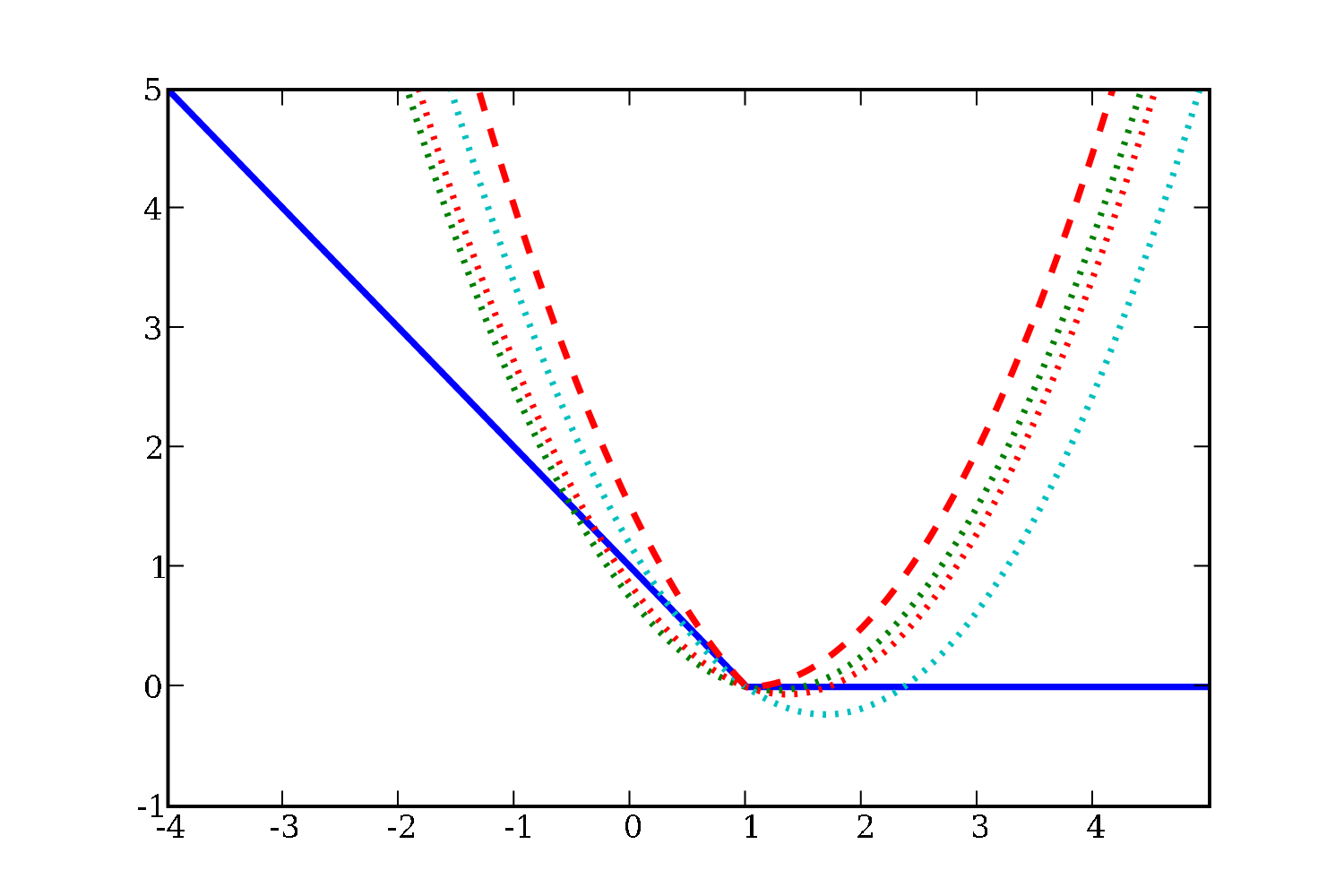}
  \caption{Left: selecting arbitrary subgradients yields many possible
    quadratic models (dotted lines) for the objective (solid blue
    line) at a subdifferentiable point. The models were built by keeping 
    $\Bmat_t$ fixed, but selecting random subgradients. Right: the tightest 
    pseudo-quadratic fit \eqref{eq:gen-quadratic-model} (bold red
    dashes); note that it is not a quadratic.}
  \label{fig:subbfgs-sup}
\end{figure}

Recall that BFGS assumes that the objective function $J$ is
differentiable everywhere so that at the current iterate $\vw_t$ it can
construct a local quadratic model \eqref{eq:quadratic-model} of
$J(\vw_t)$. For a nonsmooth objective function, such a model becomes
ambiguous at non-differentiable points
(Figure~\ref{fig:subbfgs-sup}, left). To resolve the ambiguity, we could
simply replace the gradient $\nabla J(\vw_t)$ in
\eqref{eq:quadratic-model} with an arbitrary subgradient $\vg_t
\in \partial J(\vw_t)$.  However, as will be discussed later, the
resulting quasi-Newton direction $\vp_t := -\Bmat_t\vg_t$ is not
necessarily a descent direction. To address this fundamental modeling
problem, we first generalize the local quadratic model
\eqref{eq:quadratic-model} as follows:
\begin{align}
  Q_t(\vp) & ~:=~ J(\vw_t) + M_t(\vp), \mbox{~~where} \nonumber\\
  M_t(\vp) & ~:=~ \half \vp^{\top} \!\Bmat_t^{-1} \vp ~+
    \!\!\!\sup_{\vg \in \partial J(\vw_t)}\!\!\! \vg^{\top} \!\vp.
  \label{eq:gen-quadratic-model}
\end{align}
Note that where $J$ is differentiable, \eqref{eq:gen-quadratic-model}
reduces to the familiar BFGS quadratic model
\eqref{eq:quadratic-model}.  At non-differentiable points, however,
the model is no longer quadratic, as the supremum may be attained at
different elements of $\partial J(\vw_t)$ for different directions
$\vp$. Instead it can be viewed as the tightest pseudo-quadratic fit
to $J$ at $\vw_{t}$ (Figure~\ref{fig:subbfgs-sup}, right). Although
the local model \eqref{eq:gen-quadratic-model} of subBFGS is
nonsmooth, it only incorporates non-differential points present at the
current location; all others are smoothly approximated by the
quasi-Newton mechanism.

Having constructed the model \eqref{eq:gen-quadratic-model}, we 
can minimize $Q_t(\vp)$, or equivalently $M_t(\vp)$:
\begin{align}
\min_{\vp \in \RR^d} \left( \half \vp^{\top} \!\Bmat_t^{-1} \vp ~+
    \!\!\!\sup_{\vg \in \partial J(\vw_t)}\!\!\! \vg^{\top} \!\vp \right)
\label{eq:min-gen-model}
\end{align}
to obtain a search direction. We now show that solving
\eqref{eq:min-gen-model} is closely related to the problem of finding
a \emph{normalized steepest descent} direction. A normalized steepest
descent direction is defined as the solution to the following problem
\citep[Chapter VIII]{HirLem93}:
\begin{align}
  \min_{\vp \in \RR^d} ~~ J'(\vw_t,~\vp) \mbox{~~s.t.~~} \ccc{\vp} \le 1,
  \label{eq:sd-subgrad}
\end{align}
where \[ J'(\vw_t, ~\vp) := \lim_{\eta \downarrow 0}\frac{J(\vw_t +
  \eta \vp) - J(\vw_t)}{\eta}\] is the directional derivative of $J$
at $\vw_t$ in direction $\vp$, and $\ccc{\cdot}$ is a norm
defined on $\RR^d$. In other words, the normalized steepest descent
direction is the direction of bounded norm along which the maximum
rate of decrease in the objective function value is achieved. Using
the property: $J'(\vw_t, ~\vp) = \sup_{\vg \in \partial J(\vw_{t})}
\vg^{\top} \vp$ \citep[Proposition B.24.b]{Bertsekas99}, we can
rewrite \eqref{eq:sd-subgrad} as:
\begin{align}
  \min_{\vp \in \RR^d} ~\sup_{\vg \in \partial J(\vw_t)}\!\!\! \vg^{\top}
    \!\vp \mbox{~~~~s.t.~~} \ccc{\vp} \le 1.
  \label{eq:minmax-subgrad}
\end{align}
If the matrix $\Bmat_t \succ 0$ as in \eqref{eq:min-gen-model} is used
to define the norm $\ccc{\cdot}$ as
\begin{align}
  \ccc{\vp}^2 := \vp^{\top} \!\Bmat_t^{-1} \vp, 
  \label{eq:weighted-norm}
\end{align}
then the solution to \eqref{eq:minmax-subgrad} points to the same
direction as that obtained by minimizing our pseudo-quadratic model
\eqref{eq:min-gen-model}.
To see this, we write the Lagrangian of the constrained minimization
problem \eqref{eq:minmax-subgrad}: 
\begin{align}
  L(\vp, \alpha) ~:=&~ \alpha~ \vp^{\top} \!\Bmat_t^{-1} \vp ~- \alpha ~+
  \!\!\!\sup_{\vg \in \partial J(\vw_t)}\!\!\! \vg^{\top} \!\vp
  \nonumber\\
  ~=&~  \half \vp^{\top} \!(2 \alpha\Bmat_t^{-1}) \vp ~- \alpha ~+
  \!\!\!\sup_{\vg \in \partial J(\vw_t)}\!\!\! \vg^{\top} \!\vp,
  \label{eq:dual-norm-sd}
\end{align}
where $\alpha > 0$ is a Lagrangian multiplier. It is easy to see from
\eqref{eq:dual-norm-sd} that minimizing the Lagrangian function $L$
with respect to $\vp$ is equivalent to solving
\eqref{eq:min-gen-model} with $\Bmat_t^{-1}$ scaled by a scalar
$2\alpha$, implying that the steepest descent direction obtained by
solving \eqref{eq:minmax-subgrad} with the weighted norm
\eqref{eq:weighted-norm} only differs in length from the search
direction obtained by solving \eqref{eq:min-gen-model}. Therefore, our
search direction is essentially an unnomalized steepest descent
direction with respect to the weighted norm
\eqref{eq:weighted-norm}. 

Ideally, we would like to solve \eqref{eq:min-gen-model} to obtain the
best search direction. This is generally intractable due to the
presence a supremum over the entire subdifferential set $\partial
J(\vw_t)$. In many machine learning problems, however, $\partial
J(\vw_t)$ has some special structure that simplifies the calculation
of that supremum. In particular, the subdifferential of all the
problems considered in this paper is a convex and compact polyhedron
characterised as the convex hull of its extreme points. This
dramatically reduces the cost of calculating $\sup_{\vg \in \partial
  J(\vw_t)} \inner{\vg}{\vp}$ since the supremum can only be attained
at an extreme point of the polyhedral set $\partial J(\vw_t)$
\citep[Proposition B.21c]{Bertsekas99}. In what follows, we develop an
iterative procedure that is guaranteed to find a quasi-Newton descent
direction, assuming an oracle that supplies $\argsup_{\vg \in \partial
  J(\vw_t)} \inner{\vg}{\vp}$ for a given direction $\vp\in\RR^d$.
Efficient oracles for this purpose can be derived for many machine
learning settings; we provides such oracles for $L_2$-regularized risk
minimization with the binary hinge loss (Section~\ref{sec:oracle}),
multiclass and multilabel hinge losses
(Section~\ref{sec:subbfgs-multiloss}), and $L_1$-regularized logistic
loss (Section~\ref{sec:l1loss}).

\subsection{Finding a Descent Direction}
\label{sec:find-descent-dir}

A direction $\vp_t$ is a descent direction if and only if~ $\inner{\vg}{\vp_t}
< 0 ~~\forall {\vg}\in\partial J(\vw_t)$ \citep[Theorem VIII.1.1.2]{HirLem93},
or equivalently
\begin{align}
  \label{eq:descent-dir}
  \sup_{{\vg}\in\partial J(\vw_t)}\!\!\! \inner{\vg}{\vp_t} ~<~ 0.
\end{align}
For a smooth convex function, the quasi-Newton direction
\eqref{eq:quasi-newton-dir} is always a descent direction because
\begin{align*}
\nabla J(\vw_t)^{\top}\vp_t ~=\, -\nabla J(\vw_t)^{\top} \!\Bmat_t\nabla
J(\vw_t) ~<~ 0
\end{align*}
holds due to the positivity of $\Bmat_t$.

For nonsmooth functions, however, the quasi-Newton direction $\vp_t:=
-\Bmat_t\vg_t$ for a given $\vg_t \in\partial J(\vw_t)$ may not
fulfill the descent condition \eqref{eq:descent-dir}, making it
impossible to find a step size $\eta > 0$ that obeys the Wolfe conditions
(\ref{eq:wolfe1},\,\ref{eq:wolfe2}), thus causing a failure of the
line search. We now present an iterative approach to finding a
quasi-Newton \emph{descent} direction.

\begin{algorithm}[t]
  \caption{~$\vp_t = {\tt descentDirection}(\vg^{(1)} , \epsilon, k_{\text{max}})$}
  \label{alg:find-descent-dir-cg}
  \begin{algorithmic}[1]
    \INPUT (sub)gradient $\vg^{(1)}  \in \partial J(\vw_t)$, tolerance
    $\epsilon \ge 0$, iteration limit $k_{\text{max}} > 0$, \\
    ~~~~~~~~~~and an oracle to calculate $\argsup_{\vg \in \partial J(\vw)}
    \,\vg^{\top}\vp$ for any given $\vw$ and $\vp$
    \OUTPUT descent direction $\vp_t$
    \STATE Initialize: $i = 1, ~\bar{\vg}^{(1)} = \vg^{(1)}, ~\vp^{(1)}
    = -\Bmat_t \vg^{(1)}$ 
    \STATE $\vg^{(2)} = \argsup_{\vg \in \partial J(\vw_t)} \vg^{\top}
    \vp^{(1)}$ 
    \STATE $\epsilon^{(1)}  :=  \vp^{(1) \top} \vg^{(2)} - \vp^{(1)
      \top} \bar{\vg}^{(1)}$ 
    \WHILE{$(\vg^{(i+1) \top} \vp^{(i)}  > 0$ or
     $\epsilon^{(i)} > \epsilon)$ and $\epsilon^{(i)} > 0$ and $i < k_{\text{max}}$}
   \vspace{1ex}
   \STATE $\mu^* := \min\left[1, \frac{(\bar{\vg}^{(i)}  - \vg^{(i+1)} 
        )^{\top} \Bmat_{t} \bar{\vg}^{(i)}  }{(\bar{\vg}^{(i)}  -
        \vg^{(i+1)} )^{\top} \Bmat_{t} (\bar{\vg}^{(i)}  - \vg^{(i+1)} )} \right]$;
    ~\cf \eqref{eq:meta-step-simple2}
\vspace{1ex}
    \STATE $\bar{\vg}^{(i+1)} = (1 - \mu^*) \bar{\vg}^{(i)} + \mu^* \vg^{(i+1)}$
    \STATE $\vp^{(i+1)} = (1 - \mu^*) \vp^{(i)} - \mu^* \Bmat_t
    \vg^{(i+1)}$; ~\cf \eqref{eq:pi-update}
    \STATE $\vg^{(i+2)} = \argsup_{\vg \in \partial J(\vw_t)}
    \vg^{\top} \vp^{(i+1)}$ \vspace{.5ex}
    \STATE $\epsilon^{(i+1)}  := \min_{j\le (i+1)} \left[ \vp^{(j) \top} \vg^{(j+1)} -
  \half (\vp^{(j) \top} \bar{\vg}^{(j)} +\vp^{(i+1) \top}
  \bar{\vg}^{(i+1)}) \right]$
    \STATE $i := i + 1$
    \ENDWHILE
    \STATE $\vp_t = \argmin_{j \le i} M_t(\vp^{(j)})$
    \IF{$ \sup_{\vg \in \partial J(\vw_t)} \vg^{\top}\vp_t \ge 0$}
    \STATE{\bf return} failure;
    \ELSE \STATE{\bf return} $\vp_t$.
    \ENDIF
  \end{algorithmic}
\end{algorithm}

Our goal is to minimize the pseudo-quadratic model
\eqref{eq:gen-quadratic-model}, or equivalently  minimize
$M_t(\vp)$. Inspired by bundle methods \citep{TeoVisSmoLe09}, we
achieve this by minimizing convex lower bounds of $M_t(\vp)$ that are
designed to progressively approach $M_t(\vp)$ over iterations. At iteration
$i$ we build the following convex lower bound on $M_{t}(\vp)$:
\begin{align}
  \label{eq:quadratic-model-relax}
  M_t^{(i)} (\vp) ~:=~ \half \vp^{\top} \!\Bmat_t^{-1} \vp \,+\, \sup_{j \le i}
  \vg^{(j) \top} \vp,
\end{align}
where $i,j \in \NN$ and $\vg^{(j)} \in \partial J(\vw_t) ~ \forall j \le
i$. Given a $\vp^{(i)} \in\RR^d$ the lower bound
\eqref{eq:quadratic-model-relax} is successively tightened by
computing
\begin{align}
  \label{eq:gi}
  \vg^{(i+1)}  & := \argsup_{\vg \in \partial J(\vw_t)} \inner{\vg}{\vp^{(i)}},
\end{align}
such that $M_t^{(i)}(\vp) \le M_t^{(i+1)}(\vp) \le M_t(\vp)~ \forall
\vp \in \RR^d$. Here we set $\vg^{(1)} \in \partial J(\vw_t)$
arbitrarily, and assume that \eqref{eq:gi} is provided by an oracle
(\eg as described in Section~\ref{sec:oracle}). To solve
$\min_{\vp \in \RR^d} M_t^{(i)}(\vp)$, we rewrite it as a constrained
optimization problem:
\begin{align}
  \label{eq:constrained-opt-relax}
  &\min_{\vp, \xi}\; \left( \half \vp^{\top} \!\Bmat_t^{-1} \vp + \xi \right)
  \mbox{~~s.t.~~} \vg^{(j) \top} \vp \leq \xi ~~\forall j \le i.
\end{align}
This problem can be solved exactly via quadratic programming, but 
doing so may incur substantial computational expense. Instead we adopt
an alternative approach (Algorithm~\ref{alg:find-descent-dir-cg})
which does not solve \eqref{eq:constrained-opt-relax} to
optimality. The key idea is to write the proposed descent direction at
iteration $i+1$ as a convex combination of $\vp^{(i)}$ and $-\Bmat_{t}
\vg^{(i+1)}$ (\DIRUPDATE~of Algorithm~\ref{alg:find-descent-dir-cg}); and
as will be shown in Appendix~\ref{sec:LineSearchConvergence}, the
returned search direction takes the form
\begin{align}
\vp_t = -\Bmat_t \bar{\vg}_t,
\label{eq:subqn-dir}
\end{align}
where $\bar{\vg}_t$ is a subgradient in $\partial J(\vw_t)$ that allows
$\vp_t$ to satisfy the descent condition \eqref{eq:descent-dir}. The
optimal convex combination coefficient $\mu^*$ can be computed exactly
(\DIRCOE~of Algorithm~\ref{alg:find-descent-dir-cg}) using an argument
based on maximizing the dual objective of $M_t(\vp)$; see
Appendix~\ref{sec:LineSearchDescent} for details.

The weak duality theorem \citep[Theorem XII.2.1.5]{HirLem93} states
that the optimal primal value is no less than any dual value, \ie if
$D_t(\val)$ is the dual of $M_t(\vp)$, then $\min_{\vp \in \RR^d}
M_t(\vp) \ge D_t(\val)$ holds for all feasible dual solutions $\val$.
Therefore, by iteratively increasing the value of the dual objective we
close the gap to optimality in the primal. Based on this
argument, we use the following upper bound on the duality gap as our
measure of progress:
\begin{align}
  \epsilon^{(i)} := \min_{j\le i} \left[ \vp^{(j) \top} \vg^{(j+1)} -
    \half (\vp^{(j) \top} \bar{\vg}^{(j)} +\vp^{(i) \top}
    \bar{\vg}^{(i)}) \right] \ge \min_{\vp \in \RR^d} M_t(\vp) - D_t(\val^{*}),
  \label{eq:epsilon}
\end{align}
where $\bar{\vg}^{(i)}$ is an aggregated subgradient
(\AGGREGATEGRAD~of Algorithm~\ref{alg:find-descent-dir-cg}) which lies
in the convex hull of $\vg^{(j)} \in \partial J(\vw_t) ~\forall j \le
i$, and $\val^*$ is the optimal dual solution; equations
\ref{eq:def-eps}--\ref{eq:eps-up} in
Appendix~\ref{sec:LineSearchDescent} provide intermediate steps that
lead to the inequality in
\eqref{eq:epsilon}. Theorem~\ref{th:updateguarantee}
(Appendix~\ref{sec:LineSearchConvergence}) shows that $\epsilon^{(i)}$
is monotonically decreasing, leading us to a practical stopping
criterion (\DIRSTOP~of Algorithm~\ref{alg:find-descent-dir-cg}) for
our direction-finding procedure.

A detailed derivation of Algorithm~\ref{alg:find-descent-dir-cg} is
given in Appendix~\ref{sec:LineSearchDescent}, where we also prove that
at a non-optimal iterate a direction-finding tolerance $\epsilon \ge 0$
exists such that the search direction produced by
Algorithm~\ref{alg:find-descent-dir-cg} is a descent direction; in
Appendix~\ref{sec:LineSearchConvergence} we prove that
Algorithm~\ref{alg:find-descent-dir-cg} converges to a solution with
precision $\epsilon$ in $O(1/\epsilon)$ iterations. Our proofs are
based on the assumption that the spectrum (eigenvalues) of
BFGS' approximation $\Bmat_t$ to the inverse Hessian is bounded from above
and below. This is a reasonable assumption if simple safeguards 
such as those described in Section~\ref{sec:bounded-b} are employed
in the practical implementation.

\subsection{Subgradient Line Search}
\label{sec:gen-ls}

\begin{figure}
  \centering
  \includegraphics[width=0.60\textwidth]{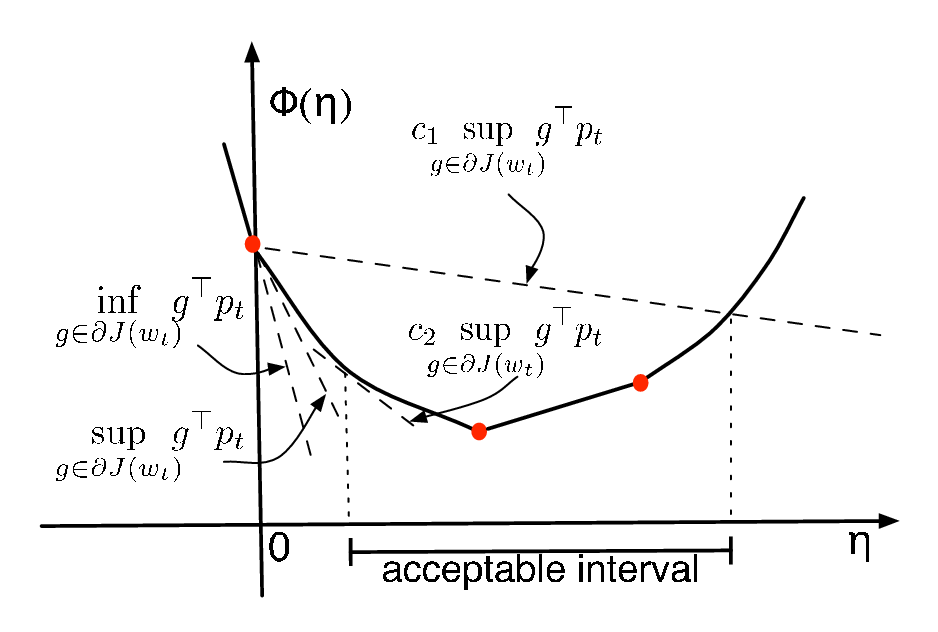}
  \caption{Geometric illustration of the subgradient Wolfe conditions
    \eqref{eq:subwolfe-decrease} and \eqref{eq:subwolfe-curvature}. Solid
    disks are subdifferentiable points; the slopes of dashed lines are
    indicated.} 
  \label{fig:subwolfe}
\end{figure}

Given the current iterate $\vw_{t}$ and a search direction $\vp_t$,
the task of a line search is to find a step size $\eta > 0$ which
reduces the objective function value along the line $\vw_t + \eta\vp_t$:
\begin{align}
  \mini ~\Phi(\eta) \,:=\, J(\vw_t + \eta\vp_t).
  \label{eq:step-fun}
\end{align}
Using the chain rule, we can write
\begin{align}
\partial \,\Phi(\eta) \,:=\, \{\inner{\vg}{\vp_t} : \vg \in \partial J(\vw_t
+ \eta \vp_t)\}.
\label{eq:step-grad}
\end{align}
Exact line search finds the optimal step size $\eta^*$ by minimizing
$\Phi(\eta)$, such that $0 \in \partial \Phi(\eta^*)$; inexact
line searches solve \eqref{eq:step-fun} approximately while enforcing
conditions designed to ensure convergence. The Wolfe conditions
\eqref{eq:wolfe1} and \eqref{eq:wolfe2}, for instance, achieve this by
guaranteeing a sufficient decrease in the value of the objective and
excluding pathologically small step sizes, respectively
\citep{Wolfe69, NocWri99}. The original Wolfe conditions, however,
require the objective function to be smooth; to extend them to
nonsmooth convex problems, we propose the following subgradient
reformulation:
\begin{align}
  J(\vw_{t+1}) ~\le~ J(\vw_t) ~+~ c_1 \eta_t & \!\!\sup_{\vg \in \partial
    J(\vw_t)}\!\! \inner{\vg}{\vp_t} \mbox{~~~~~~~~(sufficient decrease)}
    \label{eq:subwolfe-decrease} \\
  \mbox{and~~~~} \sup_{\vg' \in \partial J(\vw_{t+1})}\!\!\!\!\! \vg'^{\top}
  \vp_t ~\ge~ c_2 & \!\!\sup_{\vg \in \partial J(\vw_t)}\!\! \inner{\vg}{\vp_t},
  \mbox{~~~~~~~(curvature)} \label{eq:subwolfe-curvature}
\end{align}
where $0 < c_1 < c_2 < 1$. Figure~\ref{fig:subwolfe} illustrates how these
conditions enforce acceptance of non-trivial step sizes that decrease the
objective function value. In Appendix~\ref{sec:PositiveStepSize} we formally
show that for any given descent direction we can always find a positive step
size that satisfies \eqref{eq:subwolfe-decrease} and
\eqref{eq:subwolfe-curvature}. Moreover, Appendix~\ref{sec:convergence-proof}
shows that the sufficient decrease condition \eqref{eq:subwolfe-decrease}
provides a necessary condition for the global convergence of subBFGS.

Employing an exact line search is a common strategy to speed up
convergence, but it drastically increases the probability of landing
on a non-differentiable point (as in Figure~\ref{fig:lbfgs}, left).
In order to leverage the fast convergence provided by an exact line
search, one must therefore use an optimizer that can handle
subgradients, like our subBFGS.

A natural question to ask is whether the optimal step size $\eta^*$
obtained by an exact line search satisfies the reformulated Wolfe
conditions (\emph{resp.} the standard Wolfe conditions when $J$
is smooth). The answer is no: depending on the choice of $c_1$,
$\eta^*$ may violate the sufficient decrease condition
\eqref{eq:subwolfe-decrease}. For the function shown in
Figure~\ref{fig:subwolfe}, for instance, we can increase the value of $c_1$
such that the acceptable interval for the step size excludes $\eta^*$.
In practice one can set $c_1$ to a small value, \eg $10^{-4}$, to
prevent this from happening.

The curvature condition \eqref{eq:subwolfe-curvature}, on the other hand,
is always satisfied by $\eta^*$, as long as $\vp_t$ is a descent direction
\eqref{eq:descent-dir}:
\begin{align}
  \sup_{\vg' \in J(\vw_t + \eta^* \vp_t)}\!\!\!\!\!\!\!\!\! \vg'^{\top} \vp_t ~=
  \!\!\!\sup_{g \in \partial \Phi(\eta^*)}\!\!\! g ~\ge~ 0 ~> \!\!\sup_{\vg
    \in \partial J(\vw_t)}\!\!\! \inner{\vg}{\vp_t}
\end{align}
because $0 \in \partial \,\Phi(\eta^*)$.

\subsection{Bounded Spectrum of SubBFGS'  Inverse Hessian Estimate} 
\label{sec:bounded-b}

Recall from Section~\ref{sec:intro} that to ensure 
positivity of BFGS' estimate $\Bmat_t$ of the inverse Hessian, we
must have $(\forall t) ~\vs_t^{\top} \vy_t > 0$. Extending this
condition to nonsmooth functions, we require 
\begin{align}
  (\vw_{t+1} - \vw_t)^{\top}(\vg_{t+1} - \vg_t) > 0, \mbox{~~where~~}
  \vg_{t+1} \in \partial J(\vw_{t+1}) \mbox{~~and~~} \vg_{t}
  \in \partial J(\vw_{t}).
  \label{eq:nonsmooth-pos-sy}
\end{align}
If $J$ is strongly convex,\footnote{If $J$ is strongly convex, then $(\vg_2 -
  \vg_1)^{\top}(\vw_2 - \vw_1) \ge c\, \norm{\vw_2 - \vw_1}^2, \mbox{~~with~~} c
  >0,~ \vg_i \in \partial J(\vw_i),~ i = 1, 2$.\label{ftn:strong-convex}} and
$\vw_{t+1} \neq \vw_t$, then \eqref{eq:nonsmooth-pos-sy} holds for any choice of
$\vg_{t+1}$ and $\vg_t$.\footnote{We found empirically that no qualitative
  difference between using random subgradients versus choosing a particular
  subgradient when updating the $\Bmat_t$ matrix.} For general convex functions,
$\vg_{t+1}$ need to be chosen (\SUBBFGSGRAD~of Algorithm~\ref{alg:subbfgs}) to
satisfy \eqref{eq:nonsmooth-pos-sy}. The existence of such a subgradient is
guaranteed by the convexity of the objective function. To see this, we first use
the fact that $\eta_t\vp_t = \vw_{t+1} - \vw_t$ and $\eta_t > 0$ to rewrite
\eqref{eq:nonsmooth-pos-sy} as
\begin{align}
  \vp_t^{\top}\vg_{t+1}  > \vp_{t}^{\top}\vg_t,  \mbox{~~where~~}
\vg_{t+1} \in \partial J(\vw_{t+1}) \mbox{~~and~~} \vg_{t}
\in \partial J(\vw_{t}).
\label{eq:nonsmooth-pos-sy2}
\end{align}
It follows from \eqref{eq:step-grad} that both sides of inequality
\eqref{eq:nonsmooth-pos-sy2} are subgradients of $\Phi(\eta)$ at
$\eta_t$ and $0$, respectively. The monotonic property of $\partial
\Phi(\eta)$ given in Theorem~\ref{th:increasing-subgrad} (below) ensures
that $\vp_t^{\top}\vg_{t+1}$ is no less than $\vp_{t}^{\top}\vg_t$ for
any choice of $\vg_{t+1}$ and $\vg_t$, \ie
\begin{align}
  \inf_{\vg \in \partial J(\vw_{t+1})} \! \vp_t^{\top}\vg ~\ge
  \sup_{\vg \in \partial J(\vw_{t})}\vp_{t}^{\top}\vg.
\label{eq:phi-eta-0}
\end{align}
This means that the only case where inequality
\eqref{eq:nonsmooth-pos-sy2} is violated is when both terms of
\eqref{eq:phi-eta-0} are equal, and 
\begin{align}
  \vg_{t+1} = \arginf_{\vg \in \partial J(\vw_{t+1})} \vg^{\top}\vp_t
  \mbox{~~and~~} \vg_{t} = \argsup_{\vg \in \partial J(\vw_{t})}
  \vg^{\top}\vp_t,
\end{align}
\ie in this case
$\vp_t^{\top}\vg_{t+1} = \vp_{t}^{\top}\vg_t$. To avoid this, we
simply need to set $\vg_{t+1}$ to a different subgradient in $\partial
J(\vw_{t+1})$.

\begin{theorem}
 {\rm \citep[Theorem I.4.2.1]{HirLem93}}\\
Let $\Phi$ be a one-dimensional convex function on its domain, then
$\partial \Phi(\eta)$ is increasing in the sense that~
$g_1 \le g_2 \mbox{~~whenever~~} g_1 \in \partial \Phi(\eta_1), ~g_2
\in \partial \Phi(\eta_2), \mbox{~and~} \eta_1 < \eta_2.$
\label{th:increasing-subgrad}
\end{theorem}

Our convergence analysis for the direction-finding procedure
(Algorithm~\ref{alg:find-descent-dir-cg}) as well as the global
convergence proof of subBFGS in Appendix \ref{sec:convergence-proof}
require the spectrum of $\Bmat_t$ to be bounded from above and
below by a positive scalar:
\begin{align}
  \exists \,(h, H : 0 < h \leq H < \infty) : (\forall t)~ h \preceq
  \Bmat_t \preceq H.
  \label{eq:bound-B}
\end{align}
From a theoretical point of view it is difficult to guarantee
\eqref{eq:bound-B} \citep[page 212]{NocWri99}, but based on the fact
that $\Bmat_t$ is an approximation to the inverse Hessian
$\Hmat^{-1}_t$, it is reasonable to expect \eqref{eq:bound-B} to be
true if
\begin{align}
   (\forall t)~1/H \preceq \Hmat_t \preceq 1/h. 
\end{align}
Since BFGS ``senses'' the Hessian via \eqref{eq:bfgs-update} only
through the parameter and gradient displacements $\vs_t$ and $\vy_t$,
we can translate the bounds on the spectrum of $\Hmat_t$ into
conditions that only involve $\vs_t$ and $\vy_t$:
\begin{align}
  (\forall t)~~\frac{\vs^{\top}_t\vy_t}{\vs^{\top}_t\vs_t} \ge \frac{1}{H}
  \mbox{~~and~~} \frac{\vy^{\top}_t\vy_t}{\vs^{\top}_t\vy_t} \le
  \frac{1}{h}, \mbox{~~with~~} 0 < h \leq H < \infty.
  \label{eq:bound-sy}
\end{align}
This technique is used in \citep[Theorem 8.5]{NocWri99}. If $J$ is strongly
convex$^{\text{\ref{ftn:strong-convex}}}$ and $\vs_t \neq \vzero$, then there
exists an $H$ such that the left inequality in \eqref{eq:bound-sy} holds. On
general convex functions, one can skip BFGS' curvature update if
$(\vs^{\top}_t\vy_t/\vs^{\top}_t\vs_t)$ falls below a threshold. To establish
the second inequality, we add a fraction of $\vy_t$ to $\vs_t$ at \SUBBFGSS~of
Algorithm~\ref{alg:subbfgs} (though this modification is never actually invoked
in our experiments of Section~\ref{sec:results}, where we set $h = 10^{-8}$).

\subsection{Limited-Memory Subgradient BFGS}
\label{sec:sublbfgs}

It is straightforward to implement an LBFGS variant of our subBFGS algorithm: we
simply modify Algorithms~\ref{alg:subbfgs} and \ref{alg:find-descent-dir-cg} to
compute all products between $\Bmat_t$ and a vector by means of the standard
LBFGS matrix-free scheme \citep[Algorithm 9.1]{NocWri99}. We call the resulting
algorithm subLBFGS.

\subsection{Convergence of Subgradient (L)BFGS}
\label{sec:ConvergenceAnalysis}

\begin{figure}[tb]
 \centering
  \begin{tabular}{cc}
    \includegraphics[width=0.49\linewidth]{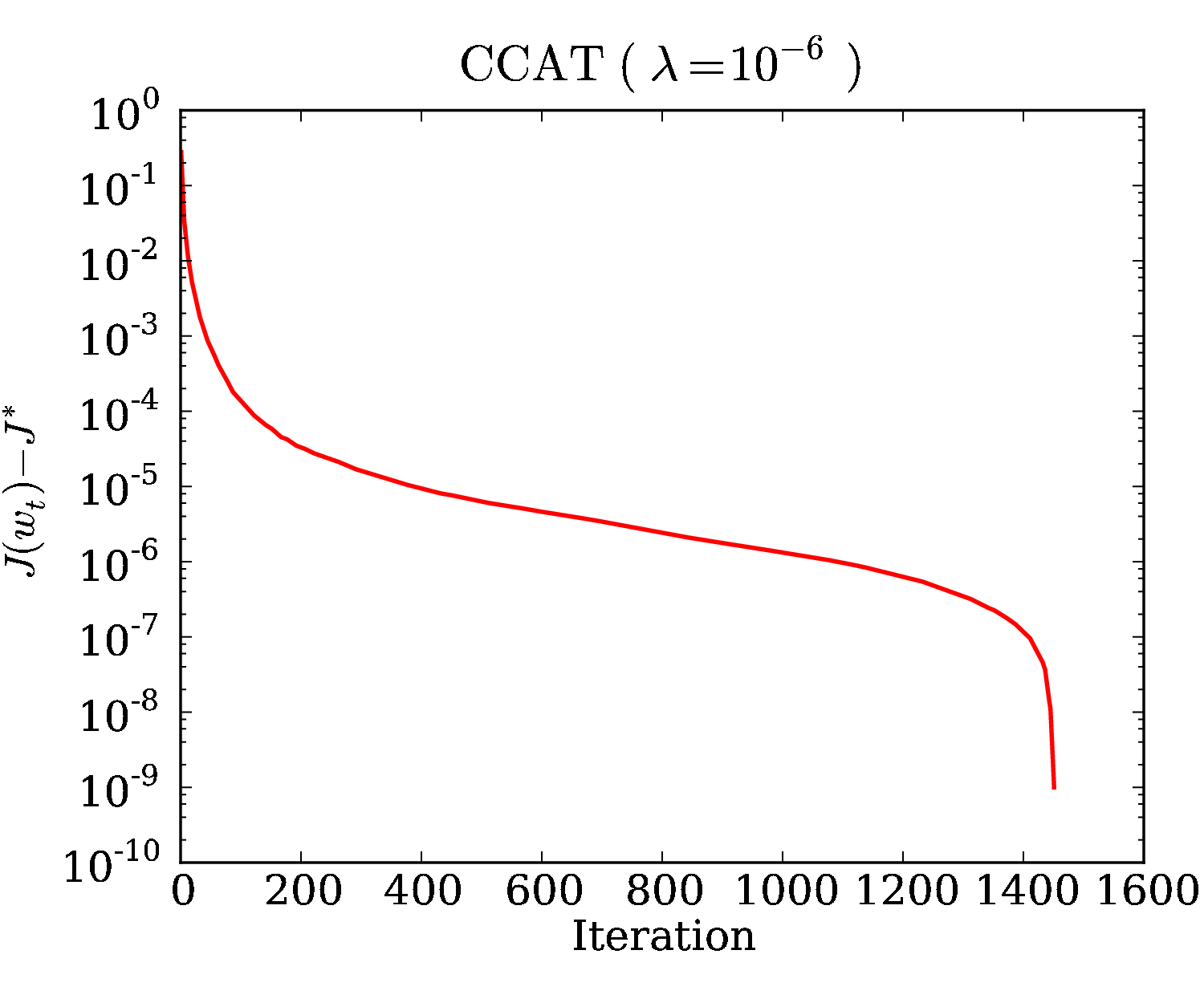} &
    \includegraphics[width=0.49\linewidth]{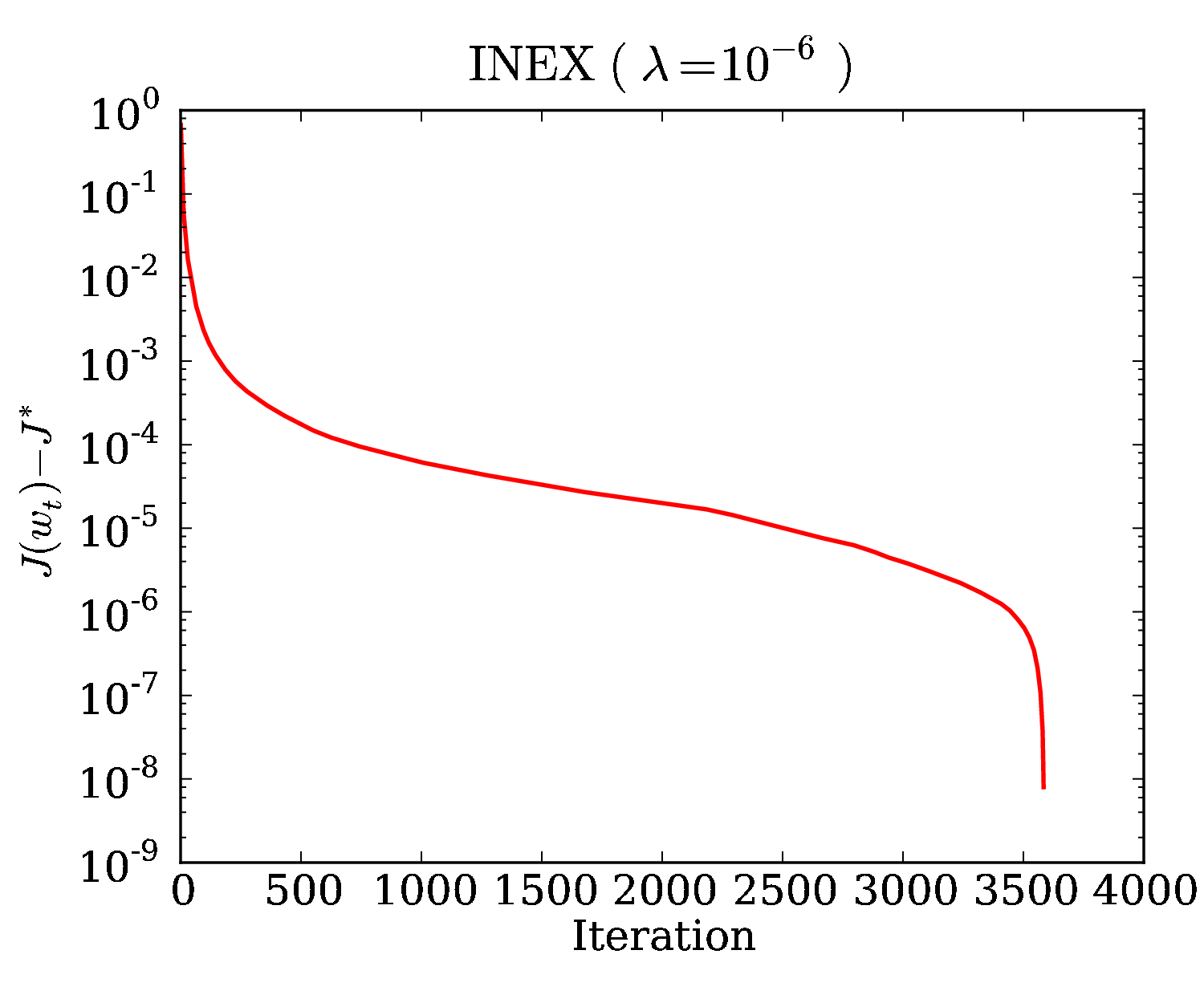} \\
 \end{tabular}
  \caption{Convergence of subLBFGS in objective function value on sample
    $L_2$-regularized risk minimization problems with binary (left) and
    multiclass (right) hinge losses.}
  \label{fig:lbfgs-convergence}
\end{figure}

In Section~\ref{sec:bounded-b} we have shown that the spectrum
of subBFGS' inverse Hessian estimate is bounded. From this and
other technical assumptions, we prove in
Appendix~\ref{sec:convergence-proof} that subBFGS is globally
convergent in objective function value, \ie $J(\vw) \rightarrow \inf_{\vw}
J(\vw)$. Moreover, in Appendix~\ref{sec:counterexample} we show
that subBFGS converges for all counterexamples we could find in the
literature used to illustrate the non-convergence of existing
optimization methods on nonsmooth problems.


We have also examined the convergence of subLBFGS empirically.
In most of our experiments of Section~\ref{sec:results}, we observe that
after an initial transient, subLBFGS observes a period of linear convergence,
until close to the optimum it exhibits superlinear convergence behavior.
This is illustrated in Figure~\ref{fig:lbfgs-convergence}, where we plot
(on a log scale) the excess objective function value $J(\vw_t)$ over its
``optimum'' $J^*$\footnote{Estimated empirically by running
subLBFGS for $10^4$ seconds, or until the relative improvement over 5
iterations was less than $10^{-8}$.} against the iteration number in two
typical runs.\ The same kind of convergence behavior was observed by
\citet[Figure 5.7]{LewOve08a}, who applied the classical BFGS algorithm
with a specially designed line search to nonsmooth functions. They
caution that the apparent superlinear convergence may be an artifact
caused by the inaccuracy of the estimated optimal value of the objective.

\section{SubBFGS for $L_2$-Regularized Binary Hinge Loss} 
\label{sec:subbfgs-hingeloss}

Many machine learning algorithms can be viewed as minimizing the
$L_2$-regularized risk
\begin{align}
  \label{eq:regrisk}
  J(\vw) ~:=~ \frac{\lambda}{2} \|\vw\|^2 \,+\, \frac{1}{n}
  \sum_{i=1}^{n} l(\vx_i, z_i, \vw),
\end{align}
where $\lambda > 0$ is a regularization constant, $\vx_i \in \Xcal
\subseteq \RR^{d}$ are the input features, $z_i \in \Zcal
\subseteq \ZZ$ the corresponding labels, and the loss $l$ is a
non-negative convex function of $\vw$ which measures the discrepancy
between $z_i$ and the predictions arising from using $\vw$. A loss
function commonly used for binary classification is the binary hinge
loss
\begin{align}
  \label{eq:binaryloss}
  l(\vx, z, \vw) ~:=~ \max(0, 1 - z\, \vw^{\top} \vx),
\end{align}
where $z \in \{\pm 1\}$. $L_2$-regularized risk minimization with
the binary hinge loss is a convex but nonsmooth optimization problem; in
this section we show how subBFGS (Algorithm~\ref{alg:subbfgs}) can be
applied to this problem.

Let $\Ecal$, $\Mcal$, and $\Wcal$ index the set of points which
are in error, on the margin, and well-classified, respectively:
\begin{align*}
  \Ecal  & := \{i \in \{1, 2, \ldots, n\}: 1 - z_i \vw^{\top} \vx_i > 0\}, \nonumber \\
  \Mcal & := \{i \in \{1, 2, \ldots, n\}: 1 - z_i \vw^{\top} \vx_i = 0\}, \\
  \Wcal & := \{i \in \{1, 2, \ldots, n\}: 1 - z_i \vw^{\top} \vx_i < 0
  \}. \nonumber
\end{align*}
Differentiating \eqref{eq:regrisk} after plugging in
\eqref{eq:binaryloss} then yields
\begin{align}
  \partial J(\vw) & ~=~ \lambda \,\vw - \frac{1}{n} \sum_{i=1}^{n}
  \beta_i z_i \vx_i
  ~=~ \bar{\vw} - \frac{1}{n} \sum_{i \in \Mcal} \beta_i z_i
  \vx_i, \label{eq:subdifferential} \\[1ex] 
  \mbox{where~~~} \bar{\vw} & := \lambda \,\vw - \frac{1}{n} \sum_{i
    \in \Ecal} z_i \vx_i \mbox{~~~and~~~} \beta_i := \left\{
    \begin{array}{cllr}
      1 & \mbox{if} & i \in \Ecal, \\
      \left [0, 1\right ] &\mbox{if} & i \in \Mcal, \\
      0 & \mbox{if} & i \in \Wcal.
    \end{array}
  \right. \nonumber
\end{align}

\subsection{Efficient Oracle for the Direction-Finding Method}
\label{sec:oracle}

Recall that subBFGS requires an oracle that provides
$\argsup_{\vg \in \partial J(\vw_t)} \vg^{\top} \vp$ for a given
direction $\vp$. For $L_2$-regularized risk minimization with the binary
hinge loss we can implement such an oracle at a computational cost of
$O(d\,|\Mcal_t|)$, where $d$ is the dimensionality of $\vp$ and $|\Mcal_t|$
the number of current margin 
points, which is normally much less than $n$.
Towards this end, we use \eqref{eq:subdifferential}
to obtain
\begin{align}
  \sup_{\vg \in \partial J(\vw_t)} \vg^{\top} \vp~~
  &= \sup_{\beta_i, i \in \Mcal_t} \left( \bar{\vw}_t -
    \frac{1}{n} \sum_{i \in \Mcal_t} \beta_i z_i \vx_i
  \right)^{\!\!\top} \!\!\vp \nonumber \\
  & =~ \bar{\vw}_t^{\top} \vp \,- \frac{1}{n} \sum_{i \in \Mcal_t}
  \inf_{\beta_i \in [0,1]} (\beta_i z_i \vx_i^{\top} \vp).
  \label{eq:sup}
\end{align}
Since for a given $\vp$ the first term of the right-hand side of
\eqref{eq:sup} is a constant, the supremum is attained when we
set $\beta_i ~\forall i\in \Mcal_{t}$ via the following strategy:
\begin{align*}
  \beta_i :=
  \begin{cases}
    0 & \text{if~~} z_i \vx_i^{\top} \vp_t \,\geq\, 0, \\
    1 & \text{if~~} z_i \vx_i^{\top} \vp_t \,<\, 0.
  \end{cases}
\end{align*}

\subsection{Implementing the  Line Search}
\label{sec:line-search}

\begin{figure}[tb]
  \centering
    \begin{tabular}{cc}
      \includegraphics[width=0.45\textwidth]{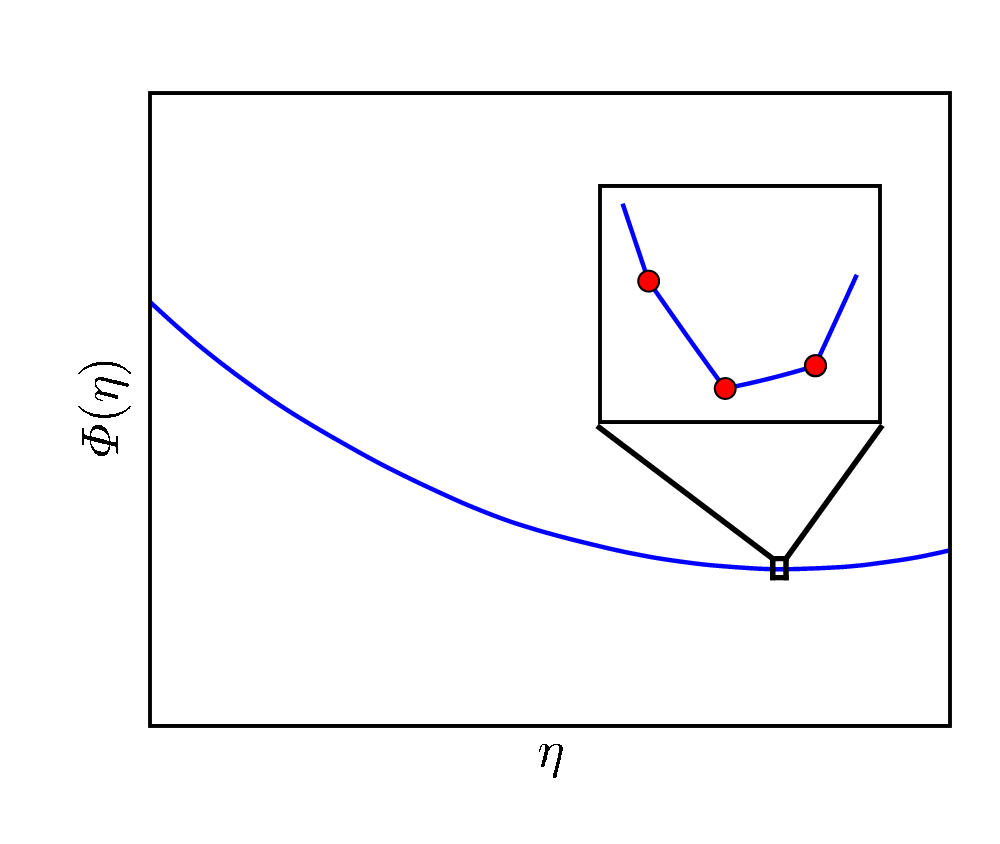} &
      \includegraphics[width=0.45\textwidth]{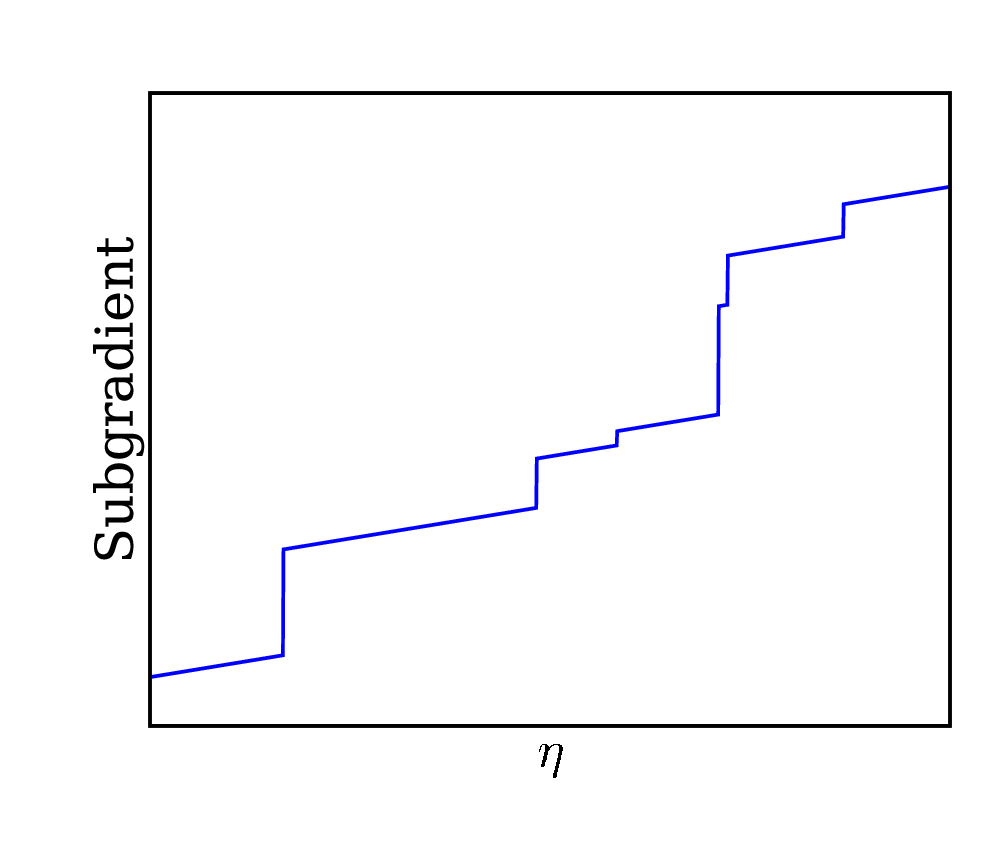}
    \end{tabular}
  \caption{Left: Piecewise quadratic convex function $\Phi$ of step size $\eta$;
    solid disks in the zoomed inset are subdifferentiable points.
    Right: The subgradient of $\Phi(\eta)$ increases monotonically
    with $\eta$, and jumps discontinuously at subdifferentiable points.}
  \label{fig:stepsize-func}
\end{figure}

The one-dimensional convex function $\Phi(\eta) := J(\vw + \eta \vp)$
(Figure~\ref{fig:stepsize-func}, left)
obtained by restricting \eqref{eq:regrisk} to a line can be evaluated
efficiently. To see this, rewrite \eqref{eq:regrisk} as
\begin{align}
  \label{eq:vec-regrisk}
  J(\vw) ~:=~ \frac{\lambda}{2} \,\|\vw\|^2 \,+~ \frac{1}{n} \,\vone^{\top}
  \!\max(\vzero, ~\vone - \vz \cdot \Xmat \vw),
\end{align}
where $\vzero$ and $\vone$ are column vectors of zeros and ones,
respectively, $\cdot$ denotes the Hadamard (component-wise) product,
and $\vz \in \RR^n$ collects correct labels corresponding to each row
of data in $\Xmat : = [\vx_1,\vx_2,\cdots,\vx_n]^{\top}\in
\RR^{n\times d}$. Given a search direction $\vp$ at a point $\vw$,
\eqref{eq:vec-regrisk} allows us to write
\begin{align}
  \label{eq:eva-regrisk}
  \Phi(\eta) & =~ \frac{\lambda}{2} \,\|\vw\|^2 +\, \lambda \,\eta \,\vw^{\top}
  \vp \,+\, \frac{\lambda \,\eta^{2}}{2} \,\|\vp\|^2 \,+~ \frac{1}{n}
  \,\vone^{\top}\max\left[0, ~(\vone - (\vf + \eta \,\Delta
    \vf))\right],
\end{align}
where $\vf := \vz \cdot \Xmat \vw$ and $\Delta \vf := \vz \cdot \Xmat
\vp$. Differentiating \eqref{eq:eva-regrisk} with respect to $\eta$
gives the subdifferential of $\Phi$:
\begin{align}
  \label{eq:grad-stepFunc}
  \partial \,\Phi(\eta) = \lambda \vw^{\top}\vp +
  \eta\lambda\norm{\vp}^2 - \frac{1}{n}\vde(\eta)^{\top}\Delta \vf,
\end{align}
where $\vde:\RR \rightarrow \RR^n$ outputs a column vector
$[\delta_1(\eta), \delta_2(\eta), \cdots, \delta_n(\eta)]^{\top}$ with
\begin{align}
  \delta_i(\eta) := \left\{
    \begin{array}{cll}
      1 & \text{if} & f_i + \eta \,\Delta f_i \,<\, 1, \\
      \left[0, 1\right] & \text{if} & f_i + \eta \,\Delta f_i \,=\, 1, \\
      0 & \text{if} & f_i + \eta \,\Delta f_i \,>\, 1.
    \end{array} \right.
  \label{eq:regrisk-para}
\end{align}

We cache $\vf$ and $\Delta \vf$, expending $O(nd)$ computational
effort and using $O(n)$ storage. We also cache the scalars
$\frac{\lambda}{2} \|\vw\|^2$, $\lambda \,\vw^{\top}\vp$, and
$\frac{\lambda}{2} \|\vp\|^2$, each of which requires $O(d)$ work. The
evaluation of $\vone - (\vf + \eta \,\Delta \vf)$, $\vde(\eta)$, and
the inner products in the final terms of \eqref{eq:eva-regrisk} and
\eqref{eq:grad-stepFunc} all take $O(n)$ effort. Given the cached
terms, all other terms in \eqref{eq:eva-regrisk} can be computed in
constant time, thus reducing the cost of evaluating $\Phi(\eta)$
(\emph{resp.}\ its subgradient) to $O(n)$. Furthermore, from
\eqref{eq:regrisk-para} we see that $\Phi(\eta)$ is differentiable
everywhere except at
\begin{align}
  \label{eq:binary-hinge}
  \eta_i := (1 - f_i)/\Delta f_i \mbox{~~with~~} \Delta f_i \neq 0,
\end{align}
where it becomes subdifferentiable. At these points an element of the
indicator vector \eqref{eq:regrisk-para} changes from $0$ to $1$ or vice
versa (causing the subgradient to jump, as shown in
Figure~\ref{fig:stepsize-func}, right); otherwise $\vde(\eta)$ remains
constant. Using this property of $\vde(\eta)$, we can update the last
term of \eqref{eq:grad-stepFunc} in constant time when passing a hinge
point (Line 25 of Algorithm~\ref{alg:exact-ls}). We are now in a
position to introduce an exact line search which takes advantage of this
scheme.

\begin{figure}[tb]
  \centering
  \begin{tabular}{cc}
    \includegraphics[width=0.45\textwidth]{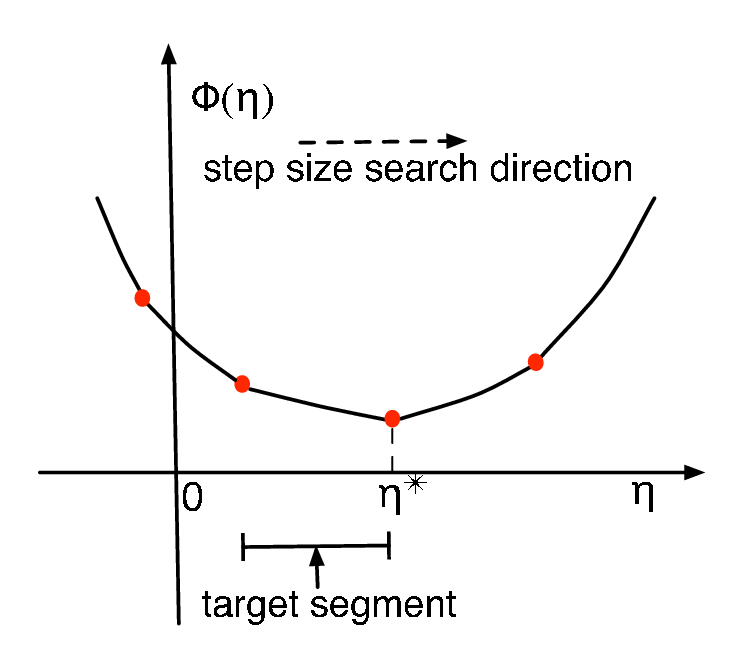} &
    \includegraphics[width=0.45\textwidth]{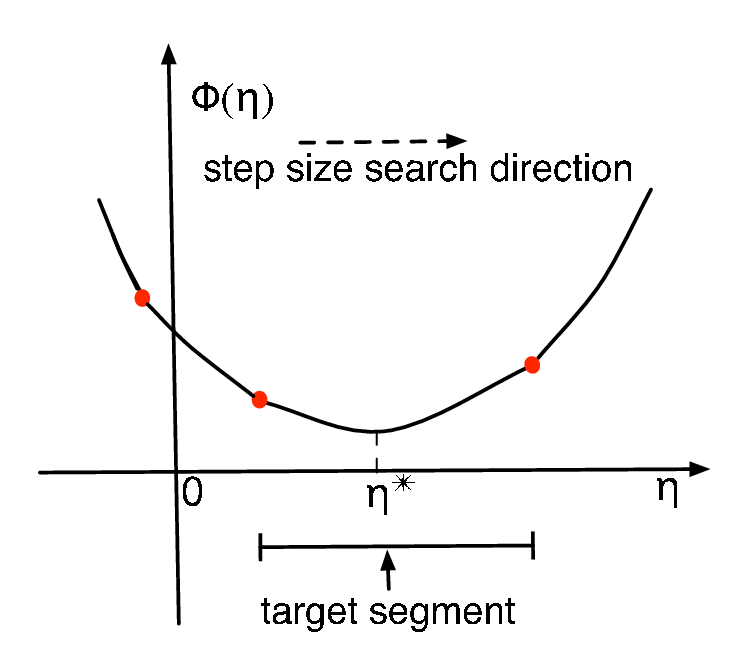}
  \end{tabular}
  \caption{Nonsmooth convex function $\Phi$ of step size $\eta$.
    Solid disks are subdifferentiable points; the optimal step
    $\eta^*$ either falls on such a point (left), or lies between two
    such points (right).}
  \label{fig:linesearch}
\end{figure}

\begin{algorithm}[t]
\caption{~Exact Line Search for $L_2$-Regularized Binary Hinge Loss}
\label{alg:exact-ls}
\begin{small}
\begin{algorithmic}[1]
  \INPUT $\vw, \vp, \lambda, \vf,$ and $\Delta \vf$ as in \eqref{eq:eva-regrisk}
  \OUTPUT optimal step size
  \STATE $ h = \lambda \|\vp\|^2, ~j := 1$
  \STATE $\veta := [(\vone - \vf)./\Delta \vf, 0]$ \hfill (vector of subdifferentiable points \& zero)
   \STATE $\vpi = {\tt argsort}(\veta)$ \hfill (indices sorted by non-descending value of $\veta$)
  \WHILE{$\eta_{\pi_j} \leq 0$}
    \STATE $j := j + 1$
  \ENDWHILE
  \STATE $\eta := \eta_{\pi_j}/2$  
  \FOR {$i := 1$ to $ \vf.{\tt size}$}
  \STATE \rule{0pt}{4ex} $\delta_i := \left\{ \begin{array}{rl}
      1 & \text{if~~} f_i + \eta \,\Delta f_i < 1 \\
      0 & \text{otherwise}\end{array} \right.$ \hfill (value of
  $\vde(\eta)$ \eqref{eq:regrisk-para} for any $\eta
  \in (0, \eta_{\pi_j})$)
  \ENDFOR
  \STATE $\varrho := \vde^{\top} \Delta \vf/n -  \lambda
  \vw^{\top}\!\vp$
  \STATE $\eta : = 0, ~ \varrho' := 0$
  \STATE $g := -\varrho$ \hfill (value of $\sup \partial \,\Phi (0)$)
  \WHILE{$ g < 0$}
  \STATE $\varrho' := \varrho$
  \IF{ $j > \vpi.{\tt size}$}
  \STATE $\eta := \infty$ \hfill (no more subdifferentiable points)
  \STATE {\bf break}
  \ELSE
  \STATE $\eta := \eta_{\pi_{j}}$
  \ENDIF
  \REPEAT
  \STATE $\varrho := \left\{ \begin{array}{rl@{\hspace{19ex}}r}
      \varrho - \Delta f_{\pi_j}/n  & \text{if~~} \delta_{\pi_j} = 1 & \text{(move to next subdifferentiable~} \\
      \varrho + \Delta f_{\pi_j}/n & \text{otherwise} & \text{point and update $\varrho$ accordingly)}
      \end{array}\right.$
  \STATE $j := j + 1$
  \UNTIL{$\eta_{\pi_j} \ne \eta_{\pi_{j-1}}$ and $j \le \vpi.{\tt size}$ }
  \STATE $ g := \eta h - \varrho $ \hfill (value of  $\sup \partial \,\Phi(\eta_{\pi_{j-1}})$)
  \ENDWHILE
  \RETURN $\min(\eta,~ \varrho'/h)$ \hfill (\cf equation~\ref{eq:eta-star})
\end{algorithmic}
\end{small}
\end{algorithm}
\subsubsection{Exact Line Search}
\label{sec:exact-ls}

Given a direction $\vp$, exact line search finds the optimal step size
$\eta^*:= \argmin_{\eta \ge 0} \Phi(\eta)$ that satisfies $0
\in \partial \,\Phi(\eta^*)$, or equivalently
\begin{align}
\label{eq:optimality-step}
  \inf \partial \,\Phi(\eta^*) \le 0 \le \sup \partial \,\Phi(\eta^*).
\end{align}
By Theorem~\ref{th:increasing-subgrad}, $\sup \partial \,\Phi(\eta)$
is monotonically increasing with $\eta$. Based on this property, our
algorithm first builds a list of all possible sub\-differentiable
points and $\eta = 0$, sorted by non-descending value of $\eta$
(\BLSCALHINGE~of Algorithm~\ref{alg:exact-ls}). Then, it starts with
$\eta = 0$, and walks through the sorted list until it locates the
``target segment'', an interval $[\eta_a, \eta_b]$ between two
subdifferential points with $\sup \partial \,\Phi(\eta_a) \leq 0$ and
$\sup \partial \,\Phi(\eta_b) \ge 0$. We now know that the optimal
step size either coincides with $\eta_b$ (Figure~\ref{fig:linesearch},
left), or lies in $(\eta_a, \eta_b)$ (Figure~\ref{fig:linesearch},
right). If $\eta^*$ lies in the smooth interval $(\eta_a, \eta_b)$,
then setting \eqref{eq:grad-stepFunc} to zero gives
\begin{align}
  \label{eq:eta-star}
  \eta^* = \frac{\vde(\eta')^{\top}\Delta \vf/n - \lambda
    \,\vw^{\top} \vp}{\lambda \,\norm{\vp}^2}, ~~\forall \eta' \in (\eta_{a},
  \eta_b).
\end{align}
Otherwise, $\eta^* = \eta_{b}$. See Algorithm~\ref{alg:exact-ls} for
the detailed implementation.

\section{Segmenting the Pointwise Maximum of 1-D Linear Functions}
\label{sec:Minimization1DConvex}

\begin{figure}
\centering
  \subfigure[Pointwise maximum of lines]{
    \includegraphics[width=0.31\linewidth]{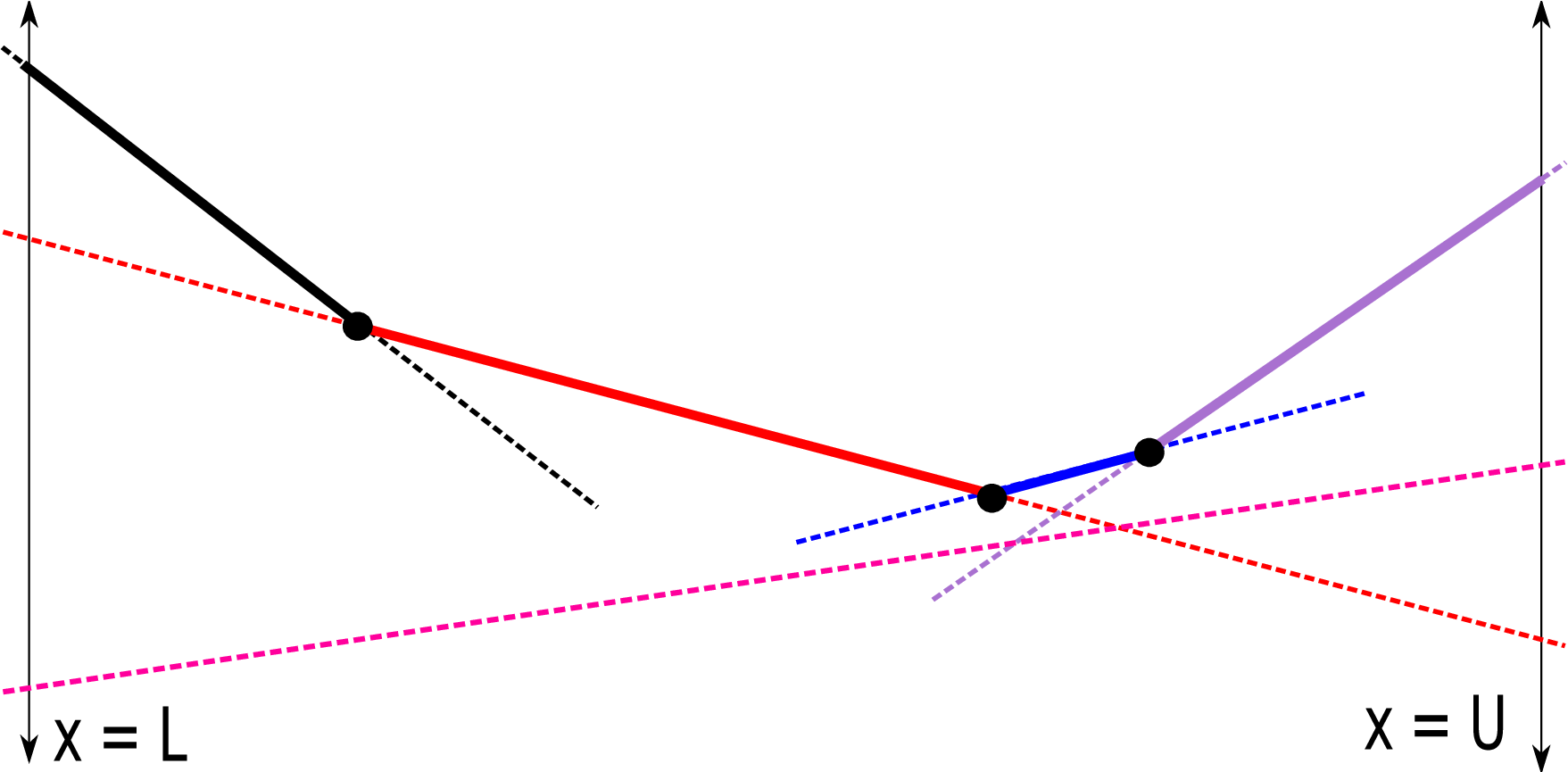} \label{fig:func}}
  \subfigure[Case 1]{
   \includegraphics[width=0.31\linewidth]{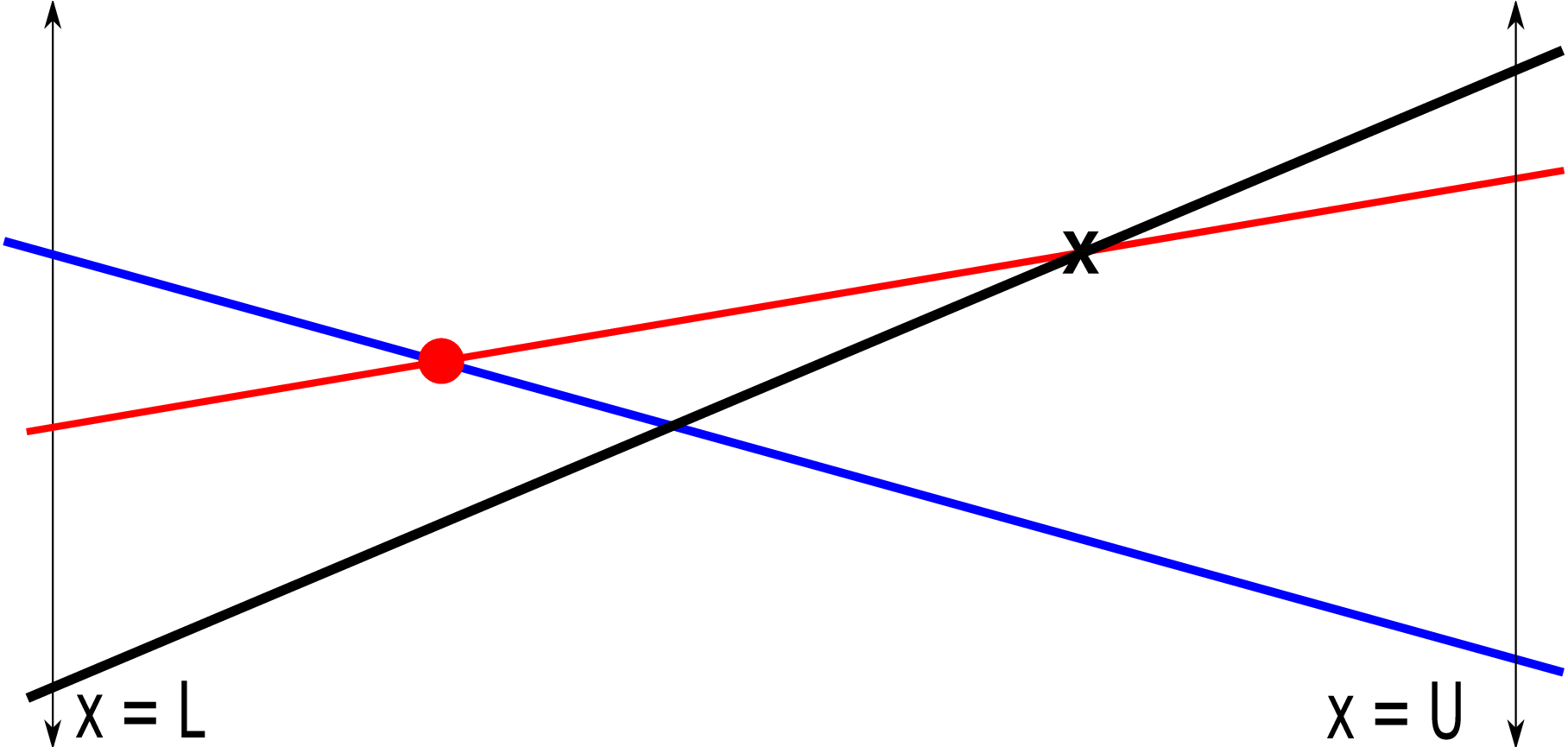} \label{fig:case1}}
  \subfigure[Case 2]{
    \includegraphics[width=0.31\linewidth]{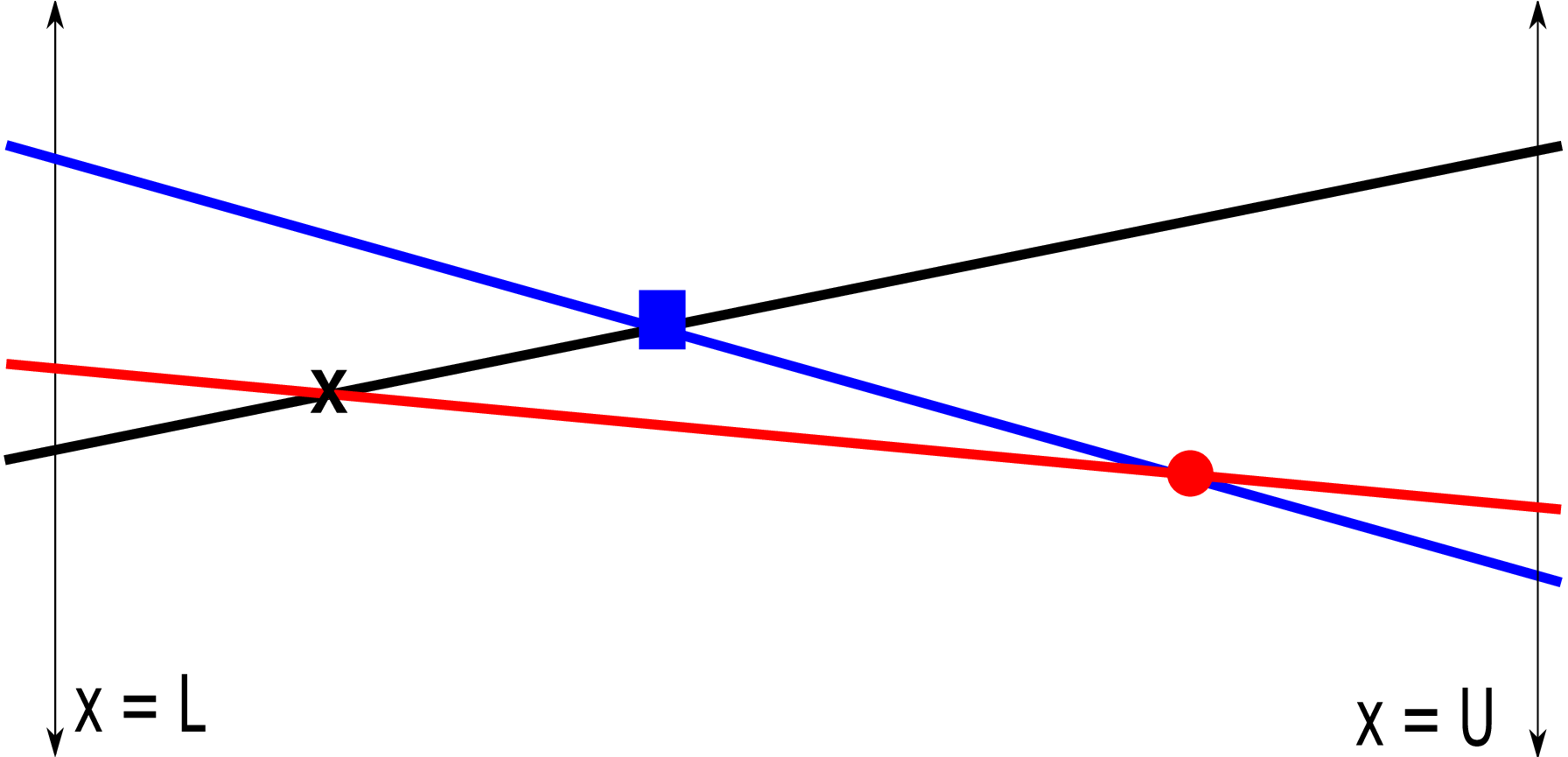} \label{fig:case2}}
  \caption{(a) Convex piecewise linear function defined as the maximum
    of 5 lines, but comprising only 4 active line segments (bold) separated
    by 3 subdifferentiable
    points (black dots). (b, c) Two cases encountered by our algorithm:
    (b) The new intersection (black cross) lies to the right of the
    previous one (red dot) and is therefore pushed onto the stack;
    (c) The new intersection lies to the left of the previous one.
     In this case the latter is popped from the stack, and a third
     intersection (blue  square) is computed and pushed onto it.} 
  \label{fig:cplf}
\end{figure}

The line search of Algorithm~\ref{alg:exact-ls} requires a vector
$\veta$ listing the subdifferentiable points along the line $\vw +
\eta \vp$, and sorts it in non-descending order (Line 5).  For an
objective function like \eqref{eq:regrisk} whose nonsmooth component
is just a sum of hinge losses \eqref{eq:binaryloss}, this vector is
very easy to compute (\cf \eqref{eq:binary-hinge}). In order to apply
our line search approach to multiclass and multilabel losses, however,
we must solve a more general problem: we need to efficiently find the
subdifferentiable points of a one-dimensional piecewise linear
function $\varrho: \RR \to \RR$ defined to be the pointwise maximum of
$r$ lines:
\begin{align}
  \label{eq:cplf}
  \varrho(\eta) = \max_{1 \leq p \leq r}(b_p + \eta \,a_p),
\end{align}
where $a_{p}$ and $b_{p}$ denote the slope and offset of the $p^{\rm
  th}$ line, respectively.  Clearly, $\varrho$ is convex since it is
the pointwise maximum of linear functions \citep[][Section
3.2.3]{BoyVan04}, \cf Figure~\ref{fig:func}.  The difficulty here is
that although $\varrho$ consists of at most $r$ line segments bounded
by at most $r-1$ subdifferentiable points, there are $r(r-1)/2$
candidates for these points, namely all intersections between any two
of the $r$ lines. A naive algorithm to find the subdifferentiable
points of $\varrho$ would therefore take $O(r^2)$ time.  In what
follows, however, we show how this can be done in just $O(r \log r)$
time. In Section~\ref{sec:subbfgs-multiloss} we will then use this
technique (Algorithm~\ref{alg:mincplf}) to perform efficient exact
line search in the multiclass and multilabel settings.

\begin{algorithm}[t]
  \caption{~Segmenting a Pointwise Maximum of 1-D Linear Functions}
  \label{alg:mincplf}
    \begin{algorithmic}[1]
      \INPUT vectors $\vec{a}$ and $\vec{b}$ of slopes and offsets \\
      ~~~~~~~~~lower bound $L$, upper bound $U$, with $0 \leq L < U < \infty$ 
      \OUTPUT sorted stack of subdifferentiable points $\veta$ \\
      ~~~~~~~~~~~and corresponding active line indices $\vxi$
      \STATE $\ylow := \vec{b} + L\vec{a}$
      \STATE $\vpi := {\tt argsort}(-\ylow)$
      \hfill (indices sorted by non-ascending value of $\ylow$)
      \STATE $S.\mathrm{\tt push} ~(L, \pi_1)$
      \hfill(initialize stack)
      \FOR{$q := 2 \;\; \mathbf{ to } \;\; \ylow.{\tt size}$}
      \WHILE{not $S.{\tt empty}$}
      \STATE $(\eta, \xi) := S.{\tt top}$
      \STATE $\displaystyle\rule{0pt}{4ex} \eta' := \frac{b_{\pi_q} - b_{\xi}}{
        \text{\raisebox{2ex}{$a_{\xi} - a_{\pi_q}$}}}$  \hfill
      (intersection of two lines)
      \IF{$L < \eta' \leq \eta$ or ($\eta' = L$ and $a_{\pi_q} > a_{\xi}$)}
      \STATE $S.{\tt pop}$ \hfill (\cf Figure~\ref{fig:case2})~~
      \ELSE
      \STATE {\bf break}
      \ENDIF
      \ENDWHILE
      \IF{$L < \eta' \leq U$  or ($\eta' = L$ and $a_{\pi_q} > a_{\xi}$) }
      \STATE $S.{\tt push} ~(\eta', \pi_q)$ \hfill (\cf Figure~\ref{fig:case1})~~
      \ENDIF
      \ENDFOR
      \RETURN $S$ 
    \end{algorithmic}
\end{algorithm}

We begin by specifying an interval $[L, U] ~(0 \leq L < U < \infty)$ in
which to find the sub\-differentiable points of $\varrho$, and set
$\ylow := \vec{b} + L \vec{a}$, where $\vec{a} = [a_1, a_2, \cdots,
a_r]$ and $\vec{b} = [b_1, b_2, \cdots,
b_r]$. In other words, $\ylow$ contains the
intersections of the $r$ lines defining $\varrho(\eta)$ with the
vertical line $\eta = L$. Let $\vpi$ denote the permutation that sorts
$\ylow$ in non-ascending order, \ie $p < q \implies y_{\pi_p} \geq
y_{\pi_q}$, and let $\varrho^{(q)}$ be the function obtained by
considering only the top $q \leq r$ lines at $\eta = L$, \ie the first
$q$ lines in $\vpi$:
\begin{align}
  \label{eq:cplf-partial}
  \varrho^{(q)}(\eta) = \max_{1 \leq p \leq q}(b_{\pi_p} + \eta \,a_{\pi_p}). 
\end{align}
It is clear that $\varrho^{(r)} = \varrho$. Let $\veta$ contain all
$q' \leq q-1$ subdifferentiable points of $\varrho^{(q)}$ in $[L, U]$
in ascending order, and $\vxi$ the indices of the corresponding
\emph{active} lines, \ie the maximum in \eqref{eq:cplf-partial} is
attained for line $\xi_{j-1}$ over the interval $[\eta_{j-1},
\eta_{j}]$: $\xi_{j-1} := \pi_{p^{*}}$, where $p^{*} = \argmax_{1 \leq
  p \leq q} (b_{\pi_p} + \eta a_{\pi_p}) \text{~for~} \eta \in
[\eta_{j-1}, \eta_{j}]$, and lines $\xi_{j-1}$ and $\xi_{j}$ intersect
at $\eta_{j}$.

Initially we set $\eta_0 := L$ and $\xi_0 := \pi_1$, the leftmost bold
segment in Figure~\ref{fig:func}. Algorithm~\ref{alg:mincplf} goes
through lines in $\vpi$ sequentially, and maintains a
Last-In-First-Out stack $S$ which at the end of the $q^{\rm th}$
iteration consists of the tuples
\begin{align}
  (\eta_0, \xi_0), (\eta_1, \xi_1), \ldots, (\eta_{q'}, \xi_{q'})
\end{align}
in order of ascending $\eta_i$, with $(\eta_{q'}, \xi_{q'})$ at the
top.  After $r$ iterations $S$ contains a sorted list of all
subdifferentiable points (and the corresponding active lines) of
$\varrho = \varrho^{(r)}$ in $[L, U]$, as required by our line
searches.

In iteration $q + 1$ Algorithm~\ref{alg:mincplf} examines
the intersection $\eta'$ between lines $\xi_{q'}$ and $\pi_{q+1}$:
If $\eta' > U$, line $\pi_{q+1}$ is irrelevant, and we proceed to the
next iteration. If $\eta_{q'} < \eta' \leq U$ as in Figure~\ref{fig:case1},
then line $\pi_{q+1}$ is becoming active at $\eta'$, and we simply
push $(\eta'\!, \pi_{q+1})$ onto the stack.
If $\eta' \leq \eta_{q'}$ as in Figure~\ref{fig:case2}, on the other hand, 
then line $\pi_{q+1}$ dominates line $\xi_{q'}$ over the interval
$(\eta'\!, \infty)$ and hence over $(\eta_{q'}, U] \subset (\eta'\!, \infty)$,
so we pop $(\eta_{q'}, \xi_{q'})$ from the stack (deactivating line
$\xi_{q'}$), decrement $q'\!$, and repeat the comparison.

\begin{theorem}
  The total running time of Algorithm~\ref{alg:mincplf} is $O(r \log
  r)$. 
\end{theorem}
\begin{proof}
  Computing intersections of lines as well as
  pushing and popping from the stack require $O(1)$ time. Each of the
  $r$ lines can be pushed onto and popped from the stack at most
  once; amortized over $r$ iterations the running time is therefore
  $O(r)$. The time complexity of Algorithm~\ref{alg:mincplf} is thus
  dominated by the initial sorting of $\ylow$ (\ie the computation
  of $\vpi$), which takes $O(r \log r)$ time.
\end{proof}

\section{SubBFGS for Multiclass and Multilabel Hinge Losses}
\label{sec:subbfgs-multiloss}

We now use the algorithm developed in
Section~\ref{sec:Minimization1DConvex} to generalize the subBFGS
method of Section~\ref{sec:subbfgs-hingeloss} to the multiclass and
multilabel settings with finite label set $\Zcal$. We assume that
given a feature vector $\vx$ our classifier predicts the label
\begin{align*}
z^{*} =
\argmax_{z \in \Zcal} f(\vw, \vx, z),
\end{align*}
where $f$ is a linear function
of $\vw$, \ie $f(\vw, \vx, z) = \inner{\vw}{\phi(\vx, z)}$ for some
feature map $\phi(\vx, z)$.

\comment{
\emph{Jin's version}
\subsection{Multiclass Hinge Loss}

A variety of multiclass hinge losses have been proposed in the literature
that generalize the binary hinge loss and enforce a margin of separation
between the true label $z_i$ and every other label. We
focus on the following rather general variant
\citep{TasGueKol04}:\footnote{Our algorithm can also deal with
  the slack-rescaled variant of \citet{TsoJoaHofAlt05}.}
\begin{align}
  \label{eq:multi-lossdef}
  l(\vx_i, z_i, \vw) ~:=~ \max_{z \neq z_i} \,[0,~\Delta(z, z_i) + f(\vw,
  \vx_i, z) - f(\vw, \vx_i, z_i)],
\end{align}
where $\Delta(z, z_i) \geq 0$ is the \emph{label loss} specifying the margin
required between labels $z$ and $z_i$.
For instance, a uniform margin of separation is achieved by
setting $\Delta(z, z') := \tau > 0 ~\forall z \neq z'$ \citep{CraSin03}.
Adapting \eqref{eq:regrisk} to the multiclass hinge loss
\eqref{eq:multi-lossdef} we obtain
\begin{align}
  \label{eq:multi-obj}
  J(\vw) ~:=~ \frac{\lambda}{2} \|\vw\|^{2} + \frac{1}{n} \sum_{i=1}^{n}
  \max_{z \neq z_i} \,[0, ~\Delta(z, z_i) + f(\vw, \vx_i, z) - f(\vw,
    \vx_i, 
z_i)].
\end{align}

For a given $\vw$, consider the set
\begin{align*}
  \Zcal_i^* :=\, \argmax_{z \neq z_i}
  [\Delta(z, z_i) + f(\vw, \vx_i, z) - f(\vw, \vx_i, z_i)]
\end{align*}
of maximum-loss labels (possibly more than one)
for the $i^{\rm th}$ training instance, and define 
\begin{align*}
  l_i^* ~:=~ \max_{z \neq z_i}
  [\Delta(z, z_i) + f(\vw, \vx_i, z) - f(\vw, \vx_i, z_i)].
\end{align*}
Since $f(\vw, \vx, z) = \inner{\vw}{\phi(\vx, z)}$, the
subdifferential of \eqref{eq:multi-obj} can then be written as
\begin{align}
  \label{eq:multi-gradient}
  \partial J(\vw)  ~=~ \lambda \vw + \frac{1}{n} \sum_{i=1}^{n}
  \sum_{z
    \in \Zcal} \beta_{i,z} \,\phi(\vx_i,z),
\end{align} 
where $\beta_{i,z}$ has the following choices of values:
\begin{align}
 \label{eq:multi-grad-cases}
\left \{
    \begin{array}{c | c c c}
      &  z = z_i & z \in \Zcal^*_i & \mbox{otherwise} \\
      \hline
      l^*_i < 0 & 0 & 0 & 0 \\
      l^*_i = 0 & [-1, 0] & [0,1] & 0 \\
      l^*_i > 0 & -1 & [0,1] & 0 \\
    \end{array}
  \right \}~~\mbox{s.t.}~~ \sum_{z\in\Zcal} \beta_{i,z} = 0.
\end{align}
Note that the constrain $\sum_{z\in \Zcal} \beta_{i,z} =0$ in
\eqref{eq:multi-grad-cases} effectively ensures that $\sum_{z \in \Zcal_i^*}
\beta_{i,z} = -\beta_{i, z_i}$. This enables us to simplify
\eqref{eq:multi-gradient} as
\begin{align}
  \label{eq:multi-gradient-simple}
  \partial J(\vw) & ~=~ \lambda \vw + \frac{1}{n} \sum_{i=1}^{n}
  \left(\sum_{z \in \Zcal_i^*} \beta_{i,z} \,\phi(\vx_i,z) +
    \beta_{i,z_i} \,\phi(\vx_i,z_i) \right) \nonumber \\
  &~=~ \lambda \vw + \frac{1}{n} \sum_{i=1}^{n} \sum_{z \in \Zcal_i^*}
  \beta_{i,z} \left [ \phi(\vx_i,z) - \, \phi(\vx_i,z_i). \right ]
\end{align}

\subsection{Efficient Multiclass Direction-Finding Oracle}
\label{sec:OracleMulticlas}

For $L_2$-regularized risk minimization with multiclass hinge loss, we
can use a similar scheme as described in Section~\ref{sec:oracle} to
implement an efficient oracle that provides $\argsup_{\vg \in \partial
  J(\vw)} \vg^{\!\top} \!\vp$ for the direction-finding procedure
(Algorithm~\ref{alg:find-descent-dir-cg}). Using
\eqref{eq:multi-gradient-simple}, we can write
\begin{align*}
  \sup_{\vg \in \partial J(\vw)} \!\vg^{\!\top} \!\vp ~=~ \lambda
  \vw^{\!\top} \!\vp \,+\, \frac{1}{n}\sum^n_{i=1} \,\sum_{z \in \Zcal^*_i}
  \,\sup_{\beta_{i, z} \in [0, 1]} \beta_{i, z} \,\left [ \phi(\vx_i, z) - \phi(\vx_i,
  z_i)\right ] ^{\!\top} \!\vp.
\end{align*}
Let $z^* := \argmax_{z \in \Zcal^*_i} [\phi(\vx_i, z) - \phi(\vx_i,
z_i) ]^{\top} \vp$ be the \emph{steepest} maximal margin violator, and
$d^* := [\phi(\vx_i, z^*_i) - \phi(\vx_i, z_i) ]^{\top}\vp$ the
associated slope. The supremum of \eqref{eq:multi-sup} is attained
when we pick, from the choices offered by
\eqref{eq:multi-grad-cases},
\begin{align*}
  \beta_{i,z} :=
  \begin{cases}
    0 & \text{if~~}  l^*_i < 0 \vee (l^*_i = 0 \wedge d^*_i \le 0),\\
    1 & \text{if~~}  l^*_i > 0 \vee (l^*_i = 0 \wedge d^*_i > 0),
  \end{cases}
  \mbox{~~for~~} z \in \Zcal^*_i, ~ 1\le i \le n.
\end{align*}
}

\subsection{Multiclass Hinge Loss}

A variety of multiclass hinge losses have been proposed in the literature
that generalize the binary hinge loss, and enforce a margin of separation
between the true label $z_i$ and every other label. We
focus on the following rather general variant
\citep{TasGueKol04}:\footnote{Our algorithm can also deal with
  the slack-rescaled variant of \citet{TsoJoaHofAlt05}.}
\begin{align}
  \label{eq:multi-lossdef}
  l(\vx_i, z_i, \vw) ~:=~ \max_{z \in \Zcal} \,[\Delta(z, z_i) + f(\vw,
  \vx_i, z) - f(\vw, \vx_i, z_i)],
\end{align}
where $\Delta(z, z_i) \geq 0$ is the \emph{label loss} specifying the margin
required between labels $z$ and $z_i$.
For instance, a uniform margin of separation is achieved by
setting $\Delta(z, z') := \tau > 0 ~\forall z \neq z'$ \citep{CraSin03}.
By requiring that $\forall z \in \Zcal: \Delta(z, z) = 0$ we ensure that
\eqref{eq:multi-lossdef} always remains non-negative.
Adapting \eqref{eq:regrisk} to the multiclass hinge loss
\eqref{eq:multi-lossdef} we obtain
\begin{align}
  \label{eq:multi-obj}
  J(\vw) ~:=~ \frac{\lambda}{2} \|\vw\|^{2} + \frac{1}{n} \sum_{i=1}^{n}
  \max_{z \in \Zcal} \,[\Delta(z, z_i) + f(\vw, \vx_i, z) - f(\vw,
    \vx_i, z_i)].
\end{align}

For a given $\vw$, consider the set
\begin{align}
\label{eq:worst-label}
  \Zcal_i^* :=\, \argmax_{z \in \Zcal}
  [\Delta(z, z_i) + f(\vw, \vx_i, z) - f(\vw, \vx_i, z_i)]
\end{align}
of maximum-loss labels (possibly more than one)
for the $i^{\rm th}$ training instance.
Since $f(\vw, \vx, z) = \inner{\vw}{\phi(\vx, z)}$, the
subdifferential of \eqref{eq:multi-obj} can then be written as 
\begin{align}
  \label{eq:multi-gradient}
  \partial J(\vw) & ~=~ \lambda \vw + \frac{1}{n} \sum_{i=1}^{n} \sum_{z
   \in \Zcal} \beta_{i,z} \,\phi(\vx_i,z) \\
  \text{with~~~}
  \label{eq:multi-grad-cases}
  \beta_{i,z} & ~=~ \left\{
  \begin{array}{rl}
    [0,1] & \text{if~} z \in \Zcal_i^* \\
    0 & \text{otherwise}
  \end{array}
  \right\} \,-\, \delta_{z,z_i}
  \text{~~s.t.~~} \sum_{z \in \Zcal} \beta_{i,z} = 0,
\end{align}
where $\delta$ is the Kronecker delta: $\delta_{a,b} = 1$ if $a = b$, and
0 otherwise.\footnote{Let
  $ l_i^* := \max_{z \neq z_i} [\Delta(z, z_i) + f(\vw, \vx_i, z) - f(\vw, \vx_i, z_i)].$
  Definition \eqref{eq:multi-grad-cases} allows the following values of $\beta_{i,z}$: 
\begin{align*}
\left \{
    \begin{array}{c | c c c}
      &  z = z_i & z \in \Zcal^*_i \setminus \{z_i\} & \mbox{otherwise} \\
      \hline
      l^*_i < 0 & 0 & 0 & 0 \\
      l^*_i = 0 & [-1, 0] & [0,1] & 0 \\
      l^*_i > 0 & -1 & [0,1] & 0 \\
    \end{array}
  \right \}~~\mbox{s.t.}~~ \sum_{z\in\Zcal} \beta_{i,z} = 0.
\end{align*}
}

\subsection{Efficient Multiclass Direction-Finding Oracle}
\label{sec:OracleMulticlass}

For $L_2$-regularized risk minimization with multiclass hinge loss, we
can use a similar scheme as described in Section~\ref{sec:oracle} to
implement an efficient oracle that provides $\argsup_{\vg \in \partial
  J(\vw)} \vg^{\!\top} \!\vp$ for the direction-finding procedure
(Algorithm~\ref{alg:find-descent-dir-cg}). Using \eqref{eq:multi-gradient}, we
can write
\begin{align}
  \sup_{\vg \in \partial J(\vw)} \!\vg^{\!\top} \!\vp ~=~ \lambda
  \vw^{\!\top} \!\vp \,+\, \frac{1}{n}\sum^n_{i=1} \,\sum_{z \in \Zcal}
  \,\sup_{\beta_{i, z}} \left(\beta_{i, z} \,\phi(\vx_i, z)^{\!\top} \!\vp\right).
  \label{eq:multi-sup}
\end{align}
The supremum in \eqref{eq:multi-sup} is attained when we pick, from the
choices offered by \eqref{eq:multi-grad-cases},
\begin{align*}
  \beta_{i,z} \,:=\, \delta_{z, z_i^*} - \delta_{z,z_i} ,
  \text{~~where~~~}
  z_i^* \,:=\, \argmax_{z \in \Zcal_i^*} \phi(\vx_i,z)^{\!\top} \!\vp.
\end{align*}

\subsection{Implementing the Multiclass Line Search}
\label{sec:linesearch-multi}

Let $\Phi(\eta) := J(\vw + \eta \vp)$ be the
one-dimensional convex function obtained by restricting
\eqref{eq:multi-obj} to a line along direction $\vp$. Letting
$\varrho_i(\eta) := l(\vx_i, z_i, \vw + \eta \vp)$, we can write
\begin{align}
  \label{eq:multi-oned}
  \Phi(\eta) ~=~ \frac{\lambda}{2} \| \vw \|^2 +\, \lambda \eta \inner{\vw}{\vp}
  \,+ \frac{\lambda \eta^2}{2} \| \vp \|^2 \,+\, \frac{1}{n} \sum_{i=1}^{n} \varrho_i(\eta).
\end{align}
Each $\varrho_i(\eta)$ is a piecewise linear convex function.
To see this, observe that 
\begin{align}
f(\vw + \eta \vp, \vx, z) :=
\inner{(\vw + \eta \vp)}{\phi(\vx, z)} = f(\vw, \vx, z) + \eta f(\vp,
\vx, z)
\end{align}
and hence
\begin{align}
\varrho_i(\eta) \,:=\, \max_{z \in \Zcal} \; [\underbrace{\Delta(z, z_i)
    + f(\vw, \vx_i, z) - f(\vw, \vx_i, z_i)}_{=: \:b_z^{(i)}} \,+~ \eta
  \underbrace{(f(\vp, \vx_i, z) - f(\vp, \vx_i, z_i))}_{=:
    \:a_z^{(i)}}],
  \label{eq:multi-lossdef-rw}
\end{align}
which has the functional form of \eqref{eq:cplf} with $r = |\!\Zcal\!|$.
Algorithm~\ref{alg:mincplf} can therefore be used to compute a sorted
vector $\veta^{(i)}$ of all subdifferentiable points of
$\varrho_i(\eta)$ and corresponding active
lines $\vxi^{(i)}\!$ in the interval $[0, \infty)$
in $O(|\!\Zcal\!| \log |\!\Zcal\!|)$ time. With some abuse of notation,
we now have
\begin{align}
\eta \in [\eta_j^{(i)}, \eta_{j+1}^{(i)}] ~\implies~
\varrho_i(\eta) \,=\, b_{\xi_j^{(i)}} +\, \eta \,a_{\xi_j^{(i)}}.
\label{eq:activeline}
\end{align}

The first three terms of \eqref{eq:multi-oned} are constant, linear,
and quadratic (with non-negative coefficient) in $\eta$, respectively. The
remaining sum of piecewise linear convex functions $\varrho_i(\eta)$
is also piecewise linear and convex, and so $\Phi(\eta)$ is a piecewise
quadratic convex function.

\subsubsection{Exact Multiclass Line Search}
\label{sec:ExactLineSearch}

Our exact line search employs a similar two-stage strategy as
discussed in Section~\ref{sec:exact-ls} for locating its minimum
$\eta^* := \argmin_{\eta > 0} \Phi(\eta)$: we first find the first
\emph{subdifferentiable} point $\check{\eta}$ past the minimum, then
locate $\eta^*$ within the differentiable region to its left. We
precompute and cache a vector $\vec{a}^{(i)}$ of all the slopes
$a_z^{(i)}$ (offsets $b_z^{(i)}\!$ are not needed), the
subdifferentiable points $\veta^{(i)}$ (sorted in ascending order
via Algorithm~\ref{alg:mincplf}), and the corresponding
indices $\vxi^{(i)}$ of active lines of $\varrho_i$ for all training
instances $i$, as well as $\|\vw\|^2\!$, $\inner{\vw}{\vp}$, and
$\lambda\|\vp\|^2\!$.

\begin{algorithm}[t]
\caption{~Exact Line Search for $L_2$-Regularized Multiclass Hinge Loss}
\label{alg:exact-multi-ls}
\begin{algorithmic}[1]
  \INPUT base point $\vw$, descent direction $\vp$,
  regularization parameter $\lambda$, vector $\vec{a}$ of \\
  ~~~~~~~~~~all slopes as defined in \eqref{eq:multi-lossdef-rw},
  for each training instance $i$: sorted stack $S_i$ of \\
  ~~~~~~~~~subdifferentiable  points and active lines,
  as produced by Algorithm~\ref{alg:mincplf}
  \OUTPUT optimal step size
  \STATE $\vec{a} := \vec{a}/n, ~h := \lambda \|\vp\|^2$
  \STATE $\varrho := \lambda \inner{\vw}{\vp}$
  \FOR {$i := 1$ to $n$}
    \WHILE{not $S_i.{\tt empty}$}
      \STATE $R_i.{\tt push} ~S_i.{\tt pop}$ \hfill (reverse the stacks)~~
    \ENDWHILE
    \STATE $(\cdot, \xi_i) := R_i.{\tt pop}$
    \STATE $\varrho := \varrho + a_{\xi_i}$
  \ENDFOR
  \STATE $\eta := 0, ~\varrho' =  0$
  \STATE $g := \varrho$ \hfill (value of $\sup \partial \,\Phi(0)$)
  \WHILE{$g < 0$}
    \STATE $\varrho' := \varrho$
    \IF{$\forall i : R_i.{\tt empty}$}
      \STATE $\eta := \infty$ \hfill (no more subdifferentiable points)
      \STATE {\bf break}
    \ENDIF
    \STATE $\Ical := \argmin_{1 \le i\le n}  ~\eta'
    : ~(\eta', \cdot) =  R_i.{\tt top}$ \hfill (find the next subdifferentiable point)
    \STATE $\varrho := \varrho - \sum_{i \in \Ical}a_{\xi_{i}}$
    \STATE $\Xi := \{\xi_{i}: (\eta, \xi_{i}\!) := R_{i}.{\tt
      pop},~~ i \in \Ical\}$
    \STATE $\varrho := \varrho + \sum_{\xi_{i} \in \Xi}a_{\xi_{i}}$
    \STATE $g := \varrho \!+ \eta \,h$ \hfill (value of $\sup\partial \Phi(\eta)$)~~ 
  \ENDWHILE
\RETURN{$\min(\eta, \, -\varrho'\!/h)$}
\end{algorithmic}
\end{algorithm}

Since $\Phi(\eta)$ is convex, any point $\eta < \eta^*$ cannot have
a non-negative subgradient.\footnote{If $\Phi(\eta)$ has a flat optimal
  region, we define $\eta^*$ to be the infimum of that region.}  The
first subdifferentiable point $\check{\eta} \geq \eta^*$ therefore
obeys
\begin{align} 
  \nonumber
  \check{\eta} & ~:=~ \min \eta \in \{\veta^{(i)}\!,\, i = 1, 2, \ldots, n\} : \eta \geq \eta^* \\
  & ~~=~ \min \eta \in \{\veta^{(i)}\!,\, i = 1, 2, \ldots, n\} : \sup
  \,\partial \,\Phi(\eta) \geq 0.
  \label{eq:eta-check}
\end{align}
We solve \eqref{eq:eta-check} via a simple linear search: Starting from
$\eta = 0$, we walk from one subdifferentiable point to the next until
$\sup \,\partial \,\Phi(\eta) \geq 0$. To perform this walk efficiently,
define a vector $\vpsi \in \NN^n$ of indices into the sorted vector
$\veta^{(i)}$ \emph{resp.}\ $\vxi^{(i)}$; initially $\vpsi :=
\vzero$, indicating that $(\forall i) ~\eta_0^{(i)} = 0$. Given the current index
vector $\vpsi$, the next subdifferentiable point is then
\begin{align}
  \eta' := \eta_{(\psi_{i'} + 1)}^{(i')}, \text{~~where~~} i' =
  \argmin_{1\le i \le n} \eta_{(\psi_i + 1)}^{(i)};
  \label{eq:etamerge}
\end{align}
the step is completed by incrementing $\psi_{i'}$, \ie $\psi_{i'} :=
\psi_{i'} + 1$ so as to remove $\eta_{\psi_{i'}}^{(i')}$ from future
consideration.\footnote{For ease of exposition, we assume $i'$ in
  \eqref{eq:etamerge} is unique, and deal with multiple choices of
  $i'$ in Algorithm~\ref{alg:exact-multi-ls}.} Note that computing the
$\argmin$ in \eqref{eq:etamerge} takes $O(\log n)$ time (\eg
using a priority queue). Inserting \eqref{eq:activeline} into
\eqref{eq:multi-oned} and differentiating, we find that
\begin{align}
  \sup \,\partial \,\Phi(\eta') ~=~ \lambda \inner{\vw}{\vp} + \lambda
  \eta' \|\vp\|^2 + \frac{1}{n} \sum_{i=1}^n a_{\xi_{\psi_i}^{(i)}}.
  \label{eq:thesup}
\end{align}
The key observation here is that after the initial calculation of
$\sup \partial \,\Phi(0) = \lambda \inner{\vw}{\vp} + \frac{1}{n}
\sum_{i=1}^n a_{\xi_{0}^{(i)}}$ for $\eta = 0$, the sum in
\eqref{eq:thesup} can be updated incrementally in constant time
through the addition of $a_{\xi_{\psi_{i'}}^{(i')}} -
a_{\xi_{(\psi_{i'} - 1)}^{(i')}}\!$ (\MLSUPDATESLOPE~of
Algorithm~\ref{alg:exact-multi-ls}).

Suppose we find $\check{\eta} = \eta_{\psi_{i'}}^{(i')}$ for some
$i'$.  We then know that the minimum $\eta^*$ is either equal to
$\check{\eta}$ (Figure~\ref{fig:linesearch}, left), or found within the
quadratic segment immediately to its left
(Figure~\ref{fig:linesearch}, right). We thus decrement $\psi_{i'}$
(\ie take one step back) so as to index the segment in question, set
the right-hand side of \eqref{eq:thesup} to zero, and solve for
$\eta'$ to obtain
\begin{align}
  \eta^* ~=~ \min \!\left( \check{\eta}, ~~\frac{\lambda
      \inner{\vw}{\vp} + \frac{1}{n}\sum^n_{i=1}
      a_{\xi_{\psi_i}^{(i)}}}{-\lambda \|\vp\|^2} \right).
  \label{eq:multietastar}
\end{align}
This only takes constant time: we have cached $\inner{\vw}{\vp}$ and
$\lambda \|\vp\|^2$, and the sum in \eqref{eq:multietastar} can be obtained
incrementally by adding $a_{\xi_{\psi_{i'}}^{(i')}} - a_{\xi_{(\psi_{i'} +
    1)}^{(i')}}\!$ to its last value in \eqref{eq:thesup}.

To locate $\check{\eta}$ we have to walk at most $O(n |\!\Zcal\!|)$
steps, each requiring $O(\log n)$ computation of $\argmin$ as in
\eqref{eq:etamerge}. Given $\check{\eta}$, the exact minimum $\eta^*$
can be obtained in $O(1)$. Including the preprocessing cost of $O(n
|\!\Zcal\!|\log|\!\Zcal\!|)$ (for invoking
Algorithm~\ref{alg:mincplf}), our exact multiclass line search
therefore takes $O(n |\!\Zcal\!|(\log n |\!\Zcal\!|))$ time in the
worst case.  Algorithm~\ref{alg:exact-multi-ls} provides an
implementation which instead of an index vector $\vpsi$ directly uses
the sorted stacks of subdifferentiable points and active lines
produced by Algorithm~\ref{alg:mincplf}. (The cost of reversing those
stacks in \MULTILSREVERSESTACK~of Algorithm~\ref{alg:exact-multi-ls}
can easily be avoided through the use of double-ended queues.)

\subsection{Multilabel Hinge Loss}
\label{sec:DealwithMult}

Recently, there has been interest in extending the concept of the hinge
loss to multilabel problems. Multilabel problems generalize the multiclass
setting  in that each training instance $\vx_i$ is associated with a set of
labels $\Zcal_i  \subseteq \Zcal$ \citep{CraSin03b}. For a uniform margin
of separation $\tau$, a hinge loss can be defined in this setting as follows:
\begin{align}
  \label{eq:cr-lossdef}
  l(\vx_i, \Zcal_i, \vw) \,:=\, \max[ \,0, ~\tau + \max_{z' \notin \Zcal_i}
  f(\vw, \vx_i, z') - \min_{z \in \Zcal_i} f(\vw, \vx_i, z)].
\end{align}
We can generalize this to a not necessarily uniform label loss
$\Delta(z', z) \geq 0$ as follows:
\begin{align}
  \label{eq:multilabel-lossdef}
  l(\vx_i, \Zcal_i, \vw) \,:=\, \max_{\substack{(z,z'): \:z \in \Zcal_i \\
  z' \notin\, \Zcal_i \!\backslash \{z\}}}
  [\Delta(z', z) + f(\vw, \vx_i, z') - f(\vw, \vx_i, z)],
\end{align}
where as before we require that $\Delta(z,z) = 0 ~\forall z \in
\Zcal$ so that by explicitly allowing $z' = z$ we can ensure that
\eqref{eq:multilabel-lossdef} remains non-negative. For a uniform
margin $\Delta(z', z) = \tau ~~\forall z' \neq z$ our multilabel hinge
loss \eqref{eq:multilabel-lossdef} reduces to the decoupled version
\eqref{eq:cr-lossdef}, which in turn reduces to the multiclass hinge
loss \eqref{eq:multi-lossdef} if $\Zcal_i := \{z_i\}$ for all $i$.

For a given $\vw$, let
\begin{align}
\label{eq:worst-labelpair}
  \Zcal_i^* := \argmax_{\substack{(z,z'): \:z \in \Zcal_i \\
  z' \notin\, \Zcal_i \!\backslash \{z\}}}
  [\Delta(z', z) + f(\vw, \vx_i, z') - f(\vw, \vx_i, z)]
\end{align}
be the set of worst label pairs (possibly more than one)
for the $i^{\rm th}$ training instance.
The subdifferential of the multilabel analogue of $L_2$-regularized
multiclass objective \eqref{eq:multi-obj} can then be written just as in
\eqref{eq:multi-gradient}, with coefficients 
\begin{align}
  \beta_{i,z} := \!\!\!\!\!\sum_{z': \,(z'\!,z) \in \Zcal_i^*} \!\!\!\!\!\gamma^{(i)}_{z'\!,z}
    ~- \!\!\!\!\!\sum_{z': \,(z,z') \in \Zcal_i^*} \!\!\!\!\!\gamma^{(i)}_{z,z'} \,,
  \text{~~where~~~} (\forall i) \!\!\sum_{(z,z') \in \Zcal_i^*}
  \!\!\!\gamma^{(i)}_{z,z'} \,=\, 1 \mbox{~~and~~} \gamma^{(i)}_{z,z'}
  \ge 0.
  \label{eq:multilabel-subdiff}
\end{align}

Now let $(z_i,z'_i) := \argmax_{(z,z') \in \Zcal_i^*} [\phi(\vx_i, z')
- \phi(\vx_i, z)]^{\top} \!\vp$ be a single steepest worst label pair
in direction $\vp$.
We obtain $\argsup_{\vg \in \partial J(\vw)}\vg^{\top}\vp$ for our
direction-finding procedure by picking, from the choices offered
by \eqref{eq:multilabel-subdiff},
$\gamma^{{}_{(i)}}_{z,z'} := \delta_{z,z_i} \delta_{z'\!, z'_i}$.

Finally, the line search we described in Section~\ref{sec:linesearch-multi}
for the multiclass hinge loss can be extended in a straightforward manner
to our multilabel setting. The only caveat is that now
$\varrho_i(\eta) := l(\vx_i, \Zcal_i, \vw + \eta \vp)$ must be written as
\begin{align}
  \label{eq:cr-lossdef-rw}
  \varrho_i(\eta) \,:= \!\!\max_{\substack{(z,z'): \:z \in \Zcal_i \\
  z' \notin\, \Zcal_i \!\backslash \{z\}}}
  [\underbrace{\Delta(z', z) + f(\vw, \vx_i, z') - f(\vw,
      \vx_i, z)}_{=: \:b^{(i)}_{z, z'}} + \,\eta \underbrace{(f(\vp,
      \vx_i, z') - f(\vp, \vx_i, z))}_{=: \:a^{(i)}_{z, z'}}]\,.
\end{align}
In the worst case, \eqref{eq:cr-lossdef-rw} could be the piecewise
maximum of $O(|\!\Zcal\!|^2)$ lines, thus increasing the overall
complexity of the line search. In practice, however, the set of true
labels $\Zcal_i$ is usually small, typically of size 2 or 3
\citep[\emph{cf.}][Figure 3]{CraSin03b}. As long as $\forall i:
|\!\Zcal_i\!| = O(1)$, our complexity estimates of
Section~\ref{sec:ExactLineSearch} still apply.

\section{Related Work}
\label{sec:related-work}

We discuss related work in two areas: nonsmooth convex optimization,
and the problem of segmenting the pointwise maximum of a set of
one-dimensional linear functions.
 
\subsection{Nonsmooth Convex Optimization}
\label{sec:other-nonsmooth-solvers}

There are four main approaches to nonsmooth convex
optimization: quasi-Newton methods, bundle methods, stochastic dual
methods, and smooth approximation. We discuss each of these briefly,
and compare and contrast our work with the state of the art.




\subsubsection{Nonsmooth Quasi-Newton Methods}

These methods try to find a descent quasi-Newton direction at every
iteration, and invoke a line search to minimize the one-dimensional
convex function along that direction.  We note that the
line search routines we describe in Sections
\ref{sec:subbfgs-hingeloss}--\ref{sec:subbfgs-multiloss} are
applicable to all such methods. An example of this class of algorithms
is the work of \citet{LukVlc99}, who propose an extension of BFGS to
nonsmooth convex problems. Their algorithm samples subgradients
around non-differentiable points in order to obtain a descent
direction. In many machine learning problems evaluating the objective
function and its (sub)gradient is very expensive, making such an
approach inefficient. In contrast, given a current iterate $\vw_{t}$,
our direction-finding routine (Algorithm~\ref{alg:find-descent-dir-cg})
samples subgradients from the set
$\partial J(\vw_{t})$ via the oracle.  Since this avoids the cost of
explicitly evaluating new (sub)gradients, it is computationally more
efficient.

Recently, \citet{AndGao07} introduced a variant of LBFGS, the
Orthant-Wise Limited-memory Quasi-Newton (OWL-QN) algorithm, suitable
for optimizing $L_{1}$-regularized log-linear models:
\begin{align}
  J(\vw) := \lambda \|\vw\|_1 + \underbrace{\frac{1}{n}\sum_{i=1}^{n}
  \ln(1+e^{-z_i \inner{\vw}{\vx_i}})}_{\text{logistic loss}},
  \label{eq:l1logistic}
\end{align}
where the logistic loss is smooth, but the regularizer is only
subdifferentiable at points where $\vw$ has zero
elements. From the optimization viewpoint this
objective is very similar to $L_{2}$-regularized hinge
loss; the direction finding and line search methods that we
discussed in Sections~\ref{sec:find-descent-dir} and \ref{sec:gen-ls},
respectively, can be applied to this problem with slight modifications.

OWL-QN is based on the observation that the $L_{1}$ regularizer is
linear within any given orthant. Therefore, it maintains an
approximation $\Bmat^{\text{ow}}$ to the inverse Hessian of the logistic
loss, and uses an efficient scheme to select orthants for
optimization. In fact, its success greatly depends on its
direction-finding subroutine, which demands a specially chosen
subgradient $\vg^{\text{ow}}$ \citep[Equation 4]{AndGao07} to produce
the quasi-Newton direction, $\vp^{\text{ow}} = \pi(\vp,
\vg^{\text{ow}})$, where $\vp:=-\Bmat^{\text{ow}}\vg^{\text{ow}}$ and
the projection $\pi$ returns a search direction by setting the
$i^{\text{th}}$ element of $\vp$ to zero whenever $p_i g^{\text{ow}}_i >
0$.  As shown in Section~\ref{sec:l1loss}, the direction-finding
subroutine of OWL-QN can be replaced by our
Algorithm~\ref{alg:find-descent-dir-cg}, which makes OWL-QN more
robust to the choice of subgradients.

\subsubsection{Bundle Methods}

Bundle method solvers \citep{HirLem93} use past (sub)gradients to build
a model of the objective function.  The (sub)gradients are used to
lower-bound the objective by a piecewise linear function which is
minimized to obtain the next iterate. This fundamentally differs from
the BFGS approach of using past gradients to approximate the (inverse)
Hessian, hence building a quadratic model of the objective function.

Bundle methods have recently been adapted to the machine learning context,
where they are known as SVMStruct \citep{TsoJoaHofAlt05}
\emph{resp.}\ BMRM \citep{SmoVisLe07}. One notable feature of
these variants is that they do not employ a line search. This is
justified by noting that a line search involves computing the value of
the objective function multiple times, a potentially expensive
operation in machine learning applications. 

\citet{FraSon08} speed up the convergence of SVMStruct for
$L_{2}$-regularized binary hinge loss. The main idea of their optimized
cutting plane algorithm, OCAS, is to perform a line search along the
line connecting two successive iterates of a bundle method
solver. Although developed independently, their line search is very
similar to the method we describe in Section~\ref{sec:exact-ls}.

\subsubsection{Stochastic Dual Methods}
\label{sec:DualMethods}

Distinct from the above two classes of primal algorithms are methods
which work in the dual domain.
A prominent member of this class is the LaRank algorithm
of \citet{BorBotGalWes07}, which achieves state-of-the-art results on
multiclass classification problems. While dual algorithms are very
competitive on clean datasets, they tend to be slow when given noisy
data. 


\subsubsection{Smooth Approximation}
\label{sec:ApprwithSmooth}

Another possible way to bypass the complications caused by the
nonsmoothness of an objective function is to work on a smooth
approximation instead\,---\,see for instance the recent
work of \citet{Nesterov05} and \citet{Nemirovski05}. Some machine
learning applications have also been pursued along these lines
\citep{Chapelle06, ZhaOle01}. Although this approach can
be effective, it is unclear how to build a smooth approximation in
general. Furthermore, smooth approximations often sacrifice
dual sparsity, which often leads to better generalization performance
on the test data, and also may be needed to prove generalization
bounds.

\subsection{Segmenting the Pointwise Maximum of 1-D Linear Functions}

The problem of computing the line segments that comprise the pointwise
maximum of a given set of line segments has received attention in the area
of computational geometry; see \citet{AgaSha00} for a survey.
\citet{Hershberger89} for instance proposed a divide-and-conquer
algorithm for this problem with the same time complexity as our
Algorithm~\ref{alg:mincplf}. The \citet{Hershberger89} algorithm solves
a slightly harder problem\,---\,his function is the pointwise maximum of line
segments, as opposed to our lines\,---\,but our algorithm is
conceptually simpler and easier to implement.

A similar problem has also been studied under the banner of kinetic data
structures by \citet{Basch99}, who proposed a heap-based algorithm for
this problem and proved a worst-case $O(r \log^{2} r)$ bound, where $r$
is the number of line segments. \citet{Basch99} also claims that the lower
bound is $O(r \log r)$; our Algorithm~\ref{alg:mincplf} achieves this bound.

\section{Experiments}
\label{sec:results}

We evaluated the performance of our subLBFGS algorithm with, and
compared it to other state-of-the-art nonsmooth optimization methods
on $L_2$-regularized binary, multiclass, and multilabel hinge loss
minimization problems. We also compared OWL-QN with a variant that
uses our direction-finding routine on $L_1$-regularized logistic loss
minimization tasks. On strictly convex problems such as these every
convergent optimizer will reach the same solution; comparing
generalisation performance is therefore pointless. Hence we
concentrate on empirically evaluating the convergence behavior
(objective function value \emph{vs.} CPU seconds). All experiments
were carried out on a Linux machine with dual 2.4\,GHz Intel Core 2
processors and 4\,GB of RAM.

In all experiments the regularization parameter was chosen from the
set $10^{\{-6, -5, \cdots, -1\}}$ so as to achieve the highest
prediction accuracy on the test dataset, while convergence behavior
(objective function value \emph{vs.} CPU seconds) is reported on the
training dataset. To see the influence of the regularization
parameter $\lambda$, we also compared the time required by each
algorithm to reduce the objective function value to within $2\%$ of
the optimal value.\footnote{For $L_1$-regularized logistic loss
  minimization, the ``optimal'' value was the final objective function
  value achieved by the OWL-QN$^*$ algorithm (\cf
  Section~\ref{sec:l1loss}). In all other experiments, it was found by
  running subLBFGS for $10^4$ seconds, or until its relative
  improvement over 5 iterations was less than $10^{-8}$.} For all
algorithms the initial iterate $\vw_0$ was set to $\vzero$. Open
source C++ code implementing our algorithms and experiments is
available for download from
{\small\url{http://www.cs.adelaide.edu.au/~jinyu/Code/nonsmoothOpt.tar.gz}}.

The subgradient for the construction of the subLBFGS search direction
(\cf \SUBBFGSGRAD~of Algorithm~\ref{alg:subbfgs}) was chosen
arbitrarily from the subdifferential. For the binary hinge loss
minimization (Section~\ref{sec:binary-results}), for instance, we
picked an arbitrary subgradient by randomly setting the coefficient
$\beta_i~\forall i \in \Mcal$ in \eqref{eq:subdifferential} to either
0 or 1.

\subsection{Convergence Tolerance of the Direction-Finding Procedure}
\label{sec:dir-epsilon}

\begin{figure}
  \begin{tabular}{@{$\!\!\!$}c@{}c@{}c}
    \includegraphics[width=0.34\linewidth]{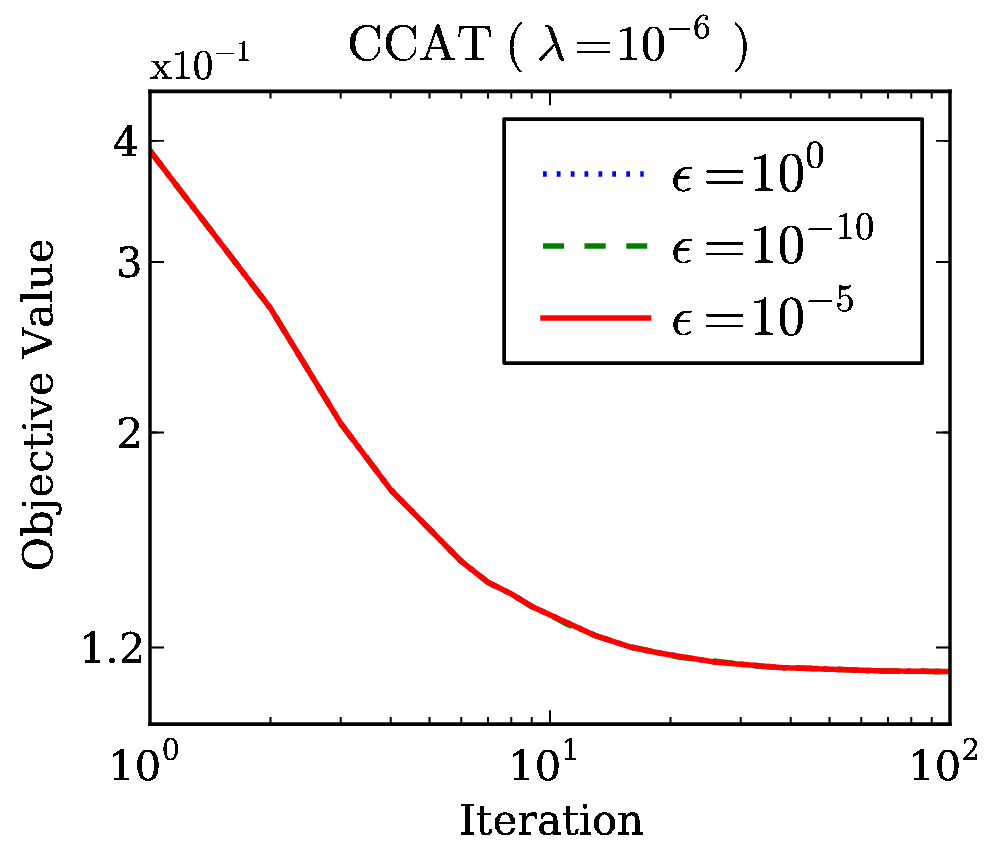} &
    \includegraphics[width=0.34\linewidth]{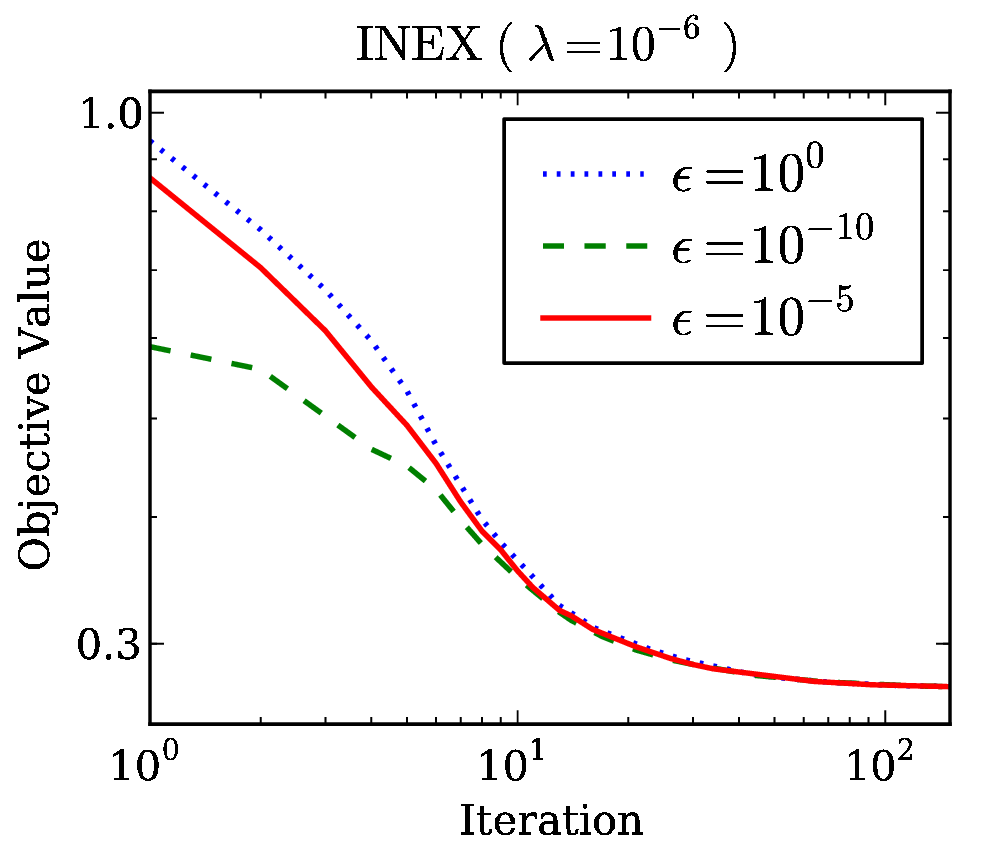} &
    \includegraphics[width=0.34\linewidth]{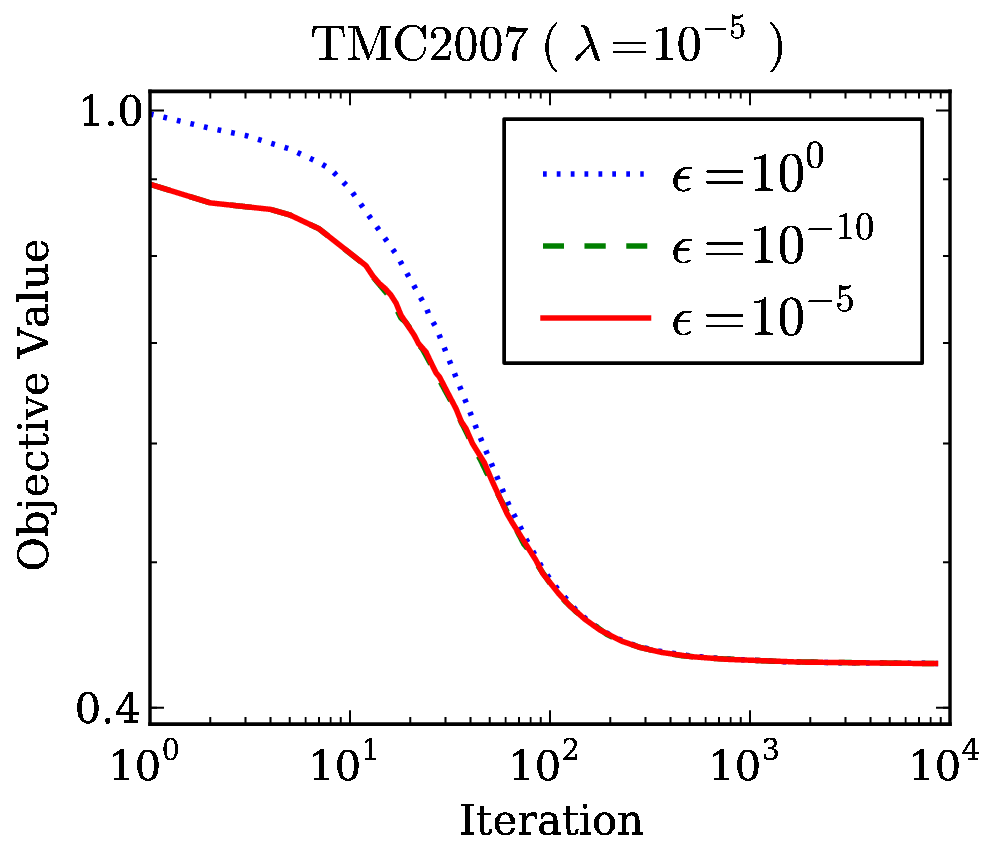} \\
    \includegraphics[width=0.34\linewidth]{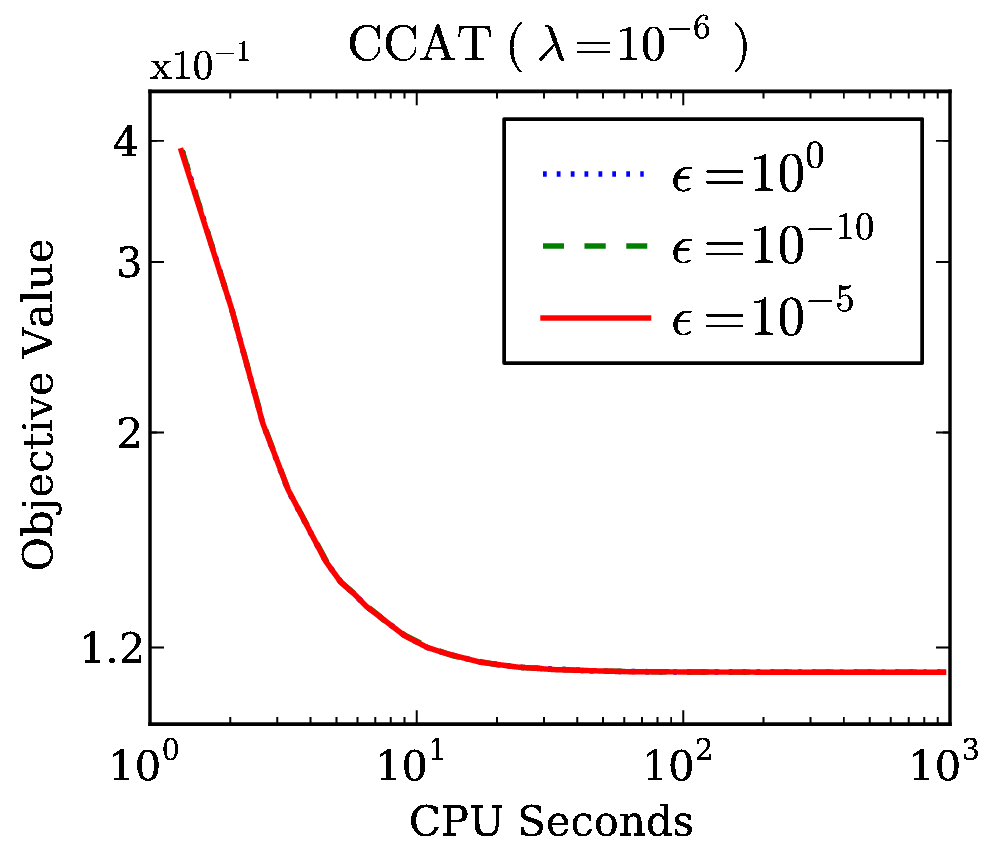} &
    \includegraphics[width=0.34\linewidth]{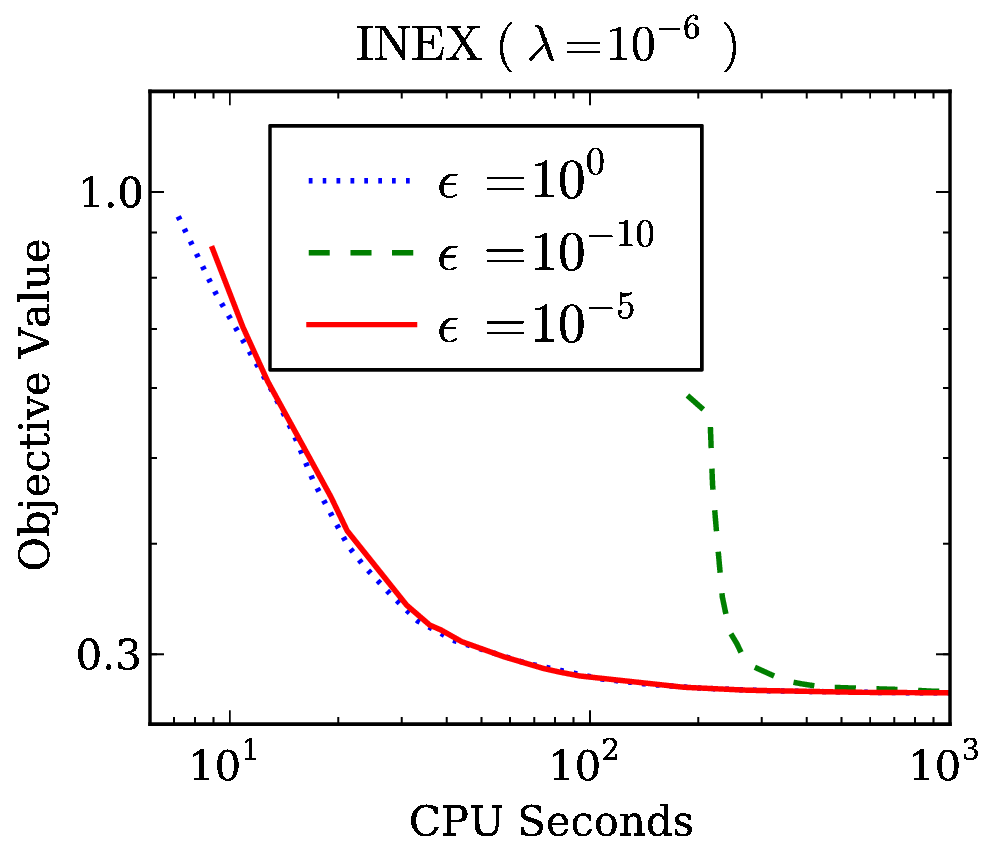} &
    \includegraphics[width=0.34\linewidth]{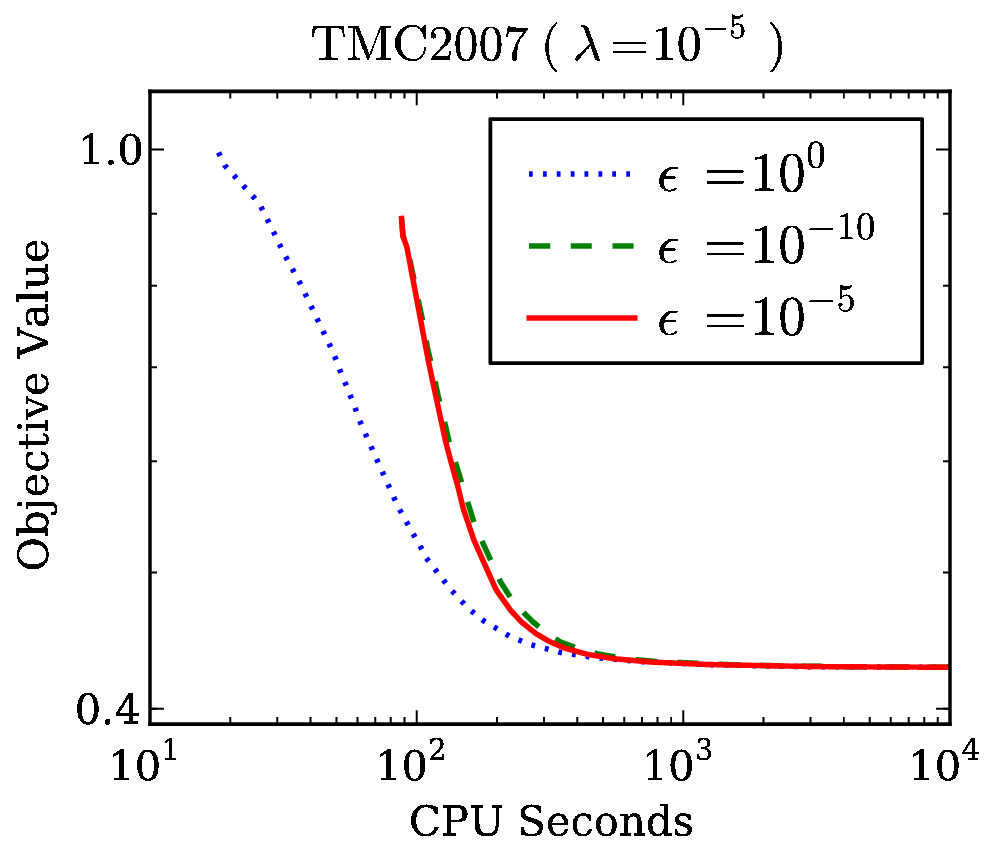} \\
  \end{tabular}
  \caption{Performance of subLBFGS with varying direction-finding
    tolerance $\epsilon$ in terms of objective
    function value \emph{vs.}\ number of iterations (top row)
    \emph{resp.}\ CPU seconds (bottom row) on sample $L_2$-regularized
    risk minimization problems with binary (left), multiclass
    (center), and multilabel (right) hinge losses.}
  \label{fig:dir-tolerance}
\end{figure}

The convergence tolerance $\epsilon$ of
Algorithm~\ref{alg:find-descent-dir-cg} controls the precision of the
solution to the direction-finding problem \eqref{eq:min-gen-model}:
lower tolerance may yield a better search direction. Figure
\ref{fig:dir-tolerance} (left) shows that on binary classification
problems, subLBFGS is not sensitive to the choice of $\epsilon$ (\ie
the quality of the search direction). This is due to the fact that
$\partial J(\vw)$ as defined in \eqref{eq:subdifferential} is usually
dominated by its constant component $\bar{\vw}$; search directions
that correspond to different choices of
$\epsilon$ therefore can not differ too much from each other. In the
case of multiclass and multilabel classification, where the structure
of $\partial J(\vw)$ is more complicated, we can see from Figure
\ref{fig:dir-tolerance} (top center and right) that a better
search direction can lead to faster convergence in terms of iteration
numbers. However, this is achieved at the cost of more CPU time spent
in the direction-finding routine. As shown in Figure
\ref{fig:dir-tolerance} (bottom center and right), extensively
optimizing the search direction actually slows down convergence in
terms of CPU seconds. We therefore used an intermediate value of
$\epsilon = 10^{-5}$ for all our experiments, except that for
multiclass and multilabel classification problems we relaxed the
tolerance to $1.0$ at the initial iterate $\vw = \vzero$, where the
direction-finding oracle $\arg\sup_{\vg \in \partial J(\vzero)}
\vg^{\top}\vp$ is expensive to compute, due to the large number of
extreme points in $\partial J(\vzero)$.

\subsection{Size of  SubLBFGS Buffer}
\label{sec:memory}

\begin{figure}
  \begin{tabular}{@{$\!\!\!$}c@{}c@{}c}
    \includegraphics[width=0.34\linewidth]{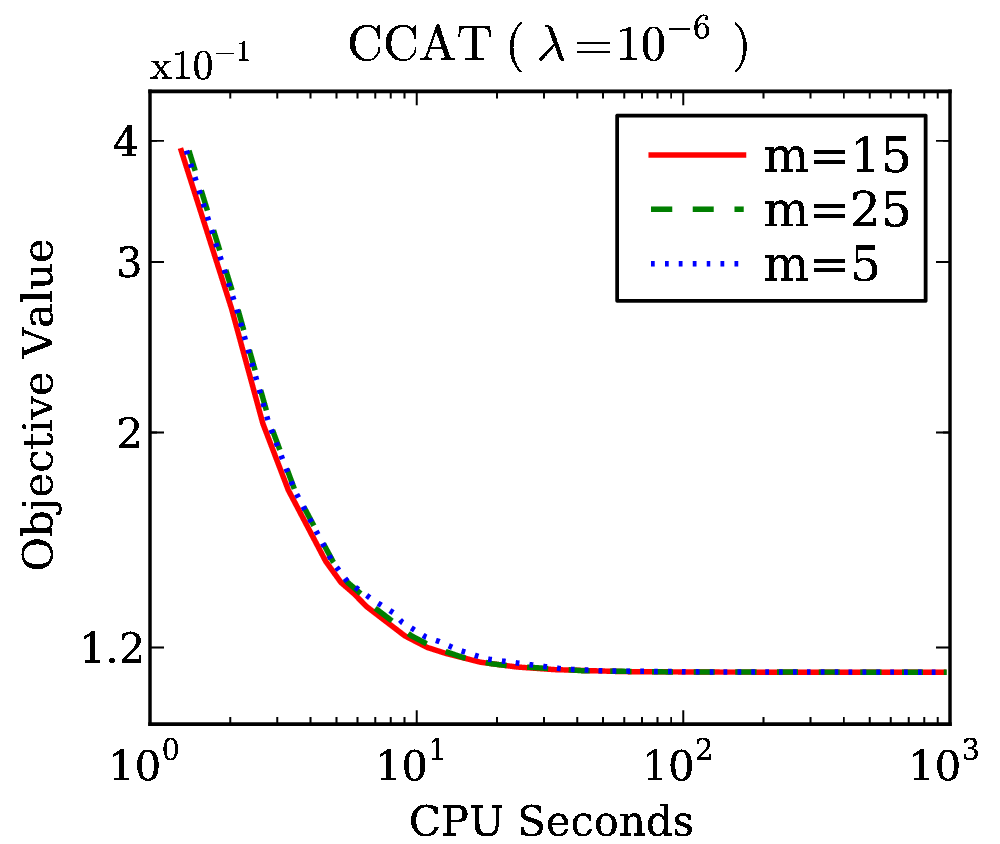} &
    \includegraphics[width=0.34\linewidth]{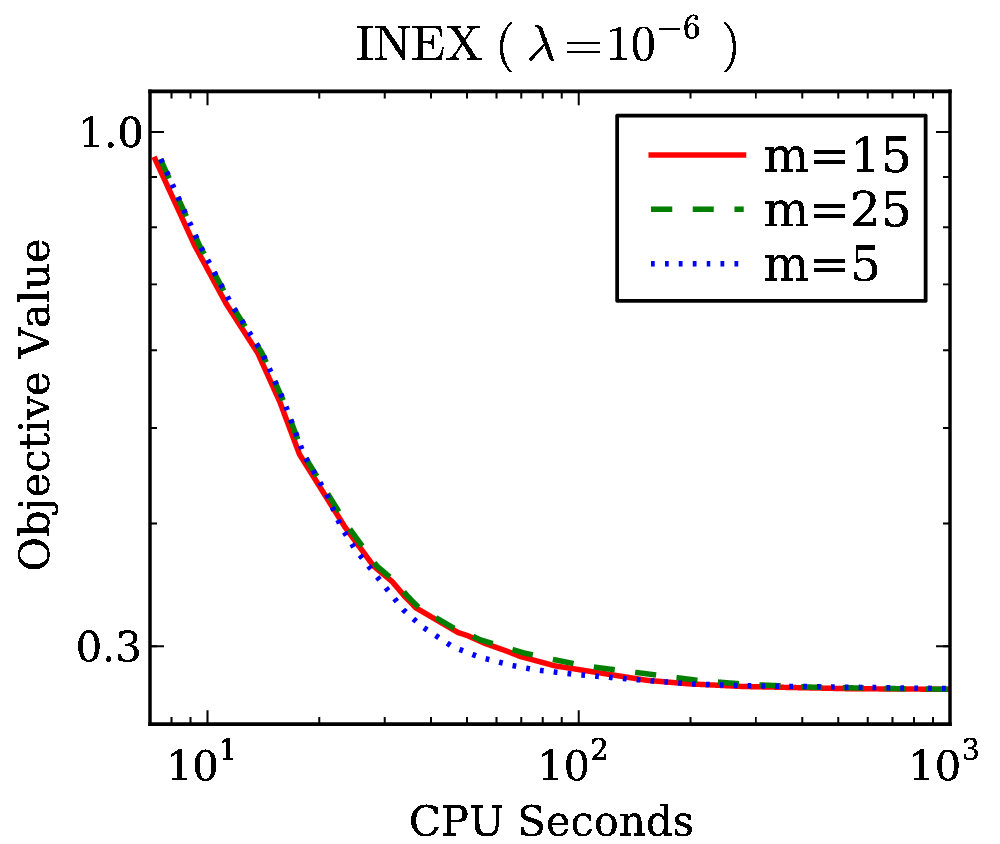} &
    \includegraphics[width=0.34\linewidth]{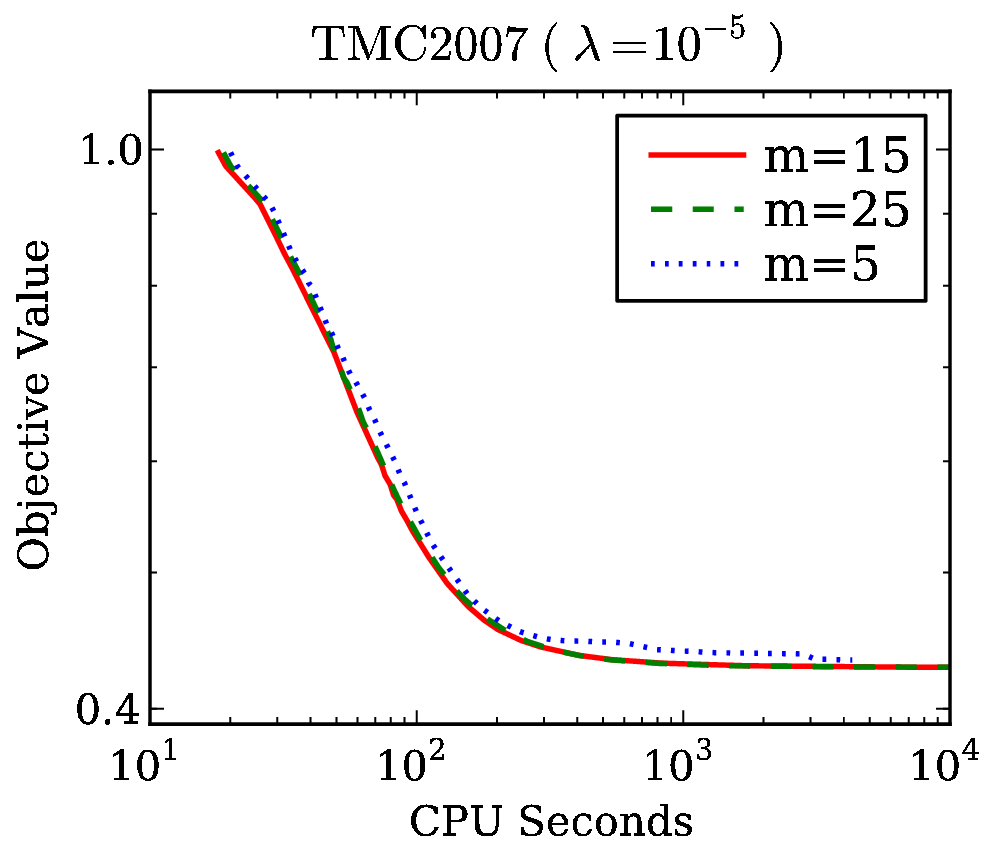} \\
  \end{tabular}
  \caption{Performance of subLBFGS with varying buffer size on sample
    $L_2$-regularized risk minimization problems with binary
    (left), multiclass (center), and multilabel hinge losses (right).}
  \label{fig:buffersize}
\end{figure}

The size $m$ of the subLBFGS buffer determines the number of parameter
and gradient displacement vectors $\vs_t$ and $\vy_t$ used in the
construction of the quasi-Newton
direction. Figure~\ref{fig:buffersize} shows that the performance of
subLBFGS is not sensitive to the particular value of $m$ within the
range $5 \le m \le 25$.  We therefore simply set $m=15$ \emph{a
  priori} for all subsequent experiments; this is a typical value for
LBFGS \citep{NocWri99}.

\subsection{$L_2$-Regularized Binary Hinge Loss}
\label{sec:binary-results}

\begin{table}[b]
  \centering
  \caption{The binary datasets used in our experiments of Sections~\ref{sec:motivation},
  \ref{sec:binary-results}, and \ref{sec:l1loss}.}
  \vspace{2ex}
  \begin{tabular}{|l||r|r|r|r|}
    \hline
    Dataset & Train/Test Set Size & Dimensionality & Sparsity \\
    \hline\hline
    Covertype\rule{0pt}{2.5ex} & 522911/58101~ & 54~ & 77.8\%~ \\
    CCAT & 781265/23149~ & 47236~ &  99.8\%~ \\
    Astro-physics & 29882/32487~ & 99757~  & 99.9\%~ \\
    MNIST-binary & 60000/10000~ & 780~ & 80.8\%~  \\
    Adult9 & 32561/16281~ & 123~ & 88.7\%~  \\
    Real-sim & 57763/14438~ & 20958~ & 99.8\%~  \\
    Leukemia & 38/34~ & 7129~ & 00.0\%~ \\ 
    \hline
  \end{tabular}
  \label{tab:datasets}
\end{table}

For our first set of experiments, we applied subLBFGS with exact line
search (Algorithm~\ref{alg:exact-ls}) to the task of $L_2$-regularized
binary hinge loss minimization. Our control methods are the bundle
method solver BMRM \citep{TeoVisSmoLe09} and the optimized cutting plane
algorithm OCAS \citep{FraSon08},\footnote{The source code
  of OCAS (version 0.6.0) was obtained from \url{http://www.shogun-toolbox.org}.}
both of which were shown
to perform competitively on this task. SVMStruct \citep{TsoJoaHofAlt05} is
another well-known bundle method solver that is widely used in the
machine learning community.  For $L_{2}$-regularized optimization
problems BMRM is identical to SVMStruct, hence
we omit comparisons with SVMStruct.

\begin{table}
  \centering
  \caption{Regularization parameter $\lambda$ and overall number $k$ of
    direction-finding iterations in our experiments of Sections~\ref{sec:binary-results}
    and \ref{sec:l1loss}, respectively.}
  \vspace{2ex}
  \begin{tabular}{|l||r|r|r||c|c|}
    \hline
    & \multicolumn{3}{c||}{$L_1$-reg.\ logistic loss}
    & \multicolumn{2}{c|}{$L_2$-reg.\ binary loss} \\
    Dataset &
    $\lambda_{L_{1}}$ & $k_{L_1}$ & $k_{L_1 r}$ &
    ~~~~$\lambda_{L_2}$~~~~ & $k_{L_2}$ \\
    \hline\hline
    Covertype\rule{0pt}{2.5ex} & $10^{-5}$ &1 & 2
    & $10^{-6}$  & 0 
    \\
    CCAT & $10^{-6}$  & 284 & 406 &
    $10^{-6}$  & 0 
    \\
    Astro-physics & $10^{-5}$ & 1702 & 1902
    & $10^{-4}$  & 0 
    \\
    MNIST-binary & $10^{-4}$  & 55 & 77
    & $10^{-6}$ & 0 
    \\
    Adult9 & $10^{-4}$ & 2 & 6
    & $10^{-5}$ & 1 
    \\
    Real-sim & $10^{-6}$ & 1017 & 1274
    & $10^{-5}$ & 1  
    \\
    \hline
  \end{tabular}
  \label{tab:parameters}
\end{table}

\begin{figure}[b]
   \begin{tabular}{@{$\!\!\!$}c@{}c@{}c}
     \includegraphics[width=0.34\linewidth]{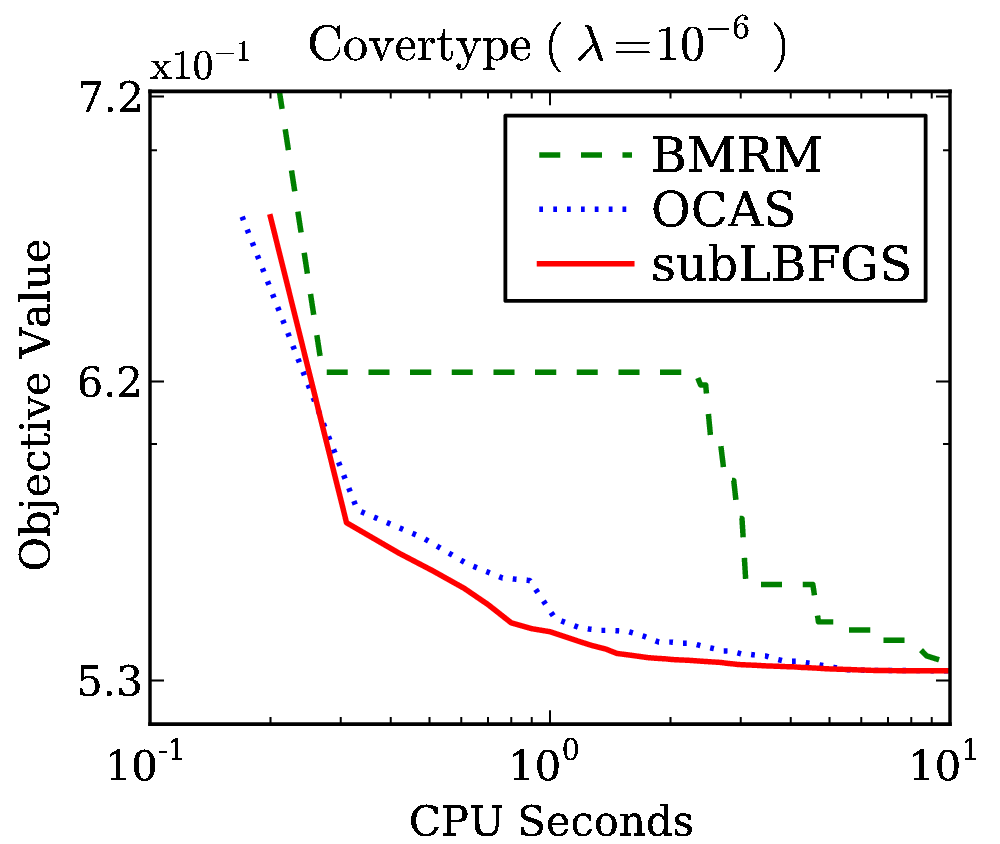} &
     \includegraphics[width=0.34\linewidth]{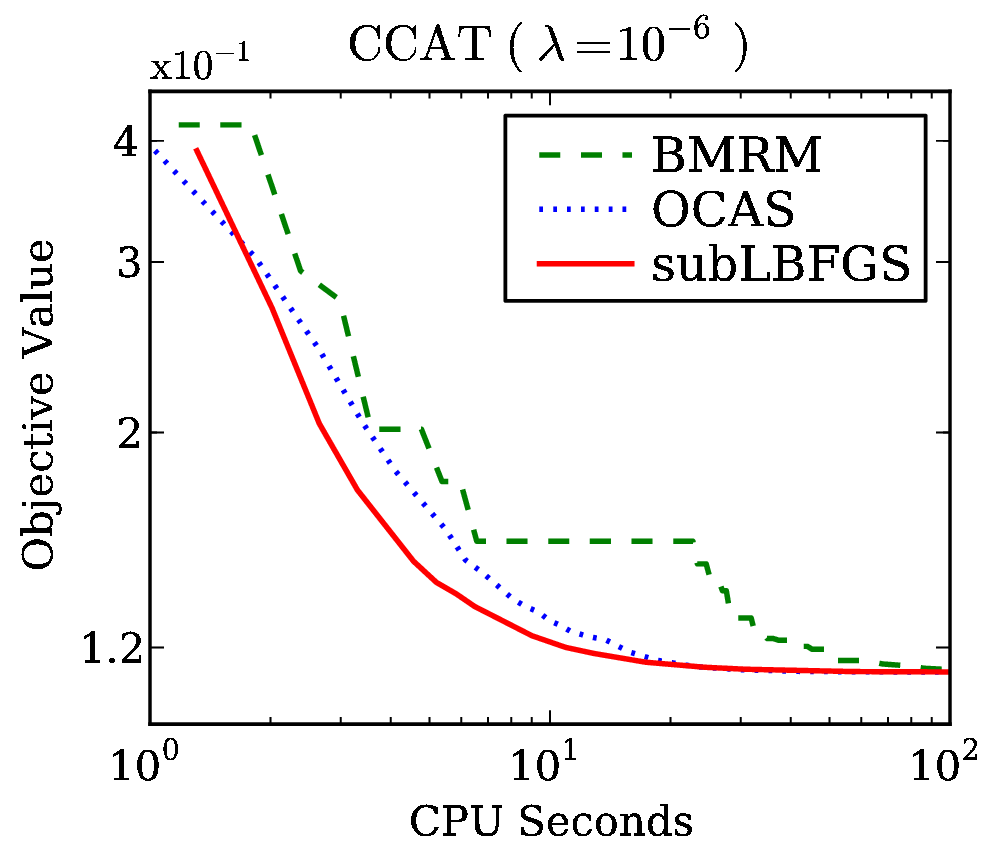} &
     \includegraphics[width=0.34\linewidth]{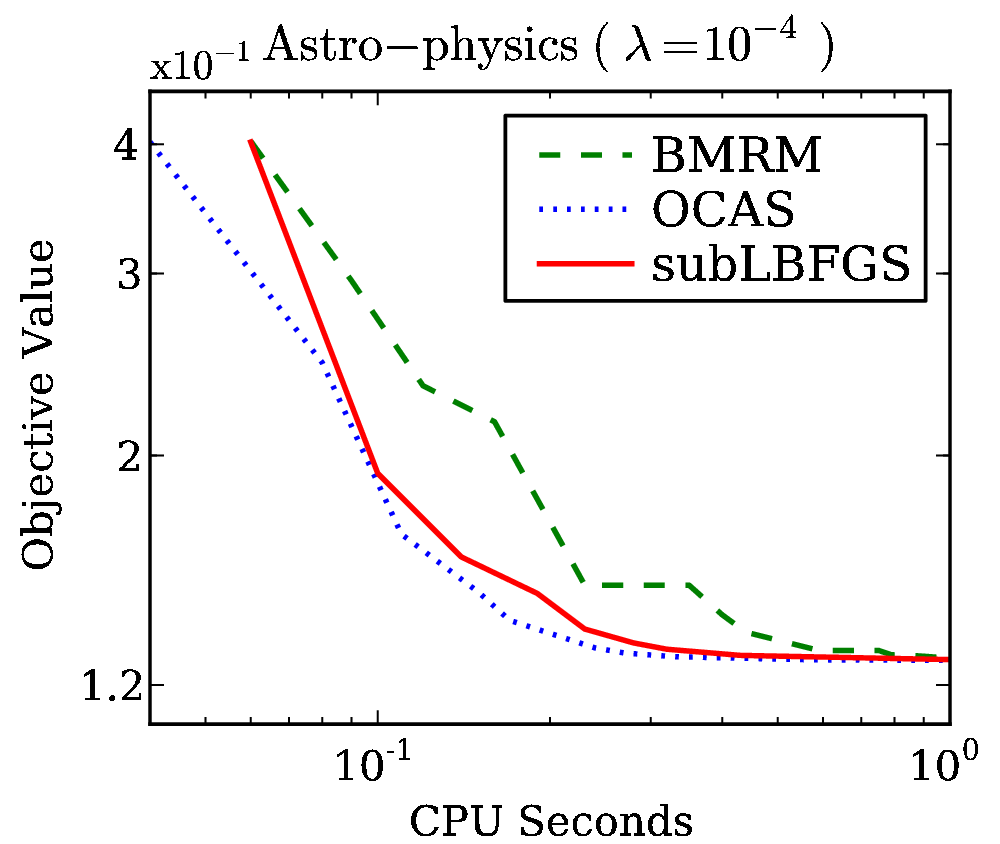} \\
     \includegraphics[width=0.34\linewidth]{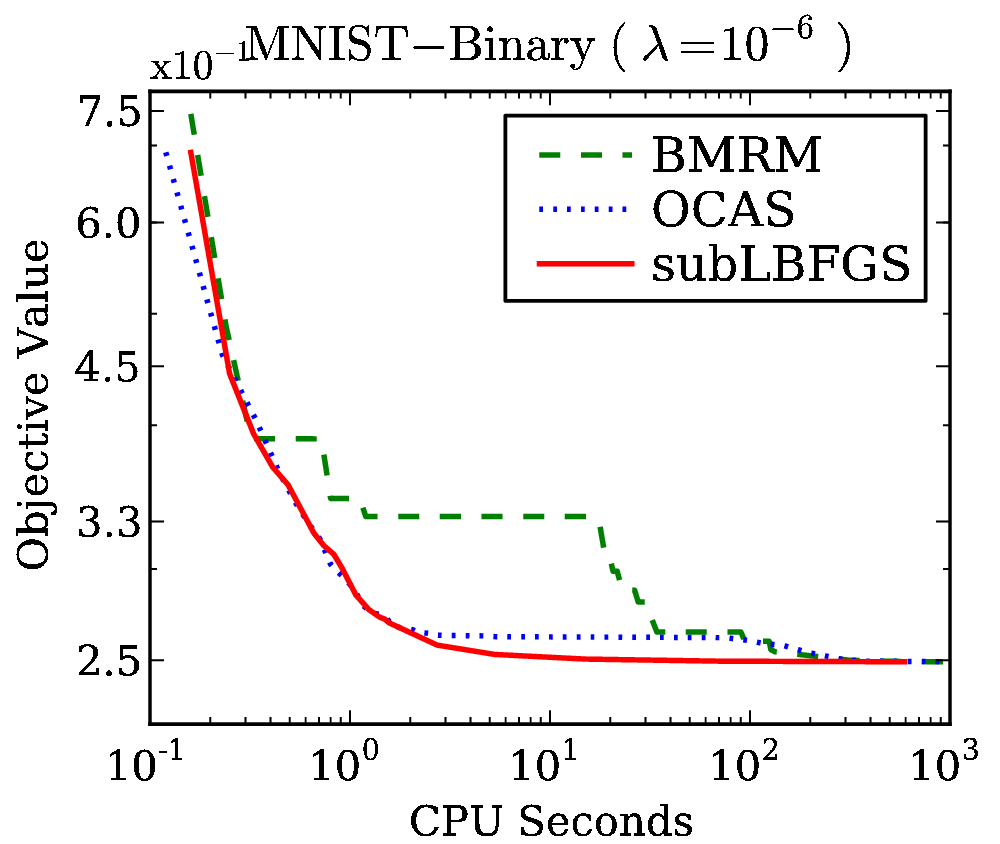} &
     \includegraphics[width=0.34\linewidth]{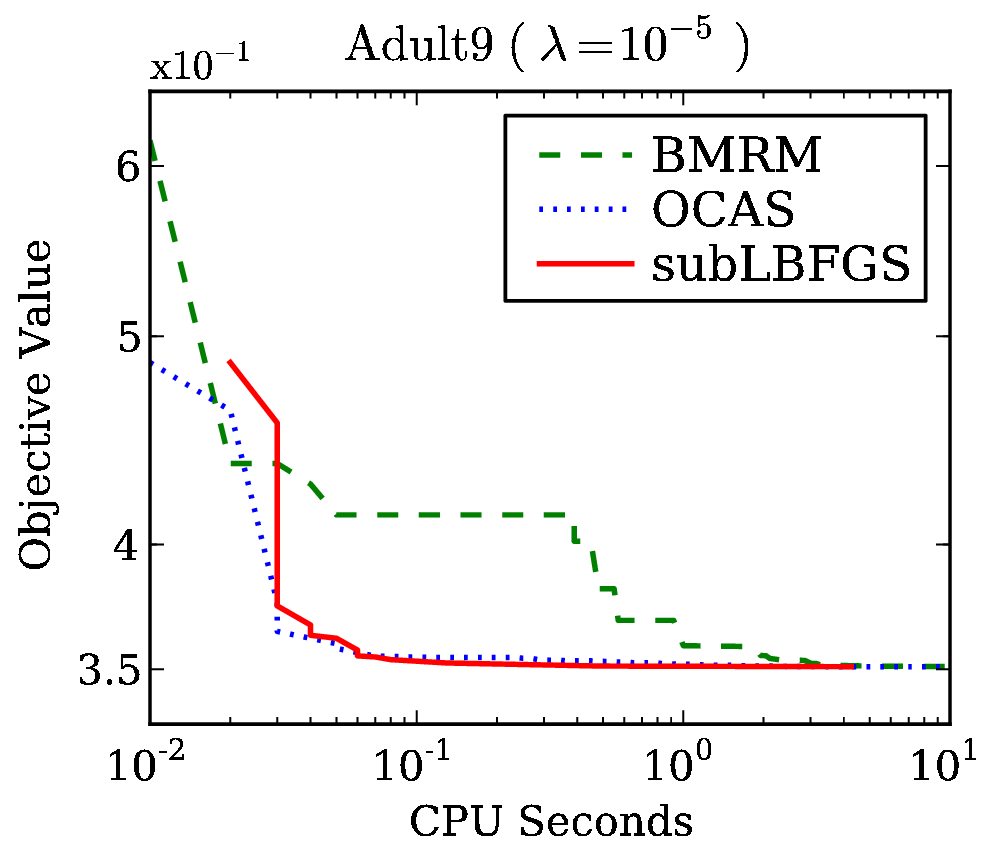} &
     \includegraphics[width=0.34\linewidth]{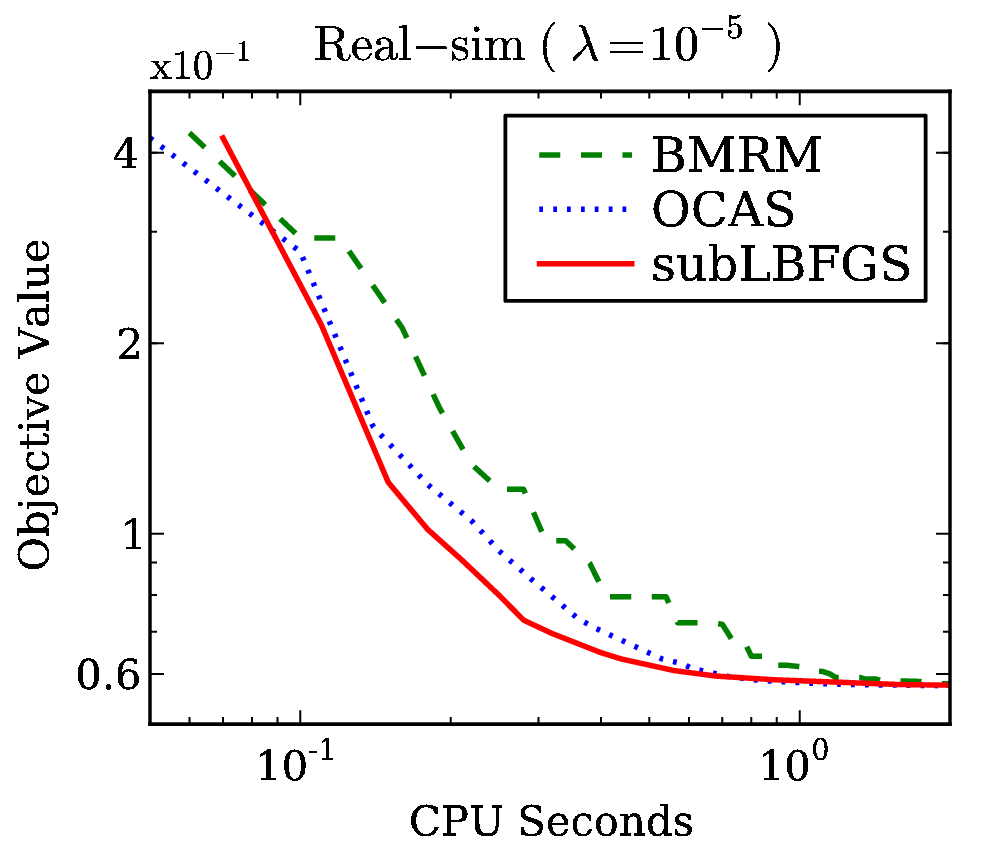} \\
   \end{tabular}
  \caption{Objective function value \emph{vs.}\ CPU seconds on $L_2$-regularized
   binary hinge loss minimization tasks. 
   }
 \label{fig:l2hinge}
\end{figure}

\begin{figure}
   \begin{tabular}{@{$\!\!\!$}c@{}c@{}c}
     \includegraphics[width=0.34\linewidth]{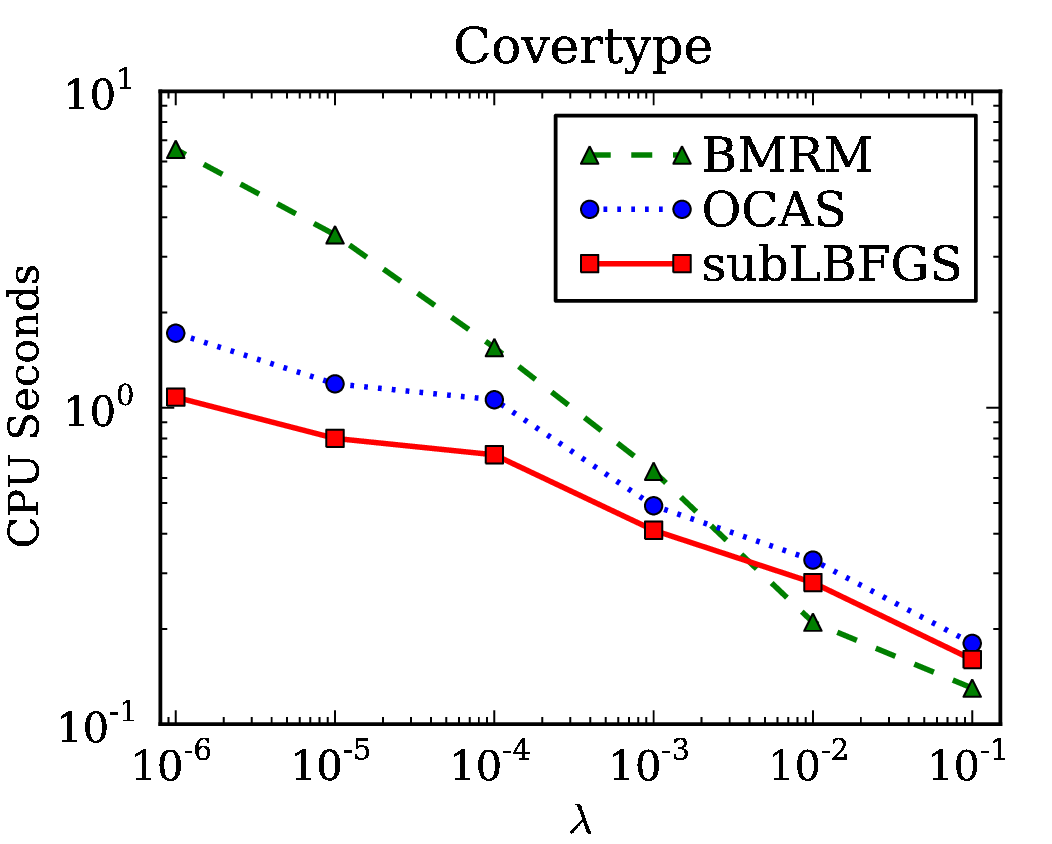} &
     \includegraphics[width=0.34\linewidth]{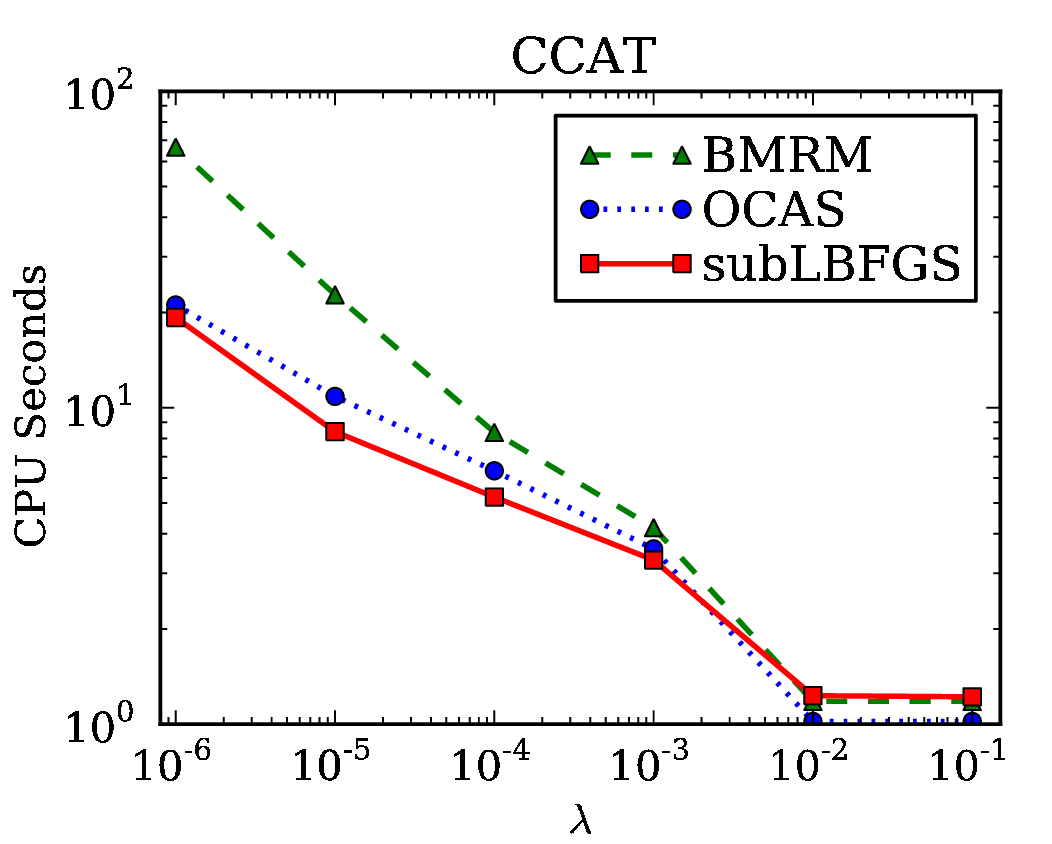} &
     \includegraphics[width=0.34\linewidth]{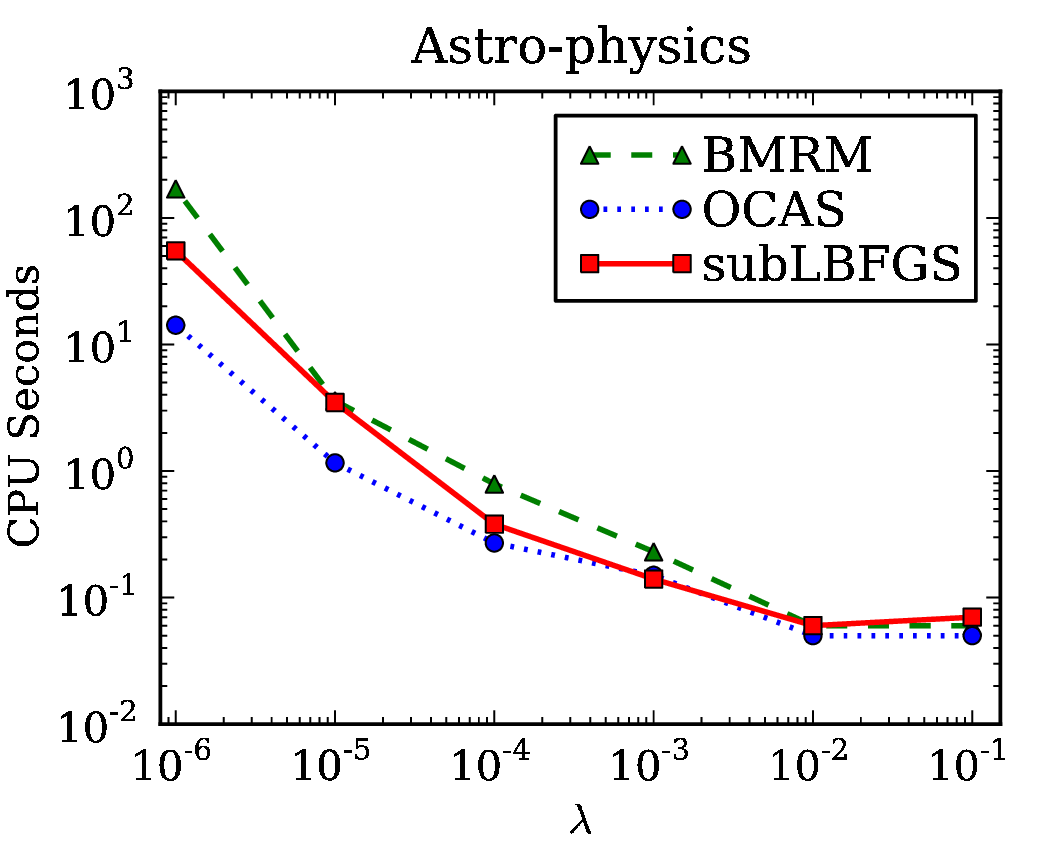} \\
     \includegraphics[width=0.34\linewidth]{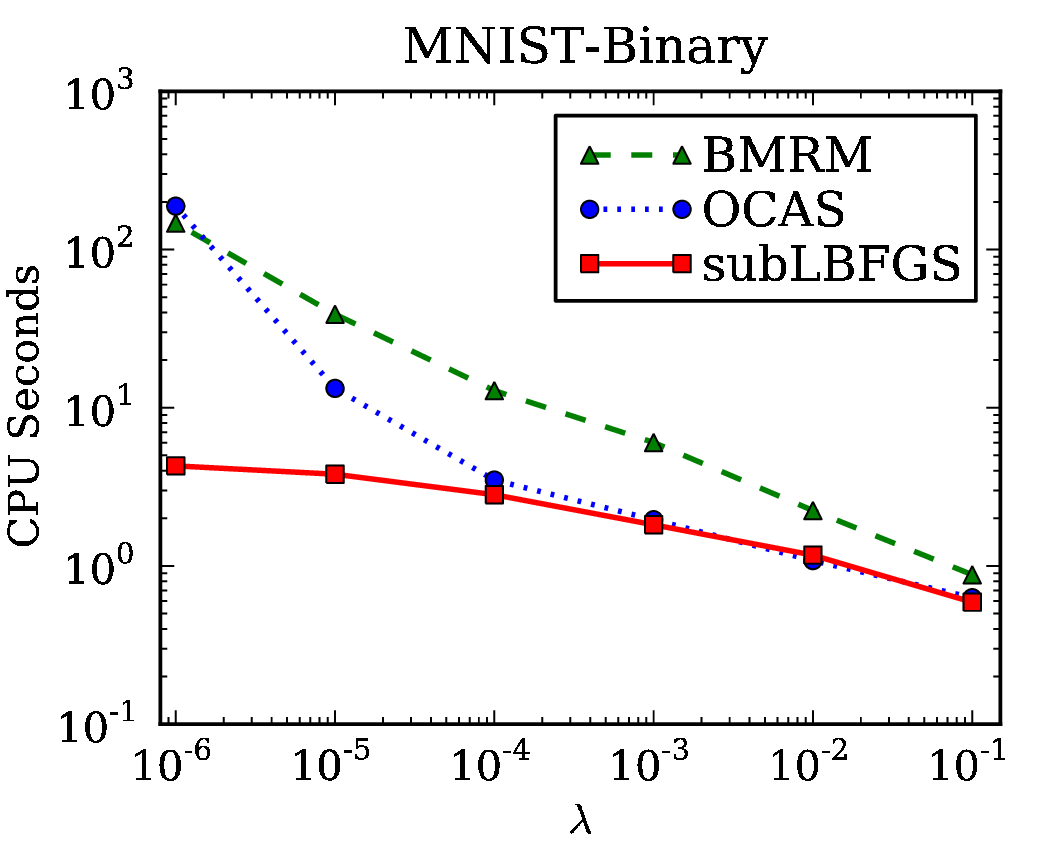} &
     \includegraphics[width=0.34\linewidth]{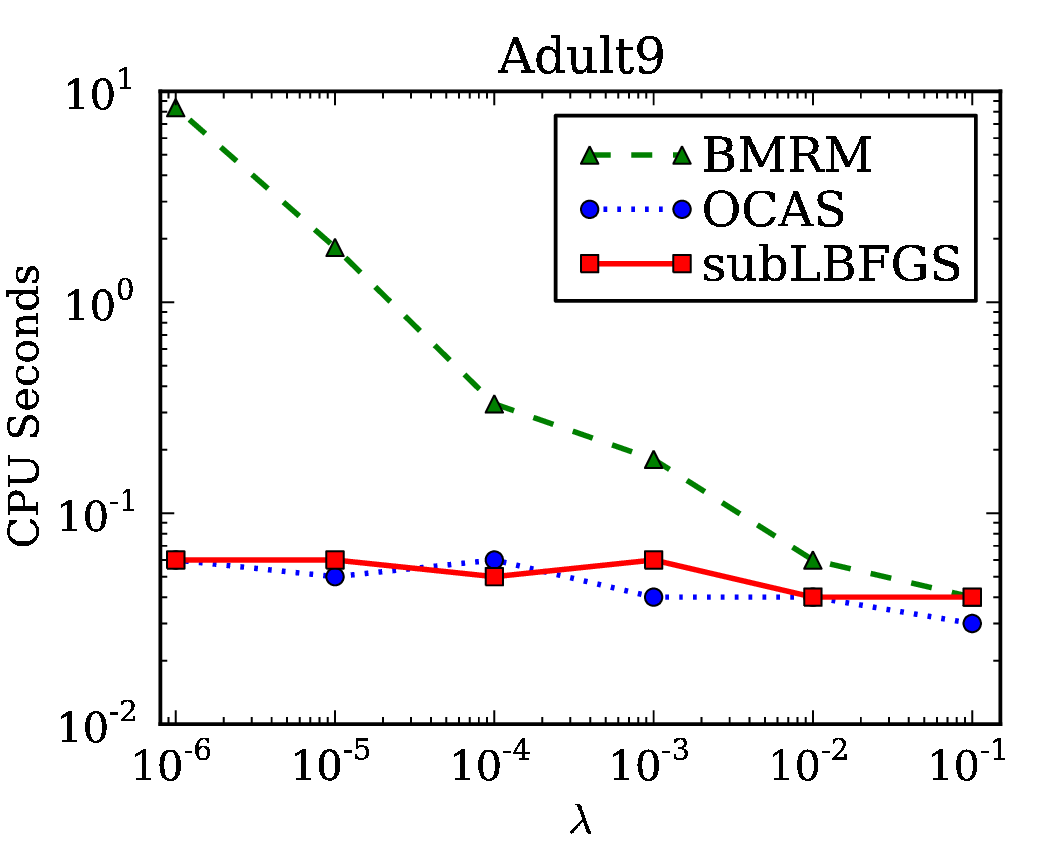} &
     \includegraphics[width=0.34\linewidth]{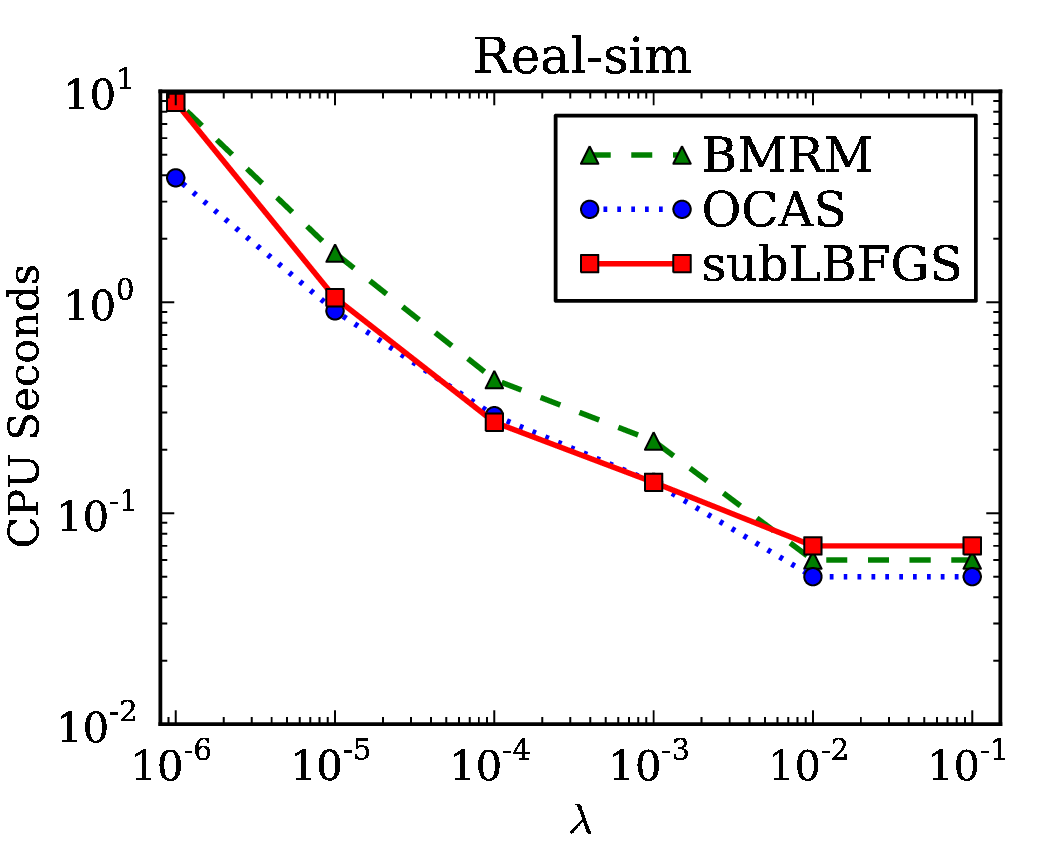} \\
   \end{tabular}
  \caption{Regularization parameter $\lambda \in \{10^{-6},\cdots,
    10^{-1}\}$ \emph{vs.}\ CPU seconds taken to reduce the objective
    function to within 2\% of the optimal value on $L_2$-regularized
   binary hinge loss minimization tasks. 
   }
 \label{fig:l2hinge-lambdas}
\end{figure}

Table~\ref{tab:datasets} lists the six datasets we used:
The Covertype dataset of Blackard, Jock \& Dean,\footnote{%
\url{http://kdd.ics.uci.edu/databases/covertype/covertype.html}}
CCAT from the Reuters RCV1 collection,\footnote{%
\url{http://www.daviddlewis.com/resources/testcollections/rcv1}}
the Astro-physics dataset of abstracts of scientific papers from the
Physics ArXiv \citep{Joachims06}, the MNIST dataset of handwritten
digits\footnote{\url{http://yann.lecun.com/exdb/mnist}} with two
classes: even and odd digits, the Adult9 dataset of census income
data,\footnote{%
\url{http://www.csie.ntu.edu.tw/~cjlin/libsvmtools/datasets/binary.html}}
and the Real-sim dataset of real \emph{vs.}\ simulated data.\samefootnote{}
Table~\ref{tab:parameters} lists our parameter settings, and reports the
overall number $k_{L_{2}}$ of iterations through the direction-finding loop
(Lines 6--13 of Algorithm~\ref{alg:find-descent-dir-cg}) for each
dataset. The very small values of $k_{L_{2}}$ indicate that on these problems
subLBFGS only rarely needs to correct its initial guess of a descent direction.  

It can be seen from Figure~\ref{fig:l2hinge} that subLBFGS (solid)
reduces the value of the objective considerably faster than BMRM
(dashed). On the binary MNIST dataset, for instance, the objective function
value of subLBFGS after 10 CPU seconds is 25\% lower than
that of BMRM. In this set of experiments the performance of subLBFGS
and OCAS (dotted) is very similar.

Figure~\ref{fig:l2hinge-lambdas} shows that all algorithms generally
converge faster for larger values of the regularization constant
$\lambda$.  However, in most cases subLBFGS converges faster than BMRM
across a wide range of $\lambda$ values, exhibiting a speedup of up to
more than two orders of magnitude. SubLBFGS and OCAS show similar
performance here: for small values of $\lambda$, OCAS converges
slightly faster than subLBFGS on the Astro-physics and Real-sim
datasets but is outperformed by subLBFGS on the Covertype, CCAT, and
binary MNIST datasets.

\subsection{$L_1$-Regularized Logistic Loss}
\label{sec:l1loss}

To demonstrate the utility of our direction-finding routine
(Algorithm~\ref{alg:find-descent-dir-cg}) in its own right, we plugged
it into the OWL-QN algorithm \citep{AndGao07}\footnote{The source code
  of OWL-QN (original release) was obtained from Microsoft Research
  through \url{http://tinyurl.com/p774cx}.}  as an
alternative direction-finding method such that $\vp^{\text{ow}} =
{\tt descentDirection}(\vg^{\text{ow}}, \epsilon, k_{\text{max}})$,
and compared this variant (denoted OWL-QN*) with the original (\cf
Section~\ref{sec:other-nonsmooth-solvers}) on $L_1$-regularized
minimization of the logistic loss \eqref{eq:l1logistic}, on the same
datasets as in Section~\ref{sec:binary-results}.

An oracle that supplies $\argsup_{\vg\in\partial J(\vw)} \vg^{\top}\vp$
for this objective is easily constructed by noting that
\eqref{eq:l1logistic} is nonsmooth whenever at least one component of
the parameter vector $\vw$ is zero. Let $w_i = 0$ be such a component;
the corresponding component of the subdifferential $\partial \,\lambda
\|\vw\|_1$ of the $L_1$ regularizer is the interval
$[-\lambda,\lambda]$.  The supremum of $\vg^{\top}\vp$ is attained at
the interval boundary whose sign matches that of the corresponding
component of the direction vector $\vp$, \ie at $\lambda \sgn(p_i)$.

\begin{figure}
   \begin{tabular}{@{$\!\!\!$}c@{}c@{}c}
     \includegraphics[width=0.34\linewidth]{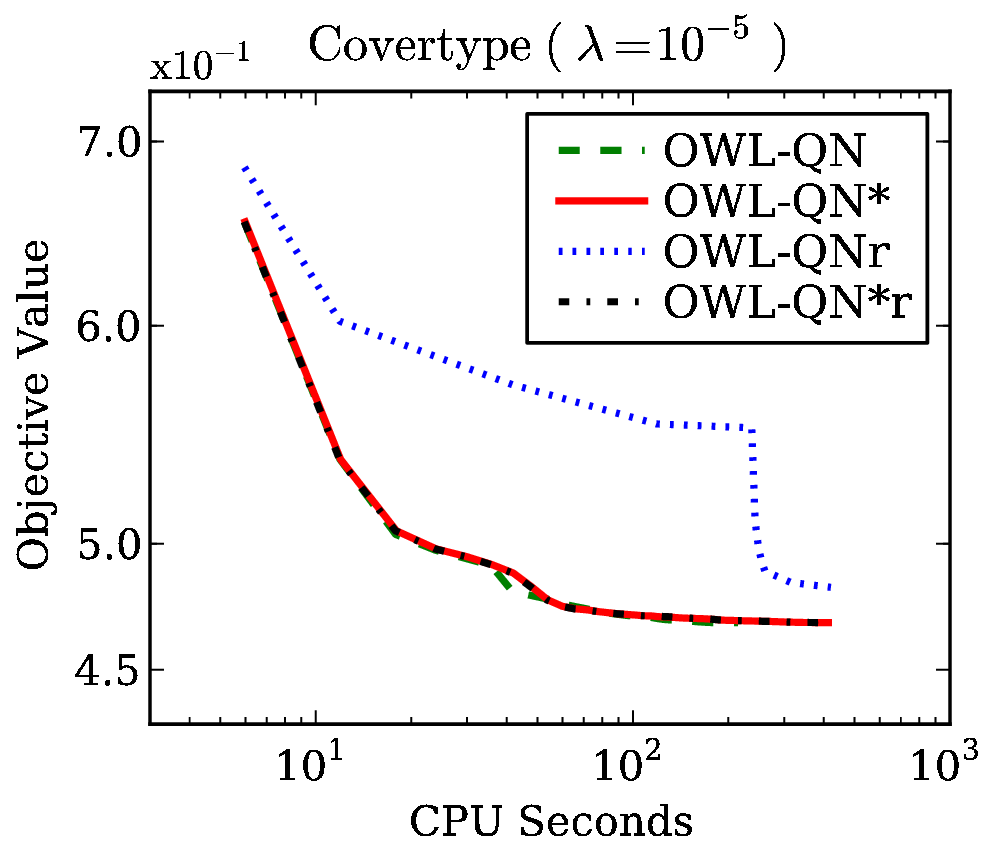} &
     \includegraphics[width=0.34\linewidth]{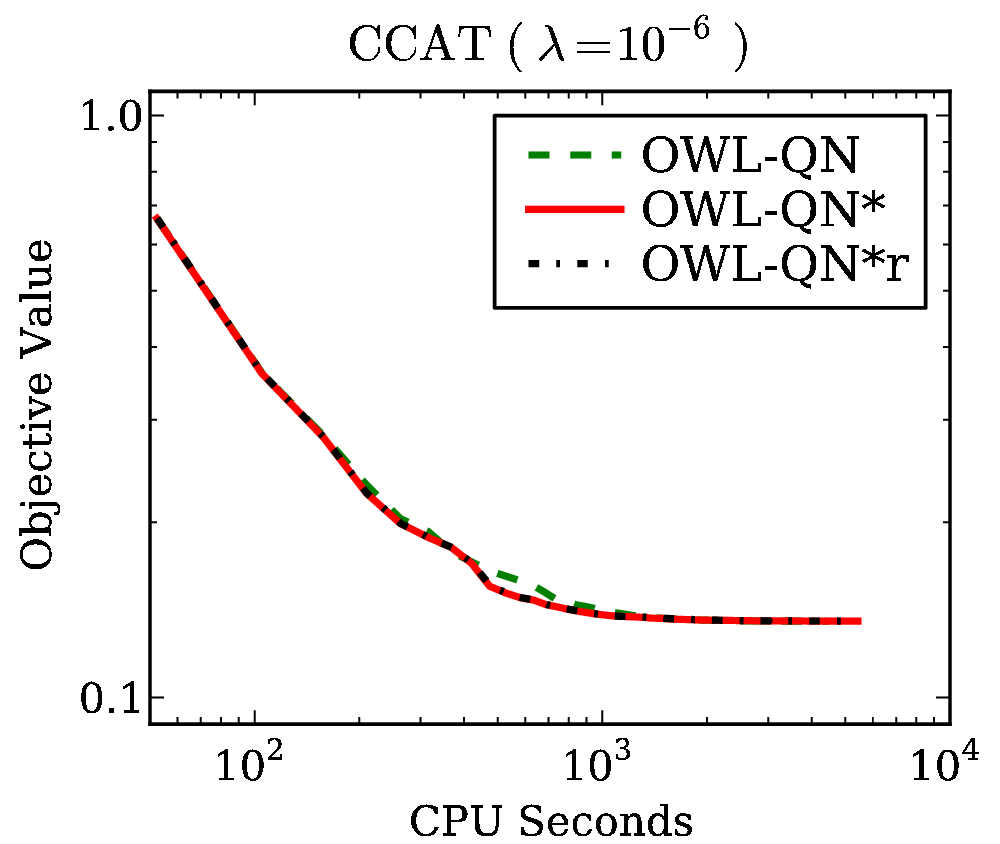} &
     \includegraphics[width=0.34\linewidth]{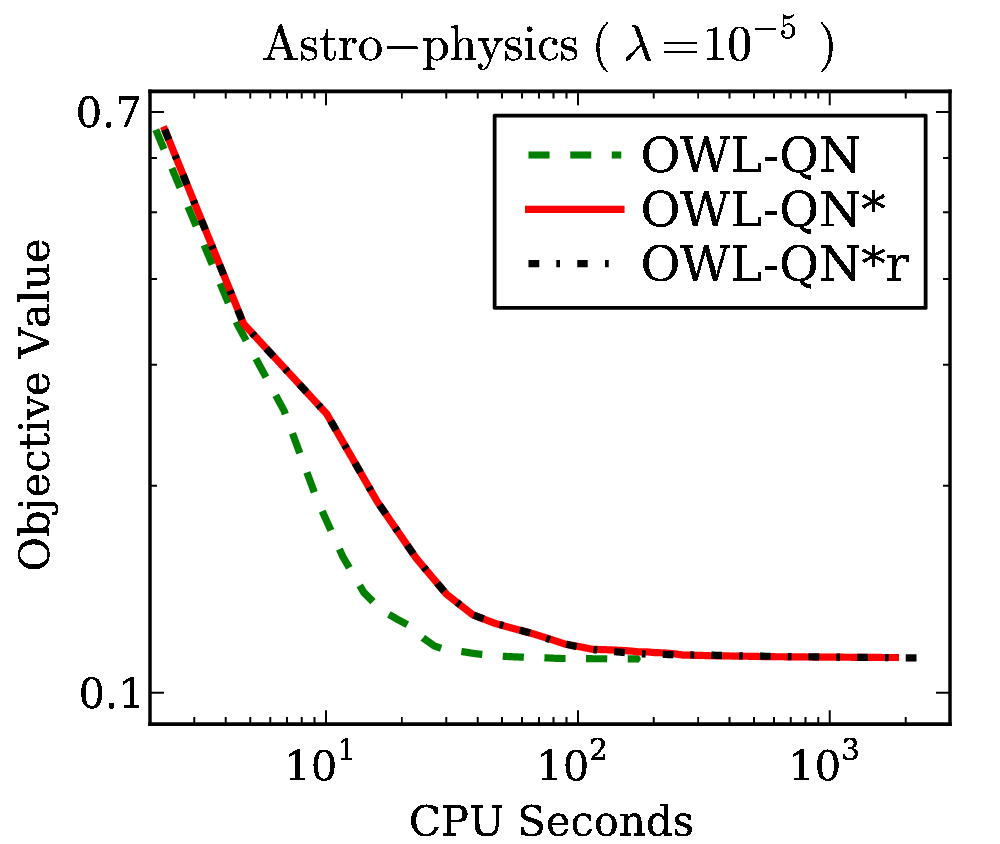} \\
     \includegraphics[width=0.34\linewidth]{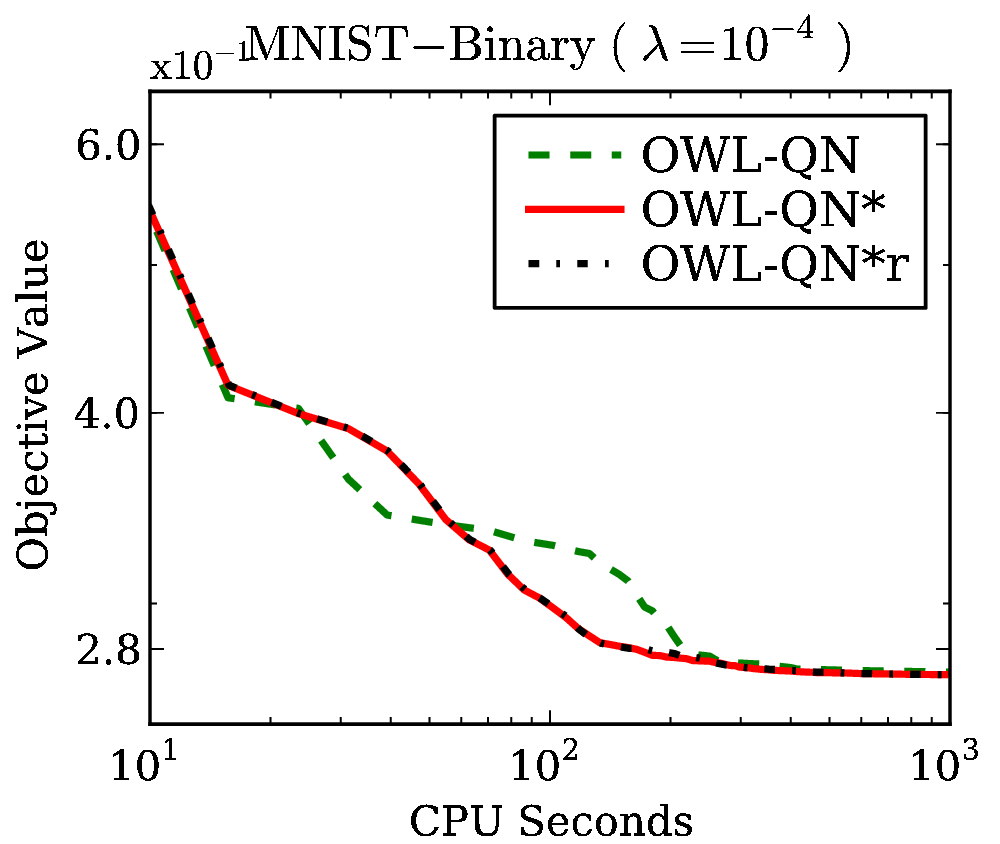} &
     \includegraphics[width=0.34\linewidth]{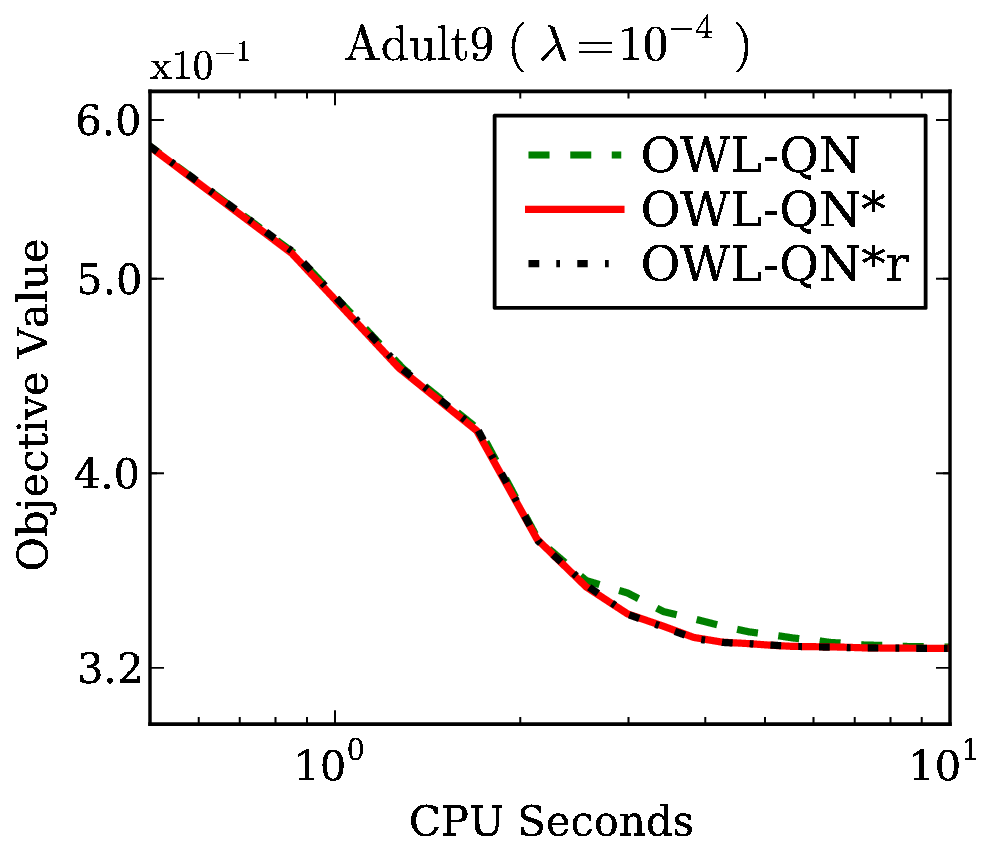} &
     \includegraphics[width=0.34\linewidth]{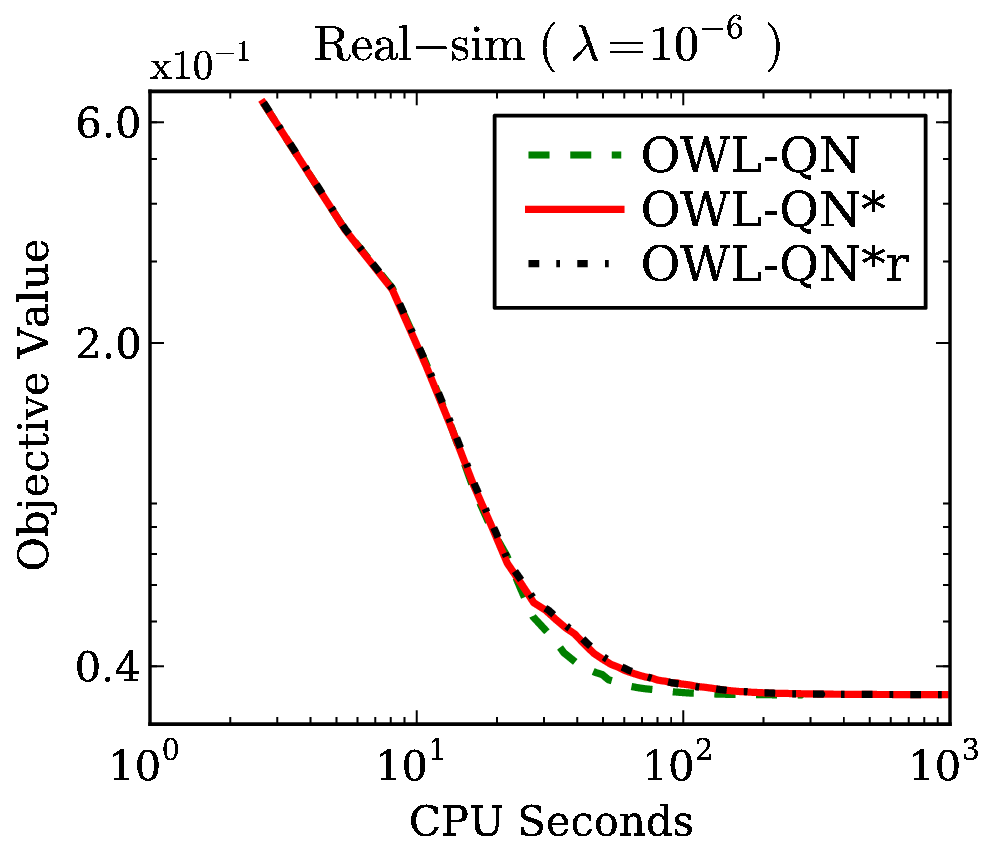} \\
   \end{tabular}
 \caption{Objective function value \emph{vs.}\ CPU seconds on $L_1$-regularized
   logistic loss minimization tasks.}
 \label{fig:l1logistic}
\end{figure}

\begin{figure}
   \begin{tabular}{@{$\!\!\!$}c@{}c@{}c}
     \includegraphics[width=0.34\linewidth]{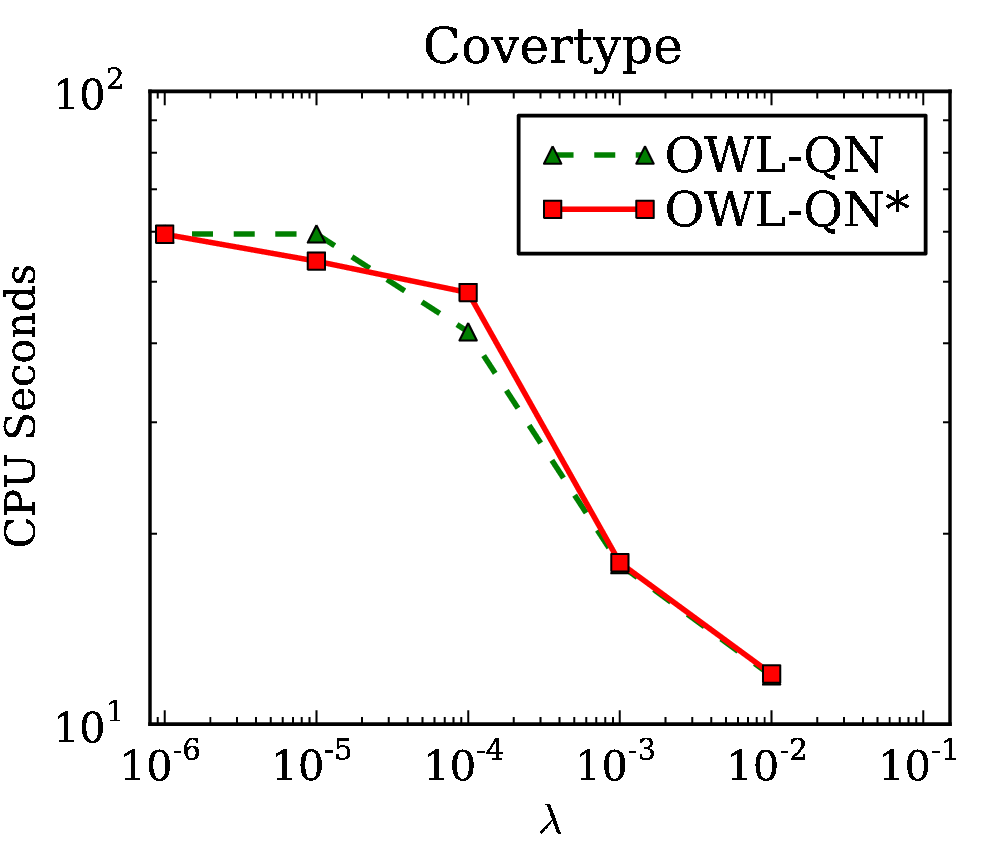} &
     \includegraphics[width=0.34\linewidth]{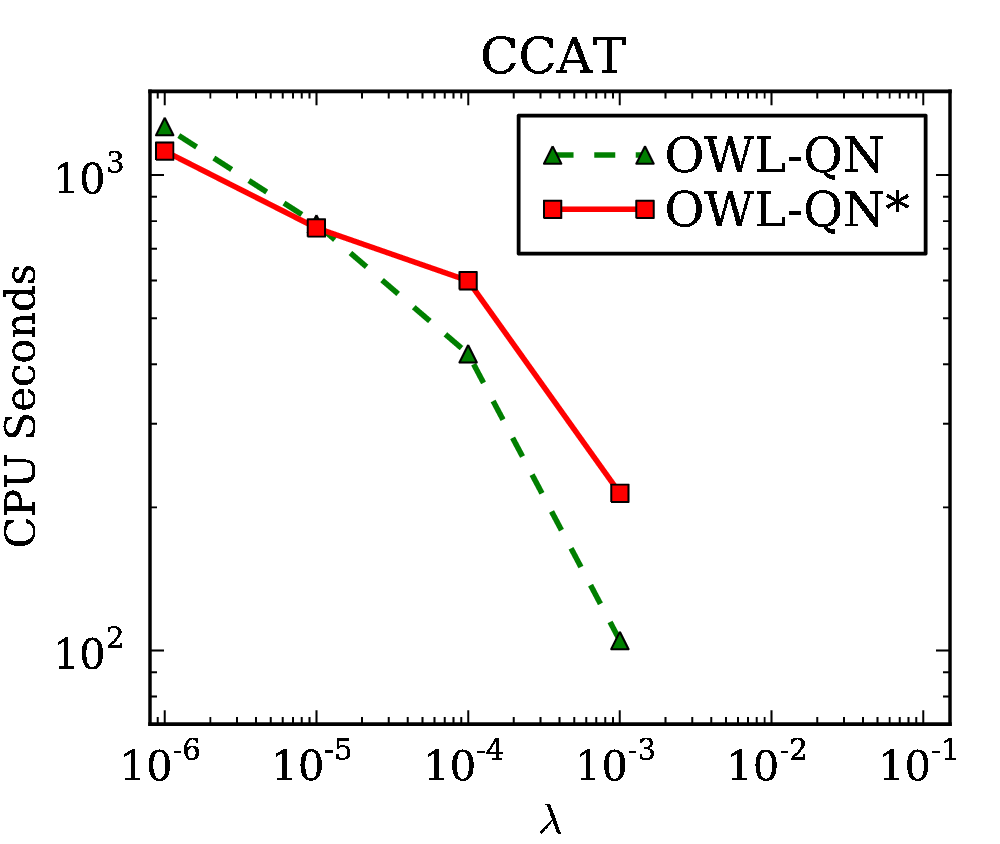} &
     \includegraphics[width=0.34\linewidth]{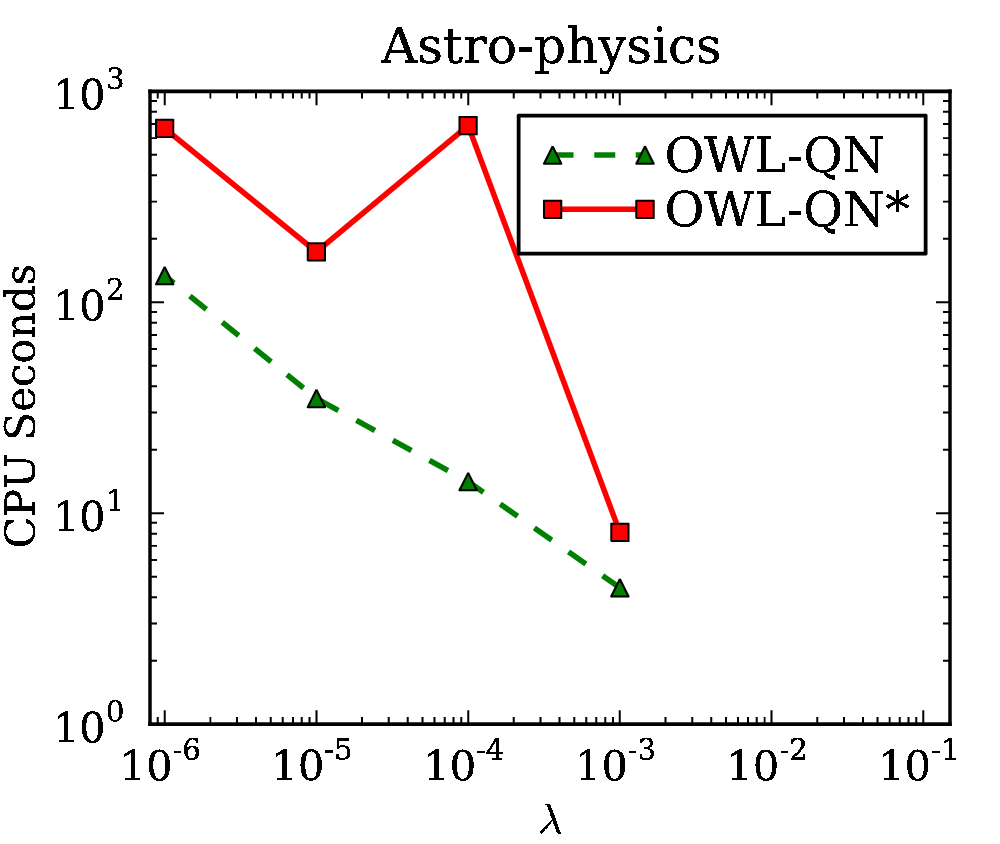} \\
     \includegraphics[width=0.34\linewidth]{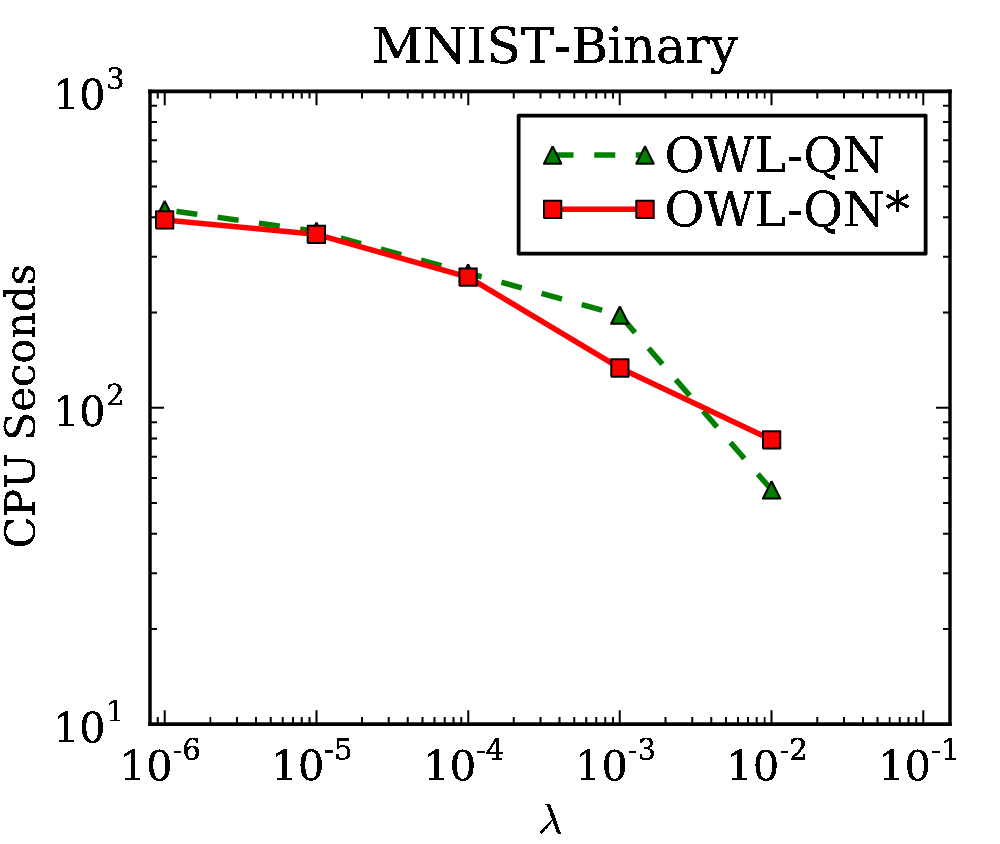} &
     \includegraphics[width=0.34\linewidth]{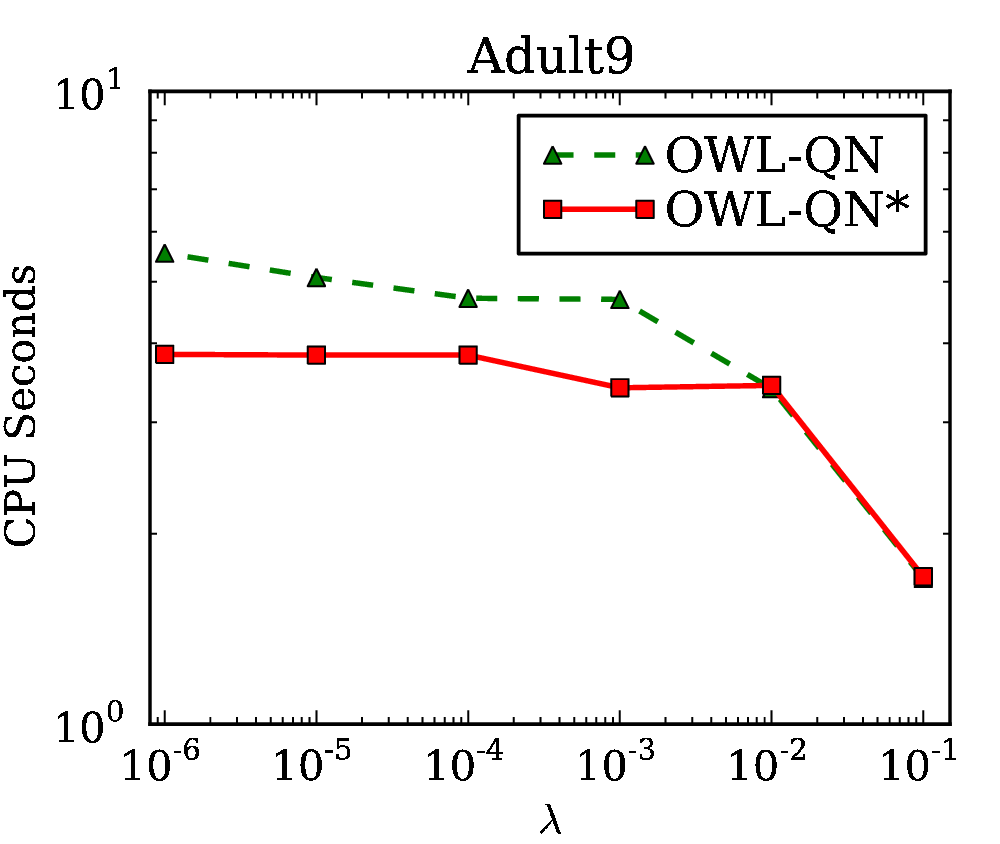} &
     \includegraphics[width=0.34\linewidth]{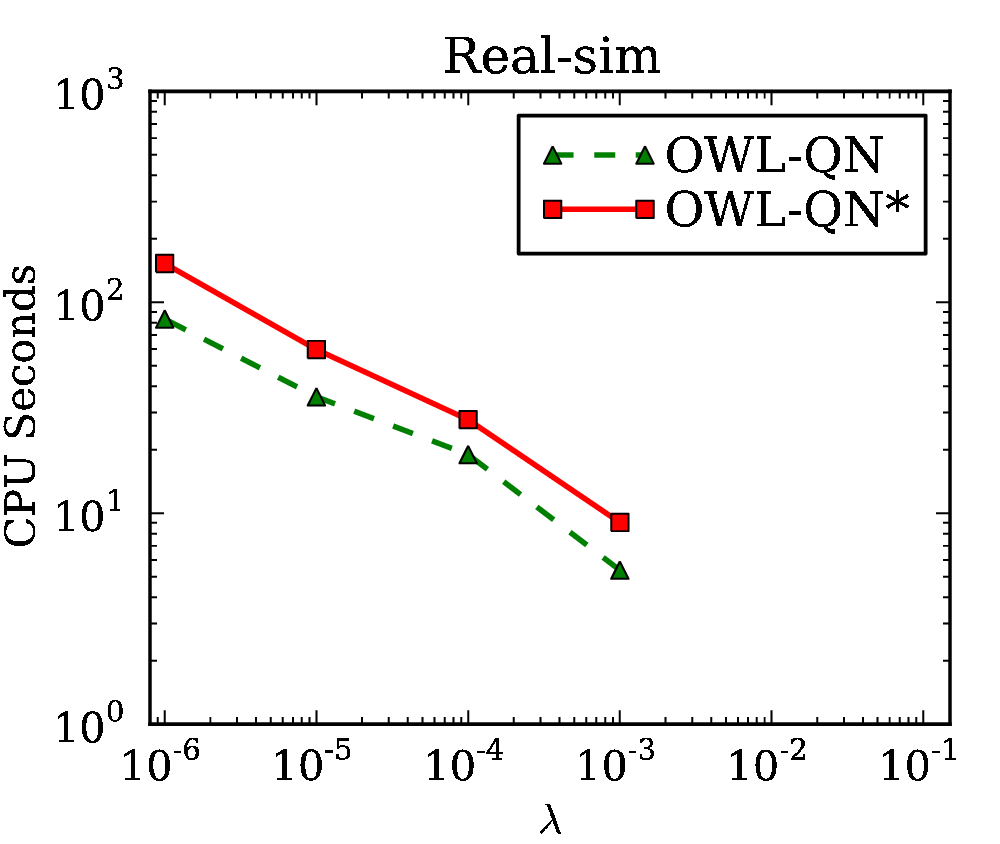} \\
   \end{tabular}
  \caption{Regularization parameter
    $\lambda \in \{10^{-6},\cdots,10^{-1}\}$
    \emph{vs.}\ CPU seconds taken to reduce the objective
    function to within 2\% of the optimal value on $L_1$-regularized
   logistic loss minimization tasks.
   (No point is plotted if the initial parameter $\vw_0 = \vzero$ is
   already optimal.) 
   }
 \label{fig:l1log-lambdas}
\end{figure}

Using the stopping criterion suggested by \citet{AndGao07}, we ran
experiments until the averaged relative change in objective function
value over the previous 5 iterations fell below $10^{-5}$. As shown in
Figure~\ref{fig:l1logistic}, the only clear difference in convergence
between the two algorithms is found on the Astro-physics dataset where
OWL-QN$^*$ is outperformed by the original OWL-QN method. This is
because finding a descent direction via
Algorithm~\ref{alg:find-descent-dir-cg} is particularly difficult on the
Astro-physics dataset (as indicated by the large inner loop iteration number
$k_{L_1}$ in Table~\ref{tab:parameters}); the slowdown on this dataset
can also be found in Figure~\ref{fig:l1log-lambdas} for other values
of $\lambda$. Although finding a descent direction can be challenging
for the generic direction-finding routine of OWL-QN$^*$, in the following
experiment we show that this routine is very robust to the choice of
initial subgradients.

To examine the algorithms' sensitivity to the choice of
subgradients, we also ran them with subgradients randomly chosen from
the set $\partial J(\vw)$ (as opposed to
the specially chosen subgradient $\vg^{\text{ow}}$ used in the
previous set of experiments) fed to their corresponding
direction-finding routines. OWL-QN relies heavily on its particular
choice of subgradients, hence breaks down completely under these
conditions: the only dataset where we could even plot its (poor)
performance was Covertype (dotted ``OWL-QNr'' line in
Figure~\ref{fig:l1logistic}). Our direction-finding routine, by
contrast, is self-correcting and thus not affected by this
manipulation: the curves for OWL-QN*r lie on top of those
for OWL-QN*. Table~\ref{tab:parameters} shows that in
this case more direction-finding iterations are needed though:
$k_{L_1 r} > k_{L_1}$. This empirically confirms that as long as
$\argsup_{\vg \in \partial J(\vw)} \vg^{\top} \vp$ is given,
Algorithm~\ref{alg:find-descent-dir-cg} can indeed be used as a
generic quasi-Newton direction-finding routine that is able to
recover from a poor initial choice of subgradients.

\subsection{$L_2$-Regularized Multiclass and Multilabel Hinge Loss}
\label{sec:multi-results}

\begin{table}
  \caption{The multiclass (top 6 rows) and multilabel (bottom 3 rows)
    datasets used, values of the regularization parameter, and 
    overall number $k$ of direction-finding iterations in our
    experiments of Section~\ref{sec:multi-results}.} 
  \centering
  \begin{center}
      \begin{tabular}{|l||r|r|r|r||r|r|}
        \hline
        Dataset & Train/Test Set Size & Dimensionality &
        $|\!\Zcal\!|$ & Sparsity &  $\lambda$ & $k$
        \\
        \hline\hline
        Letter\rule{0pt}{2.2ex} & 16000/4000~ & 16~ & 26 & 0.0\%~  & 
        $10^{-6}$  & 65 \\
        USPS & 7291/2007~ & 256~ & 10 & 3.3\%~ & $10^{-3}$ & 
         14 \\
        Protein & 14895/6621~ & 357~ & 3 & 70.7\%~  & $10^{-2}$ &
         1 \\
        MNIST & 60000/10000~ & 780~ & 10 & 80.8\%~ &$10^{-3}$ &
        1  \\
        INEX & 6053/6054~ &167295~ & 18  & 99.5\%~  & $10^{-6}$ &
        5 \\
        News20 & 15935/3993~ &62061~ & 20 & 99.9\%~  & $10^{-2}$ &
        12 \\ \hline
        Scene\rule{0pt}{2.2ex} & 1211/1196~ & 294~ & 6 & 0.0\%~ &
         $10^{-1}$ & 14 \\
        TMC2007 & 21519/7077~ & 30438~ & 22 & 99.7\%~ &
       $10^{-5}$ & 19 \\
        RCV1 & 21149/2000~ &  47236~ & 103 & 99.8\%~ &
        $10^{-5}$ & 4 \\
        \hline
      \end{tabular}
  \end{center}
  \label{tab:datasets2}
\end{table}

\begin{figure}[b]
   \begin{tabular}{@{$\!\!\!$}c@{}c@{}c}
     \includegraphics[width=0.34\linewidth]{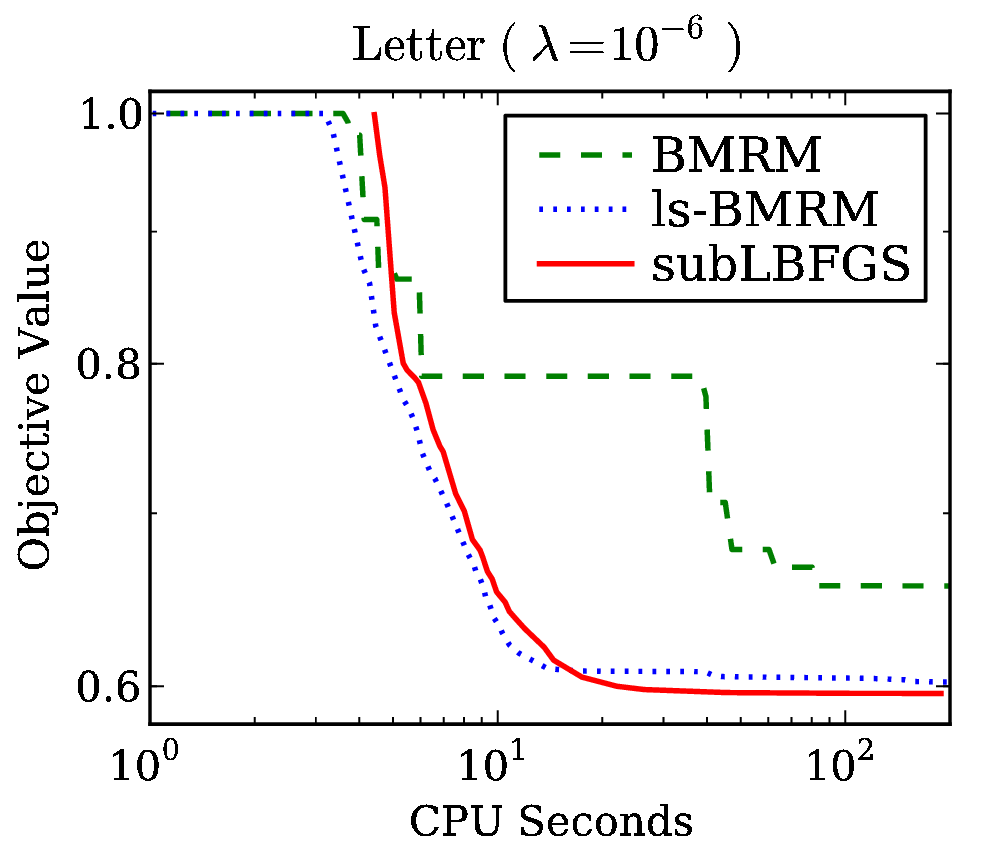} &
     \includegraphics[width=0.34\linewidth]{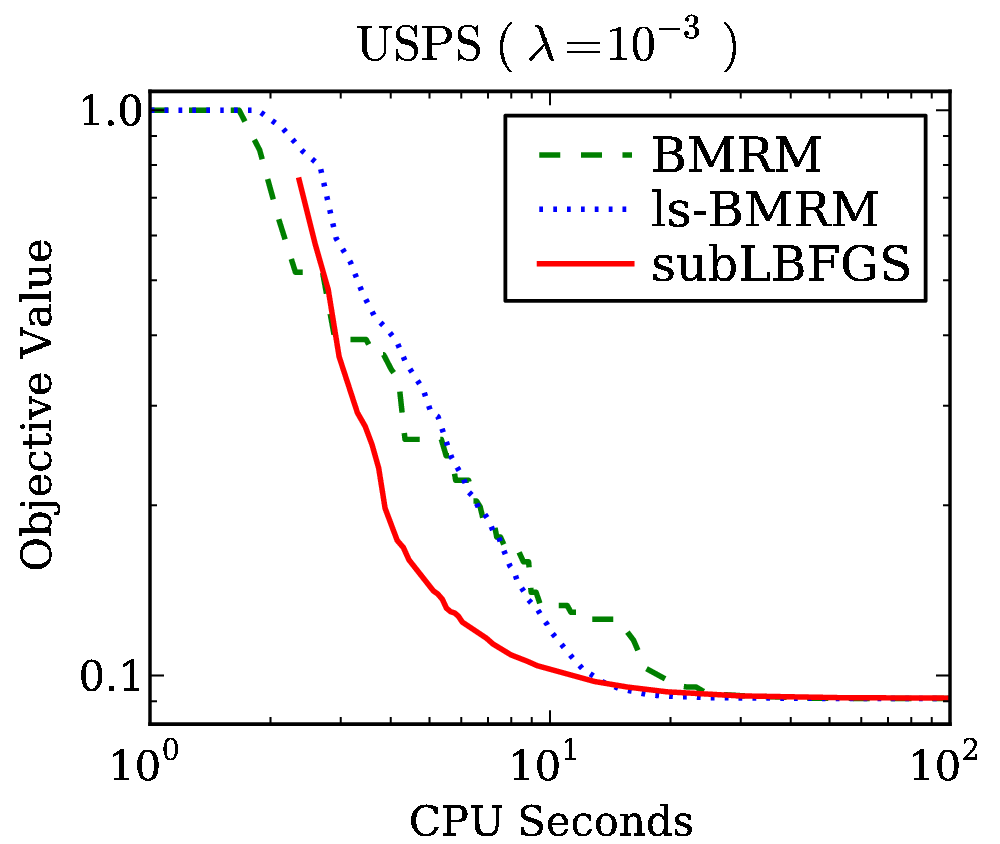} &
     \includegraphics[width=0.34\linewidth]{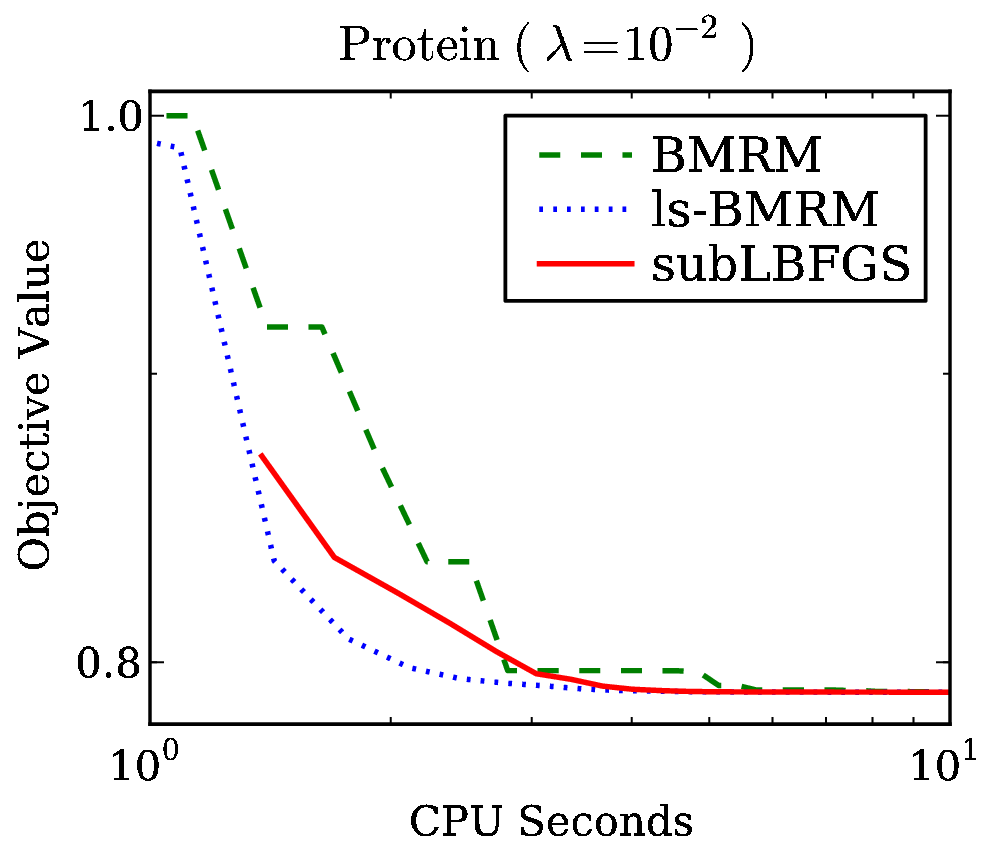} \\
     \includegraphics[width=0.34\linewidth]{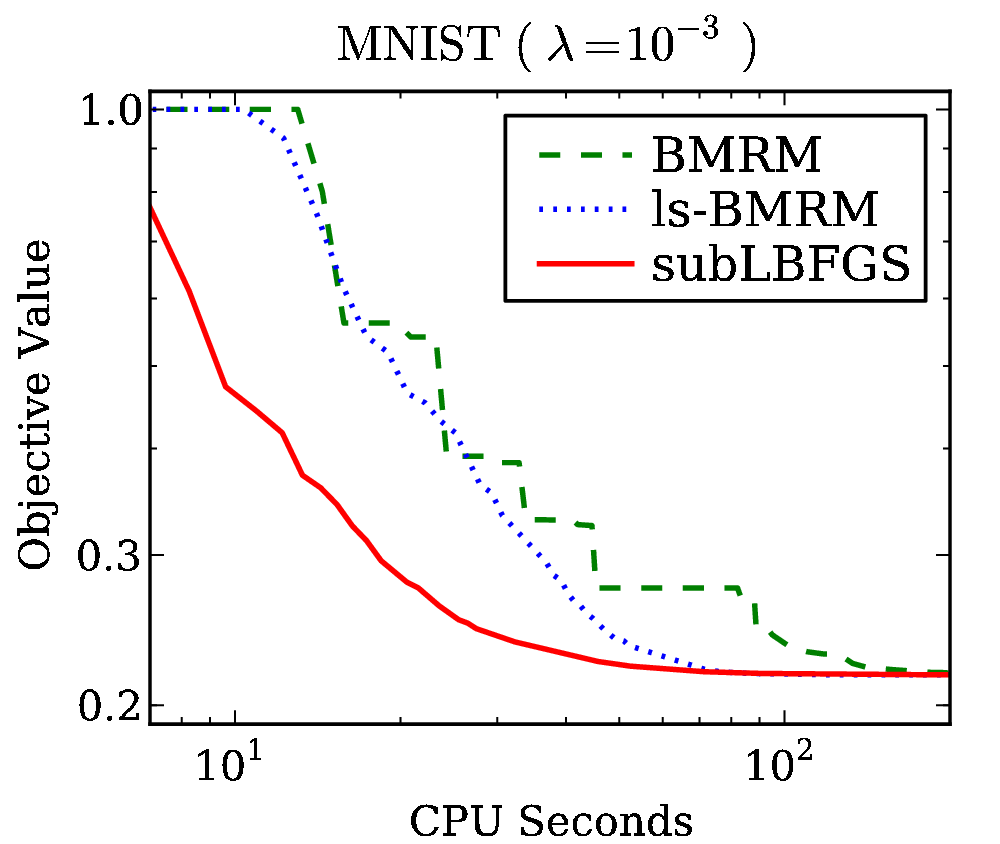} &
     \includegraphics[width=0.34\linewidth]{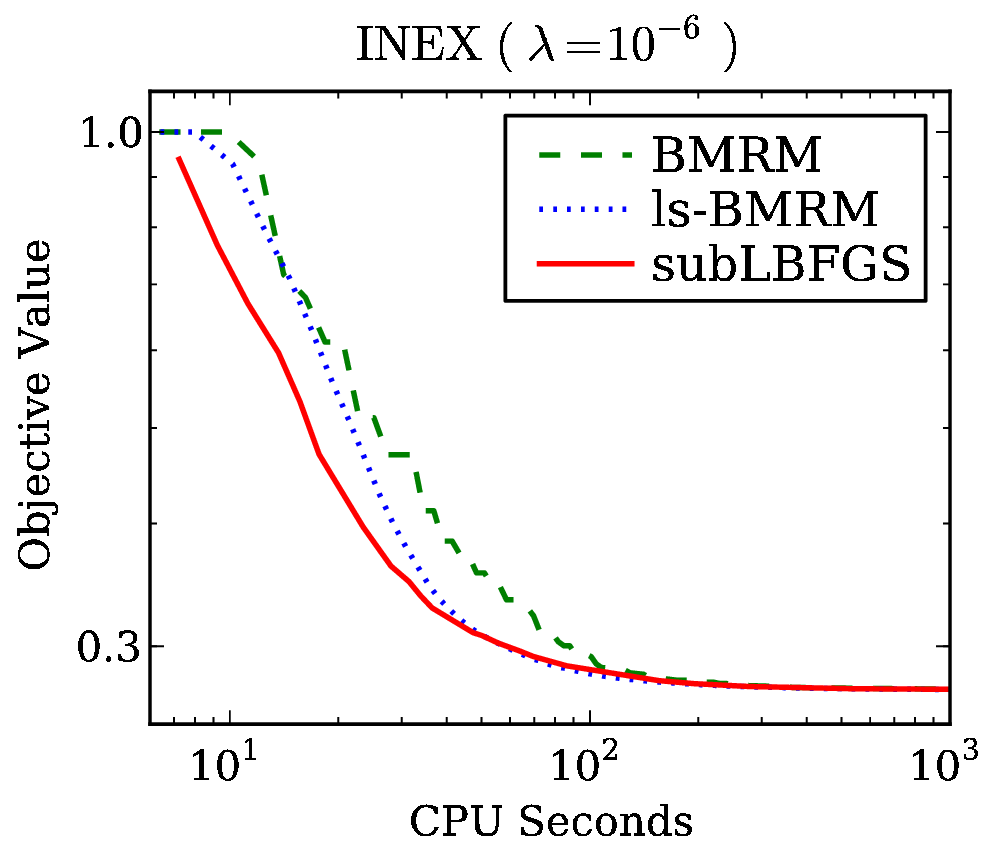} &
     \includegraphics[width=0.34\linewidth]{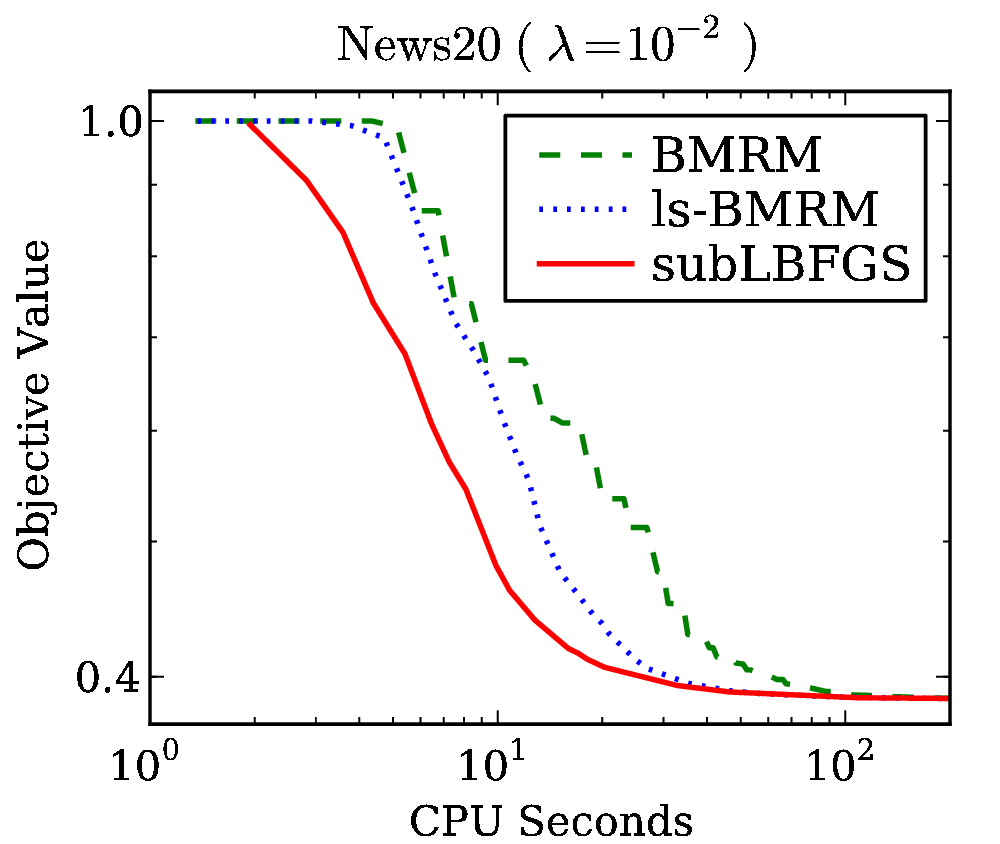} \\
   \end{tabular}
 \caption{Objective function value \emph{vs.}\ CPU seconds on $L_2$-regularized
   multiclass hinge loss minimization tasks.}
 \label{fig:l2hinge-multi}
\end{figure}

We incorporated our exact line search of
Section~\ref{sec:ExactLineSearch} into both subLBFGS and OCAS
\citep{FraSon08}, thus enabling them to deal with multiclass and
multilabel losses. We refer to our generalized version of OCAS as line
search BMRM (ls-BMRM). Using the variant of the multiclass and
multilabel hinge loss which enforces a uniform margin of separation
($\Delta(z, z') = 1 ~\forall z \neq z'$), we experimentally evaluated
both algorithms on a number of publicly available datasets
(Table~\ref{tab:datasets2}). All multiclass datasets except INEX were
downloaded from
{\small\url{http://www.csie.ntu.edu.tw/~cjlin/libsvmtools/datasets/multiclass.html}},
while the multilabel datasets were obtained from
{\small\url{http://www.csie.ntu.edu.tw/~cjlin/libsvmtools/datasets/multilabel.html}}.
INEX \citep{MaeDenGal07} is available from
{\small\url{http://webia.lip6.fr/~bordes/mywiki/doku.php?id=multiclass_data}}.
The original RCV1 dataset
consists of 23149 training instances, of which we used 21149 instances
for training and the remaining 2000 for testing.

\subsubsection{Performance on Multiclass Problems}

\begin{figure}
    \begin{tabular}{@{$\!\!\!$}c@{}c@{}c}
      \includegraphics[width=0.34\linewidth]{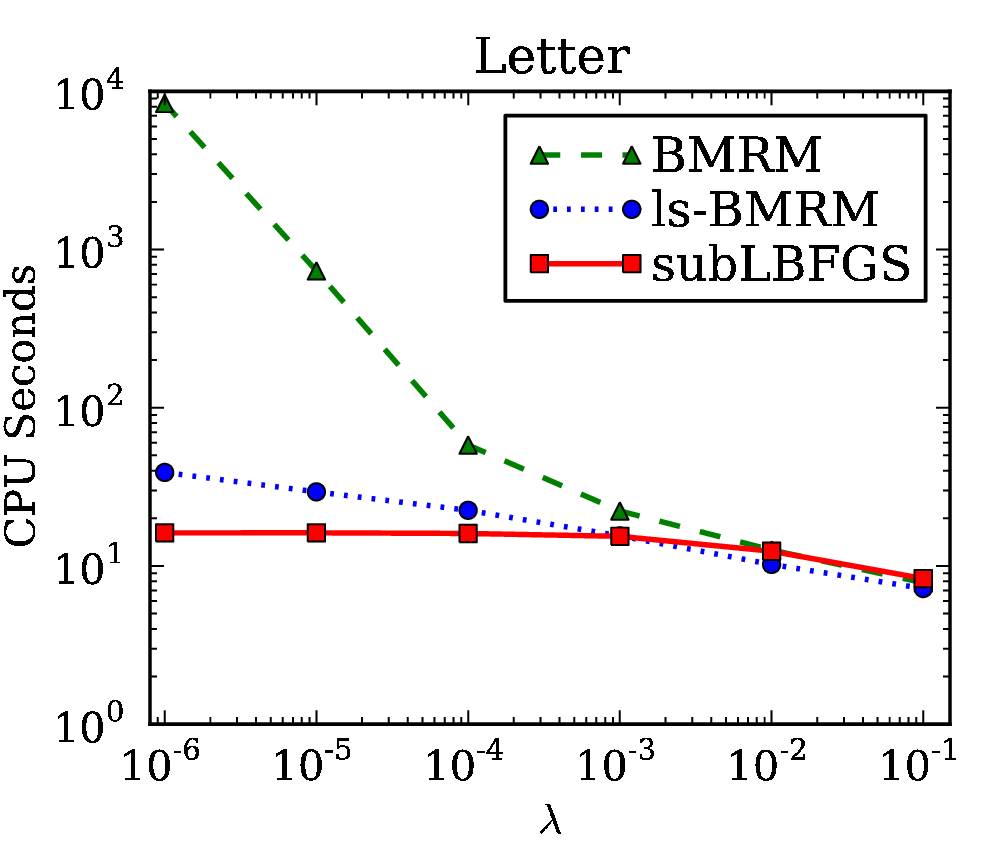} &
      \includegraphics[width=0.34\linewidth]{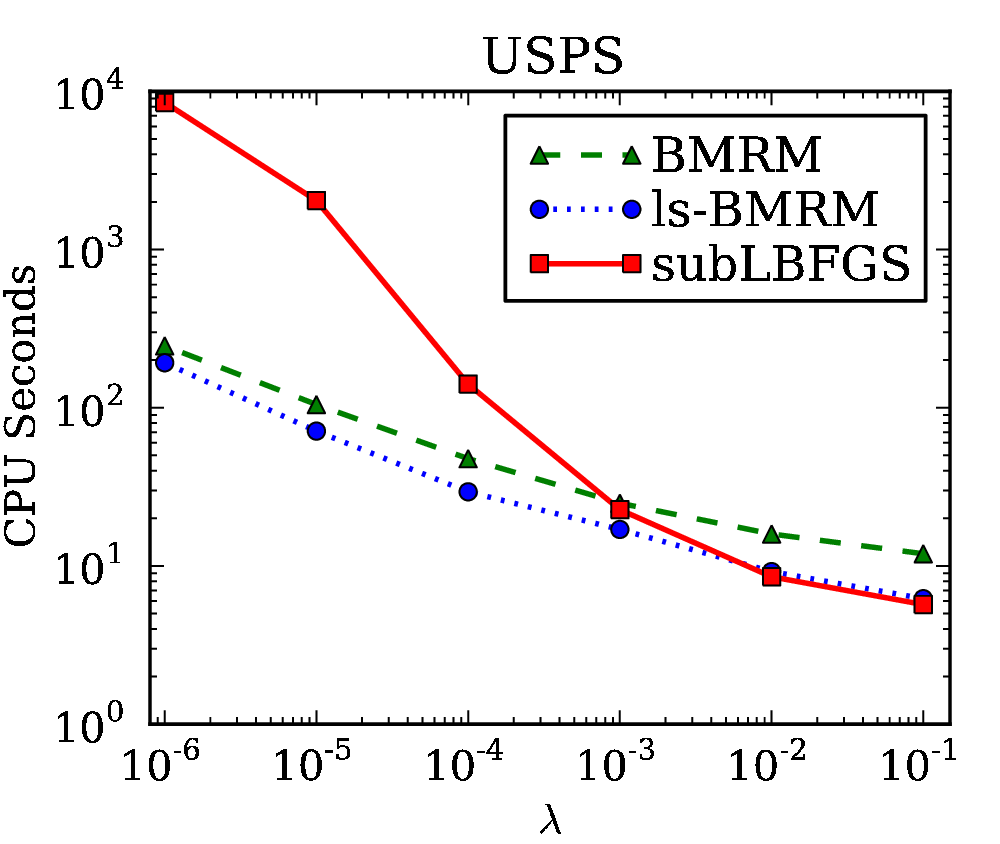} &
      \includegraphics[width=0.34\linewidth]{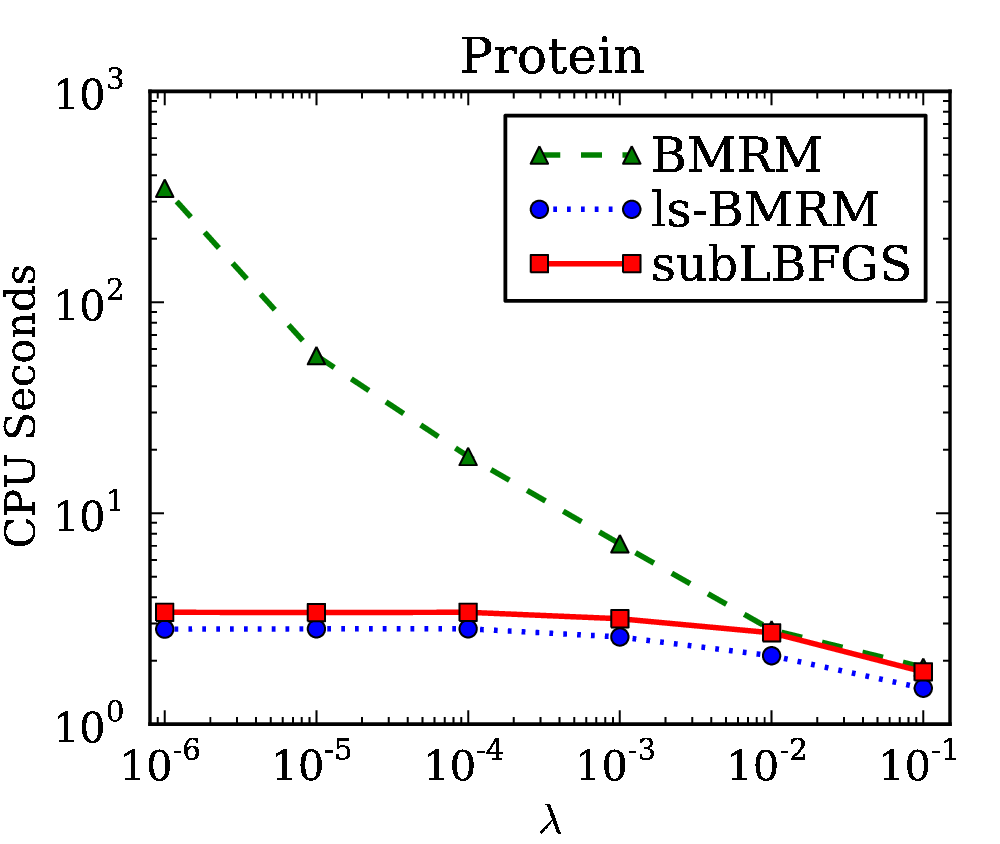} \\
     \includegraphics[width=0.34\linewidth]{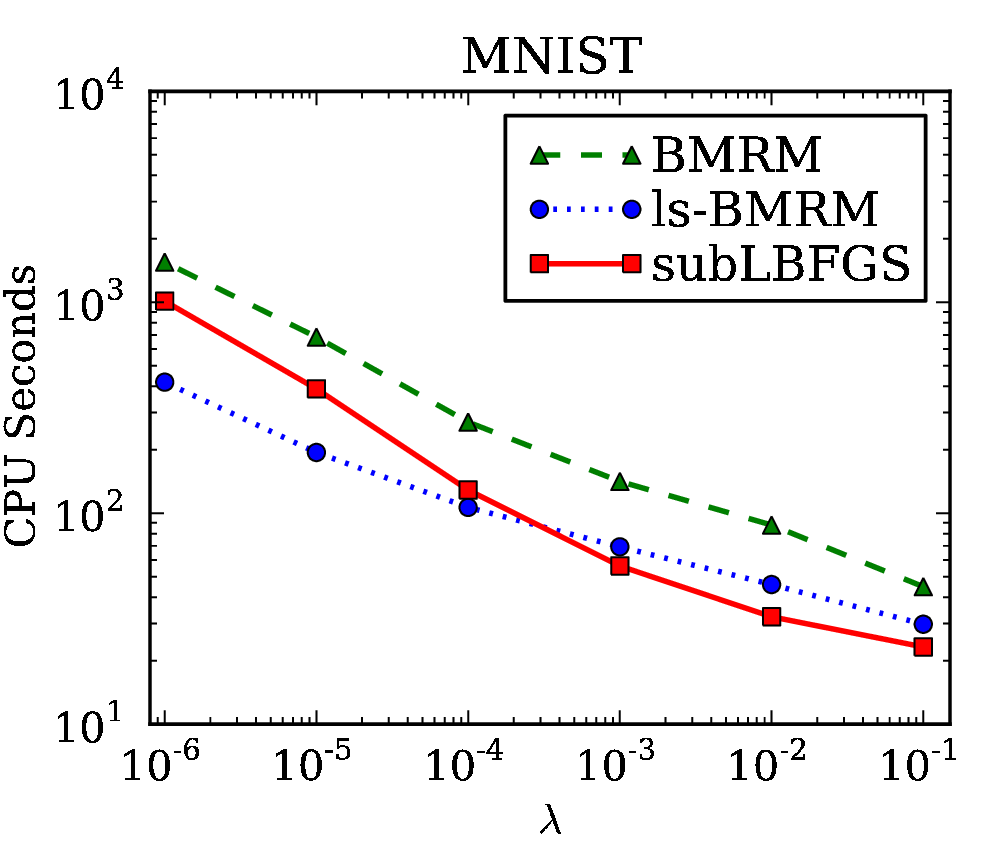} &
     \includegraphics[width=0.34\linewidth]{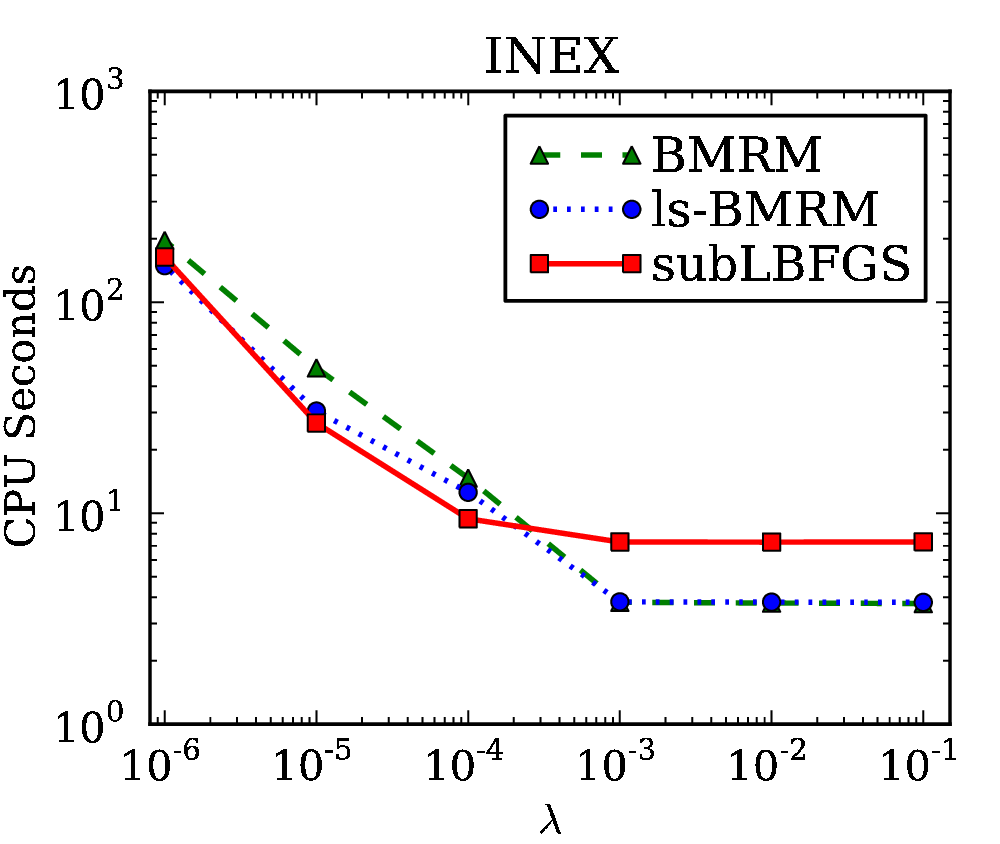} &
      \includegraphics[width=0.34\linewidth]{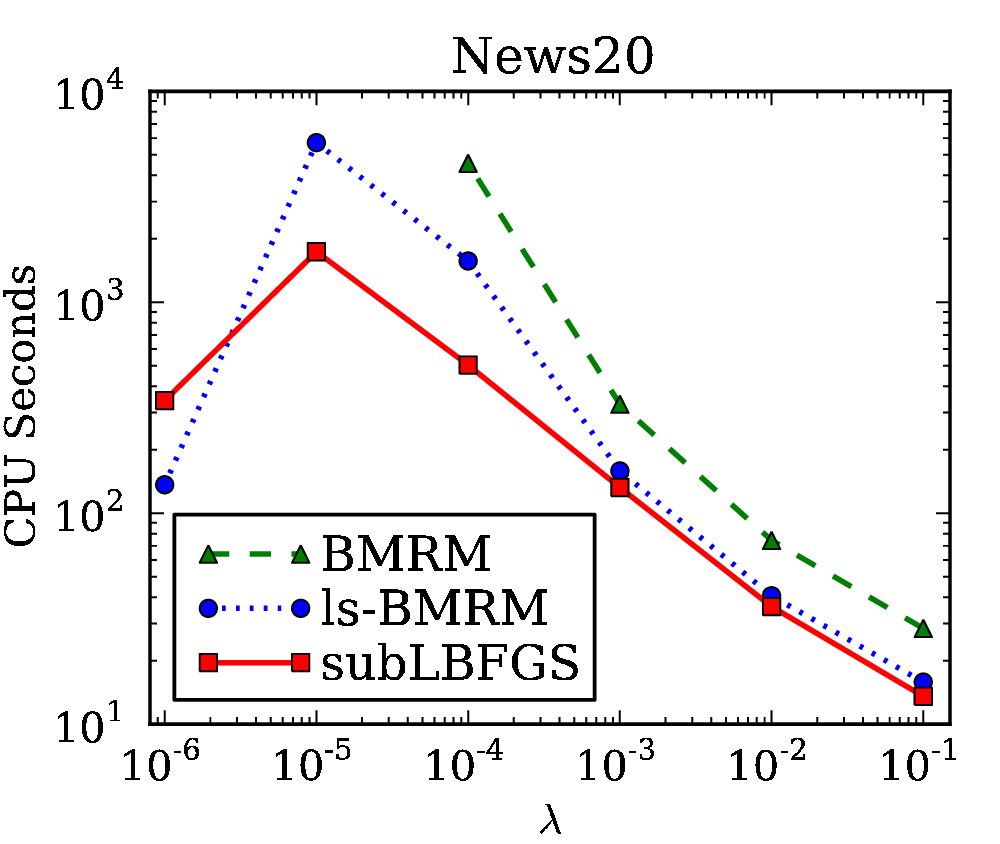} \\
    \end{tabular}
  \caption{Regularization parameter $\lambda \in \{10^{-6},\cdots,
    10^{-1}\}$ \emph{vs.}\ CPU seconds taken to reduce the objective
    function to within 2\% of the optimal value. (No point is plotted if
    an algorithm failed to reach the threshold value within $10^4$
    seconds.)
  }
  \label{fig:l2mclass-lambdas}
\end{figure}

This set of experiments is designed to demonstrate the convergence
properties of multiclass subLBFGS, compared to the BMRM bundle
method \citep{TeoVisSmoLe09} and ls-BMRM.
Figure~\ref{fig:l2hinge-multi} shows that subLBFGS
outperforms BMRM on all datasets.
On 4 out of 6 datasets, subLBFGS outperforms ls-BMRM as well
early on but slows down later, for an overall performance comparable
to ls-BMRM. On the MNIST dataset, for instance, subLBFGS takes only
about half as much CPU time as ls-BMRM to reduce the objective
function value to 0.3 (about 50\% above the optimal value), yet both
algorithms reach within 2\% of the optimal value at about the same time
(Figure~\ref{fig:l2mclass-lambdas}, bottom left). We hypothesize
that subLBFGS' local model \eqref{eq:gen-quadratic-model} of the
objective function facilitates rapid early improvement but is less
appropriate for final convergence to the optimum (\cf the discussion
in Section~\ref{sec:discuss}).  Bundle methods, on the other hand, are
slower initially because they need to accumulate a sufficient number
of gradients to build a faithful piecewise linear model of the
objective function.  These results suggest that a hybrid approach that
first runs subLBFGS then switches to ls-BMRM may be promising.

Similar to what we saw in the binary setting
(Figure~\ref{fig:l2hinge-lambdas}), Figure~\ref{fig:l2mclass-lambdas}
shows that all algorithms tend to converge faster for large values of
$\lambda$. Generally, subLBFGS converges faster than BMRM across a
wide range of $\lambda$ values; for small values of $\lambda$ it can
greatly outperform BMRM (as seen on Letter, Protein, and News20). The
performance of subLBFGS is worse than that of BMRM in two instances:
on USPS for small values of $\lambda$, and on INEX for large values of
$\lambda$. The poor performance on USPS may be caused by a limitation
of subLBFGS' local model \eqref{eq:gen-quadratic-model} that causes it
to slow down on final convergence.  On the INEX dataset, the initial
point $\vw_0 = \vzero$ is nearly optimal for large values of
$\lambda$; in this situation there is no advantage in using subLBFGS.

Leveraging its exact line search (Algorithm~\ref{alg:exact-multi-ls}), ls-BMRM
is competitive on all datasets and across all $\lambda$ values, exhibiting
performance comparable to subLBFGS in many cases.
From Figure~\ref{fig:l2mclass-lambdas} we find that BMRM never
outperforms both subLBFGS and ls-BMRM.

\subsubsection{Performance on Multilabel Problems}

\begin{figure}
  \begin{tabular}{@{$\!\!\!$}c@{}c@{}c}
    \includegraphics[width=0.34\linewidth]{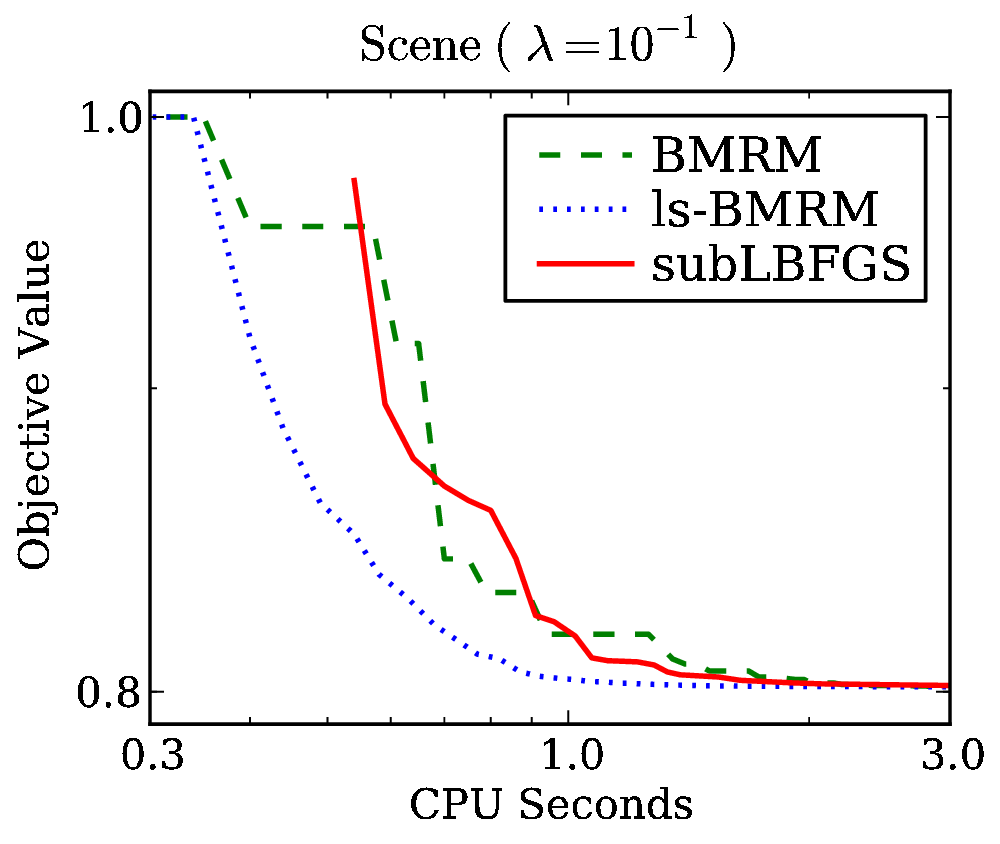} &
    \includegraphics[width=0.34\linewidth]{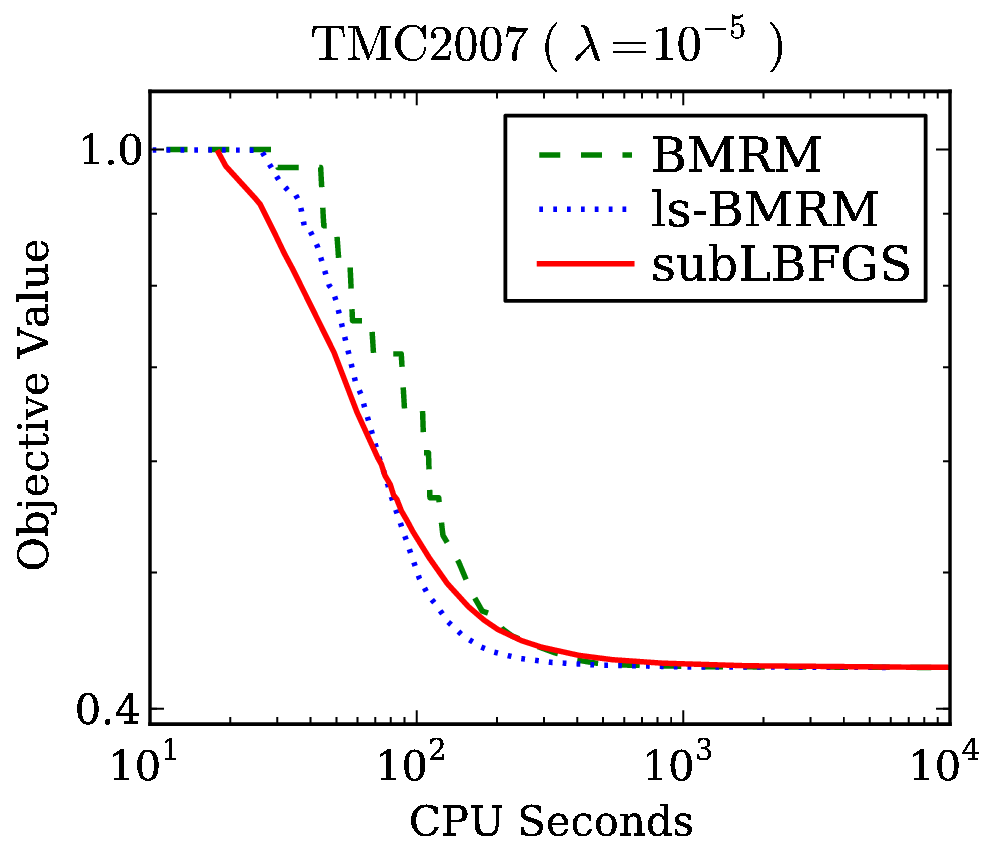} &
    \includegraphics[width=0.34\linewidth]{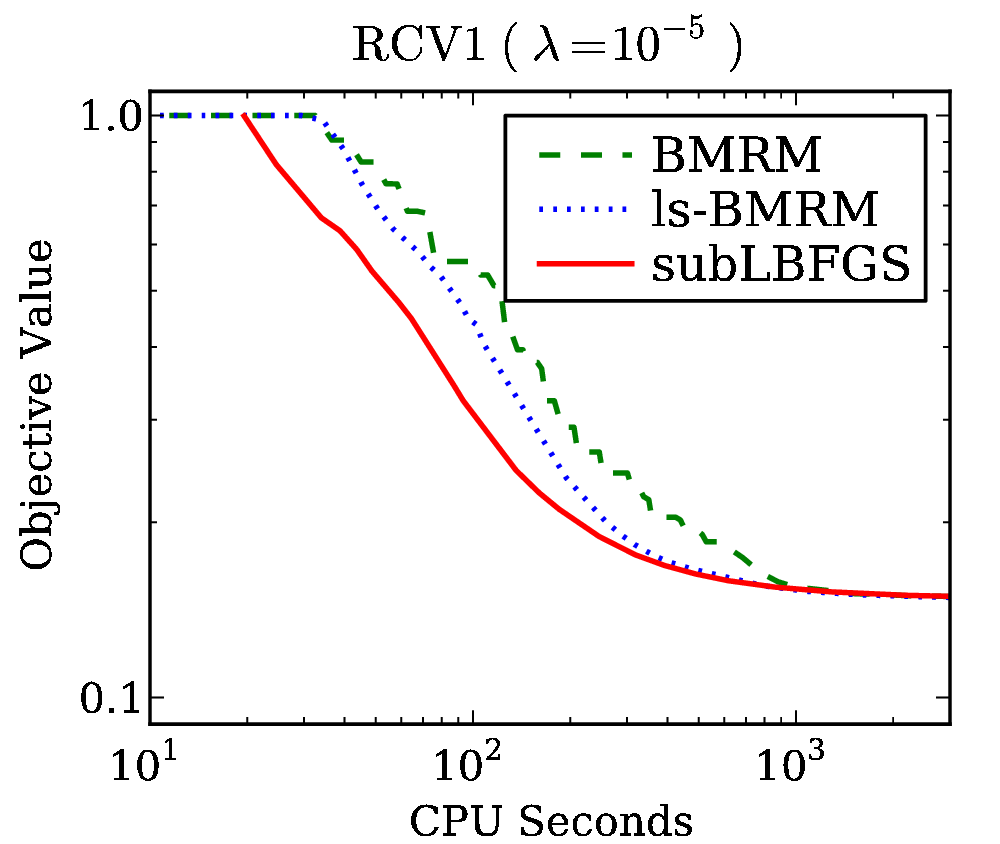} \\
  \end{tabular}
  \caption{Objective function value \emph{vs.}\ CPU seconds in
    $L_2$-regularized multilabel hinge loss minimization tasks.}
  \label{fig:multilabel}
\end{figure}

\begin{figure}
  \begin{tabular}{@{$\!\!\!$}c@{}c@{}c}
    \includegraphics[width=0.34\linewidth]{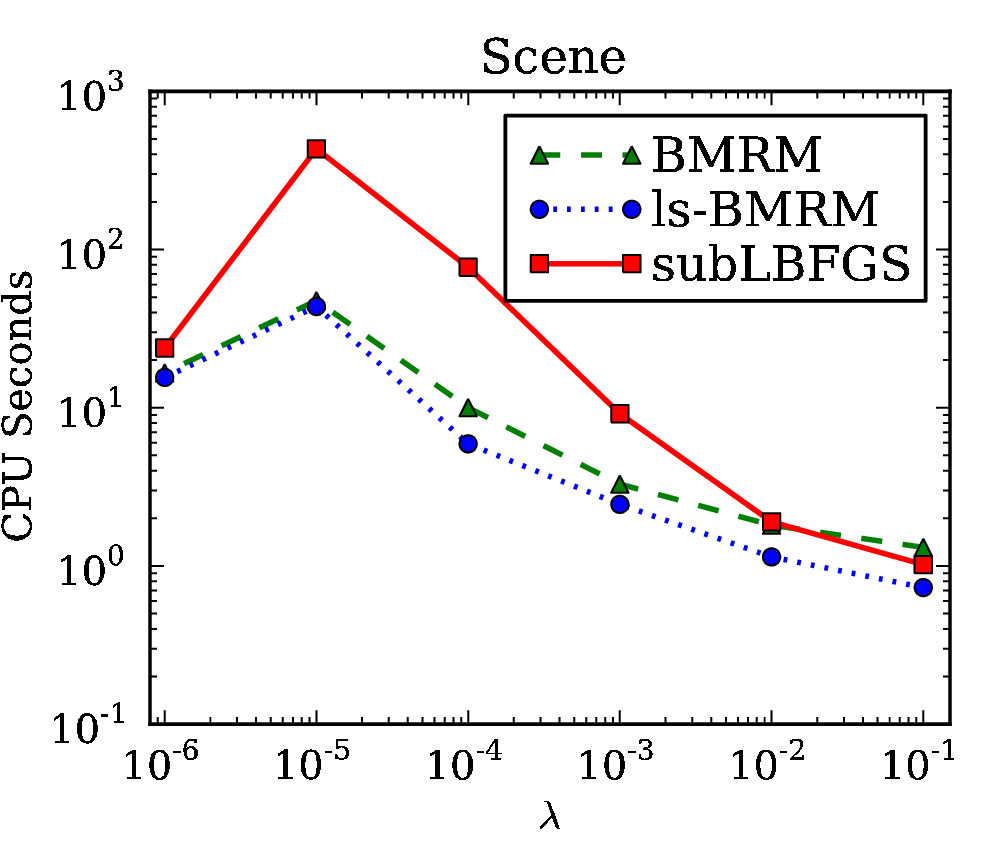} &
    \includegraphics[width=0.34\linewidth]{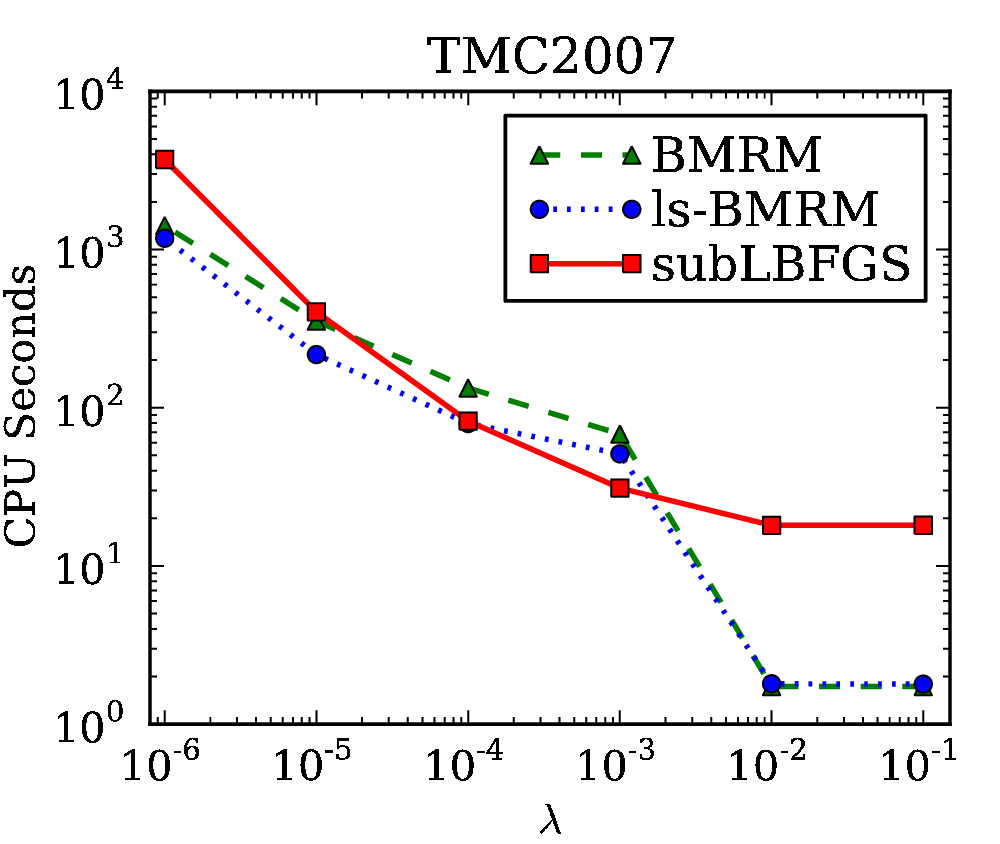} &
    \includegraphics[width=0.34\linewidth]{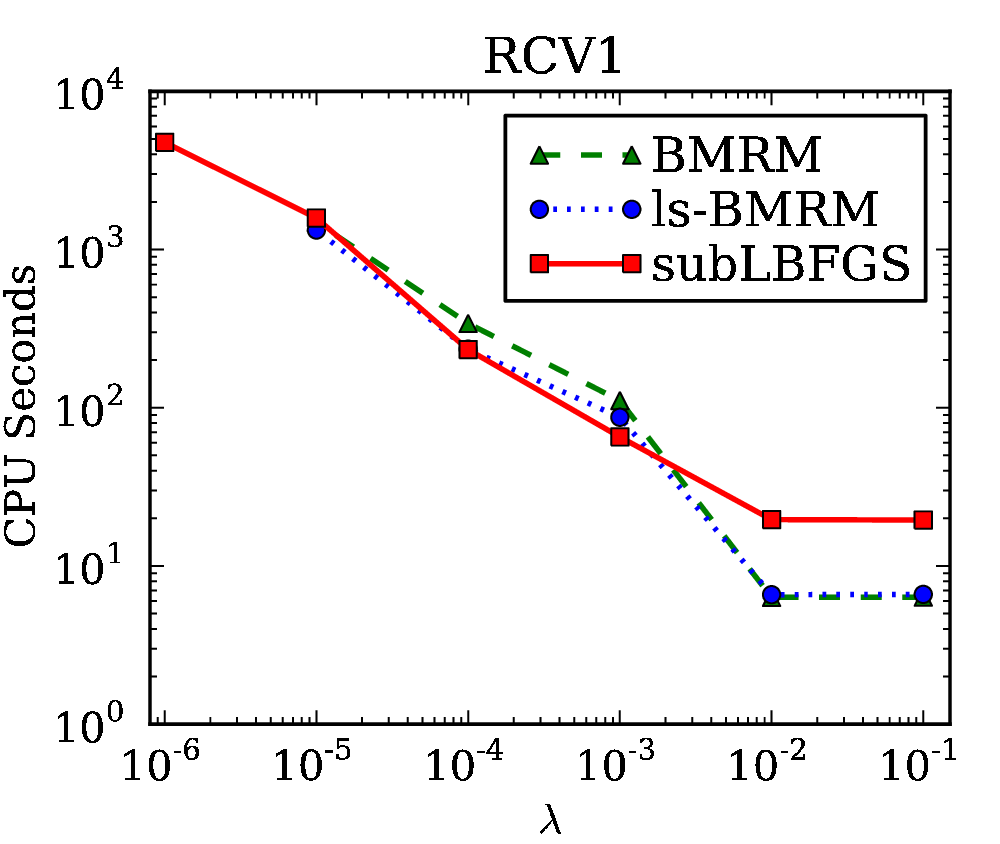} \\
  \end{tabular}
  \caption{Regularization parameter $\lambda \in \{10^{-6},\cdots,
    10^{-1}\}$ \emph{vs.}\ CPU seconds taken to reduce the objective
    function to within 2\% of the optimal value. (No point is plotted if
    an algorithm failed to reach the threshold value within $10^4$
    seconds.)}
  \label{fig:mlabel-lambdas}
\end{figure}

For our final set of experiments we turn to the multilabel setting.
Figure~\ref{fig:multilabel} shows that on the Scene dataset the
performance of subLBFGS is similar to that of BMRM, while on the
larger TMC2007 and RCV1 sets, subLBFGS outperforms both of its
competitors initially but slows down later on, resulting in
performance no better than BMRM. Comparing performance across
different values of $\lambda$ (Figure~\ref{fig:mlabel-lambdas}), we
find that in many cases subLBFGS requires more time than its
competitors to reach within 2\% of the optimal value, and in contrast
to the multiclass setting, here ls-BMRM only performs marginally better
than BMRM. The primary reason for this is that the exact line search
used by ls-BMRM and subLBFGS requires substantially more computational
effort in the multilabel than in the multiclass setting. There is an
inherent trade-off here: subLBFGS and ls-BMRM expend computation in an
exact line search, while BMRM focuses on improving its local model of
the objective function instead. In situations where the line search is
very expensive, the latter strategy seems to pay off.

\section{Discussion and Outlook}
\label{sec:discuss}

We proposed subBFGS (\emph{resp.}\ subLBFGS), an extension of the BFGS
quasi-Newton method (\emph{resp.}\ its limited-memory variant), for
handling nonsmooth convex optimization problems, and proved its global
convergence in objective function value. We applied our algorithm to a variety
of machine learning problems employing the $L_2$-regularized binary
hinge loss and its multiclass and multilabel generalizations, as well
as $L_1$-regularized risk minimization with logistic loss. Our
experiments show that our algorithm is versatile, applicable to many
problems, and often outperforms specialized solvers.

Our solver is easy to parallelize: The master node computes the search
direction and transmits it to the slaves. The slaves compute the
(sub)gradient and loss value on subsets
of data, which is aggregated at the master node. This information is
used to compute the next search direction, and the process
repeats. Similarly, the line search, which is the expensive part of the
computation on multiclass and multilabel problems, is easy to
parallelize: The slaves run Algorithm~\ref{alg:mincplf} on subsets of
the data; the results are fed back to the master which can then run
Algorithm~\ref{alg:exact-multi-ls} to compute the step size. 

In many of our experiments we observe that subLBFGS decreases the
objective function rapidly at the beginning but slows down closer to
the optimum. We hypothesize that this is due to an averaging effect:
Initially (\ie when sampled sparsely at a coarse scale) a
superposition of many hinges looks sufficiently similar to a smooth
function for optimization of a quadratic local model to work well (\cf
Figure~\ref{fig:bfgs-model}). Later on, when the objective is sampled
at finer resolution near the optimum, the few nearest hinges begin to
dominate the picture, making a smooth local model less appropriate.

Even though the local model \eqref{eq:gen-quadratic-model} of
sub(L)BFGS is nonsmooth, it only explicitly models the hinges at its
present location\,---\,all others are subject to smooth quadratic
approximation. Apparently this strategy works sufficiently well during early
iterations to provide for rapid improvement on multiclass problems,
which typically comprise a large number of hinges. The exact
location of the optimum, however, may depend on individual nearby
hinges which are not represented in \eqref{eq:gen-quadratic-model},
resulting in the observed slowdown. 

Bundle method solvers, by contrast, exhibit slow initial progress but
tend to be competitive asymptotically. This is because they build a
piecewise linear lower bound of the objective function, which initially
is not very good but through successive tightening eventually becomes a
faithful model. To take advantage of this we are contemplating hybrid
solvers that switch over from sub(L)BFGS to a bundle method as
appropriate.

While bundle methods like BMRM have an exact, implementable stopping
criterion based on the duality gap, no such stopping criterion exists
for BFGS and other quasi-Newton algorithms. Therefore, it is customary
to use the relative change in function value as an implementable
stopping criterion. Developing a stopping criterion for sub(L)BFGS based on
duality arguments remains an important open question. 

sub(L)BFGS relies on an efficient exact line search. We
proposed such line searches for the multiclass hinge loss and its
extension to the multilabel setting, based on a conceptually simple yet
optimal algorithm to segment the pointwise maximum of lines. A crucial
assumption we had to make is that the number $|\!\Zcal\!|$ of labels is
manageable, as it takes $O(|\!\Zcal\!| \log |\!\Zcal\!|)$ time to
identify the hinges associated with each training instance. In certain
structured prediction problems \citep{TsoJoaHofAlt05} which have
recently gained prominence in machine learning, the set $\Zcal$ could be
exponentially large\,---\,for instance, predicting binary labels on a chain
of length $n$ produces $2^n$ possible labellings. Clearly our line
searches are not efficient in such cases; we are investigating trust
region variants of sub(L)BFGS to bridge this gap.

 
Finally, to put our contributions in perspective, recall that we
modified three aspects of the standard BFGS algorithm, namely the
quadratic model (Section~\ref{sec:gen-model}), the descent direction
finding (Section~\ref{sec:find-descent-dir}), and the Wolfe conditions
(Section~\ref{sec:gen-ls}). Each of these modifications is versatile
enough to be used as a component in other nonsmooth optimization
algorithms. This not only offers the promise of improving existing
algorithms, but may also help clarify connections between them. We hope
that our research will focus attention on the core subroutines that need
to be made more efficient in order to handle larger and larger datasets.

\section*{Acknowledgments}

A short version of this paper was presented at the 2008 ICML conference
\citep{YuVisGunSch08b}.  We thank Choon Hui Teo for many useful
discussions and help with implementation issues, Xinhua Zhang for
proofreading our manuscript, and the anonymous reviewers of both ICML
and JMLR for their useful feedback which helped improve this paper.
We thank John R.\ Birge for pointing us to his work \citep{BirQiWei98} which
led us to the convergence proof in Appendix~\ref{sec:convergence-proof}.

This publication only reflects the authors' views.
All authors were with NICTA and the Australian National University for parts
of their work on it. NICTA is funded by the Australian Government's
Backing Australia's Ability and Centre of Excellence programs.
This work was also supported in part by the IST Programme of the European
Community, under the PASCAL2 Network of Excellence, IST-2007-216886.

\newpage
\bibliography{subBFGS}
\newpage
\appendix

\section{Bundle Search for a Descent Direction}
\label{sec:LineSearchDescent}

Recall from Section~\ref{sec:find-descent-dir} that at a subdifferential
point $\vw$ our goal is to find a descent direction $\vp^*$ which
minimizes the pseudo-quadratic model:\footnote{For ease of
  exposition we are suppressing the iteration index $t$ here.}
\begin{align}
  M(\vp) := \half \vp^{\top} \Bmat^{-1} \vp + \sup_{\vg
  \in \partial J(\vw)} \vg^{\top} \vp.
\label{eq:model-without-t}
\end{align}
This is generally intractable due to the presence of a supremum
over the entire subdifferential $\partial J(\vw)$. We therefore propose
a bundle-based descent direction finding procedure
(Algorithm~\ref{alg:find-descent-dir-cg}) which progressively
approaches $M(\vp)$ from below via a series of convex functions
$M^{(1)} (\vp),\cdots, M^{(i)} (\vp)$, each taking the same form as $M(\vp)$
but with the supremum defined over a countable subset of $\partial J(\vw)$.
At iteration $i$ our convex lower bound $M^{(i)} (\vp)$ takes the form
\begin{align}
  \label{eq:subgrad-set}
  M^{(i)} (\vp) &:= \half \vp^{\top} \Bmat^{-1} \vp + \sup_{\vg
    \in \Vcal^{(i)} } \vg^{\top} \vp, \mbox{~where~} \nonumber \\
  \Vcal^{(i)}  &:= \{\vg^{(j)}: j \leq i, ~i,j~\in \NN\} \subseteq \partial J(\vw).
\end{align}
Given an iterate $\vp^{(j-1)}\in \RR^{d}$ we find a
\emph{violating subgradient} $\vg^{(j)}$ via
\begin{align}
  \label{eq:violating-grad}
  \vg^{(j)} :=\argsup_{\vg \in \partial J(\vw)} \vg^{\top} \vp^{(j-1)}. 
\end{align} 
Violating subgradients recover the true objective $M(\vp)$ at the iterates $\vp^{(j-1)}$:
\begin{align}
  \label{eq:M-meet-M}
  M(\vp^{(j-1)}) = M^{(j)}(\vp^{(j-1)}) = \half  \vp^{(j-1) \top} \Bmat^{-1}
  \vp^{(j-1)} + \vg^{(j) \top} \vp^{(j-1)}.
\end{align}

To produce the iterates $\vp^{(i)}$, we rewrite $\min_{\vp \in \RR^d} M^{(i)}
(\vp)$ as a constrained optimization problem \eqref{eq:constrained-opt-relax},
which allows us to write the Lagrangian of \eqref{eq:subgrad-set} as
\begin{align}
  \label{eq:lagrange}
  L^{(i)} (\vp, \xi, \val) := \half \vp^{\top}\Bmat^{-1}\vp + \xi -
  {\val}^{\top}(\xi\vone - {{\Gmat}^{(i)} }^{\top}\vp), \hspace{5pt}
\end{align}
where $\Gmat^{(i)} := [\vg^{(1)}\! ,\, \vg^{(2)}\!,\, \ldots,\, \vg^{(i)} ] \in
\RR^{d\times i}$ collects past violating subgradients, and $\val$ is a column
vector of non-negative Lagrange multipliers. Setting the derivative of
\eqref{eq:lagrange} with respect to the primal variables $\xi$ and $\vp$ to zero
yields, respectively,
\begin{align}
  {\val}^{\top}\vone = 1 \text{~~~and}
  \label{eq:grad-xi}
\end{align}
\begin{align}
  \vp = -\Bmat\Gmat^{(i)} \val.
  \label{eq:grad-p}
\end{align}
The primal variable $\vp$ and the dual variable $\val$ are related via
the dual connection \eqref{eq:grad-p}. To eliminate the primal variables
$\xi$ and $\vp$, we plug \eqref{eq:grad-xi} and \eqref{eq:grad-p} back
into the Lagrangian to obtain the dual of $M^{(i)} (\vp)$:
\begin{align}
  \label{eq:dual}
  &D^{(i)} (\val):= -\half  (\Gmat^{(i)} \val)^{\top} \Bmat
  (\Gmat^{(i)} \val),\\
  &\mbox{~s.t.~} \val \in [0,1]^i,~ \norm{\val}_1 = 1. \nonumber
\end{align}
The dual objective $D^{(i)} (\val)$ (\emph{resp.}\ primal objective
$M^{(i)} (\vp)$) can be maximized (\emph{resp.}\ minimized) exactly via
quadratic programming. However, doing so may incur substantial
computational expense. Instead we adopt an iterative scheme which is
cheap and easy to implement yet guarantees dual improvement.

Let $\val^{(i)} \in [0,1]^i$ be a feasible solution for $D^{(i)}
(\val)$.\footnote{Note that $\val^{(1)}\!  = \vone$ is a feasible solution for
  $D^{(1)} (\val)$.} The corresponding primal solution $\vp^{(i)} $ can be found
by using \eqref{eq:grad-p}. This in turn allows us to compute the next violating
subgradient $\vg^{(i+1)} $ via \eqref{eq:violating-grad}. With the new violating
subgradient the dual becomes
\begin{align}
  \nonumber
  D^{(i+1)}(\val) ~:=\, & -\half  (\Gmat^{(i+1)} \val)^{\top} \Bmat
  (\Gmat^{(i+1)} \val),\\
  \mbox{s.t.} & ~~\val \in [0,1]^{i+1},~ \norm{\val}_1 = 1,
  \label{eq:dual-iplusone}
\end{align}
where the subgradient matrix is now extended:
\begin{align}
\Gmat^{(i+1)}  = [\Gmat^{(i)}\!,\, \vg^{(i+1)} ].
\label{eq:subgrad-matrix}
\end{align}
Our iterative strategy constructs a new feasible solution
$\val \in [0,1]^{i+1}$ for \eqref{eq:dual-iplusone} by
constraining it to take the following form:
\begin{align}
  \val = \left[ \!\begin{array}{c} (1-\mu)\val^{(i)} \\ \mu \end{array}\! \right],
  \text{~~where~~} \mu \in [0,1].
  \label{eq:alpha-update}
\end{align}
In other words, we maximize a one-dimensional function 
$\bar{D}^{(i+1)} : [0,1] \rightarrow \RR$:
\begin{align}
  \label{eq:def-Dbar}
  \bar{D}^{(i+1)} (\mu) & ~:=~
  -\half  \left(\Gmat^{(i+1)} \val\right)^{\top}
  \Bmat\left(\Gmat^{(i+1)} \val\right) \\
  & ~~=~ -\half \left((1-\mu)\bar{\vg}^{(i)}  +
    \mu\vg^{(i+1)} \right)^{\top}\Bmat\left((1-\mu)\bar{\vg}^{(i)}  +
    \mu\vg^{(i+1)} \right),\nonumber
\end{align}
where
\begin{align}
\bar{\vg}^{(i)}  := \Gmat^{(i)} \val^{(i)}  \in \partial J(\vw)
\label{eq:gi-bar}
\end{align}
lies in the convex hull of $~\vg^{(j)} \in \partial J(\vw)~\forall j \le i$
(and hence in the convex set $\partial J(\vw)$) because $~\val^{(i)} 
\in [0,1]^i$ and $\norm{\val^{(i)} }_1 = 1$.  Moreover, $\mu\in[0,1]$ ensures
the feasibility of the dual solution. Noting that $\bar{D}^{(i+1)} (\mu)$
is a concave quadratic function, we set
\begin{align}
\partial \bar{D}^{(i+1)} (\mu) & = \left(\bar{\vg}^{(i)} -
\vg^{(i+1)} \right)^{\top}\Bmat\left((1-\eta)\bar{\vg}^{(i)}  +
\eta\vg^{(i+1)} \right)  = 0
\label{eq:mu-grad}
\end{align}
to obtain the optimum
\begin{align}
  \label{eq:meta-step}
  \!\!\!\mu^* := \argmax_{\mu \in [0,1]} \bar{D}^{(i+1)} (\mu) =
  \min\left(1, \max\left (0, \frac{(\bar{\vg}^{(i)} - \vg^{(i+1)}
        )^{\top} \Bmat \bar{\vg}^{(i)} }{(\bar{\vg}^{(i)} -
        \vg^{(i+1)} )^{\top} \Bmat (\bar{\vg}^{(i)} - \vg^{(i+1)}
        )}\right ) \right ).\!\!
\end{align}
Our dual solution at step $i+1$ then becomes
\begin{align}
\val^{(i+1)}  := \left[ \!\begin{array}{c} (1-\mu^*)\val^{(i)} \\
    \mu^* \end{array}\! \right]. 
\label{eq:alpha-iplusone}
\end{align} 
Furthermore, from \eqref{eq:subgrad-matrix}, \eqref{eq:alpha-update},
and \eqref{eq:gi-bar} it follows that $\bar{\vg}^{(i)} $ can be
maintained via an incremental update (Line 8 of
Algorithm~\ref{alg:find-descent-dir-cg}):
\begin{align}
  \bar{\vg}^{(i+1)}  := \Gmat^{(i+1)} \val^{(i+1)} = (1-\mu^*)\bar{\vg}^{(i)} +
  \mu^* \vg^{(i+1)} ,
  \label{eq:gbar-incr}
\end{align}
which combined with the dual connection \eqref{eq:grad-p}
yields an incremental update for the primal solution (Line 9 of
Algorithm~\ref{alg:find-descent-dir-cg}):
\begin{align}
  \nonumber
  \vp^{(i+1)} := -\Bmat\bar{\vg}^{(i+1)} & = -(1-\mu^*)\Bmat\bar{\vg}^{(i)} 
  -\mu^* \Bmat\vg^{(i+1)} \\
  & = (1-\mu^*)\vp^{(i)}  -\mu^* \Bmat\vg^{(i+1)} .
  \label{eq:pi-update}
\end{align}
Using \eqref{eq:gbar-incr} and \eqref{eq:pi-update}, computing a primal
solution (Lines 7--9 of Algorithm~\ref{alg:find-descent-dir-cg}) costs a
total of $O(d^2)$ time (\emph{resp.}\ $O(md)$ time for LBFGS with buffer
size $m$), where $d$ is the dimensionality of the optimization problem.
Note that maximizing $D^{(i+1)} (\val)$ directly via quadratic programming
generally results in a larger progress than that obtained by our approach.

In order to measure the quality of our solution at iteration $i$, we define
the quantity
\begin{align}
  \epsilon^{(i)} ~:=~ \min_{j\le i} M^{(j+1)}(\vp^{(j)}) - D^{(i)}(\val^{(i)})
  ~=~ \min_{j\le i} M(\vp^{(j)}) - D^{(i)}(\val^{(i)}),
  \label{eq:def-eps}
\end{align}
where the second equality follows directly from
\eqref{eq:M-meet-M}. Let $D(\val)$ be the corresponding dual problem
of $M(\vp)$, with the
property $D\left( \left[ \!\!\begin{array}{l} {}^{\val^{(i)}} \\[-2.2ex]
      {}_{~\vzero} \end{array}\!\!\! \right] \right) = D^{(i)}
(\val^{(i)} )$, and let $\val^*$ be the optimal solution to
$\argmax_{\val \in \Acal}D(\val)$ in some domain $\Acal$ of
interest. As a consequence of the weak duality theorem \citep[Theorem
XII.2.1.5]{HirLem93}, $\min_{\vp \in \RR^d}M(\vp) \ge
D(\val^*)$. Therefore \eqref{eq:def-eps} implies that
\begin{align}
 \epsilon^{(i)} ~\geq~ \min_{\vp \in \RR^d}M(\vp) - D^{(i)}
  (\val^{(i)}) ~\ge~ \min_{\vp \in \RR^d}M(\vp) -  D(\val^*)
  ~\ge~ 0.
  \label{eq:eps-nonnegative}
\end{align}
The second inequality essentially says that $\epsilon^{(i)}$ is an
upper bound on the duality gap. In fact,
Theorem~\ref{th:updateguarantee} below shows that $(\epsilon^{(i)}
-\epsilon^{(i+1)} )$ is bounded away from 0, \ie $\epsilon^{(i)} $ is
monotonically decreasing. This guides us to design a practical
stopping criterion (\DIRSTOP~of Algorithm~\ref{alg:find-descent-dir-cg})
for our direction-finding procedure. Furthermore, using the dual
connection \eqref{eq:grad-p}, we can derive an implementable formula
for $\epsilon^{(i)} $:
\begin{align}
  \epsilon^{(i)}~  & =~ \min_{j\le i}\left[\half \vp^{(j) \top} \Bmat^{-1} \vp^{(j)}
  + \vp^{(j) \top} \vg^{(j+1)} + \half  (\Gmat^{(i)} \val^{(i)} )^{\top}
  \Bmat
  (\Gmat^{(i)} \val^{(i)} ) \right ] \nonumber \\
  & =~ \min_{j\le i}\left[-\half \vp^{(j) \top} \bar{\vg}^{(j)} +
  \vp^{(j) \top} \vg^{(j+1)} - \half \vp^{(i) \top} \bar{\vg}^{(i)} \right] \nonumber \\
  & =~ \min_{j\le i}\left[ \vp^{(j) \top} \vg^{(j+1)} - \label{eq:eps-up}
  \half (\vp^{(j) \top} \bar{\vg}^{(j)} + {\vp^{(i)} }^{\top}\bar{\vg}^{(i)} )\right],\\
  & \text{~~~~~where~~} \vg^{(j+1)} := \argsup_{\vg \in \partial J(\vw)}
  \vg^{\top} \vp^{(j)} \mbox{~~and~~\,} \bar{\vg}^{(j)} :=
  \Gmat^{(j)} \val^{(j)} ~~\forall j \le i . \nonumber
\end{align}
It is worth noting that continuous progress in the dual objective
value does not necessarily prevent an increase in the primal objective
value, \ie it is possible that $M(\vp^{(i+1)} ) \ge M(\vp^{(i)}
)$. Therefore, we choose the best primal solution so far, 
\begin{align}
\label{eq:best_p}
\vp := \argmin_{j\le i } M(\vp^{(j)}),
\end{align}
as the search direction (\DIRRETURN~of
Algorithm~\ref{alg:find-descent-dir-cg}) for the parameter update
\eqref{eq:update}. This direction is a direction of descent as long as
the last iterate $\vp^{(i)} $ fulfills the descent condition
\eqref{eq:descent-dir}. To see this, we use
(\ref{eq:g-iplusone2}--\ref{eq:Di-at-ali}) below to get $\sup_{\vg
  \in \partial J(\vw)} \vg^{\top}\vp^{(i)} = M(\vp^{(i)}) +
D^{(i)}(\val^{(i)})$, and since
\begin{align}
M(\vp^{(i)}) \ge \min_{j \le i}
M(\vp^{(j)})  \mbox{~~and~~} D^{(i)}(\val^{(i)}) \ge D^{(j)}(\val^{(j)})~
~\forall j \le i,
\end{align} 
definition \eqref{eq:best_p} immediately gives
$\sup_{\vg \in \partial J(\vw)} \vg^{\top}\vp^{(i)} ~\ge~ \sup_{\vg
  \in \partial J(\vw)} \vg^{\top}\vp$. Hence if $\vp^{(i)}$ is a
descent direction, then so is $\vp$.

We now show that if the current parameter vector $\vw$ is not optimal,
then a direction-finding tolerance $\epsilon \ge 0$ exists for
Algorithm~\ref{alg:find-descent-dir-cg} such that the returned search
direction $\vp$ is a descent direction, \ie $\sup_{\vg \in \partial
  J(\vw)} \vg^{\top}\vp < 0$.

\begin{lemma}
  \label{lem:gp-upperbound}
  Let $\Bmat$ be the current approximation to the inverse Hessian maintained by
  Algorithm~\ref{alg:subbfgs}, and $h > 0$ a lower bound on the
  eigenvalues of $\Bmat$.  If the current iterate $\vw$ is not optimal: $\zero
  \notin \partial J(\vw)$, and the number of direction-finding iterations is
  unlimited ($k_{\text{max}} = \infty$), then there exists a
  direction-finding tolerance $\epsilon \ge 0$ such that the descent direction
  $\vp = -\Bmat\bar{\vg},~\bar{\vg} \in \partial J(\vw)$ returned by
  Algorithm~\ref{alg:find-descent-dir-cg} at $\vw$ satisfies $\sup_{\vg
    \in \partial J(\vw)} \vg^{\top}\vp < 0$.
\end{lemma}
\begin{proof}
  Algorithm~\ref{alg:find-descent-dir-cg} returns $\vp$ after $i$
  iterations when $\epsilon^{(i)} \le \epsilon$, where $\epsilon^{(i)} =
  M(\vp) - D^{(i)}(\val^{(i)})$ by definitions \eqref{eq:def-eps} and
  \eqref{eq:best_p}. Using definition \eqref{eq:dual} of $ 
  D^{(i)}(\val^{(i)})$, we have
  \begin{align}
    -D^{(i)}(\val^{(i)}) ~=~ \half (\Gmat^{(i)} \val^{(i)})^{\top} \!\Bmat
    (\Gmat^{(i)} \val^{(i)}) ~=~ \half \bar{\vg}^{(i)\top} \!\Bmat \bar{\vg}^{(i)},
    \label{eq:Dai-gBg}
  \end{align}
  where $\bar{\vg}^{(i)} = \Gmat^{(i)} \val^{(i)}$ is a subgradient in
  $\partial J(\vw)$. On the other hand, using
  \eqref{eq:model-without-t} and \eqref{eq:pi-update}, one can write
  \begin{align}
    M(\vp) &~= \sup_{\vg \in \partial
      J(\vw)} \vg^{\top}\vp ~+~ \half \vp^{\top}\Bmat^{-1}\vp  \nonumber \\
    &~= \sup_{\vg \in \partial J(\vw)} \vg^{\top}\vp ~+~\half
    \bar{\vg}^{\top} \Bmat \bar{\vg}, \mbox{~~~where~~~} \bar{\vg}
    \in \partial J(\vw).
    \label{eq:Mp-neg}
  \end{align}
  Putting together \eqref{eq:Dai-gBg} and \eqref{eq:Mp-neg}, and using
  $\Bmat \succ h$, one obtains 
  \begin{align}
    \epsilon^{(i)}  \,= \!\!\sup_{\vg \in \partial J(\vw)}\!\! \vg^{\top} \!\vp
    \,+\, \half \bar{\vg}^{\top} \!\Bmat \bar{\vg} \,+\, \half
    \bar{\vg}^{(i)\top} \!\Bmat
    \bar{\vg}^{(i)} \,\geq \!\!\sup_{\vg \in \partial J(\vw)}\!\! \vg^{\top} \!\vp \,+\,
    \frac{h}{2} \|\bar{\vg}\|^2 + \,\frac{h}{2} \|\bar{\vg}^{(i)}\|^2.
    \label{eq:epsilon-i}
  \end{align}
  Since $\vzero \notin \partial J(\vw)$, the last two terms of
  \eqref{eq:epsilon-i} are strictly positive; and by \eqref{eq:eps-nonnegative},
  $\epsilon^{(i)} \ge 0$ . The claim follows by choosing an
  $\epsilon$ such that $(\forall i)~~\frac{h}{2} ( \|\bar{\vg}\|^{2} +
  \|\bar{\vg}^{(i)}\|^{2}) > \epsilon \geq \epsilon^{(i)} \ge 0$.
\end{proof}

Using the notation from Lemma~\ref{lem:gp-upperbound}, we
show in the following corollary that a stricter upper bound on
$\epsilon$ allows us to bound $\sup_{\vg \in \partial J(\vw)}
\vg^{\top}\vp$ in terms of $\bar{\vg}^{\top} \Bmat \bar{\vg}$ and
$\norm{\bar{\vg}}$. This will be used in
Appendix~\ref{sec:convergence-proof} to establish the global convergence
of the subBFGS algorithm.

\begin{corollary}
  Under the conditions of Lemma~\ref{lem:gp-upperbound}, there exists an
  $\epsilon \ge 0$ for Algorithm~\ref{alg:find-descent-dir-cg} such that the
  search direction $\vp$ generated by Algorithm~\ref{alg:find-descent-dir-cg}
  satisfies
  \begin{align}
    \sup_{\vg \in \partial J(\vw)}\!\! \vg^{\top} \!\vp \,\le -\half
    \bar{\vg}^{\top} \!\Bmat \bar{\vg} \,\le -\frac{h}{2} \norm{\bar{\vg}}^2 <~ 0.
    \label{eq:supgp-pos-bound}
  \end{align}
\label{cor:supgp-pos-bound}
\end{corollary}

\begin{proof}
  Using \eqref{eq:epsilon-i}, we have
  \begin{align}
    (\forall i)~~\epsilon^{(i)} \ge \!\!\sup_{\vg \in \partial J(\vw)}\!\! \vg^{\top} \!\vp \,+\,
    \half \bar{\vg}^{\top} \!\Bmat \bar{\vg} \,+ \frac{h}{2} \|\bar{\vg}^{(i)}\|^{2}.
  \end{align}
  The first inequality in \eqref{eq:supgp-pos-bound} results from choosing
  an $\epsilon$ such that
  \begin{align}
    (\forall i)~~\frac{h}{2} \|\bar{\vg}^{(i)}\|^{2} \,\geq~ \epsilon ~\geq~
    \epsilon^{(i)} \geq~ 0.
    \label{eq:tighter-bound-eps}
  \end{align}
  The lower bound $h >0$ on the spectrum of $\Bmat$ yields the second
  inequality in \eqref{eq:supgp-pos-bound}, and the third follows from the
  fact that $\norm{\bar{\vg}} > 0$ at non-optimal iterates.
\end{proof}

\section{Convergence of the Descent Direction Search}
\label{sec:LineSearchConvergence}

Using the notation established in
Appendix~\ref{sec:LineSearchDescent}, we now prove the convergence of
Algorithm~\ref{alg:find-descent-dir-cg} via several technical
intermediate steps. The proof shares similarities with the proofs
found in \citet{SmoVisLe07}, \citet{ShaSin08}, and
\citet{WarGloVis08}. The key idea is that at each iterate
Algorithm~\ref{alg:find-descent-dir-cg} decreases the upper bound
$\epsilon^{(i)}$ on the distance from the optimality, and the decrease
in $\epsilon^{(i)} $ is characterized by the recurrence
$\epsilon^{(i)} - \epsilon^{(i+1)} ~\geq~ c (\epsilon^{(i)})^2$ with
$c >0$ (Theorem \ref{th:updateguarantee}). Analysing this recurrence
then gives the convergence rate of the algorithm (Theorem
\ref{th:main}).

We first provide two technical lemmas (Lemma \ref{le:eps-less-grad} and
\ref{lem:linquad}) that are needed to prove Theorem
\ref{th:updateguarantee}.  
\begin{lemma}
  Let $\bar{D}^{(i+1)} (\mu)$ be the one-dimensional function defined in
  \eqref{eq:def-Dbar}, and $\epsilon^{(i)}$ the positive measure defined in
  \eqref{eq:def-eps}. Then
  $\epsilon^{(i)}  \,\leq~ \partial \bar{D}^{(i+1)}(0)$.
\label{le:eps-less-grad}
\end{lemma}

\begin{proof}
  Let $\vp^{(i)} $ be our primal solution at iteration $i$, derived from the
  dual solution $\val^{(i)} $ using the dual connection
  \eqref{eq:grad-p}. We then have
  \begin{align}
    \vp^{(i)}  = -\Bmat\bar{\vg}^{(i)}\!,\, \mbox{~~where~~~} \bar{\vg}^{(i)}  :=\,
    \Gmat^{(i)} \val^{(i)} .
    \label{eq:pi}
  \end{align}
  Definition \eqref{eq:model-without-t} of $M(\vp)$ implies that
  \begin{align}
    M(\vp^{(i)} ) ~=~ \half {\vp^{(i)} }^{\top}\Bmat^{-1}\vp^{(i)}  +
    {{\vp}^{(i)} }^{\top}{\vg^{(i+1)} },
    \label{eq:mpi}
  \end{align}
  where 
\begin{align}
  \vg^{(i+1)} \, :=~ \argsup_{\vg \in \partial J(\vw)}
  \vg^{\top}\vp^{(i)} .
\label{eq:g-iplusone2}
\end{align}
Using \eqref{eq:pi}, we have $\Bmat^{-1}\vp^{(i)}  =
-\Bmat^{-1}\Bmat\bar{\vg}^{(i)}  = -\bar{\vg}^{(i)}$\!,\, and hence
\eqref{eq:mpi} becomes
  \begin{align}
    M(\vp^{(i)} ) ~=~ {{\vp}^{(i)} }^{\top}{\vg^{(i+1)} } -
    \half {\vp^{(i)} }^{\top}\bar{\vg}^{(i)} .
    \label{eq:M-at-pi}
  \end{align}
  Similarly, we have
\begin{align}
  D^{(i)} (\val^{(i)} ) ~=~ -\half  (\Gmat^{(i)} \val^{(i)} )^{\top} \Bmat
  (\Gmat^{(i)} \val^{(i)} ) ~=~  \half {\vp^{(i)} }^{\top}\bar{\vg}^{(i)} .
  \label{eq:Di-at-ali}
\end{align}
From \eqref{eq:mu-grad} and \eqref{eq:pi} it follows that
\begin{align}
  \partial \bar{D}^{(i+1)} (0) ~=~ (\bar{\vg}^{(i)}  - \vg^{(i+1)} )^{\top} \Bmat
  {\bar{\vg}^{(i)} } ~=~ (\vg^{(i+1)}  -\bar{\vg}^{(i)} )^{\top}\vp^{(i)} ,
  \label{eq:partial-D-at-0}
\end{align}
where $\vg^{(i+1)} $ is a violating subgradient chosen via
\eqref{eq:violating-grad}, and hence coincides with
\eqref{eq:g-iplusone2}.  Using
\eqref{eq:M-at-pi}\,--\,\eqref{eq:partial-D-at-0}, 
we obtain
\begin{align}
  M(\vp^{(i)} ) - D^{(i)} (\val^{(i)} ) ~=~ \left({\vg^{(i+1)} } -
    \bar{\vg}^{(i)} \right)^{\top}{\vp^{(i)}} ~=~ \partial \bar{D}^{(i+1)} (0).
  \label{eq:mpi-di}
\end{align}
Together with definition \eqref{eq:def-eps} of $\epsilon^{(i)}$,
\eqref{eq:mpi-di}  implies that
\begin{align*}
\nonumber
\epsilon^{(i)} & \,=~ \min_{j\le i} M(\vp^{(j)}) - D^{(i)} \left( \val^{(i)} \right) \\
  & \,\leq~ M(\vp^{(i)}) - D^{(i)}(\val^{(i)}) ~=~ \partial \bar{D}^{(i+1)}(0).
\end{align*}
\end{proof}

\begin{lemma}
  \label{lem:linquad}
  Let $f: [0,1] \to \RR$ be a concave quadratic function with $f(0) =
  0$, $\partial f(0) \in [0, a]$, and $\partial f^2(x) \geq -a$ for some
  $a \geq 0$. Then $\max_{x \in [0,1]} f(x) \geq \frac{ (\partial
    f(0))^2}{2a}$.
\end{lemma}

\begin{proof}
  Using a second-order Taylor expansion around 0, we have $f(x) \ge \partial
  f(0)x - \frac{a}{2} x^2$. $x^* = \partial f(0)/a$ is the
  unconstrained maximum of the lower bound. Since $\partial f(0) \in
  [0, a]$, we have $x^* \in [0,1]$.  Plugging $x^*$ into the lower
  bound yields $(\partial f(0))^2/(2a)$.
\end{proof}

\begin{theorem}
  \label{th:updateguarantee}
  Assume that at $\vw$ the convex objective function $J: \RR^d
  \rightarrow \RR$ has bounded subgradient: $\norm{\partial
    J(\vw)} \leq G$, and that the approximation $\Bmat$ to the inverse
    Hessian has bounded eigenvalues: $\Bmat \preceq H$. Then
  \begin{align*}
    \epsilon^{(i)}  - \epsilon^{(i+1)}  ~\geq~ \frac{(\epsilon^{(i)} )^2}{8G^2H}.
  \end{align*}
\end{theorem}

\begin{proof}
  Recall that we constrain the form of feasible dual solutions for
  $D^{(i+1)} (\val)$ as in \eqref{eq:alpha-update}.
  Instead of $D^{(i+1)} (\val)$, we thus work
  with the one-dimensional concave quadratic function
  $\bar{D}^{(i+1)}(\mu)$ \eqref{eq:def-Dbar}.
  It is obvious that
  $\left[ \!\!\begin{array}{l} {}^{\val^{(i)}} \\[-2.2ex]  {}_{~0} \end{array}\!\!\! \right]$
  is a feasible solution for
  $D^{(i+1)} (\val)$. In this case, $\bar{D}^{(i+1)} (0) = D^{(i)} (\val^{(i)} )$.
  \eqref{eq:alpha-iplusone} implies that $\bar{D}^{(i+1)} (\mu^*) =
  D^{(i+1)} (\val^{(i+1)} )$. Using the definition \eqref{eq:def-eps} of
  $\epsilon^{(i)}$, we thus have
  \begin{align}
    \epsilon^{(i)}  - \epsilon^{(i+1)}  ~\geq~  D^{(i+1)} (\val^{(i+1)} ) -
    D^{(i)} (\val^{(i)} ) ~=~ \bar{D}^{(i+1)} (\mu^*) - \bar{D}^{(i+1)} (0).
    \label{eq:eps-diff}
  \end{align}
  It is easy to see from \eqref{eq:eps-diff} that $\epsilon^{(i)}  - \epsilon^{(i+1)}$
  are upper bounds on the maximal value of the concave quadratic function
  $f(\mu) := \bar{D}^{(i+1)} (\mu)-\bar{D}^{(i+1)} (0)$ with $\mu\in [0,1]$
  and $f(0) = 0$. Furthermore, the definitions of $\bar{D}^{(i+1)} (\mu)$ and
  $f(\mu)$ imply that
  \begin{align}
    \label{eq:f-grad-hessian}
    \partial f(0) & ~=~~ \partial\bar{D}^{(i+1)} (0) \, ~=~ (\bar{\vg}^{(i)}  -
    \vg^{(i+1)} )^{\top} \Bmat
    {\bar{\vg}^{(i)} } \text{~~~and} \\
    \partial^2 f(\mu) & ~=~ \partial^2\bar{D}^{(i+1)} (\mu) ~=~ -(
    \bar{\vg}^{(i)} -\vg^{(i+1)} )^{\top} \Bmat(\bar{\vg}^{(i)}  -
    \vg^{(i+1)} ). \nonumber
  \end{align}
  Since $\norm{\partial J(\vw)} \le G$ and $\bar{\vg}^{(i)}  \in \partial
  J(\vw)$ \eqref{eq:gi-bar}, we have $\norm{\bar{\vg}^{(i)}  - \vg^{(i+1)}}
  \leq 2G$. Our upper bound on the spectrum of $\Bmat$ then gives
  $|\partial f(0)| \leq 2G^2H$ and $\left|\partial^2f(\mu)\right| \leq
  4G^2H$. Additionally, Lemma~\ref{le:eps-less-grad} and the fact that
  $\Bmat \succeq 0$ imply that
  \begin{align}
    \label{eq:f-grad-hessian-sign}
    \partial f(0) ~=~ \partial\bar{D}^{(i+1)} (0) \geq 0 \mbox{~~~and~~~}
    \partial^2 f(\mu) ~=~ \partial^2\bar{D}^{(i+1)} (\mu) \leq 0,
  \end{align}
  which means that
  \begin{align}
    \label{eq:bounds}
  \partial f(0) ~\in~ [0,2G^2H] \subset [0, 4G^2H] \mbox{~~~and~~~}
  \partial^2f(\mu) ~\ge\, -4G^2H.
  \end{align}
  Invoking Lemma~\ref{lem:linquad}, we immediately get
  \begin{align}
    \epsilon^{(i)}  - \epsilon^{(i+1)}  ~\geq~ \frac{(\partial
        f(0))^2}{8G^2H} ~=~ \frac{(\partial\bar{D}^{(i+1)} (0))^2}{8G^2H}.
    \label{eq:eps-eps}
  \end{align}
  Since $\epsilon^{(i)}  \leq \partial \bar{D}^{(i+1)} (0)$ by
  Lemma~\ref{le:eps-less-grad}, the inequality
  \eqref{eq:eps-eps} still holds when $\partial\bar{D}^{(i+1)} (0)$ is
  replaced with $\epsilon^{(i)}$.
\end{proof}

\eqref{eq:f-grad-hessian} and \eqref{eq:f-grad-hessian-sign}
imply that the optimal combination coefficient $\mu^*$
\eqref{eq:meta-step} has the property
\begin{align}
  \mu^* = \min\left[1,
    \frac{\partial\bar{D}^{(i+1)} (0)}{-\partial^2\bar{D}^{(i+1)} (\mu)}\right].
  \label{eq:meta-step-simple}
\end{align}
Moreover, we can use \eqref{eq:grad-p} to reduce the cost of
computing $\mu^*$ by setting $\Bmat\bar{\vg}^{(i)}$ in
\eqref{eq:meta-step} to be $-\vp^{(i)}$ (Line 7 of
Algorithm~\ref{alg:find-descent-dir-cg}), and calculate 
\begin{align}
  \mu^* = \min\left[1, \frac{\vg^{(i+1) \top} \vp^{(i)} -\bar{\vg}^{(i)
        \top} \vp^{(i)} } { \vg^{(i+1) \top} \Bmat_t \vg^{(i+1)} +
      2~\vg^{(i+1) \top}\vp^{(i)} - \bar{\vg}^{(i) \top} \vp^{(i)}} \right],
  \label{eq:meta-step-simple2}
\end{align}
where $\Bmat_t \vg^{(i+1)}$ can be cached for the update of the primal
solution at Line 9 of Algorithm~\ref{alg:find-descent-dir-cg}.

To prove Theorem~\ref{th:main}, we use the following lemma proven by
induction by \citet[Sublemma 5.4]{AbeTakWar01}:
\begin{lemma}
  Let $\{\epsilon^{(1)}\!,\, \epsilon^{(2)}\!,\, \cdots \}$ be a
  sequence of non-negative numbers satisfying $\forall i \in \NN$ the
  recurrence
  \begin{align*}
    \epsilon^{(i)} - \epsilon^{(i+1)} \,\geq~ c \,(\epsilon^{(i)} )^2,
  \end{align*}
  where $c \in \RR_+$ is a positive constant. Then $\forall i \in \NN$
  we have
  \begin{align*}
    \epsilon^{(i)} ~\leq~ \frac{1}{c\left(i + \frac{1}{\epsilon^{(1)}
          c}\right)}.
  \end{align*}
  \label{le:eps-sequence}
\end{lemma}
We now show that Algorithm~\ref{alg:find-descent-dir-cg} decreases
$\epsilon^{(i)} $ to a pre-defined tolerance $\epsilon$ in
$O(1/\epsilon)$ steps:
\begin{theorem}
  \label{th:main}
  Under the assumptions of Theorem~\ref{th:updateguarantee},
  Algorithm~\ref{alg:find-descent-dir-cg} converges to
  the desired precision $\epsilon$ after
  \begin{align*}
   1 ~\leq~  t ~\leq~ \frac{8 G^2 H}{\epsilon} - 4
  \end{align*}
  steps for any $\epsilon < 2 G^2 H$.
\end{theorem}
\begin{proof}
  Theorem~\ref{th:updateguarantee} states that
  \begin{align}
  \label{eq:update-guarantee}
    \epsilon^{(i)}  - \epsilon^{(i+1)} \,\geq~ \frac{(\epsilon^{(i)} )^2}{8G^2H},
  \end{align}
  where $\epsilon^{(i)}$ is non-negative  $\forall i \in \NN$ by
  \eqref{eq:eps-nonnegative}. Applying
  Lemma~\ref{le:eps-sequence} we thus obtain
  \begin{align}
    \label{eq:eps-seq}
    \epsilon^{(i)}  ~\leq~ \frac{1}{c \left(i + \frac{1}{\epsilon^{(1)} 
          c}\right)}, \mbox{~~~where~~~} c := \frac{1}{8G^2H}.
  \end{align}
  Our assumptions on $\norm{\partial J(\vw)}$ and the spectrum of $\Bmat$
  imply that
  \begin{align}
    \label{eq:Dbound}
    \bar{D}^{(i+1)} (0) ~=~ (\bar{\vg}^{(i)}  - \vg^{(i+1)} )^{\top} \Bmat
    \bar{\vg}^{(i)} ~\leq~ 2G^2H. 
  \end{align}
  Hence $\epsilon^{(i)}  \leq 2G^2H$ by
  Lemma~\ref{le:eps-less-grad}. This means that \eqref{eq:eps-seq}
  holds with $\epsilon^{(1)}  = 2G^2H$. Therefore we can solve
  \begin{align}
    \epsilon \le \frac{1}{c \left(t + \frac{1}{\epsilon^{(1)}  c}\right)}
    \mbox{~~~with~~~} c := \frac{1}{8G^2H} \mbox{~~and~~} \epsilon^{(1)}  :=
    2G^2H
    \label{eq:solve-upperbound}
  \end{align}
  to obtain an upper bound on $t$ such that $(\forall i \ge t)$
  $\epsilon^{(i)} \le \epsilon < 2G^2H$. The solution to
  \eqref{eq:solve-upperbound} is $t \leq \frac{8 G^2 H}{\epsilon} - 4$.
\end{proof}

\section{Satisfiability of the Subgradient Wolfe Conditions}
\label{sec:PositiveStepSize}

To formally show that there always is a positive step
size that satisfies the subgradient Wolfe conditions
(\ref{eq:subwolfe-decrease},\,\ref{eq:subwolfe-curvature}),
we restate a result of \citet[Theorem VI.2.3.3]{HirLem93}
in slightly modified form:
\begin{lemma}
  Given two points $\vw \neq \vw'$ in $\RR^{d}$, define $\vw_{\eta} =
  \eta \vw' + (1-\eta) \vw$. Let $J: \RR^{d} \to \RR$ be convex.
  There exists $\eta \in (0, 1)$ and $\tilde{\vg} \in \partial
  J(\vw_{\eta})$ such that
  \begin{align*}
    J(\vw') - J(\vw) ~=~ \tilde{\vg}^{\top} \!(\vw' - \vw) ~\leq~
    \hat{\vg}^{\top} \!(\vw' - \vw),
  \end{align*}
  where $\hat{\vg} := \argsup_{\vg \in \partial J(\vw_{\eta})}
  \,\vg^{\top} \!(\vw' - \vw)$.
 \label{lem:conv-mvt}
\end{lemma}

\begin{theorem}
  \label{th:gen-wolfe}
  Let $\vp$ be a descent direction at an iterate $\vw$.
  If $\Phi(\eta) := J(\vw + \eta\vp)$ is bounded below,
  then there exists a step size $\eta > 0$ which satisfies the
  subgradient Wolfe conditions
  (\ref{eq:subwolfe-decrease},\,\ref{eq:subwolfe-curvature}).
\end{theorem}
\begin{proof}
  Since $\vp$ is a descent direction, the line
  $J(\vw) + c_1 \eta \sup_{\vg \in \partial J(\vw)} \vg^{\top} \vp$
  with $c_1 \in (0, 1)$ must intersect $\Phi(\eta)$ at least once at
  some $\eta > 0$ (see Figure~\ref{fig:wolfe} for geometric intuition).
  Let $\eta'$ be the smallest such intersection point; then
  \begin{align}
    J(\vw + \eta' \vp) ~=~ J(\vw) ~+~ c_1 \eta'
    \!\!\sup_{\vg \in \partial J(\vw)}\!\! \vg^{\top} \vp.
    \label{eq:intersect}
  \end{align}
  Since $\Phi(\eta)$ is lower bounded, the sufficient decrease condition
  \eqref{eq:subwolfe-decrease} holds for all $\eta'' \in [0, \eta']$.
  Setting $\vw' = \vw + \eta' \vp$ in Lemma~\ref{lem:conv-mvt}
  implies that there exists an $\eta'' \in (0, \eta')$ such that
  \begin{align}
    \label{eq:j-conv-mvt}
    J(\vw + \eta' \vp) \,-\, J(\vw) ~\leq~ \eta'
    \!\!\!\sup_{\vg \in \partial J(\vw + \eta''\vp)}\!\!\! \vg^{\top} \vp.
  \end{align}
  Plugging \eqref{eq:intersect} into \eqref{eq:j-conv-mvt} and
  simplifying it yields
  \begin{align}
    \label{eq:simp-wolfe}
    c_1 \!\!\sup_{\vg \in \partial J(\vw)}\!\! \vg^{\top} \vp ~\leq
    \!\!\sup_{\vg \in \partial J(\vw + \eta''\vp)}\!\!\! \vg^{\top}
    \vp.
  \end{align}
  Since $\vp$ is a descent direction, $\sup_{\vg \in \partial J(\vw)}
  \vg^{\top}\vp < 0$, and thus \eqref{eq:simp-wolfe} also holds when
  $c_1$ is replaced by $c_2 \in (c_1, 1)$.
\end{proof}

\section{Global Convergence of  SubBFGS}
\label{sec:convergence-proof}

There are technical difficulties in extending the classical BFGS
convergence proof to the nonsmooth case. This route was taken by
\citet{AndGao07}, which unfortunately left their proof critically flawed:
In a key step \citep[Equation 7]{AndGao07} they seek to establish the
non-negativity of the directional derivative $f'(\bar{x}; \bar{q})$ of a
convex function $f$ at a point $\bar{x}$ in the direction $\bar{q}$,
where $\bar{x}$ and $\bar{q}$ are the limit points of convergent
sequences $\{x^{k}\}$ and $\{\hat{q}^{k}\}_{\kappa}$, respectively.
They do so by taking the limit for $k \in \kappa$ of
\begin{align}
  f'(x^{k} + \tilde{\alpha}^{k}\hat{q}^{k}; \hat{q}^{k}) ~>~ \gamma
  f'(x^{k}; \hat{q}^{k}), \mbox{~~where~~} \{\tilde{\alpha}^{k}\}
  \rightarrow 0 \mbox{~~and~~} \gamma \in (0, 1) \,,
  \label{eq:wrong-ineq1}
\end{align}
which leads them to claim that
\begin{align}
   f'(\bar{x}; \bar{q}) ~\ge~ \gamma f'(\bar{x}; \bar{q}) \,,
   \label{eq:wrong-ineq2}
\end{align}
which would imply $f'(\bar{x}; \bar{q}) \ge 0$ because $\gamma \in (0, 1)$.
However, $f'(x^{k}, \hat{q}^{k})$ does not necessarily converge to
$f'(\bar{x}; \bar{q})$ because the directional derivative of a nonsmooth
convex function is not continuous, only \emph{upper semi-continuous}
\citep[Proposition B.23]{Bertsekas99}. Instead of \eqref{eq:wrong-ineq2}
we thus only have
\begin{align}
f'(\bar{x}; \bar{q}) ~\ge~ \gamma \mathop{\lim\sup}_{k \rightarrow \infty, k \in \kappa} f'(x^{k};
  \hat{q}^{k}) \,,
\end{align}
which does not suffice to establish the desired result:
$f'(\bar{x}; \bar{q}) \ge 0$.  A similar mistake is also
found in the reasoning of \citet{AndGao07} just after Equation 7.

Instead of this flawed approach, we use the technique introduced
by \citet{BirQiWei98} to prove the global convergence of subBFGS
(Algorithm~\ref{alg:subbfgs}) in objective function value, \ie $J(\vw_t)
\rightarrow \inf_{\vw} J(\vw)$, provided that the spectrum of BFGS'
inverse Hessian approximation $\Bmat_t$ is bounded from above and
below for all $t$, and the step size $\eta_t$ (obtained at
\SUBBFGSSTEP) is not summable: $\sum^{\infty}_{t = 0}\eta_t =
\infty.$

\begin{algorithm}[t]
  \caption{~Algorithm 1 of \citet{BirQiWei98}}
  \label{alg:general-eps-subgd}
  \begin{algorithmic}[1]
    \STATE Initialize: $t := 0$ and $\vw_0$ 
    \WHILE{not converged} 
    \STATE ~~Find $\vw_{t+1}$ that obeys
    \vspace{-1ex}
   \begin{align}
    J(\vw_{t+1}) & \,\le\, J(\vw_t) \,-\, a_t \,\norm{\vg_{\epsilon'_t}}^2 +\, \epsilon_t
    \label{eq:general-eps-subgd} \\ \nonumber
  \text{where~~} & \vg_{\epsilon'_t} \in \partial_{\epsilon'_t}
  J(\vw_{t+1}), ~a_t >0, ~\epsilon_t, \epsilon'_t \ge 0 \,. \hspace*{10ex}
  \end{align}
   \vspace{-4ex}
  \STATE ~~$t := t+1$
  \ENDWHILE
  \end{algorithmic}
\end{algorithm}

\citet{BirQiWei98} provide a unified framework for convergence analysis of
optimization algorithms for nonsmooth convex optimization, based on the
notion of \emph{$\epsilon$-subgradients}. Formally, $\vg$ is called an
$\epsilon$-subgradient of $J$ at $\vw$ iff \citep[Definition
XI.1.1.1]{HirLem93}
\begin{align}
  (\forall \vw')~~ J(\vw') ~\ge~ J(\vw) + \inner{(\vw' - \vw)}{\vg} -\epsilon,
  \mbox{~~where~~} \epsilon \ge 0.
\label{eq:eps-subgrad}
\end{align}
The set of all $\epsilon$-subgradients at a point $\vw$ is called the
$\epsilon$-subdifferential, and denoted $\partial_{\epsilon} J(\vw)$.  From the
definition of subgradient \eqref{eq:subgrad-def}, it is easy to see that
$\partial J(\vw) = \partial_{0} J(\vw) \subseteq \partial_{\epsilon} J(\vw)$.
\citet{BirQiWei98} propose an $\epsilon$-subgradient-based algorithm
(Algorithm~\ref{alg:general-eps-subgd}) and provide sufficient conditions
for its global convergence:

\begin{theorem} {\rm \citep[Theorem 2.1(iv), first sentence]{BirQiWei98}}\\
  Let $J:\RR^d \rightarrow \RR \cup \{\infty\}$ be a proper lower
semi-continuous\footnote{This means that there exists at least one $\vw
  \in \RR^d$ such that $J(\vw) < \infty$, and that for all $\vw \in \RR^d$,
  $J(\vw) > -\infty$ and $J(\vw) \le \lim \inf_{t \rightarrow \infty}
  J(\vw_t)$ for any sequence $\{\vw_t\}$ converging to $\vw$. All
  objective functions considered in this paper fulfill these
  conditions.}
extended-valued convex function, and let
 $\{(\epsilon_t, \epsilon'_t, a_t, \vw_{t+1}, \vg_{\epsilon'_t})\}$
 be any sequence generated by Algorithm~\ref{alg:general-eps-subgd} satisfying
\begin{align}
\sum^{\infty}_{t = 0} \epsilon_t < \infty \mbox{~~and~~} \sum^{\infty}_{t = 0} a_t =
\infty.
\label{eq:condition-eps-a}
\end{align}
If $\epsilon'_t \rightarrow 0$, and there exists a positive
number $\beta > 0$ such that, for all large $t$,
\begin{align}
\beta \,\norm{\vw_{t+1} - \vw_t} ~\le~ a_t \norm{\vg_{\epsilon'_t}},
\label{eq:normw-normg}
\end{align}
then $J(\vw_t) \rightarrow \inf_{\vw} J(\vw)$.
\label{th:general-eps-subgd}
\end{theorem}
We will use this result to establish the global convergence
of subBFGS in Theorem~\ref{th:convergence-subbfgs}. Towards this end, we
first show that subBFGS is a special case of
Algorithm~\ref{alg:general-eps-subgd}:

\begin{lemma}
  Let $\vp_t = -\Bmat_t\bar{\vg}_t$ be the descent direction produced by
  Algorithm~\ref{alg:find-descent-dir-cg} at a non-optimal iterate $\vw_t$,
  where $\Bmat_t \succeq h > 0$ and $\bar{\vg}_t \in \partial J(\vw_t)$, and let
  $\vw_{t+1} = \vw_t + \eta_t \vp_t,$ where $\eta_t > 0$ satisfies
  sufficient decrease \eqref{eq:subwolfe-decrease} with
  free parameter $c_1 \in (0,1)$. Then
  $\vw_{t+1}$ obeys \eqref{eq:general-eps-subgd} of
  Algorithm~\ref{alg:general-eps-subgd} for $a_t := \frac{c_1\eta_t h}{2}$,
  $\epsilon_t = 0$, and $\epsilon'_t := \eta_t (1 - \frac{c_1}{2})
  \,\inner{\bar{\vg}_t}{\Bmat_t \bar{\vg}_t}$.
  \label{lem:subbfgs-specialcase}
\end{lemma}

\begin{proof}
Our sufficient decrease condition \eqref{eq:subwolfe-decrease}
  and Corollary~\ref{cor:supgp-pos-bound} imply that
  \begin{align}
    \label{eq:decrease-J-gBg}
    J(\vw_{t+1}) & ~\le~ J(\vw_t) \,-\, \frac{c_1 \eta_t}{2} \,\inner{\bar{\vg}_t}{\Bmat_t
    \bar{\vg}_t} \\
    & ~\le~ J(\vw_t) \,-\, a_t \norm{\bar{\vg}_t}^2, \mbox{~~where~~} a_t := \frac{c_1
      \eta_t h}{2}.
  \end{align}
  What is left to prove is that $\bar{\vg}_t \in \partial_{\epsilon'_t} J(\vw_{t+1})$
  for an $\epsilon'_t \geq 0$.
  Using $\bar{\vg}_t \in \partial J(\vw_t)$ and the definition
  \eqref{eq:subgrad-def} of subgradient, we have
\begin{align}
  \nonumber
  (\forall \vw)~~ J(\vw) & ~\ge~ J(\vw_t) \,+\, \inner{(\vw - \vw_t)}{\bar{\vg}_t} \\
  & ~=~ J(\vw_{t+1}) \,+\, \inner{(\vw - \vw_{t+1})}{\bar{\vg}_t} \,+\,  J(\vw_t) -
  J(\vw_{t+1}) \,+\, \inner{(\vw_{t+1} - \vw_t)}{\bar{\vg}_t} \,.
\label{eq:eps-2t-subgrad}
\end{align}
Using $\vw_{t+1} - \vw_t = - \eta_t \Bmat_t\bar{\vg}_t$ and
\eqref{eq:decrease-J-gBg} gives
\begin{align*}
\nonumber
 (\forall \vw)~~ J(\vw) & ~\ge~ J(\vw_{t+1}) \,+\, \inner{(\vw -
 \vw_{t+1})}{\bar{\vg}_t} \,+\, \frac{c_1 \eta_t }{2} \,\inner{\bar{\vg}_t}{\Bmat_t
\bar{\vg}_t} \,-\, \eta_t \,\inner{\bar{\vg}_t}{\Bmat_t \bar{\vg}_t} \\ 
 & ~=~ J(\vw_{t+1}) \,+\, \inner{(\vw - \vw_{t+1})}{\bar{\vg}_t}
 \,-\, \epsilon'_t \,,
\end{align*}
where $\epsilon'_t := \eta_t (1- \frac{c_1}{2}) \,\inner{\bar{\vg}_t}{
\Bmat_t \bar{\vg}_t}$. Since $\eta_t > 0$, $c_1 < 1$, and
$\Bmat_t \succeq h > 0$, $\epsilon'_t$ is non-negative. 
By the definition \eqref{eq:eps-subgrad} of $\epsilon$-subgradient,
$\bar{\vg}_t \in \partial_{\epsilon'_t} J(\vw_{t+1})$.
\end{proof}

\begin{theorem}
  Let $J:\RR^{d} \to \RR \cup \{\infty\}$ be a proper lower
  semi-continuous\samefootnote{} extended-valued convex
  function. Algorithm~\ref{alg:subbfgs} with a line search that
  satisfies the sufficient decrease condition
  \eqref{eq:subwolfe-decrease} with $c_1 \in (0, 1)$ converges globally
  to the minimal value of $J$, provided that:
  \begin{enumerate}
  \item the spectrum of its approximation to the inverse Hessian is
    bounded above and below: $\exists \,(h, H : 0 < h \leq H < \infty) :
    (\forall t)~ h \preceq \Bmat_t \preceq H$
  \item the step size $\eta_t > 0$ satisfies $\sum^{\infty}_{t=0} \eta_t
    = \infty$, and
  \item the direction-finding tolerance $\epsilon$ for
    Algorithm~\ref{alg:find-descent-dir-cg} 
    satisfies \eqref{eq:tighter-bound-eps}.
  \end{enumerate}
  \label{th:convergence-subbfgs}
\end{theorem}

\begin{proof}
  We have already shown in Lemma~\ref{lem:subbfgs-specialcase} that
  subBFGS is a special case of Algorithm~\ref{alg:general-eps-subgd}.
  Thus if we can show that the technical conditions of
  Theorem~\ref{th:general-eps-subgd} are met, it directly establishes
  the global convergence of subBFGS.
  
  Recall that for subBFGS $a_t := \frac{c_1\eta_t h}{2}$, $\epsilon_t
  = 0$, $\epsilon'_t := \eta_t (1 - \frac{c_1}{2})
  \,\inner{\bar{\vg}_t}{\Bmat_t \bar{\vg}_t}$, and $\bar{\vg}_t =
  \vg_{\epsilon'_t}$. Our assumption on $\eta_t$ implies that
  $\sum^{\infty}_{t=0} a_t = \frac{c_1h}{2} \sum^{\infty}_{t=0} \eta_t
  = \infty$, thus establishing \eqref{eq:condition-eps-a}. We now show
  that $\epsilon'_t \rightarrow 0$. Under the third condition of
  Theorem~\ref{th:convergence-subbfgs}, it follows from the first
  inequality in \eqref{eq:supgp-pos-bound} in 
  Corollary~\ref{cor:supgp-pos-bound} that
  \begin{align}
    \sup_{\vg \in \partial J(\vw_t)}\!\!\!
    \inner{\vg}{\vp_t} ~\le\, -\half \inner{\bar{\vg}_t}{\Bmat_t
      \bar{\vg}_t} \,,  
    \label{eq:another_inequality}
  \end{align} 
  where $\vp_t = -\Bmat_t\bar{\vg}_t, ~\bar{\vg}_t \in \partial
  J(\vw_t)$ is the search direction returned by
  Algorithm~\ref{alg:find-descent-dir-cg}. Together with the
  sufficient decrease condition \eqref{eq:subwolfe-decrease},
  \eqref{eq:another_inequality} implies
  \eqref{eq:decrease-J-gBg}. Now use \eqref{eq:decrease-J-gBg}
  recursively to obtain
  \begin{align}
    J(\vw_{t+1}) ~\le~ J(\vw_0) \,-\, \frac{c_1}{2} \sum^{t}_{i=0} \eta_i
    \,\inner{\bar{\vg}_i}{\Bmat_i \bar{\vg}_i} \,. 
  \end{align}
  Since $J$ is proper (hence bounded from below), we have
  \begin{align}
    \sum^{\infty}_{t=0} \eta_i \,\inner{\bar{\vg}_i}{\Bmat_i \bar{\vg}_i}
    ~=~ \frac{1}{1 - \frac{c_1}{2}} \sum^{\infty}_{t=0} \epsilon'_i ~<~ \infty \,.
    \label{eq:summable}
  \end{align}
  Recall that $\epsilon'_i \geq 0$. The bounded sum of non-negative
  terms in \eqref{eq:summable} implies that the terms in the
  sum must converge to zero.

  Finally, to show \eqref{eq:normw-normg} we use $\vw_{t+1} - \vw_t =
  -\eta_t \Bmat_t \bar{\vg}_t$, the definition of the matrix norm:
  $\|\Bmat\| := \max_{\vx \neq 0} \frac{\|\Bmat \vx\|}{\|\vx\|}$, and the
  upper bound on the spectrum of $\Bmat_{t}$ to write:
  \begin{align}
    \norm{\vw_{t+1} - \vw_t} ~=~ \eta_t \norm{\Bmat_t \bar{\vg}_t} ~\le~
    \eta_t \norm{\Bmat_t}\norm{\bar{\vg}_t} ~\le~ \eta_t H
    \norm{\bar{\vg}_t}.
    \label{eq:lottanorms}
  \end{align}
  Recall that $\bar{\vg}_t = \vg_{\epsilon'_t}$ and $a_{t} =
  \frac{c_{1}\eta_{t} h}{2}$, and multiply both sides of
  \eqref{eq:lottanorms} by $\frac{c_{1}h}{2H}$ to obtain
  \eqref{eq:normw-normg} with $\beta := \frac{c_1 h}{2
    H}$. \hspace*{0pt}
\end{proof}

\section{SubBFGS Converges on Various Counterexamples}
\label{sec:counterexample}

We demonstrate the global convergence of subBFGS\footnote{We run
  Algorithm \ref{alg:subbfgs} with $h = 10^{-8}$ and $\epsilon = 10^{-5}$.}
with an exact line search on various counterexamples from the literature,
designed to show the failure to converge of other gradient-based algorithms.

\begin{figure}[tb]
  \centering
  \begin{tabular}{cc}
    \includegraphics[width=0.45\linewidth]{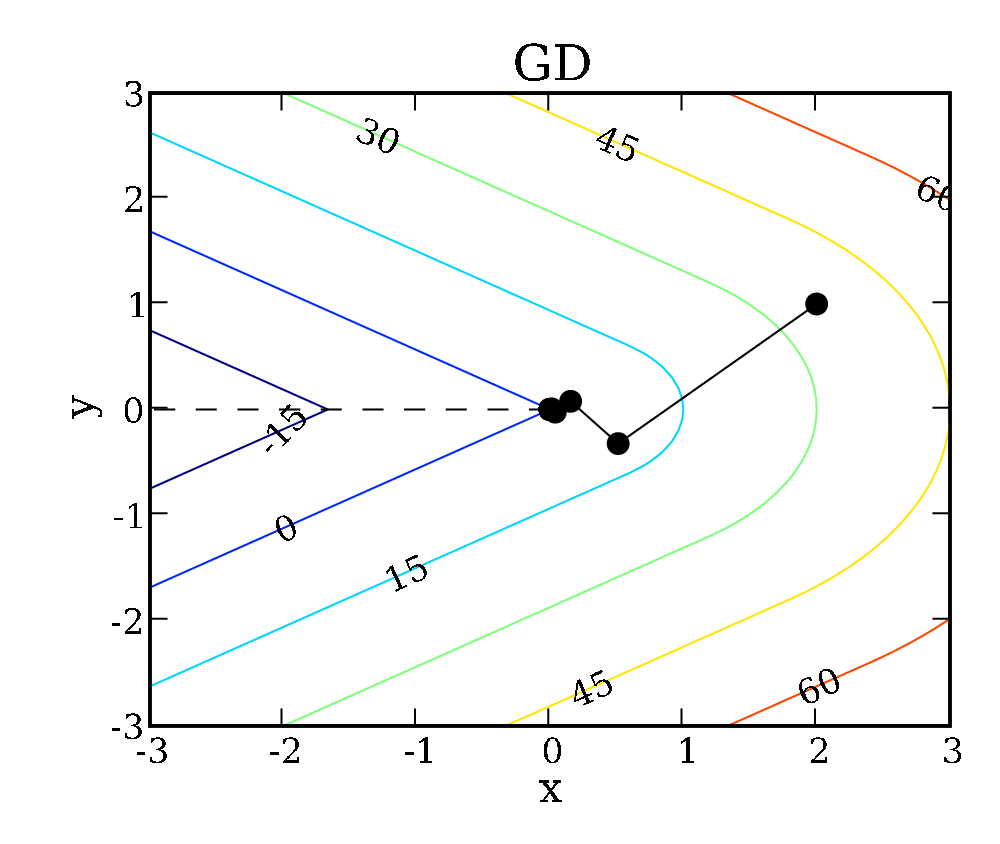} &
    \includegraphics[width=0.45\linewidth]{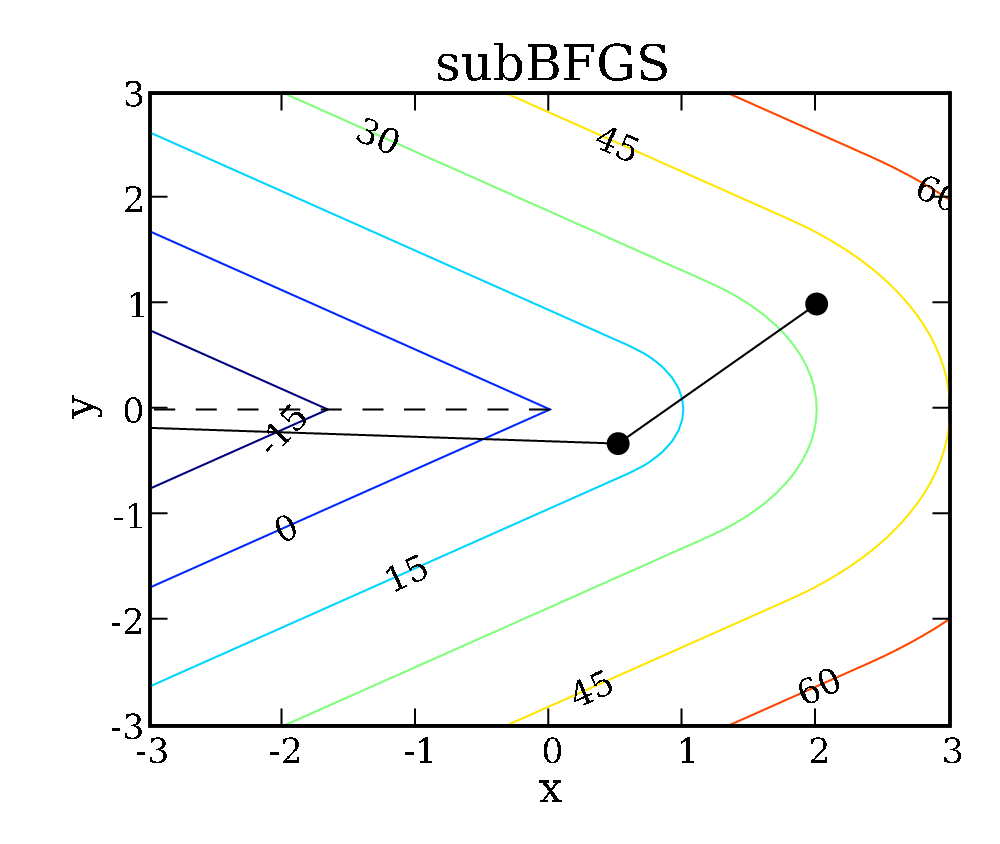} \\
  \end{tabular}
  \caption{Optimization trajectory of steepest descent (left) and
    subBFGS (right) on counterexample \eqref{eq:W-example}.}
  \label{fig:W-example}
\end{figure}

\subsection{Counterexample for Steepest Descent}

The first counterexample \eqref{eq:W-example} is given by
\citet{Wolfe75} to show the non-convergent behaviour of the steepest
descent method with an exact line search (denoted GD):
\begin{align}
  f(x, y) :=
  \begin{cases}
    5\sqrt{(9x^2 + 16y^2)} &\mbox{~~~if~~~} x \ge |y|,\\
    9x + 16|y| &\mbox{~~~otherwise}.
  \end{cases}
  \label{eq:W-example}
\end{align}
This function is subdifferentiable along $x \le
0,~y=0$ (dashed line in Figure~\ref{fig:W-example}); its minimal
value ($-\infty$) is attained for $x = -\infty$.
As can be seen in Figure \ref{fig:W-example} (left), starting from a
differentiable point $(2,1)$, GD follows successively orthogonal
directions, \ie $-\nabla f(x, y)$, and converges to the non-optimal
point $(0, 0)$. As pointed out by \citet{Wolfe75}, the failure of GD
here is due to the fact that GD does not have a global view of $f$,
specifically, it is because the gradient evaluated at each iterate
(solid disk) is not informative about $\partial f(0, 0)$, which
contains subgradients (\eg $(9,0)$), whose negative directions point
toward the minimum. SubBFGS overcomes this ``short-sightedness'' by
incorporating into the parameter update \eqref{eq:update} an estimate
$\Bmat_t$ of the inverse Hessian, whose information about the shape of
$f$ prevents subBFGS from zigzagging to a non-optimal point. Figure
\ref{fig:W-example} (right) shows that subBFGS moves to the correct
region ($x < 0$) at the second step. In fact, the second step of
subBFGS lands exactly on the hinge $x \le 0, y = 0$, where a
subgradient pointing to the optimum is
available. 

\begin{figure}[tb]
  \centering
  \begin{tabular}{cc}
    \includegraphics[width=0.45\linewidth]{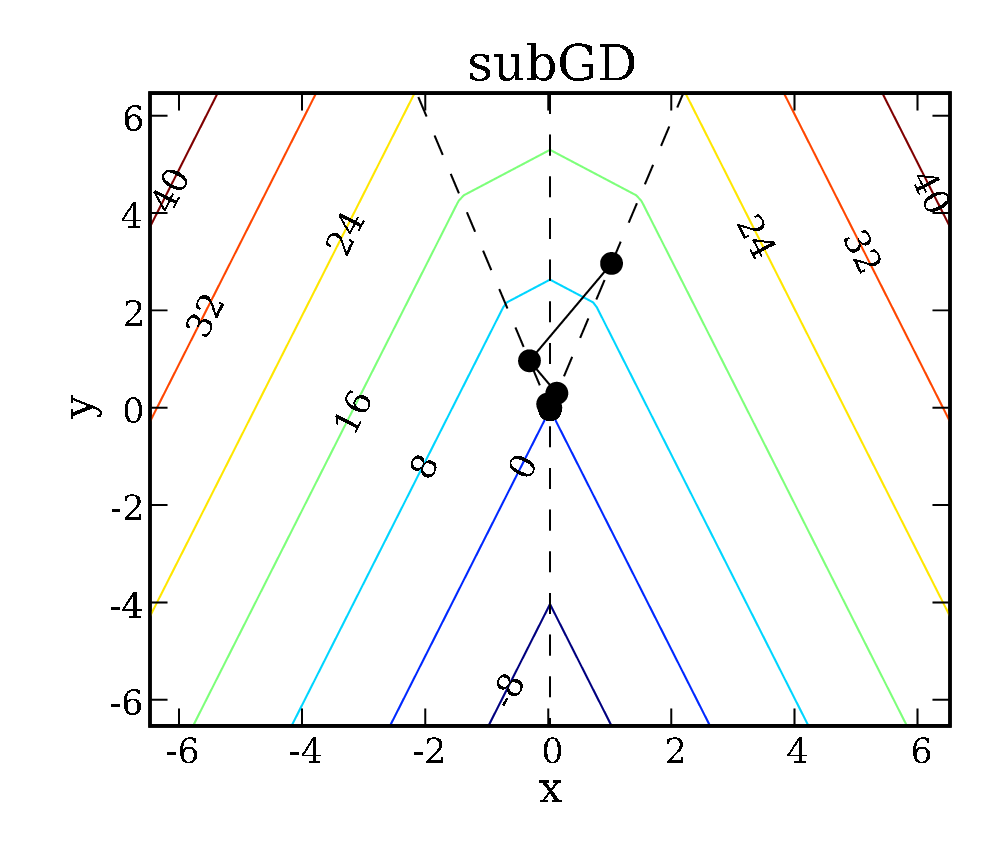} &
    \includegraphics[width=0.45\linewidth]{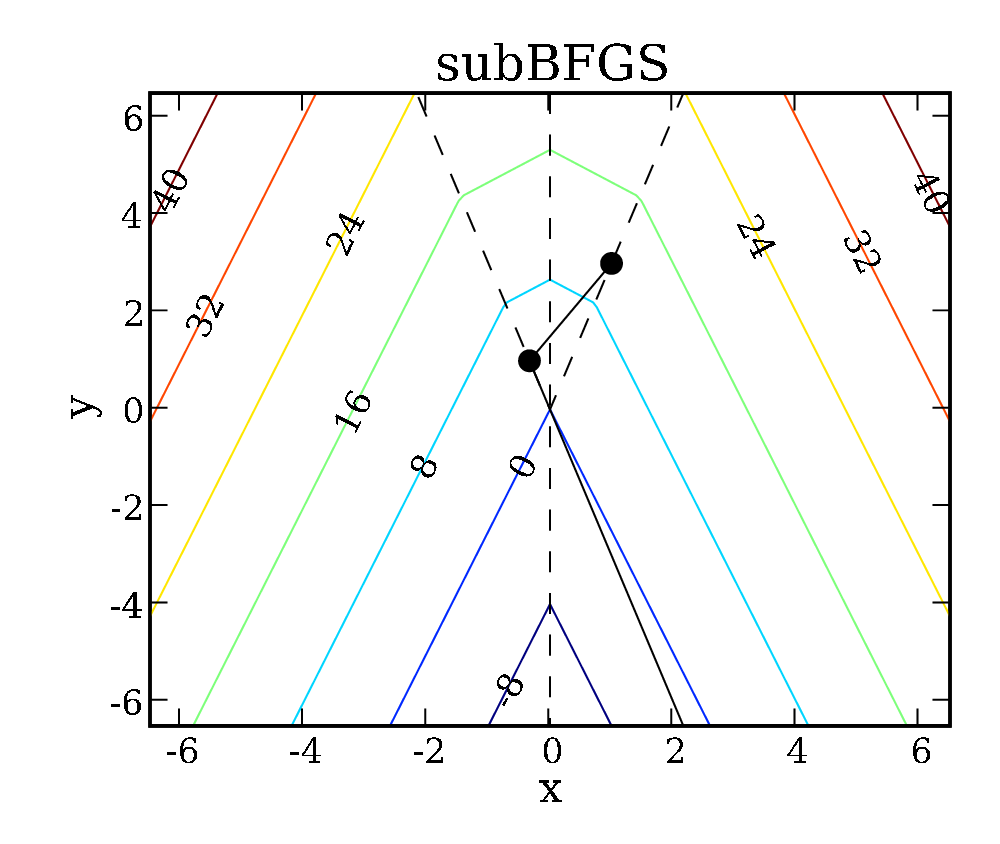} \\
  \end{tabular}
  \caption{Optimization trajectory of steepest subgradient descent
    (left) and subBFGS (right) on counterexample \eqref{eq:UL-example}.}
  \label{fig:UL-example}
\end{figure}

\subsection{Counterexample for Steepest Subgradient Descent}

The second counterexample \eqref{eq:UL-example}, due to
\citet[Section VIII.2.2]{HirLem93}, is a piecewise linear function
which is subdifferentiable along $0 \le y = \pm 3x$ and $x = 0$ (dashed
lines in Figure~\ref{fig:UL-example}):
\begin{align}
f(x,y) := \max \{-100,~ \pm 2x + 3y, ~ \pm5x + 2y \}.
\label{eq:UL-example}
\end{align}
This example shows that steepest subgradient descent with an exact
line search (denoted subGD) may not converge to the optimum of a
nonsmooth function.  Steepest subgradient descent updates parameters
along the \emph{steepest descent} subgradient direction, which is
obtained by solving the min-sup problem \eqref{eq:minmax-subgrad}
with respect to the Euclidean norm. Clearly, the minimal value of $f$ ($-100$)
is attained for sufficiently negative values of $y$. However, subGD
oscillates between two hinges $0 \le y = \pm 3x$, converging to the
non-optimal point $(0, 0)$, as shown in Figure \ref{fig:UL-example}
(left). The zigzagging optimization trajectory of subGD does not allow
it to land on any informative position such as the hinge $y = 0$,
where the steepest subgradient descent direction points to the desired
region ($y < 0$); \citet[Section VIII.2.2]{HirLem93} provide a detailed
discussion.
By contrast, subBFGS moves to the $y < 0$ region at the second step
(Figure~\ref{fig:UL-example}, right), which ends at the point $(100,
-300)$ (not shown in the
figure) where the minimal value of $f$ is attained . 

\subsection{Counterexample for BFGS}

\begin{figure}[tb]
  \centering
  \begin{tabular}{cc}
    \includegraphics[width=0.45\linewidth]{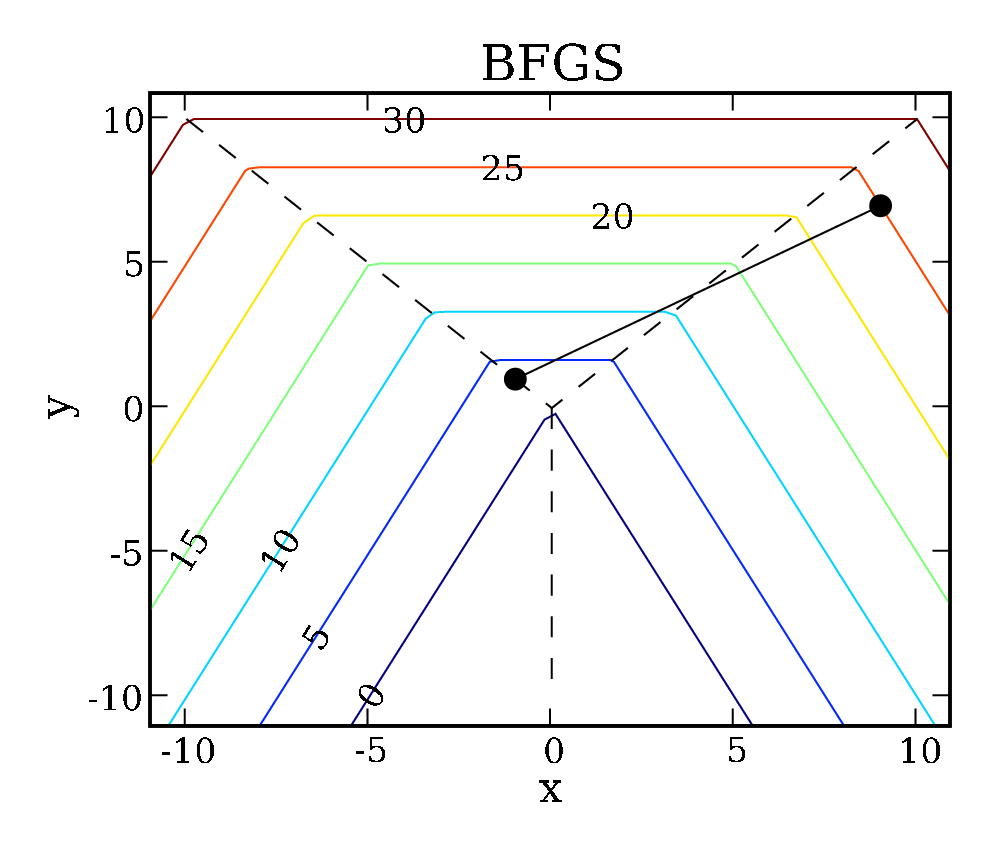} &
    \includegraphics[width=0.45\linewidth]{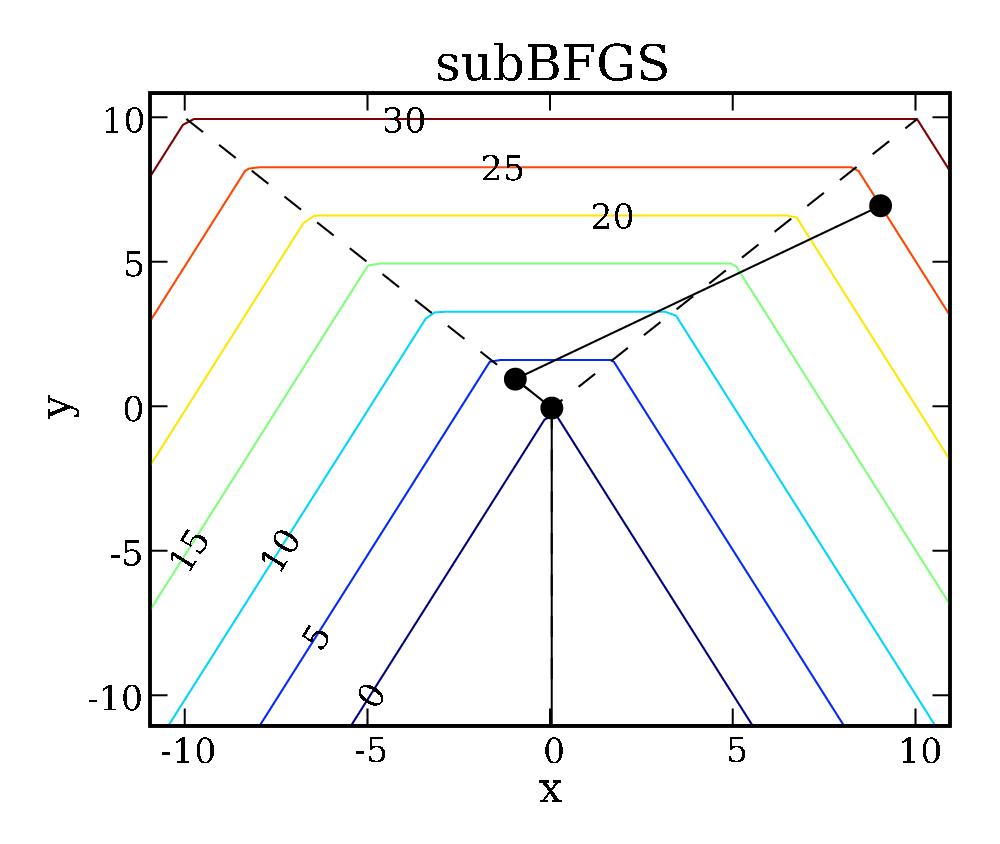} \\
  \end{tabular}
  \caption{Optimization trajectory of standard BFGS (left) and
    subBFGS (right) on counterexample \eqref{eq:LO-example}.}
  \label{fig:LO-example}
\end{figure}

The final counterexample \eqref{eq:LO-example} is given by
\citet{LewOve08b} to show that the standard BFGS algorithm with an
exact line search can break down when encountering a nonsmooth point:
\begin{align}
  f(x, y) := \max\{ 2 |x| + y, ~ 3y\}.
  \label{eq:LO-example}
\end{align}
This function is subdifferentiable along $x = 0,~ y \le 0$ and $y =
|x|$ (dashed lines in Figure \ref{fig:LO-example}). Figure
\ref{fig:LO-example} (left) shows that after the first step, BFGS
lands on a nonsmooth point, where it fails to find a descent
direction. This is not surprising because at a nonsmooth point $\vw$
the quasi-Newton direction $\vp := -\Bmat \vg$ for a given subgradient
$\vg \in \partial J(\vw)$ is not necessarily a direction of
descent. SubBFGS fixes this problem by using a direction-finding
procedure (Algorithm \ref{alg:find-descent-dir-cg}), which is
guaranteed to generate a descent quasi-Newton direction. Here subBFGS
converges to $f = -\infty$ in three iterations
(Figure~\ref{fig:LO-example}, right).

\end{document}